\documentclass{article}

\usepackage{float}\usepackage{colortbl}
\usepackage{amsmath, amssymb, amsthm, mathrsfs}
\usepackage{enumitem}
\usepackage{placeins}
\usepackage{xparse}
\usepackage{color}

\usepackage[numbers]{natbib}

\usepackage{hyperref}
\hypersetup{
	colorlinks = true,
	linkcolor = blue,
	anchorcolor = blue,
	citecolor = blue,
	filecolor = blue,
	urlcolor = blue
}
\usepackage{txfonts}

\usepackage{newtxtext}

\usepackage{caption}
\usepackage{accents}

\usepackage{booktabs}
\newcommand{\ra}[1]{\renewcommand{\arraystretch}{#1}}
\renewcommand{\arraystretch}{1.25}
\usepackage{lscape}

\usepackage{tablefootnote}

    \newtheorem{theorem}{Theorem}
    
    \newtheorem{thmy}{Result}
    \newenvironment{thmx}{\stepcounter{thm}\begin{thmy}}{\end{thmy}}
\newtheorem{definition}[theorem]{Definition}

\newtheorem{corollary}[theorem]{Corollary}
\newtheorem{lemma}[theorem]{Lemma}
\newtheorem{proposition}[theorem]{Proposition}
\newtheorem{remark}[theorem]{Remark}

\theoremstyle{remark}
\newtheorem{setting}[theorem]{Setting}

\theoremstyle{definition}
\newtheorem{examplex}[theorem]{Example}
\newenvironment{example}
  {\pushQED{\qed}\examplex}
  {\popQED\endexamplex}

    \newcommand{\fff}{\mathcal{F}}

 \newcommand{\set}[1]{\left\{#1\right\}}

 \renewcommand{\phi}{\varphi}
\newcommand{\pp}{\mathcal P}

\newcommand{\R}{\mathbb{R}}

\newcommand{\Z}{\mathbb{Z}}

\newcommand{\bes}{\begin{subequations}}
\newcommand{\ees}{\end{subequations}}
\newcommand{\eea}{\end{eqnarray}}

    \newcommand{\rr}{{\mathbb R}}

\usepackage{bbm}

\renewcommand{\epsilon}{\varepsilon}

\newcommand{\fourIdx}[5]{%
\setbox1=\hbox{\ensuremath{^{#1}}}%
 \setbox2=\hbox{\ensuremath{_{#2}}}%
 \setbox5=\hbox{\ensuremath{#5}}%
 \hspace{\ifnum\wd1>\wd2\wd1\else\wd2\fi}%
 \ensuremath{\copy5^{\hspace{-\wd1}\hspace{-\wd5}#1\hspace{\wd5}#3}%
 _{\hspace{-\wd2}\hspace{-\wd5}#2\hspace{\wd5}#4}%
 }}

\numberwithin{equation}{section}
\numberwithin{theorem}{section}

\renewcommand{\subset}{\subseteq}

\newcommand{\bra}[1]{\left( #1 \right)}

\usepackage{subcaption}
\usepackage{graphicx}

\renewcommand{\mathrm}{}

\makeatletter
\newcommand{\mylabel}[2]{#2\def\@currentlabel{#2}\label{#1}}
\makeatother

\usepackage{relsize}
\def\fcmp{\mathbin{\raise 0.6ex\hbox{\oalign{\hfil$\scriptscriptstyle \mathrm{o}$\hfil\cr\hfil$\scriptscriptstyle\mathrm{9}$\hfil}}}}

\newcommand{\nn}{\mathbb{N}}

\newcommand{\xxx}{\mathcal{X}}
\newcommand{\yyy}{\mathcal{Y}}
\newcommand{\zzz}{\mathcal{Z}}
\NewDocumentCommand{\NN}{oo}{
    \ensuremath{
        \mathcal{NN}
        \IfValueT{#1}{_{#1}}\IfValueF{#1}{_{[d]}}
        \IfValueT{#2}{^{#2}}\IfValueF{#2}{^{\sigma}}
    }
}
\NewDocumentCommand{\GT}{oo}{
    \ensuremath{
        \mathcal{GT}
        \IfValueT{#1}{_{#1}}\IfValueF{#1}{_{[d],N,q}}
        \IfValueT{#2}{^{#2}}\IfValueF{#2}{^{\sigma}}
        \left(
        \rr^d
            ,
        \yyy
        \right)
    }
}
\usepackage{wasysym} %

\usepackage{marginnote}
\usepackage{xcolor}
\definecolor{darkcyan}{rgb}{0.0, 0.55, 0.55}
\definecolor{MidnightBlue}{RGB}{25,25,112}
\definecolor{MidnightBlueComplementingGreen}{RGB}{25,112,25}
\definecolor{MidnightBlueComplementingPurple}{RGB}{112,25,112}
\definecolor{MidnightBlueComplementingRed}{RGB}{112,25,69}
\definecolor{WowColor}{rgb}{.75,0,.75}
\definecolor{MildlyAlarming}{rgb}{0.85,0.25,0.1}
\definecolor{SubtleColor}{rgb}{0,0,.50}
\definecolor{antiquefuchsia}{rgb}{0.57, 0.36, 0.51}
\definecolor{fashionfuchsia}{rgb}{0.96, 0.0, 0.63}
\definecolor{jade}{rgb}{0.0, 0.66, 0.42}
\definecolor{caribbeangreen}{rgb}{0.0, 0.8, 0.6}
\definecolor{aquamarine}{rgb}{0.5, 0.8, 0.85}
\definecolor{attentioncolor}{RGB}{152,90,81}
\definecolor{burgred}{RGB}{40,3,22}
\definecolor{darkpastelred}{RGB}{194,59,34}
\definecolor{AnnieGreen}{RGB}{17,123,92}
\definecolor{Turquoise}{RGB}{64,224,208}
\definecolor{darkjade}{RGB}{0,122,84}
\definecolor{Asparagus}{rgb}{0.53, 0.66, 0.42}
\definecolor{Darkspringgreen}{rgb}{0.09, 0.45, 0.27}

\definecolor{Window1}{RGB}{92,150,31}%
    \definecolor{Window1dark}{RGB}{41,67,13}%
\definecolor{Window2}{RGB}{255,168,28}
    \definecolor{Window2dark}{RGB}{114,75,12}
\definecolor{Window3}{RGB}{255,96,33}
    \definecolor{Window3dark}{RGB}{97,36,12}
\definecolor{InputColor}{RGB}{20,255,177}
    \definecolor{InputColorlight}{RGB}{222,237,229}

\usepackage[toc,page]{appendix}

\usepackage[colorinlistoftodos]{todonotes}

\NewDocumentCommand{\Bea}{mo}{
    \IfValueF{#2}{
                        {{
                            \textcolor{magenta}{ 
                            \textbf{B:}
                            \textit{{#1}}
                            }
                        }}
        }
    \IfValueT{#2}{
                        \marginnote{{\scriptsize
                            \textcolor{blue}{ 
                            \textbf{B:}
                            \textit{{#1}}
                            }
                        }}
        }
                    }

\NewDocumentCommand{\Gudi}{mo}{
    \IfValueF{#2}{
                        {{
                            \textcolor{caribbeangreen}{ 
                            \textbf{G:}
                            \textit{{#1}}
                            }
                        }}
        }
    \IfValueT{#2}{
                        \marginnote{{\scriptsize
                            \textcolor{caribbeangreen}{ 
                            \textbf{G:}
                            \textit{{#1}}
                            }
                        }}
        }
                    }

\usepackage[normalem]{ulem}
\usepackage{adjustbox}

\newcommand{\eqdef}{\ensuremath{
        \overset{
                \scalebox{.5}{\mbox{def.}}
            }{=}
}}

\NewDocumentCommand{\Anastasis}{mo}{
    \IfValueF{#2}{
                        {{\scriptsize
                            \textcolor{caribbeangreen}{ 
                            \textbf{A:}
                            \textit{{#1}}
                            }
                        }}
        }
    \IfValueT{#2}{
                        \marginnote{{\scriptsize
                            \textcolor{caribbeangreen}{ 
                            \textbf{A:}
                            \textit{{#1}}
                            }
                        }}
        }
                    }

\NewDocumentCommand{\Chong}{mo}{
    \IfValueF{#2}{
                        {{\scriptsize
                            \textcolor{blue}{ 
                            \textbf{C:}
                            \textit{{#1}}
                            }
                        }}
        }
    \IfValueT{#2}{
                        \marginnote{{\scriptsize
                            \textcolor{blue}{ 
                            \textbf{C:}
                            \textit{{#1}}
                            }
                        }}
        }
                    }

\NewDocumentCommand{\Matti}{mo}{
    \IfValueF{#2}{
                        {{\scriptsize
                            \textcolor{attentioncolor}{ 
                            \textbf{Matti:}
                            \textit{{#1}}
                            }
                        }}
        }
    \IfValueT{#2}{
                        \marginnote{{\scriptsize
                            \textcolor{attentioncolor}{ 
                            \textbf{Matti:}
                            \textit{{#1}}
                            }
                        }}
        }
                    }

\NewDocumentCommand{\Maarten}{mo}{
    \IfValueF{#2}{
                        {{\scriptsize
                            \textcolor{red}{ 
                            \textbf{Maarten:}
                            \textit{{#1}}
                            }
                        }}
        }
    \IfValueT{#2}{
                        \marginnote{{\scriptsize
                            \textcolor{green}{ 
                            \textbf{Maarten:}
                            \textit{{#1}}
                            }
                        }}
        }
                    }

\NewDocumentCommand{\Ivan}{mo}{
    \IfValueF{#2}{
                        {{\scriptsize
                            \textcolor{blue}{ 
                            \textbf{I:}
                            \textit{{#1}}
                            }
                        }}
        }
    \IfValueT{#2}{
                        \marginnote{{\scriptsize
                            \textcolor{blue}{ 
                            \textbf{I:}
                            \textit{{#1}}
                            }
                        }}
        }
                    }

\NewDocumentCommand{\IvanL}{mo}{
    \IfValueF{#2}{
                        {{\scriptsize
                            \textcolor{blue}{ 
                            \textbf{I:}
                            \textit{{#1}}
                            }
                        }}
        }
    \IfValueT{#2}{
                        \reversemarginpar\marginnote{{\scriptsize
                            \textcolor{blue}{ 
                            \textbf{I:}
                            \textit{{#1}}
                            }
                        }}
                        \normalmarginpar
        }
                    }

\newenvironment{denseitemize}
{ \begin{itemize}
    \setlength{\itemsep}{0pt}
    \setlength{\parskip}{0pt}
    \setlength{\parsep}{0pt}     }
{ \end{itemize}                  } 

\usepackage{tikz}
\usepackage{tikz-cd}

\title{An Approximation Theory for Metric Space-Valued Functions\\%
With A View Towards Deep Learning}

\author{Anastasis Kratsios\thanks{Department of Mathematics and Statistics, McMaster University, Canada} \thanks{Corresponding Author: kratsioa@mcmaster.ca}, Chong Liu\thanks{Institute of Mathematical Sciences, ShanghaiTech University, China}, Matti Lassas\thanks{Department of Mathematics and Statistics, University of Helsinki, Finland}, Maarten V. de Hoop\thanks{Department of Mathematics, Rice University, USA}, Ivan Dokmani\'{c}\thanks{Department of Mathematics and Computer Science, University of Basel, Switzerland}}

\begin{document}

\maketitle




\begin{abstract}
Motivated by the developing mathematics of deep learning, we build universal functions approximators of continuous maps between arbitrary Polish metric spaces $\xxx$ and $\yyy$ using elementary functions between Euclidean spaces as building blocks. Earlier results assume that the target space $\yyy$ is a topological vector space. We overcome this limitation by ``randomization'': our approximators output discrete probability measures over $\yyy$. When $\xxx$ and $\yyy$ are Polish without additional structure, we prove very general qualitative guarantees; when they have suitable combinatorial structure, we prove quantitative guarantees for H\"{o}lder-like maps, including maps between finite graphs, solution operators to rough differential equations between certain Carnot groups, and continuous non-linear operators between Banach spaces arising in inverse problems. In particular, we show that the required number of Dirac measures is determined by the combinatorial structure of $\xxx$ and $\yyy$. For barycentric $\yyy$, including Banach spaces, $\mathbb{R}$-trees, Hadamard manifolds, or Wasserstein spaces on Polish metric spaces, our approximators reduce to $\yyy$-valued functions. When the Euclidean approximators are neural networks, our constructions generalize transformer networks, providing a new probabilistic viewpoint of geometric deep learning.
\end{abstract}

\noindent\textbf{Keywords}: Approximation Theory, Rough Paths, Inverse Problems, Metric Spaces. \hfill





\section{Introduction}
\label{s:Introduction}
Universal approximation theorems are guarantees that classes of functions with favourable algorithmic properties, such as those that can be efficiently computed on modern hardware, are dense within given function spaces.  Universal approximation was pioneered in neural network theory in the early $90$s by Hornik and Cybenko \cite{hornik1989multilayer, Cybenko}, who showed that for any compact $K \subset \mathbb{R}^m$, non-polynomial%
\footnote{The set of continuous functions from a topological space $\xxx$ to a metric space $\yyy$ is denoted by $C(\xxx,\yyy)$.  Unless otherwise stated, this set is equipped with the topology of uniform convergence on compact sets, i.e.\ the compact-open topology.} 
$\sigma \in C(\mathbb{R}, \mathbb{R})$, any $f \in C(K, \mathbb{R}^n)$, and any $\epsilon > 0$, there is an integer $k \in \mathbb{N}$, a vector $b \in \mathbb{R}^k$, and matrices $A \in \mathbb{R}^{k \times m}$ and $C \in \mathbb{R}^{n \times k}$, such that   
\begin{equation}
        \sup_{x \in K} 
        ~
        \| 
            f(x) - C (\sigma \circ (Ax + b )) 
        \| 
    < 
        \epsilon
    .
\end{equation}
This motivated that the class of functions
\begin{equation}
    \mathcal{F} = \set{x \mapsto C (\sigma \circ (Ax + b ))
    :\,
    A\in \mathbb{R}^{k\times m},\, b\in \mathbb{R}^k,\, C\in \mathbb{R}^{n\times k}
    },
    \label{eq:one-hidden-layer}
\end{equation}
known in machine learning as single-hidden-layer neural networks, be called a \textit{universal approximator} for $C(\mathbb{R}^m, \mathbb{R}^n)$. There exist many other universal approximators for $C(\mathbb{R}^m, \mathbb{R}^n)$; classical examples preceding neural networks include polynomials \cite{WoodBerstein_1984_Classical,DellaMastroianniSzabados_2004_WeightedBernstein,Draganov_2019_SimultaneousapproximationBernseteinpolynomailsintegercoefficients,Dragonov_SimultaneousBersnstein_2020} and splines \cite{Dahmen_1982_AdaptiveApproxSplines,Pence_1987_BestApproxSplines,AndersonBabenkoLeskevysch_2014_ApproxiSplinesSimultaneous}.

The universal approximation theory of neural networks has since seen a spate of results focusing on the advantages of multiple hidden layers (the so-called network \textit{depth}) and sharpening guarantees for functions with various regularity priors.  While there has been limited work on approximation between Riemannian manifolds \cite{kratsios2020non,kratsios2021_GDL,puthawala2022universal} and between function spaces, \cite{chen1995universal,li2020fourier,kovachki2021neural,KovachkiLanthalerMishra_JMLR_2021_FourierNeuralOperatorsUniversal,lu2021learning,gonon2023infinite}, most results address finite-dimensional normed source and target spaces and focus on cases where approximation rates can be improved either by exploiting assumptions on the target function regularity \cite{elbrachter2021deep,mhaskar2016deep,guhring2021approximation,Haizhao_2021_DeepNetworkReLUSmoothFunctions,marcati2022exponential} or by fine-tuning the structure of the usual feed-forward network (cf. \eqref{eq:one-hidden-layer}) \cite{shen2021neural,yarotsky2021elementary,shen2021deep,zhang2022neural}.  In this paper, we propose a significant generalization of these results.

When $\xxx$ and $\yyy$ are more general than finite-dimensional normed spaces, a key difficulty in building a universal class for $C(\xxx, \yyy)$ lies between topology and efficient computation.  Whenever $\xxx$ and $\yyy$ are non-contractible, $C(\xxx, \yyy)$ has multiple (often infinitely many) connected components \cite{MilnorHomotopyTypeCWComplexMappingSpace1959}. A class $\mathcal{F}$ can only be a universal approximator for $C(\xxx, \yyy)$ if it contains universal approximators for each connected component of $C(\xxx,\yyy)$.  Since the connected components are enumerated by the homotopy classes in $C(\xxx,\yyy)$,  universal approximation by $\mathcal{F}$ demands that it exhausts all homotopy classes.  A direct verification of universality within each class would require enumerating the homotopy classes. This, in turn requires computing the homotopy groups of $\xxx$ and $\yyy$, which is in general NP-hard even when we are given simplicial complexes for $\xxx$ and $\yyy$, and even when they are as topologically simple as spheres \cite{LechugaMurillo_2000_NPHARDHOMOTOPY,LechugaGarvin_2003_HardnessBetti,matousek2013computing}.

We sidestep this topological obstruction by randomizing the approximation problem
\footnote{Randomization has often been a powerful tool to deal with problems that are ``hard to crack'' deterministically. Examples include Kantorovich's randomization of transport maps in optimal transportation \cite{Kantorovitch_1942} as opposed to Monge's deterministic construction, which need not exist, Shannon's communication over noisy channels with an arbitrarily low but non-zero probability of error \cite{Shannon_1948_MathTheoryCommun,CandesRombergTao_Sampling_2006} as opposed to earlier attempts at errorless communication, and randomized algorithms which achieve polynomial time and space complexity as opposed to their computationally hard deterministic counterparts \cite{cheriyan1991randomized}.}. 
Instead of approximating a function $f:\xxx \to \yyy$, we lift the problem to probability measures by approximating the map $\xxx \ni x \mapsto \delta_{f(x)} \in \mathcal{P}_1(\yyy)$ taking values in the 1-Wasserstein space of probability measures on $\yyy$. This construction can be seen as a far-reaching generalization of the standard practice in binary pattern classification where instead of directly approximating a classifier $f : \mathbb{R}^d \to \{0, 1\}$ (with $0$ and $1$ labelling the two ``classes''), one replaces $f$ with a Markov kernel mapping $\mathbb{R}^d$ to the set of probability measures on $\{0, 1\}$\footnote{This should be contrasted with approaches which randomize parameters of approximators \cite{gonon2020approximation}.}.  Since $\mathcal{P}_1(\yyy)$ is contractible, $C(\xxx,\mathcal{P}_1(\yyy))$ has a single homotopy class, and thus all the above topological challenges vanish.

Passing to probability measures allows us to construct \textit{randomized} universal approximators of maps $f:\xxx\rightarrow \yyy$ between \emph{arbitrary Polish metric spaces} $\xxx$ and $\yyy$, thus greatly extending the reach of comparably general theorems from approximation theory \cite{Todd_Strict_StoneWeirestrass_ProcAmerMath1965,ProllaBoshopsGeneralizedStoneWeirestrass,FalindoSanchis2004StoneWeierstrassTheoremsForGroupValued,Timofte_StoneWeirestrass_2005,Petrakis_Constructive_SWT_2016,Timofte2Liaqad2018JMathAnalApplStoneWeierstrassTheorems,kratsios2021_GDL,KovachkiLanthalerMishra_JMLR_2021_FourierNeuralOperatorsUniversal,schmocker2022universal,AcciaioKratsiosPammer2022}. 
When $\xxx$ and $\yyy$ are Polish metric spaces, without additional structure, we prove a very general qualitative guarantee which can be informally summarized as follows.

\begin{thmx}[Theorem~\ref{theorem:Unstructured_Case}, informal statement]%
\label{thm:unstructured-informal}
  Let $(\xxx,d_{\xxx})$ and $(\yyy,d_{\yyy})$ be Polish metric spaces, $\varphi:\xxx\rightarrow F$ a continuous and injective ``feature map'' into a suitable Banach ``feature space'' $F$, and suppose that $\mathcal{F}$ is a family of maps which universally approximates continuous functions between Euclidean spaces, uniformly on compact sets. 
  Then for any continuous map $f:\xxx\rightarrow \yyy$ and $\varepsilon>0$ and any compact $K\subseteq \xxx$ there exists a function $\hat T: \xxx\rightarrow \mathcal{P}_1(\yyy)$ (written down explicitly in~\eqref{eq:compressedfeaturemap_0}-\eqref{eq:theorem_determinsitic_transferprinciple__PAC_Bound___RandomizedVersion_qualitative}) which satisfies
  \[
            \sup_{x\in K}\, \mathcal{W}_1(\hat T(x),\delta_{f(x)}) 
        < 
            \varepsilon
    .
  \]
\end{thmx}

\begin{figure}[ht!]
	\centering
  \includegraphics[width=1\textwidth]{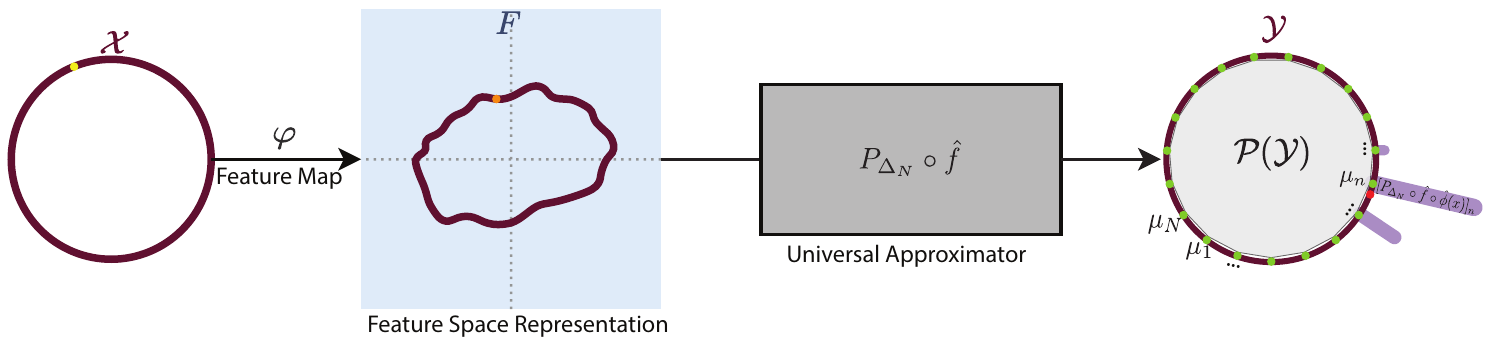}
  	\caption{Illustration of Result~\ref{thm:unstructured-informal}: An input $x$ in $\mathcal{X}$ (yellow dot) is mapped to the output $f(x)$ (red dot) in $\yyy$ by the target function $f$.  A randomized universal approximator $\hat{T}:\xxx\rightarrow \mathcal{P}_1(\yyy)$ transforms $x$ by first linearizing into a Banach feature space $F$ by a feature map $\varphi$ and then projecting the linearized feature representation onto an $n$-dimensional normed subspace identified with $\mathbb{R}^n$, for a suitable $n$. This finite-dimensional representation (orange dot) is then transformed by a universal approximator (black box, $\mathcal{F}$) mapping $\mathbb{R}^n$ to some $\mathbb{R}^N$, e.g., a deep neural network with ReLU activation function, whose output is projected onto the $N$-simplex. We identify the $N$-simplex with the set of quantized probability measures\protect\footnotemark $\mu_1$, $\dots$, $\mu_N$ (light green) in $\mathcal{P}_1(\yyy)$; which we visualize as Dirac masses on different points in $\yyy$.  The thick black box generates the probabilities, indicated as lengths of the purple bars, of sampling any of the light green points in $\yyy$, which then suitably approximate $f(x)$.}
	\label{fig:thm:unstructured-informal}
\end{figure}
\footnotetext{In the notation of Theorem~\ref{theorem:Unstructured_Case}: $\mu_n\eqdef \sum_{q=1}^Q\,[P_{\Delta_Q}(u^n)]_q\,\delta_{y^n_{{}_{\lceil z^n_q\rceil}}}$ for $n=1,\dots,N$.  }

Result~\ref{thm:unstructured-informal} implies\footnote{See Corollary~\ref{cor:uniformW1_small_implies_hpestimates} for a precise statement.} that any for any finite set $\{x_1, \ldots, x_N\}$ of inputs in $ \xxx$, we can generate $\yyy$--valued random variables $Y_1, \ldots, Y_N$ with distributions $\hat{T}(x_1), \ldots, \hat{T}(x_N)$ which probabilistically approximate $f$ in the usual sense; i.e.,
\[
        \mathbb{P}\biggl(
            \max_{n=1,\ldots,N} \,
            d_{\yyy}(Y_n, f(x_n)) \le \varepsilon
        \biggr) 
    \ge
        1-\mathcal{O}(\varepsilon)
    .
\]
Thus, topological obstructions to universal approximation imposed by $\yyy$ can be removed by sampling the measure-valued approximator $\hat T$, and this procedure is universal with arbitrarily high probability.

Beyond these topological aspects, our construction immediately yields results for approximation of operators between Banach spaces, generalizing many of the recent results on learning solution operators in forward and inverse problems for partial differential equations (Section~\ref{s:Applications__ss:InverseProblemsPDE___sss:DirichletToNeumanMap}), with favourable rates when these operators exhibit H\"older-like regularity. More generally, for barycentric $\yyy$ we can ``de-randomize'' the approximators by passing their output through a barycenter map $\beta:\mathcal{P}_1(\yyy)\rightarrow \yyy$, resulting in a map from $\xxx$ to $\yyy$. This is a special case of our main quantitative approximation result:

\begin{thmx}[Theorem~\ref{theorem:determinsitic_transferprinciple}, informal statement]
\label{thm:barycentric-informal}
 Let $\mathcal{X}$, $\mathcal{Y}$, $\phi:\xxx \to F$ and $\mathcal{F}$ be as before, and in addition assume that $\yyy$ is barycentric with a barycenter mapping $\beta: \mathcal{P}_1(\yyy)\rightarrow \yyy$. Let $\hat T$ be the mapping constructed in Result~\ref{thm:unstructured-informal}. Then $\hat t := \beta \circ \hat T$ satisfies 
 \[
            \sup_{x\in K}\, d_{\yyy}(\hat t(x),f(x)) 
        < 
            \varepsilon
    .
  \]
\end{thmx}

\begin{figure}[H]
	\centering
    \includegraphics[width=1\textwidth]{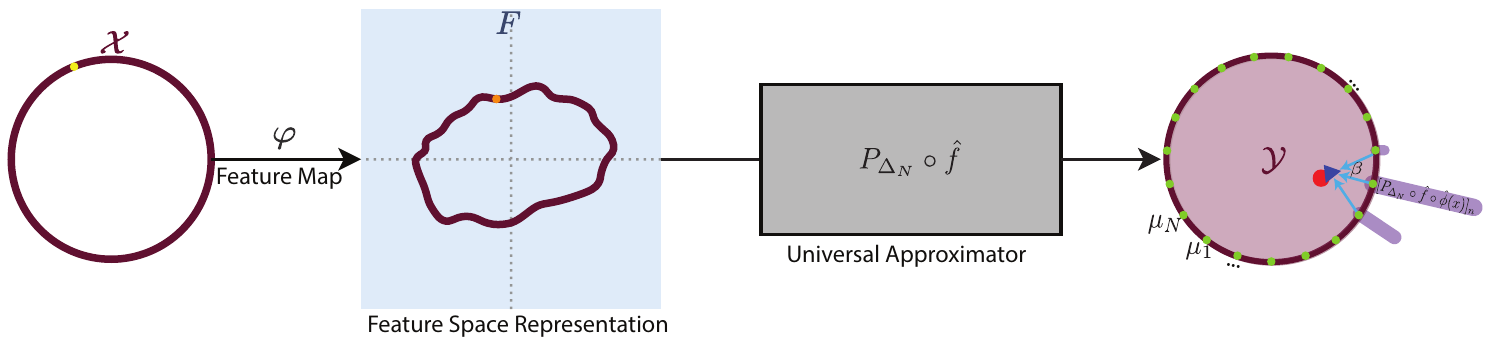}
	\caption{Illustration of Result~\ref{thm:barycentric-informal}: Consider the setting of Figure~\ref{fig:thm:unstructured-informal}, if $\yyy$ is barycentric then (e.g.\ a closed convex subset of a Banach space) then any probability measure generated black box's outputs, has a well-defined center of mass (blue triangle) which is obtained by applying the given barycenter map $\beta$ (light blue arrows) on $\yyy$.  Varying these probability measure's weights moves around the barycenter in $\yyy$ allowing for the approximation of $f(x)$ (red point).
    \hfill\\
    The approximators constructed using Result~\ref{thrm-Full} extend this construction in two ways: they implement several simultaneous local versions of this construction and relax the parameterized barycenter to a parameterized geodesic selection between the green points in $\yyy$.}
	\label{fig:thm:barycentric-informal}
\end{figure}

Finally, when $(\xxx,d_{\xxx},\mu)$ is a metric measure space, $\yyy$ admits combinatorial decompositions into barycentric parts, $f$ is a H\"{o}lder-like\footnote{Functions of H\"{o}lder-like regularity extend the class of H\"{o}lder maps to include uniformly continuous functions of sub-polynomial regularity at small distances.} functions which does not transport too much of $\mu$'s mass to the interphase of each of $\yyy$'s parts, then we derive a quantitative analogue of our Results~\ref{thm:unstructured-informal}.  The constructed approximator encodes both spaces' geometries  and it outputs a finitely supported probability measure whose number of atoms equals to the number of parts in $\yyy$; furthermore, on a set of arbitrarily high probability, the approximator can be derandomized to a $\yyy$-valued function, extending  Result~\ref{thm:barycentric-informal} to general geometries.

\begin{thmx}[Theorem~\ref{theorem:Structured}, informal statement]
\label{thrm-Full}
 Let $(\xxx,d_{\xxx})$, $(\yyy,d_{\yyy})$, and $\mathcal{F}$ be as before, and assume that $\xxx$ and $\yyy$ admit suitable combinatorial decompositions so that $\yyy$ is decomposed into ``barycentric parts''\footnote{See Setting~\ref{setting:theorems_Quantitative} for details.} and $f: \xxx \to \yyy$ is H\"{o}lder and respects that decomposition. 
 Then for any compactly supported Borel probability measure $\mu$ on $\xxx$ and every $\varepsilon>0$, there exist a Borel subset $\xxx_{\varepsilon} \subseteq \xxx$ and 
 a function $\hat{t}:\xxx_{\epsilon}\rightarrow \yyy$ explicitly constructed in~\eqref{eq:representationTransformer} and Table~\ref{tab:modelsummary}, depending only on $\varepsilon$, $f$, $K$, $\mathcal{F}$, and $F$, as quantified in Table~\ref{tab:parameterestimates}, satisfying
 \begin{equation*}
\sup_{x\in \xxx_{\epsilon}} d_{\yyy}(\hat t(x), f(x)) < \epsilon \quad 
   \text{ and } \quad  \mu\Big(
       \mathcal{X}_{\epsilon}
    \Big)
    \ge 1 - \epsilon
        .
\end{equation*}
\end{thmx}  
\paragraph{Applications} We demonstrate the generality of these results by applying them to four approximation problems from various domains for which no similar approximation theory has been available:
\begin{denseitemize}
    \item Maps between finite (discrete) metric spaces often arising in computer science, such as graph colouring problems \cite{Robin_4colourings_2014,3colourings_2016}, link prediction in network analysis \cite{zhang2018link,panneural}, and node classification \cite{monti2017geometric};
    \item Maps between smooth manifolds \cite{FisherLewisEmbleton_1993} arising in shape-space analysis \cite{Bhattacharya2_ShapeSpaceStatsBook} or in statistics when approximating the inverse operator of acting on high-dimensional non-singular covariance matrices \cite{yuan2010high} with affine-invariant metric \cite{pennec2006riemannian};
    \item Solution operators to rough differential equations which are defined on the non-smooth (sub-Riemannian) geometry of a particular Carnot group \cite{Lyons1998,HamblyLyons_RoughPath_2010_UniquenessLift};
    \item Solution operators to inverse problems between infinite-dimensional Banach spaces \cite{RulandSalo_2020_fractionalCalderon_NonLinearAnalysis} and their regularizations when reconstructions lie on finite-dimensional sub-manifolds \cite{alberti2020inverse}.
\end{denseitemize}

Our constructions are modular in the sense that they can be built around any given class of universal function approximators between Euclidean spaces, including  deep neural networks \cite{zhou2020universality,shen2022optimal,ElbachterPerekrestankoGrohsBolckei_2021_IEEETransactionsInfoTheory}, kernel regressors \cite{micchelli2006universal,gonon2020approximation}, regression trees \cite{SharpApproximation_of_RandomForestRegressors_PMLR_2021__Klusowski}, and splines \cite{BookModCOnt}. Thus, in particular, we contribute to the emerging mathematical theory of deep learning \cite{DeepLearningBook,elbrachter2021deep}.

\paragraph{Outline} We introduce the required notation and terminology in Section~\ref{s:Definitions_and_Background}. The main results are presented in Section~\ref{s:Main}. In Section~\ref{s:Applications}, we collect diverse applications of our theoretical results to the approximation of maps arising in rough differential equations, computational geometry, graph theory, and in inverse problems.  Section~\ref{s:Proofs} contains the proofs of Theorem~\ref{theorem:Unstructured_Case}, Theorem~\ref{theorem:determinsitic_transferprinciple} and Theorem~\ref{theorem:Structured} from Section~\ref{s:Main}.

\section{Definitions and Background}
\label{s:Definitions_and_Background}

We begin by reviewing the terminology and analytic tools:
\begin{itemize}
    \item In Section \ref{s:Definitions_and_Background__AnnMet} we recall the definitions of doubling spaces and quasisymmetric maps. We also introduce a class of uniformly continuous functions which mimic the geometric properties of H\"older functions but allow for much lower regularity;
    \item In Section \ref{s:Introduction__ss:QASSpaces} we give the background on quantizable and approximately simplicial spaces \cite{AcciaioKratsiosPammer2022}, a class of spaces which admit sequences of progressively finer local finite-dimensional parameterizations. These spaces are conceptually similar to multi-resolution analysis on $L^2(\mathbb{R}^d)$~\cite{Stephane_Book_Edition3_2009} or quantization of probability measures in $\mathcal{P}_1(\mathbb{R}^d)$ \cite{GrafLuschgy_2000_FoundationsQuantizationofProbMeasure};
    \item In Section \ref{s:Introduction__ss:Approximation} we formalize the definition of classical universal approximators between linear spaces which are a building block of our randomized approximators.
\end{itemize}

\subsection{Analysis on Metric Spaces and Quasi Symmetry}
\label{s:Definitions_and_Background__AnnMet}
A metric space $(\xxx,d_{\xxx})$ is said to be ($d_{\xxx}$-)\textit{doubling}, with doubling constant $C_{\xxx}>0$, if $C_{\xxx}$ is the smallest number such that for every $x\in \xxx$ and every $r>0$, the metric ball $B_{\xxx}(x,2r)\eqdef \{u \in \xxx:\,d_{\xxx}(u,x)<2r\}$ can be covered by $C_{\xxx}$ metric balls of radius $r$ in $(\xxx,d_{\xxx})$, and $C_{\xxx} < \infty$.
This implies that a doubling subset $K\subseteq \xxx$ cannot be too large, in the sense that it can be deformed into a subset of a Euclidean space with only a small perturbation to its geometry \cite[Theorem 12.1]{heinonen2001lectures}. (We note in the passing that \textit{some} perturbation is in generally necessary \cite{CheegerKleiner_2010_DifferentiatingMapsL1andgeomBVmaps,AustinNaorTessera_2013_NonembeddabilityHeisenberg}.)

\textit{Quasisymmetric} maps are a class of maps which preserve the doubling property; they generalize conformal maps from complex analysis \cite{LehtoQuasiConformal_1975}. These maps preserve the geometry of a space by encoding the relative distance in triplets of points\footnote{See Lemma~\ref{lemma_quantiative_doublinginvariance_of_quasisymmetricmaps} for a quantitative estimate of a doubling space's doubling constant under a quasisymmetric map.}.  Let $\rr_+\eqdef [0,\infty)$.  
A topological embedding $\varphi:(\xxx,d_{\xxx})\rightarrow (\yyy,d_{\yyy})$ is said to be quasisymmetric if there is a strictly increasing surjective map $\eta_{\phi}:\rr_+\rightarrow \rr_+$, satisfying
\[
d_{\xxx}(x_1,x_2)\le t d_{\xxx}(x_1,x_3) 
\mbox{ implies that }
d_{\yyy}(\varphi(x_1),\varphi(x_2))\le \eta_{\phi}(t) d_{\yyy}(\varphi(x_1),\varphi(x_3))
\]
for all $x_1,x_2,x_3\in \xxx$ and $t \in \rr_+$. 

In this paper, we study continuous functions $f:\xxx\rightarrow \yyy$ which exhibit a \textit{H\"{o}lder-like} regularity.
\begin{definition}[H\"{o}lder-Like Functions]
A uniformly continuous map $f:(\xxx,d_{\xxx})\rightarrow (\yyy,d_{\yyy})$ which admits a strictly increasing, subadditive and continuous modulus of continuity $\omega$ for which there is an increasing homeomorphism $h_{\omega}:[0,\infty)\rightarrow [0,\infty)$, satisfying
\[
        \omega(st)
    \le 
        h_{\omega}(s)
        \,
        \omega(t)
\]
for every $s,t\ge 0$, is said to be $\omega$-H\"{o}lder-like; $\omega$ is the corresponding H\"{o}lder-like modulus of continuity.
\end{definition}
When clear from the context, we omit the explicit dependence on $\omega$ and say $f$ is H\"{o}lder-like.
\begin{example}[H\"{o}lder Functions are of H\"{o}lder-Like Regularity]
\label{ex_Holderregularity}
Fix $L\ge 0$ and $0< \alpha \le 1,0 \le \beta\le 1-\alpha$.  Then $\omega(t)=L t^{\alpha}(\log(1+t))^{\beta}$ is concave\footnote{Acknowledgement: Andrew Colinet, who also noted the correct constraint on $\beta$ to ensure concavity of this modulus of continuity.}, implying the subadditivity of $\omega(\cdot)$. Thus $\omega(t)$ is a H\"{o}lder-like modulus of continuity with $h_{\omega}(s)=s^{\alpha}\max\{1,s\}^\beta$, since $h_{\omega}(s)=s^\alpha\sup_{u>0} \frac{\log(1+(us))^{\beta}}{\log(1+u)^{\beta}}$ and $\sup_{u>0} \frac{\log(1+(us))^{\beta}}{\log(1+u)^{\beta}} = 1$ when $0\le s\le 1$ and equals $s^\beta$ otherwise.
\end{example}
H\"{o}lder-like maps need not have regularity similar to any H\"{o}lder function near $0$, which is the region impacting quantitative approximation rates. The next class of examples shows that the class of H\"{o}lder-like functions contain maps whose modulus of continuity converges to $0$ slower than any H\"{o}lder function.
\begin{example}[H\"{o}lder-Like Moduli of Sub-H\"{o}lder Regularity\footnote{Acknowledgement: Andrew Colinet.}]
\label{ex:hyperlowregularity_Holderlike}
Fix a parameter $\beta>0$.  An example of a H\"{o}lder-like modulus of continuity $\omega$ with the property that $\lim\limits_{t\rightarrow 0^+}\, t^{\alpha}\omega^{-1}(t)=0$ for every $0<\alpha\le 1$ is
\begin{equation}
\omega(t)
\eqdef 
    \begin{cases}
        0 
            & \mbox{ if } t=0\\
        \frac1{|\log(t)|^{\beta}} 
            & \mbox{ if } 0<t\le \frac1{e^{\beta+1}} \\
            \frac1{(\beta+1)^{\beta}}
        +
            \frac{\beta e^{\beta+1}}{(\beta+1)^{\beta+1}}
            (t-\frac1{e^{\beta+1}})
                & \mbox{ if } \frac1{e^{\beta+1}} < t < \infty
    .
    \end{cases}
\end{equation}
The function $\omega$ is increasing, concave (therefore sub-additive), continuous with homeomorphism $h_{\omega}(s)=\max\{s^{\beta},s\}$ satisfying $\sup_{u>0}\, \frac{\omega(us)}{\omega(u)}\le h_{\omega}(s)\le s^{\beta}$ for $0\le s\le 1$ and $\sup_{u>0}\, \frac{\omega(us)}{\omega(u)}\le h_{\omega}(s)\le s$ otherwise.
\end{example}

H\"{o}lder-like maps become Lipschitz under perturbations of the geometry of the source space $\xxx$ known as the generalized snowflake\footnote{The terminology ``snowflake'' stems from the observation that $(\mathbb{R},|\cdot|^{\alpha})$ with $\alpha=\frac{\log(3)}{\log(4)}$ is isometric to the Koch snowflake with metric inherited from inclusion in the Euclidean plane.} transform \cite{Kalton2004LipHolderApplications_2004}.  
Since this perturbation is a quasi-symmetry, it preserves the doubling property of the source space.
\begin{lemma}[Generalized Snowflakes are Quasisymmetric to their Original Space]
\label{lem_snowflake}
Let $\omega$ be a H\"{o}lder-like modulus and $(\xxx,d_{\xxx})$ a metric space.  Then, $d_{\xxx}^{\omega}\eqdef \omega \circ d_{\xxx}$ defines a metric on $\xxx$ and $(\xxx,d_{\xxx})\ni x \mapsto x\in (\xxx,d_{\xxx}^{\omega})$ is a quasi-symmetry.  
Furthermore, if $K\subseteq \xxx$ is $d_{\xxx}$-doubling then it is also $d_{\xxx}^{\omega}$-doubling.
\end{lemma}
\begin{remark}
    The doubling constant with respect to $d_{\xxx}^\omega$ can be explicitly computed; see Lemma~\ref{lem_snowflake__quantitative}.
\end{remark}

Lemma~\ref{lem_snowflake} implies that approximating H\"{o}lder-like maps is equivalent to approximating Lipschitz maps with a judiciously perturbed metric. Note however that in general one cannot expect to approximate continuous functions by H\"{o}lder functions; for example, no map from $(\mathbb{R},|\cdot|)$ to $(\mathbb{R},\log(1+|\cdot|))$ is H\"{o}lder continuous.  

The generalized snowflake transform perturbs the geometry of $\xxx$ enough to make all H\"{o}lder functions Lipschitz. By further distorting $\xxx$'s metric while preserving its topology, we can approximate all continuous maps from $\xxx$ to $\yyy$ by functions which are Lipschitz for the perturbed metric. As a consequence, quantitative statements about approximation of Lipschitz functions between Polish metric spaces imply qualitative statements about general continuous functions.
\begin{lemma}[{Most Continuous Functions are Lipschitz\footnote{Acknowledgment: Joseph Van Name.}}]
\label{lemma:change_of_metric}
Let $\xxx$ be a compact Polish space, and let $(\yyy,d_{\yyy})$ be a separable metric space. There is a metric $d_{\xxx}$ on $\xxx$ which generates $\xxx$'s topology such that the set of Lipschitz maps from $\xxx$ to $\yyy$ is uniformly dense in the set $C(\xxx,\yyy)$ of continuous functions from $\xxx$ to $\yyy$.  
\end{lemma}

We are interested in uniformly approximating continuous maps on compact subsets of the source metric space. The lemmata above imply that it is sufficient to focus on the approximation theory of Lipschitz maps.  

\subsection{Quantizable and Approximately Simplicial (QAS) Spaces}
\label{s:Introduction__ss:QASSpaces}
Quantizable and approximately simplicial spaces (QAS) spaces are a class of metric spaces $(\yyy,d_{\yyy})$ which capture some of the essential metric properties of Wasserstein spaces, from optimal transport theory, as they pertain to approximation theory \cite{AcciaioKratsiosPammer2022}, which like the Wasserstein spaces over any polish spaces are asymptotically parametrizable by Euclidean vectors.  The construction can also be interpreted as a non-Euclidean analogue of multi-resolution analysis on $L^2(\mathbb{R}^d)$ used in signal processing ~\cite{Stephane_Book_Edition3_2009}, where one progressively refines their description of a set of functions by using more progressively frequencies.
As we will see in our applications section, most reasonable spaces are QAS spaces.

The $N$-simplex, which parameterizes any the space of probability measures on $N$ points is denoted by
$\Delta_N \eqdef \{w\in [0,1]^N:\,\sum_{n=1}^N w_n=1 \}$.  For any given metric space $(\yyy,d_{\yyy})$ define
\[
    \widehat\yyy
\eqdef 
    \bigcup_{N\in\mathbb{N}_+}\left(\Delta_N\times \yyy^N\right)
.
\]
Using $\widehat{\yyy}$, we can approximately parameterize $\yyy$ by inscribing simplices in $\yyy$.
\begin{definition}[Approximately Simplicial]
\label{ass_approximately_simplicial}
A metric space $(\yyy,d_{\yyy})$ is said to be approximately simplicial if there is
a function $\eta:\widehat\yyy\to\yyy$, called a mixing function, and constants $C_\eta\geq 1$ and $p\in\mathbb{N}_+$, such that
for every $N\in \mathbb{N}_+$, $w=(w_1,\dots,w_N)\in\Delta_N$ and $\mathbf{y}=(y_1,\dots,y_N) \in \yyy^N$, and for all $i \in \{1, \ldots, N\}$ one has
\begin{equation}
\label{eq:ass_approximately_simplicial}
    d_\yyy \left( 
    \eta\left( w, \mathbf{y} \right), y_i \right) 
    \le 
C_\eta
\left(
    \sum_{j = 1}^N d_\yyy(y_i,y_j)^p w_j\right)^{1 / p}.
\end{equation}
In particular, the function $\eta$ satisfies $\eta(e_i,\mathbf{y})=y_i, i=1,\dots,N$,  where $\{e_i\}_{i=1}^N$ is the standard basis of $\mathbb{R}^N$.
\end{definition}
Intuitively, the mixing function $\eta$ mixes finite sets of points in $\yyy$ by moving along geodesic segments and this mixing is parameterized by Euclidean simplices.
Quantizability can roughly be understood as a simultaneously quantitative and ``asymptotically parametric'' analogue of the otherwise qualitative separability property of a topological space \cite{AcciaioKratsiosPammer2022}.  
\begin{definition}[Quantization]
\label{defn_Quantization_modulus}
Let $(\yyy,d_{\yyy})$ be a metric space and let $Q_{\cdot}\eqdef(Q_q)_{q\in \nn_+}$ be a family of functions $Q_q:\rr^{D_q}\rightarrow \yyy$ with each $D_q\in \nn_+$ satisfying
\begin{enumerate}
\item for all $q\in \nn_+$ and each $z\in \rr^{D_q}$, there exists some $\tilde{z}\in \rr^{D_{q+1}}$ with 
$
Q_q(z)=Q_{q+1}(\tilde{z});
$
\item for every $y\in \yyy$ and each $\epsilon>0$, there exists some $q\in \nn_+$ and some $z\in \rr^{D_q}$ such that
\[
d_{\yyy}\left(
y
,
Q_q(z)
\right)
<
\epsilon
.
\]
\end{enumerate}
The family $Q$ is a \textit{quantization} of $(\yyy,d_{\yyy})$ if for any compact subset $K$ of $\yyy$ and each $\epsilon>0$ the quantization modulus
\[
\mathscr{Q}_K(\epsilon)
\eqdef
\inf\left\{
D_q:\, (\forall y \in K)\, \exists z\in \rr^{D_q} \mbox{ such that } 
d_{\yyy}(y,Q_q(z))<\epsilon
\right\}
\]
is finite.  The map $\mathscr{Q}_K$ is called the modulus of quantizability of $K$.  
\end{definition}
If $(\yyy,d_{\yyy})$ is quantizable by $Q_{\cdot} = (Q_q)_{q\in \mathbb{N}_+}$ and approximately simplicial with mixing function $\eta$, then we can approximately implement any generalized geodesic simplex in $\yyy$ by combining these two structures via
    \begin{equation}
        \label{eq:quantizedmixingmap}
        \begin{aligned}
        \hat{\eta} : \bigcup_{N,q\in \mathbb{N}_+, N\ge 2} \,
        \Delta_N \times \mathbb{R}^{N\times D_q}
            & \rightarrow \mathcal{Y}\\
    	\hat{\eta}(w , Z) 
    	    & \eqdef 
    	\eta(w,(Q_q(Z_1),\dots,Q_q(Z_N)))
    	,
        \end{aligned}
    \end{equation}
where $w\in \Delta_N$, $Z_1,\dots,Z_N\in \mathbb{R}^{ D_q}$ and $N,D_q\in \mathbb{N}_+$. The map $\hat{\eta}$, which can be interpreted as a quantized version of the mixing function $\eta$, is called the \textit{quantized mixing map}. A triple $(\yyy,d_{\yyy},\hat{\eta})$ is called a \textit{QAS space} if $(\yyy,d_{\yyy})$ is \textit{q}uantizable and \textit{a}pproximately \textit{s}implicial with $\hat{\eta}$ defined by~\eqref{eq:quantizedmixingmap}.  

\paragraph{A Prototypical Example: (First) Wasserstein Space over Separable Metric Spaces}
We fix ideas with a prototypical QAS space, namely, the (first) Wasserstein space $(\mathcal{P}_1(\xxx),W_1)$ above a separable metric space $(\xxx,d_{\xxx})$, a metric space which plays a central role throughout our analysis. The elements of $\mathcal{P}_1(\xxx)$ are probability measures $\mu$ on $\xxx$ with finite first moment, meaning that there is some $x_0 \in \xxx$ for which the integral $
\int \,d_{\xxx}(x,x_0) \mu(dx)
$ is finite. The distance between any two $\mu$ and $\nu$ therein measures the minimal amount of ``work'' required to move mass from $\mu$ and $\nu$,
\begin{equation}
\label{eq:Wassersteindistance}
W_1(\mu,\nu)
    \eqdef
\inf_{\pi}\,\int d(x_1,x_2)\, \pi(d(x_1,x_2)),
\end{equation}
where the minimization is over all Radon probability distributions $\pi$ on $\xxx\times \xxx$ whose push-forwards by the canonical projections are $\mu$ and $\nu$. We will repeatedly use the fact that $\xxx$ isometrically embeds into $\mathcal{P}_1(\xxx)$ via the map $x\mapsto \delta_x$, where $\delta_x$ is the point mass on $x$ (see \cite{Benoit_GeomWassersteinSeries_EuclideanSpaces_2010,Benoit_GeomWassersteinSeries_Hadamard_2012,Benoit_GeomWassersteinSeries_UltrametricandCompactMetricSpaces_2015} for details).  

An example of a mixing function $\eta$ on $(\mathcal{P}_1(\xxx),W_1)$ sends any weight $w$ in an $N$-simplex $\Delta_N$ and any set of $N$ probability measures $\mu_1\dots,\mu_N$ in $\mathcal{P}_1(\xxx)$ to their convex combination,
\begin{equation}
\label{eq:WassersteinMixingFunctionPrototype}
        \eta(w,(\mu_n)_{n=1}^N)
     = 
        \sum_{n=1}^N\,
            w_n\,
            \mu_n
    .
\end{equation}
It is easy to show that such $\eta$ satisfies the inequality \eqref{eq:ass_approximately_simplicial} with $C_{\eta} = 1$ and $p = 1$. 
Quantizability comes into play as probability measures in $\mathcal{P}_1(\xxx)$ need not be exactly describable as mixtures, in the sense of $\eta$, of finitely many ``elementary probability measures''.  Since $\xxx$ is separable, there is a countable dense subset $\{y_q\}_{q \in \mathbb{N}}$ of $\xxx$, and by the proof of \citep[Theorem 6.18]{VillaniOptTrans}, the set of probability measures on $\xxx$ supported on finite subsets of $\{y_q\}_{q\in \mathbb{N}}$ is dense in $\mathcal{P}_1(\xxx)$.  We can therefore define the quantization maps $\mathcal{Q}_q:\mathbb{R}^{2 \times Q}\rightarrow \mathcal{P}_1(\xxx)$ by
\[
    \mathcal{Q}_q(u,z)
        \eqdef 
    \sum_{i=1}^q\,
        [P_{\Delta_q}(u)]_i\,
        \,
        \delta_{y_{{}_{\lceil z_i \rceil}}}
    ,
\]
where $v_i$ denotes the $i^{th}$ component of a vector $v\in \mathbb{R}^q$, $P_{\Delta_q}:\mathbb{R}^q\rightarrow \Delta_q$ is the Euclidean orthogonal projection onto the $q$-simplex $\Delta_q$, and
 $y_{{}_{\lceil 
                                    z_{i}
                              \rceil}}=y_n,$ where $n=
                              \lceil 
                                    z_{i}
                              \rceil\in \mathbb Z
                              $ smallest integer satisfying $n\ge z_{i}$.
Combining $\eta$ and $\mathcal{Q}_{\cdot} \eqdef (\mathcal{Q}_q)_{q\in \mathbb{N}_+}$, we obtain our quantized mixing function
\begin{equation}
\label{eq:quantizedmixing_Wasserstein}
        \hat{\eta}\big(w,(u_{i,j},z_{i,j})_{i,j=1}^{I,q}\big)
    \eqdef
            \sum_{i=1}^{I}\,
                w_i\,
    \cdot
    \,
            \biggl(
                \sum_{j=1}^{q}\,
                    [P_{\Delta_q}(u_{i,j})]_j\,
                    \delta_{y_{{}_{\lceil 
                                    z_{i,j}
                              \rceil}}}
            \biggr)
        \,
        ,
\end{equation}
where $Z \eqdef (u_{i,j},z_{i,j})_{i,j=1}^{I,q} \in \mathbb{R}^{2\times I \times q}$ and where $\mathbb{R}^{2\times I \times q}$ is identified with $\mathbb{R}^{2 I q}$.  
The global geometry of $\xxx$ is encoded in several ways in the Wasserstein space $(\mathcal{P}_1(\xxx),W_1)$. A useful interpretation of QAS spaces arises from considering the metric analogue of the geometric realization of $\xxx$'s Vietoris--Rips complex.  For a radius parameter $r>0$, the $r$-thick metric Vietoris--Rips complex \cite{Latscev_RipsRiemannian_2001}, denoted by $\operatorname{VR}_r^{\operatorname{m}}(\xxx)$, is a metric space which is often homeomorphic to $\xxx$ but can be easier to work with from the computational topology perspective \citep[Theorem 4.6]{AdamszekAdamsFrick_2018_MetricReconstructionviaOT}. The connection with QAS spaces is from its realization of a metric subspace of $(\mathcal{P}_1(\xxx),W_1)$ defined as
\begin{equation}
\label{eq:VR_Complex}
    \operatorname{VR}_r^{\operatorname{m}}(\xxx) \eqdef 
    \left\{
        \sum_{i=0}^k\, w_i\delta_{x_i}:\,
        k\in \mathbb{N}_+,\, w\in \Delta_k,\, \max_{i,j\le k}\, d_{\xxx}(x_i,x_j)\le r
    \right\}
.
\end{equation}
Comparing~\eqref{eq:quantizedmixing_Wasserstein} and~\eqref{eq:VR_Complex}, we notice that the image of the quantized mixing function subsumes $\operatorname{VR}_r^{\operatorname{m}}(\xxx)$.   $\operatorname{VR}_r^{\operatorname{m}}(\xxx)$ are formed by convex combinations of point masses in $\xxx$ which are close enough (at a maximum distance of $r$). By comparsion, in $\hat{\eta}$'s image, we can take convex combinations of point masses\footnote{More precisely, these are instead supported on the dense subset $\{x_q\}_{q\in \mathbb{N}}$ not on any of the possibly uncountable points in $\xxx$.} with the main difference being that $\eta$ automatically adapts to the closeness of the points in $\xxx$ as it satisfies the condition~\footnote{See \citep[Example 8]{AcciaioKratsiosPammer2022}.}~\eqref{eq:ass_approximately_simplicial}.

Another way in which $\xxx$'s global geometry is reflected in $\mathcal{P}_1(\xxx)$ is through the \textit{barycentricity} property, which can be understood as a far-reaching abstraction of the geometry of the $1$-Wasserstein space over a Banach space.  
A metric space is barycentric if the isometric embedding $x \mapsto \delta_x$ of $\xxx$ in $\mathcal{P}_1(\xxx)$ has a uniformly continuous right inverse \cite{ohta2009extending,MendelAssafSectralCalcLipchitzExtensionBarycentric2013}.  This is because, as shown in \cite{BruHeinicheLootgieter1993}, any Banach space $\xxx$ admits a unique Lipschitz right-inverse $\beta$ to the map $x\mapsto \delta_x$ exists and it is simply given by the Bochner integration.  Furthermore, every barycentric metric space $\xxx$ is approximately simplicial with mixing function 
\[
        \eta(w, (x_1, \ldots,x_N)) 
    = 
        \beta \circ \biggl(
            \sum_{i=1}^N w_i \delta_{x_i}
        \biggl)
    ,
\]
consequentially, every Polish barycentric metric spaces is a QAS space.
In this case one can choose $C_{\eta} = 1$ and $p = 1$ in the inequality \eqref{eq:ass_approximately_simplicial}, details are developed in Section \ref{s:Applications__ss:NonSmooth}.  

Barycentric metric spaces are precisely those that admit conical geodesic bicombings \cite{basso_2019_ETHThesis_fixed}; we explore bicombings further in applications of our approximation theory to rough differential equations in Section \ref{s:Applications__ss:NonSmooth}. Barycentricity is a transport-theoretic analogue of a non-expansive barycenter map in non-linear Banach space theory \cite{godefroy2003lipschitz}, which is defined similarly on a certain Banach space containing a copy\footnote{This Banach space contains an isometric image of $\mathcal{P}_1(\xxx)$; see Lemma~\ref{lemma_closedconvex_embedding_Wasserstein} for details.} of the $\mathcal{P}_1(\xxx)$.  
We note that for any compact set of probability measures $K$ in $\mathcal{P}_1(\xxx)$ the modulus of quantizability $\mathscr{Q}_K(\epsilon)$ is precisely the uniform rate of quantization; this is known in the case where $\xxx$ is doubling and the measures in $K$ satisfy certain moment conditions \cite{AhidarLeGouicParis_RateConvergenceEmpiricalBarycenters_General}.  Optimal constants are known when the measures in $K$ are compactly supported and have the same (finite) Assouad dimension \cite{Kloeckner_2012_QuantizationAlhforsRegularity,GrafLuschgy_2000_FoundationsQuantizationofProbMeasure}.  

\begin{figure}[ht!]
     \centering
     \begin{subfigure}[b]{0.3\textwidth}
         \centering
         \includegraphics[width=.95\textwidth,clip]{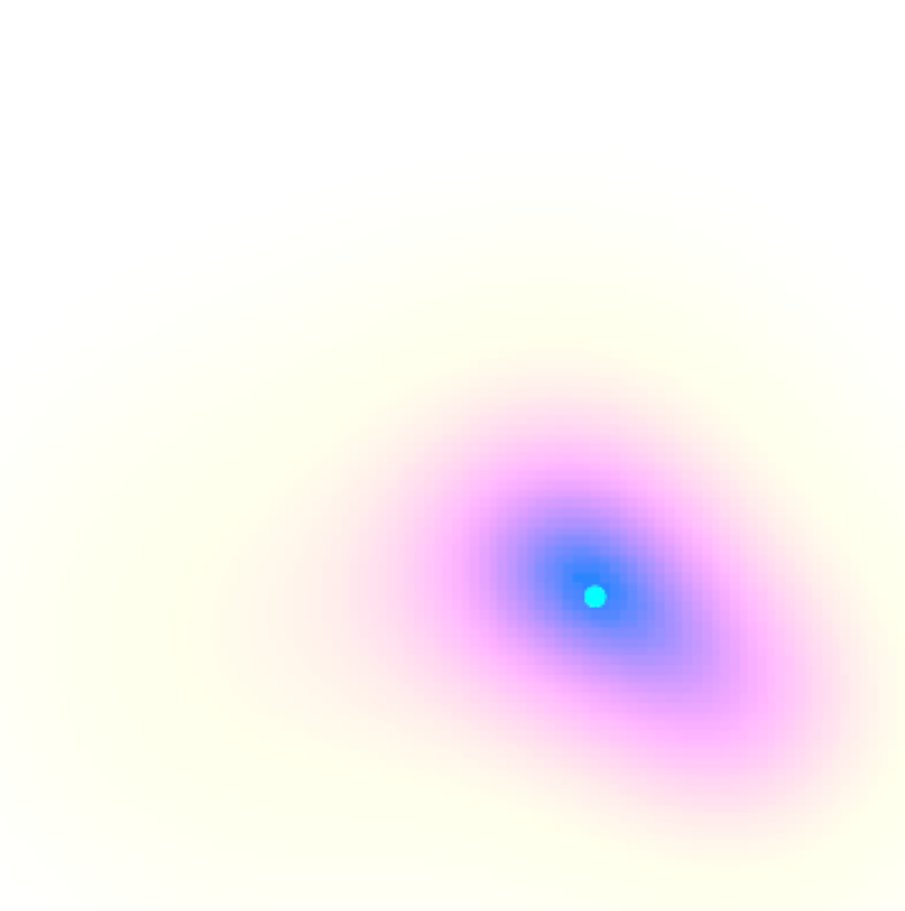}
         \caption{Approximation by probability measure with mass spread over $\yyy$.}
     \end{subfigure}
     \hfill
     \begin{subfigure}[b]{0.3\textwidth}
         \centering
         \includegraphics[width=.95\textwidth,clip]{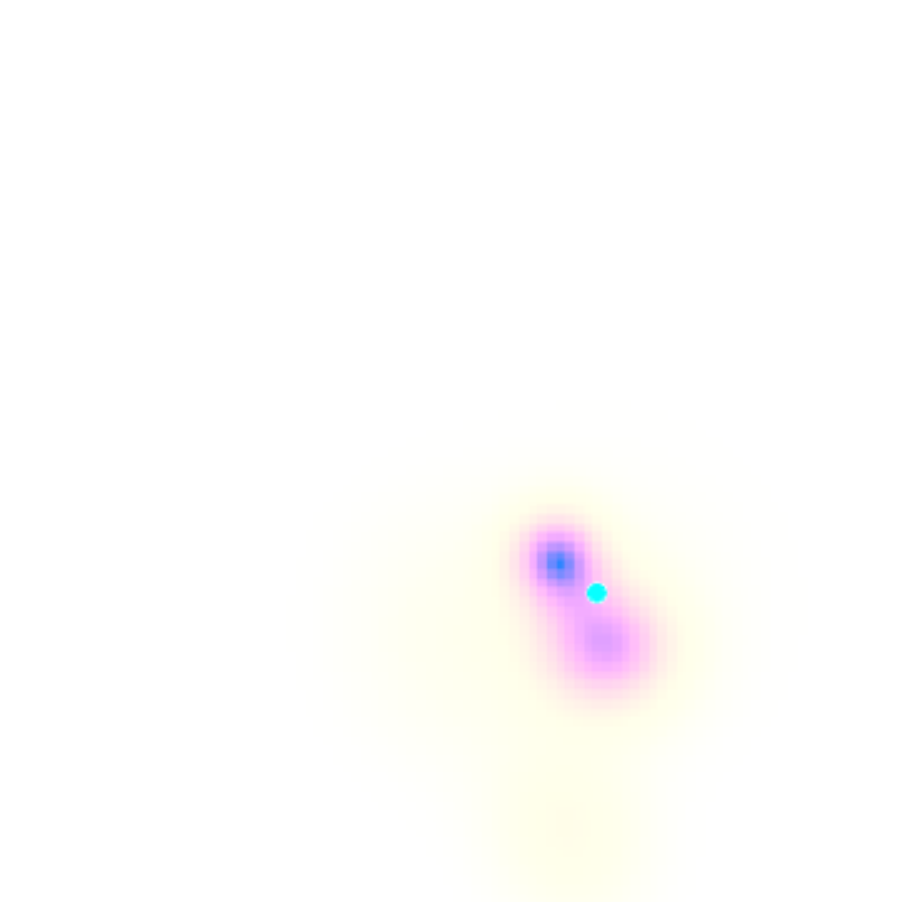}
         \caption{Approximation by concentrated multi-modal probability measure.}
     \end{subfigure}
     \hfill
     \begin{subfigure}[b]{0.3\textwidth}
         \centering
         \includegraphics[width=.95\textwidth,clip]{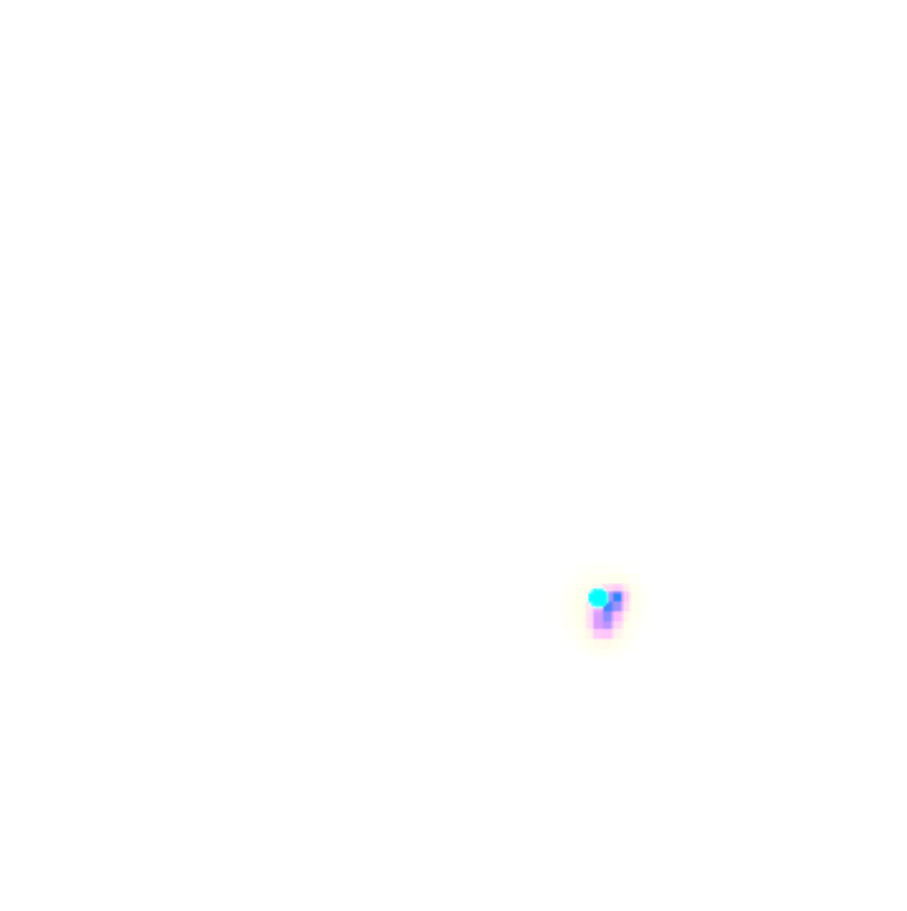}
         \caption{Approximation by probability measure with concentrated mass.}
     \end{subfigure}
        \caption{Approximation of a point (light-blue) by probability measures with progressively higher concentration. The regions of high concentration of these measures are illustrated in dark blue and the regions of low-concentration are illustrated in white, regions of intermediate concentration are depicted in purple.}
        \label{fig:ApproxPointByProbMeasure}
\end{figure}

\paragraph{Randomized Approximation of Points in Metric Spaces}
Just as the notion of closeness between two points in a metric space $(\yyy,d_{\yyy})$ may be extended to the distance between a non-empty compact subset of $\yyy$ and a point, we may give meaning to the distance between a probability measure $\mathbb{P}\in \mathcal{P}_1(\yyy)$ and a point $y$ in $\yyy$.  Figure~\ref{fig:ApproxPointByProbMeasure} illustrates this idea by showing a sequence of probability measures which intuitively approach a point. Using the isometric embedding of $(\yyy,d_{\yyy})$ into $(\mathcal{P}_1(\yyy),W_1)$ sending any point $y\in \yyy$ to the point-mass $\delta_y$, we define the distance between $\mathbb{P}$ and $y$ as
\[
        d_{\yyy}(\mathbb{P},y)
    \eqdef 
        W_1(\mathbb{P},\delta_y)
    ,
\]
noting that if $\mathbb{P}=\delta_{\tilde{y}}$ for some point $\tilde{y}\in \yyy$ then $d_{\yyy}(\delta_{\tilde{y}},y)$ is precisely $d_{\yyy}(\tilde{y},y)$.  In this way, $d_{\yyy}$ is a map from $\mathcal{P}_1(\yyy)\times \yyy\rightarrow [0,\infty)$ extending the metric $d_{\yyy}$ upon isometrically identifying $(\yyy,d_{\yyy})$ with $(\{\delta_y\}_{y\in \yyy},W_1)$.  

The $1$-Wasserstein distance to a point mass admits a simple expression reflecting Figure~\ref{fig:ApproxPointByProbMeasure}.\footnote{We note that this is not the case for general empirical distributions: even when $\yyy$ is a Euclidean space of dimension $2$ or more, fast algorithms to compute the Wasserstein-$1$ distance between empirical distributions such as the accelerated primal-dual randomized coordinate descent  \cite{MichaelJordanHoGuo_FastOT_AISTATS2020} have super-quadratic complexity.} As a result we obtain a simple expression for the distance between $\yyy$-valued and $\mathcal{P}_1(\yyy)$-valued functions.

\begin{proposition}[Closed-Form Expression of Distance Between $\mathcal{P}_1(\yyy)$-Valued and $\yyy$-Valued Functions]
\label{prop:simple_Interpretation}
Let $\xxx$ and $\yyy$ be metric spaces, $\hat{T}:\xxx\rightarrow \mathcal{P}_1(\yyy)$, and $f:\xxx\rightarrow \yyy$. Then for any $x \in \xxx$ we have
\[
            \mathcal{W}_1\big(
                \hat{T}(x)
            ,
                \delta_{f(x)}
            \big)
    = 
            \mathbb{E}_{Y\sim \hat{T}(x)}\big[
                d_{\yyy}(Y,f(x))
            \big]
,
\]
and, in particular, the suprema over $\xxx$ of both sides coincide.
\end{proposition}
Proposition~\ref{prop:simple_Interpretation} implies that if $f$ is approximated by a $\mathcal{P}_1(\yyy)$-valued function $\hat{T}$ on finite subsets of the source space $\xxx$ to $\varepsilon>0$ precision, then with high probability, it can be uniformly approximated on that finite set by independent random variables whose laws are dictated by the approximator $\hat{T}$.  
\begin{corollary}[Approximation by $\mathcal{P}_1(\yyy)$-Valued Maps Imply High-Probability Estimates on Finite Sets]
\label{cor:uniformW1_small_implies_hpestimates}
Assume the setting of Proposition~\ref{prop:simple_Interpretation} and let $\varepsilon>0$.
If $\sup_{x\in \xxx}\,\mathcal{W}_1\big(\hat{T}(x),\delta_{f(x)}\big) <\varepsilon$, $\xxx=\{x_n\}_{n=1}^N$ for a positive integer $N$, and $Y_1,\dots,Y_N$ are independent random variables on a probability space $(\Omega,\mathcal{F},\mathbb{P})$ with $Y_n$ distributed according to $\hat{T}(x_n)$, for $n=1,\dots,N$, then
\[
        \mathbb{P}\Biggl(
                    \max_{n=1,\dots,N}\,
                        d_{\yyy}\big(
                            Y_n
                        ,
                            f(x_n)
                        \big)
                \le 
                    N\sqrt{\varepsilon}
        \Biggr)
    \ge 
        \biggl(
            1 - \frac{\varepsilon}{N}
        \biggr)^N
    .
\]
\end{corollary}

\subsection{Approximation in Linear Spaces}
\label{s:Introduction__ss:Approximation}
In the following, we let $\operatorname{Comp}(\xxx)$ denote the collection of compact subsets of a given metric space $\xxx$.  

\begin{definition}[Universal Approximator]
\label{definition:UniversalApproximators}
Consider a family $\fff_{\cdot}\eqdef \{\fff_{n,m,c}\}_{n,m,c=1}^{\infty}$, where each $\{\fff_{n,m,c}\}_{c=1}^{\infty}$ is a nested family of functions mapping $\mathbb{R}^n$ to $\mathbb{R}^m$ together with a map $r: [
C(\rr_+,\rr_+)
\times 
\operatorname{Comp}(\rr^n)
\times
\nn_+^2
]
\times \nn_+
\rightarrow \rr_+$. We call $\fff_{\cdot}$ a \textit{universal approximator} with \textit{rate function} $r$ if
for any pair of positive integers $n,m$, any uniformly continuous $f:{\mathbb{R}^n}\rightarrow {\mathbb{R}^m}$ with continuous modulus of continuity $\omega$, and any non-empty compact $K\subset \rr^n$ it holds that
\begin{enumerate}
    \item for every $c>0$ there is some $\hat{f}\in \fff_{n,m,c}$ satisfying the uniform estimate
    \[
        \underset{x\in K}{\sup}\,
                \left\|
                    f(x)
                    -
                    \hat{f}(x)
                \right\|
            \le 
                r(\omega,K,n,m,c)
        ;
    \]
    \item the function $c\mapsto r(\omega,K,m,c)$ is decreasing with $\lim\limits_{c\to \infty}\, r(\omega,K,m,c)=0$.
\end{enumerate}
\end{definition}
Classical examples of universal approximators are Bernstein polynomials \cite{WoodBerstein_1984_Classical,DellaMastroianniSzabados_2004_WeightedBernstein,GalTrifa_BernseteinLp,Draganov_2019_SimultaneousapproximationBernseteinpolynomailsintegercoefficients,Dragonov_SimultaneousBersnstein_2020,AdellCardenasMorael_2022_NonAsymtpoticBersneing}, piecewise constant functions (for example trees) \cite{KonnoKuno_1988_BestPiecewiseConstantApproximation,Avakyan_BestApproxByPiecewiseConstant}, splines \cite{Dahmen_1982_AdaptiveApproxSplines,Pence_1987_BestApproxSplines,KopotunLeviatanPrymak_2009_NEarlyMonotoneandNearlyConvexApproxSplines,AndersonBabenkoLeskevysch_2014_ApproxiSplinesSimultaneous,ASplineTheoryforDeepLearning_ICML2018}, wavelets \cite{JawerthMilman_1992WaveletApprox,Bazarkhanov_2004_BesTNtermapproxWaveletsMixedNorm,Nitsce2006_BestWaveletNtermApprox,BookModCOnt}. Examples of universal approximators central to contemporary approximation theory, computer science, and machine learning are feedforward \cite{yarotsky2019phase,kratsios2022do,shen2022optimal,puthawala2022globally} and convolutional neural networks \cite{zhou2020universality,ZHOU_ATheoryOfConvNetswDownsampling__NeuralNetworks_2020}. A typical example is the class of piecewise linear functions known as ReLU networks (i.e. neural networks which activation functions are Rectified Linear Units).
\begin{example}[ReLU Networks]
\label{ex:deepfeedfowardnetworks}
The set $\mathcal{F}_{n,m,c}$ of piecewise linear maps $\hat{f}:\mathbb{R}^n\rightarrow \mathbb{R}^m$ with representation
\[
\begin{aligned}
        \hat{f}(x) 
            \eqdef 
        A^{(c+1)} x^{(c)} + b^{(c)} \,;
    \qquad 
        x^{(t+1)}
        \eqdef 
        \operatorname{ReLU}\big(A^{(t)}x^{(t)}+ b^{(t)}\big) \,\,(t=0,\dots,c);
    \qquad
        x^{(0)} \eqdef x.
\end{aligned}
\]
where each $A^{(t)}$ is a $d_{t+1}\times d_{t}$ matrix, $b^{(t)}\in \mathbb{R}^{d_{t+1}}$, the $\operatorname{ReLU}$ function acts on vectors $x\in \cup_{d\in \mathbb{N}_+}\,\mathbb{R}^{d}$ by $\operatorname{ReLU}((x_i)_{i=1}^d)\eqdef (\max\{x_i,0\})_{i=1}^d$, $d_0=n$, $d_{c+1}=m$, and $d_t\le c$ for every $t=1,\dots,c$, is such that $\mathcal{F}_{\cdot}\eqdef \{\mathcal{F}_{n,m,c}\}_{n,m,c\in \mathbb{N}_+}$ is a universal approximator.  
Its rate function $r$ is given in \cite[Theorem 1]{ALG_UniversalApproximationOperators_2022}.  
\end{example}

Our next tool does not concern the approximation of functions but of linear spaces themselves.  
We recall the definition of the \textit{bounded approximation property} (BAP) introduced in \cite{GorthendieckNuclearThesisQ}. Given a constant $C>0$, a Banach space $F$ has the $C$-\textit{Bounded Approximation Property} ($C$-BAP) if\footnote{An alternative characterization of the $C$-BAP formulation in terms of stable $\sqrt{C}$-manifold widths, as defined on \citep[page 612]{CohenDeVorePetrovaWojtaszcyk_2022_OptimalStable_FOCM}, can be found in \citep[Theorem 2.4]{CohenDeVorePetrovaWojtaszcyk_2022_OptimalStable_FOCM}.} for every non-empty compact subset $K\subseteq F$ there are finite-rank operators $\{T_n\}_{n \in \nn}$ on $F$ satisfying
\begin{equation}
	\label{eq_BAP}
	\lim\limits_{n\to \infty}\,
	\max_{x\in K}\,\left\|T_{n}(x)-x\right\|=0
	\,
	\mbox{and}
	\,
	\|T_{n}\|_{op}\le C ~ \text{for all} ~ n \in \nn
	;
\end{equation}
where $\|\cdot\|_{op}$ is the operator norm of $T$.  
We say that $T_{\cdot}\eqdef (T_n)_{n=1}^{\infty}$ realize the $C$-BAP on $K$ if~\eqref{eq_BAP} holds.  We also say that $F$ has the BAP if it has the $C$-BAP for some $C>0$.  
We quantify the rate at which $(T_n)_{n=1}^{\infty}$ approximates the identity using the map\footnote{The map $R^{T_{\cdot}:K}$ is, by definition, lower-bounded by the linear width of the compact set $K$ (see \cite{Pinkus_LinearWidths_1985} for details) and, a fortiori, by various non-linear widths of $K$ such as its Kolmogorov width \cite{MR1230257} or its Lipschitz width \cite{petrova2022lipschitz}.} $R^{T_{\cdot}:K}:(0,\infty)\rightarrow \nn_+$ defined for every $\epsilon>0$ by
\begin{equation}
	\label{eq_definition_rate_of_identityapproximation}
	R^{T_{\cdot}:K}(\epsilon)
	\eqdef 
	\min\big\{
	n\in \nn_+:\, \max_{x\in K}\,\left\|T_{n}(x)-x\right\|\le \epsilon
	\big\}
	.
\end{equation}

\subsection*{Notation and Terminology}

Table~\ref{tab:notation} aggregates the notation used throughout the paper, aside from each self-contained application.
    \begin{table}[t!]
    \centering
	\ra{1.3}
    \caption{Notation alphabetical order.  }
    \begin{adjustbox}{width=\columnwidth,center}
	\begin{tabular}{@{}lllll@{}}
		\cmidrule[0.3ex](){1-3}
		\textbf{Symbol} & \textbf{Description} & \textbf{Reference}\\
		\midrule
		$A^{\delta}$ & Contraction of a pointed set $(A\bar{y})$ towards $\bar{y}$ in a QAS Space $(\yyy,d_{\yyy},\hat{\eta})$ & Top of Section~\ref{s:Main__ss:Combinatorial_XandY} \\
		$\beta$ & Barycenter Map on a Barycentric Metric Space $(\yyy,d_{\yyy})$ & Page 6\\
		$d_{\mathbb{H}(\xxx,d_{\xxx})}$ & Hausdorff pseudo-metric on closed subsets of $(\xxx,d_{\xxx})$ & Page 9 \\
		$F$ & Feature space - A Separable Banach space with the BAP & Equation~\eqref{eq_BAP} \\ 
		$\mathcal{F}_{\cdot}$ & Universal Approximator - Dense subsets of $C(\mathbb{R}^n,\mathbb{R}^m)$ for each $n,m\in \mathbb{N}_+$ & Definition~\ref{definition:UniversalApproximators}\\
		$\{(\phi_n,\xxx_n)\}_{n=1}^N$ & Feature Decomposition on $(\xxx,d_{\xxx},\mu)$ & Definition~\ref{defn:CombinatorialStructure_X}\\
		$\varphi$ & Continuous injective \textit{feature map} into a Banach space $F$ with BAP & Setting~\ref{setting:theorems_Qualitative}\\
		$\iota_T$ & Isometry Between Image of Finite-Rank Operator and finite-dimensional Normed Space & Equation~\eqref{eq:finitedimensionalcopyBanach}\\ 
		$(\mathcal{P}_1(\xxx),W_1)$ & $1$-Wasserstein Space over a Metric Space $(\xxx,d_{\xxx})$ & Equation~\eqref{eq:Wassersteindistance}\\
		$P_{\Delta_N}$ & Orthogonal Projection of $\mathbb{R}^N$ onto the Euclidean $N$-simplex & Equation \eqref{eq:quantizedmixing_Wasserstein} \\
		$\mathcal{Q}_{\cdot}$ & Quantization & Definition~\ref{defn_Quantization_modulus}\\
		$\{T^{(n)}\}_{n=1}^{\infty}$ & Finite-Rank Operators Implementing the BAP of the Feature Space $F_n$ & Theorem~\ref{theorem:determinsitic_transferprinciple}\\
		$(\xxx,d_{\xxx},\mu)$ & Source Polish Metric Measure Space with Borel Probability Measure $\mu$ & Setting~\ref{setting:theorems_Qualitative} \\
		$(\yyy,d_{\yyy})$ & Target Polish Metric Space & Setting~\ref{setting:theorems_Qualitative} \\
		$(\yyy,d_{\yyy},\hat{\eta})$ & (QAS) Quantizable and Approximately Simplicial Space & Circa Equation~\eqref{eq:quantizedmixingmap}\\
		$(\yyy,(\mathcal{Q}_q)_{q\in \mathbb{N}},\eta)$ & Quantized Geodesic Partition of the QAS space $(\yyy,d_{\yyy},\hat{\eta})$ & Definition~\ref{defn:CombinatorialStructure_Y}\\
		$\eta$ & Mixing Function on an Approximately Simplicial Metric Space & Definition~\ref{ass_approximately_simplicial}\\
		$\hat{\eta}$ & Quantized Mixing Function on an Approximately Simplicial Metric Space & Equation~\eqref{eq:quantizedmixingmap}\\
		\bottomrule
	\end{tabular}
	\end{adjustbox}
    \label{tab:notation}
\end{table}
In addition to the notation in Table~\ref{tab:notation}, we use the following standard terminology.  
\paragraph{Generalized Inverses}
Given a monotone increasing function $f:\rr\rightarrow \rr$, we define its generalized inverse by \cite{EmbrechtsHofert}
\[
  f^{\dagger}(t)
    \eqdef
  \inf\{s\in \mathbb{R}:\, f(s)\ge t\},
\]
with the convention that the infimum of $\emptyset$ is $\infty$.  
If $f$ is strictly increasing and surjective then $f^{\dagger}=f^{-1}$. 

\paragraph{Norms Induced by Finite-Rank Operators}
\label{paragraph:identification_finite_rank_operators}
Let $T$ be a finite-rank operator on $F$ and let $\{e_n\}_{n=1}^N$ be a basis for $T(F)$.  Let $\iota_T:\rr^n\ni x \mapsto \sum_{i=1}^n\, x_i\cdot e_i \in F$. We define the norm $\|\cdot \|_{F:n}$ on $\rr^n$ as
\begin{equation}
\label{eq:finitedimensionalcopyBanach}
    \|x\|_{F:n}
    \eqdef 
\|\iota_T(x)\|_{F}.
\end{equation}
In words, $\|\cdot \|_{F:n}$ is the pullback of the restriction $\|\cdot\|_{F}$ to the finite-dimensional subspace spanned by the vectors in $T$'s image.  

\paragraph{Typical Compact Sets}
Given a metric space $(\xxx,d_{\xxx})$, one may measure the ``distance'' between non-empty closed subsets $K_1$ and $K_2$ of $\xxx$ is using the \textit{Hausdorff pseudo-metric} $d_{\mathbb{H}(\xxx,d_{\xxx})}$, defined by
\[
        d_{\mathbb{H}(\xxx,d_{\xxx})}(K_1,K_2)
    \eqdef
        \max\biggl\{
                    \sup_{x \in K_2} d(x,K_1)
                ,\, 
                    \sup_{\tilde{x} \in K_1} d(K_2,\tilde{x})
            \biggr\}
.
\]
The Hausdorff pseudo-metric defines a metric when restricted to the class of non-empty compact subsets of $(\xxx,d_{\xxx})$.  We will call a family of (non-empty) compact subsets $\mathcal{K}$ of a metric space $(\xxx,d_{\xxx})$ \textit{typical}, if for every $\epsilon>0$ and every non-empty compact subset $K\subseteq \xxx$ there is some $K_{\epsilon}\in \mathcal{K}$ satisfying the estimate:
\[
    d_{\mathbb{H}(\xxx,d_{\xxx})}(K,K_{\epsilon})
    <
        \epsilon
.
\]
Equivalently, $\mathcal{K}$ is typical precisely if and only if it is dense in the space of non-empty compact subsets of $(\xxx,d_{\xxx})$ metrized by the Hausdorff metric.

\section{Main Results}
\label{s:Main}

We begin by presenting our main qualitative result which constructs a class of randomized maps that universally approximate arbitrarily complex continuous functions between most metric spaces. In practice we expect that the parameter complexity of the approximant correlates with the complexity of the function being approximated.

We then complement this very general result by studying the case where the source and target space possess additional combinatorial structure.  This allows us to refine the analysis by building the metric geometry of source and target spaces into the approximator. We obtain efficient quantitative approximation rates for $\omega$-H\"{o}lder-like functions which respect the said combinatorial structure.  This result covers most spaces relevant to deep learning, operator learning, and learning on graphs.  

\subsection{General Case: Randomized Approximation}
\label{s:MainResults__ss:Qualitative}
We summarize the structural assumptions about $\xxx$, $\yyy$, and the \textit{feature map} $\varphi$, which we assume exists, mapping $\xxx$ into a suitable Banach space $F$, called a \textit{feature space}.
\begin{setting}[Qualitative Setting]
\label{setting:theorems_Qualitative}
\phantom{.}
\begin{enumerate}[label=(\roman*)]
    \item $(\xxx,d_{\xxx})$ and $(\yyy,d_{\yyy})$ are Polish metric spaces;
    \item $\phi:\xxx\rightarrow F$ is a continuous injective map into a Banach space $F$ with the BAP%
        \footnote{One could weaken the sequence of finite-rank \textit{linear} operators $(T^{(n)})_{n\in \mathbb{N}_+}$ associated to any compact subset $K$ of $F$, given by the BAP, to a sequence of \textit{non-linear} Lipschitz maps approximating the identity on $K$; e.g.\ studied in \cite{cohen2022optimal,petrova2022lipschitz}, without much change to the theory or proofs.}%
    ;
    \item $\mathcal{F}_{\cdot}$ is a universal approximator.
\end{enumerate}
\end{setting}

Setting~\ref{setting:theorems_Qualitative} (ii) concerns the existence of a feature map.  Feature maps are common in machine learning with examples including signature-based feature maps for irregularly sampled or continuous time-series data \cite{chevyrev2022signature,cuchiero2022signature,ExpectedSig_JMLR_2022}, the feature map associated to any kernel regressor \cite{micchelli2006universal}, Taken's delay map \cite{GrigoryevaHartOrtega_Takens}, randomly generated feature maps underpinning reservoir computing \cite{lukovsevivcius2009reservoir,GononGrigoryevaOrtega_RiskBoundsReservoir,gonon2023approximation} or in ELMs \cite{rahimi2007random}, and Riemannian logarithms used in deep learning on small compact subsets of complete Riemannian manifolds \cite{kratsios2021_GDL,fletcher2003statistics,fletcher2004principal}.

Feature maps are typically constructed on a case-by-case basis.  The next proposition establishes that a feature map satisfying Setting~\ref{setting:theorems_Qualitative} (ii), associated to the feature space $(\ell^2,\|\cdot\|_2)$, must exist on a Polish space.
\begin{proposition}[Existence of Feature Maps into A Separable Hilbert Space]
\label{prop:phi_typical}
Let $(\xxx,d_{\xxx})$ be a Polish metric space. There exists a continuous injective map $\varphi:\xxx\rightarrow (\ell^2,\|\cdot\|_2)$.
\end{proposition}
We are now ready to state our first main result which guarantees that arbitrary continuous functions between arbitrary Polish metric spaces can be approximated on compact sets by ``randomized functions''.
\begin{theorem}[Transfer Principle: Polish $\xxx$ and $\yyy$]
\label{theorem:Unstructured_Case}
Assume Setting~\ref{setting:theorems_Qualitative}.  For any compact $K\subset \xxx$, any
continuous $f:(K, d_\xxx) \rightarrow (\yyy, d_\yyy)$, and any $\epsilon>0$, there exist $c, d, N, Q \in \nn_+
$, an ``approximate feature map'' $
\hat{\phi}
:\xxx
    \rightarrow 
(\rr^{d},\|\cdot\|_{F:d})$ defined by
\begin{equation}
\label{eq:compressedfeaturemap_0}
        \hat{\phi}
    \eqdef 
    	\iota_{
    	    T_{d}
    	}^{-1}
    \circ 
    	T_{d}
    \circ 
        \phi
,
\end{equation}
where $\{T_k\}_{k=1}^{\infty}$ realize the BAP on $\phi(K)$, an approximator $\hat{f} \in  
\fff_{d,N,c}$, and vectors $(u^1,z^1),\dots,(u^N,z^N)$ in $\mathbb{R}^{2\times Q}$, such that the Borel map $\hat{T}:\xxx\rightarrow \mathcal{P}_1(\yyy)$ with representation
\begin{equation}
\label{eq:theorem_determinsitic_transferprinciple__PAC_Bound___RandomizedVersion_qualitative}
    \hat{T}(x) 
\eqdef 
            \sum_{n=1}^N\,
                [P_{\Delta_N}\circ \hat{f}\circ \hat{\phi}(x)]_n
    \,
            \biggl(
                \sum_{q=1}^Q\,
                    [P_{\Delta_Q}(u^n)]_q\,
                    \delta_{y^n_{{}_{\lceil z^n_q\rceil}}}
            \biggr)
,
\end{equation}
which satisfies the estimate,
\[
    \sup_{x\in K}
    \,
        W_1\big(
            \hat{T}(x)
                ,
            \delta_{f(x)}
        \big)
    <
        \epsilon
.
\]
Furthermore, there is a typical\footnote{This means that $\mathcal{K}$ is dense in the space of non-empty compact subsets of $\xxx$, metrized by the Hausdorff--Pompeiu metric.} family $\mathcal{K}$ of compact subsets of $\xxx$ containing all finite subsets of $\xxx$, such that if $K\in \mathcal{K}$ then the parameters $\{c,d,N,Q\}$ depend quantitatively\footnote{See Lemma~\ref{lemma:Unstructured_Case} for precise estimates.  } on $\epsilon$.
\end{theorem}

\paragraph{Relationship to Transformer Networks}
Introduced by \cite{vaswani2017attention}, transformer networks are a class of deep neural networks built using the attention mechanism of \cite{bahdanau2014neural}, (a deep learning layer build suing softmax function, discussed below).  Suppose that $\xxx=\mathbb{R}^d$, $\yyy=\mathbb{R}^D$, $\varphi=1_{\mathbb{R}^d}$, and $\mathcal{F}$ is the set of deep feedforward networks in Example~\ref{ex:deepfeedfowardnetworks}.  Note that $T_n=1_{\mathbb{R}^d}$ implements the BAP of $\mathbb{R}^d$.  Taking the ``expectation'' of a random variable distributed according to each $\hat{T}(x)$ in~\eqref{eq:theorem_determinsitic_transferprinciple__PAC_Bound___RandomizedVersion_qualitative}, in the sense of Bochner integration, yields a vector-valued function $\hat{t}:\mathbb{R}^d\rightarrow\mathbb{R}^D$ given by
\[
        \hat{t}(x)
    \eqdef 
        \mathbb{E}_{X\sim \hat{T}(x)}[X]
        =
            \sum_{n=1}^N\,
                [P_{\Delta_N}(\hat{f}(x))]_n
            \,
        V_n
    ,
\]
for each $x\in \mathbb{R}^d$, where $V$ is the $N\times d$ matrix with rows $V_n\eqdef \biggl(\sum_{q=1}^Q\,[P_{\Delta_Q}(u^n)]_q\,y^n_{\lceil z^n_q\rceil}\biggr)$.  
The projection onto the $N$-simplex is analogous to the softmax function%
\footnote{Most of our analysis goes through with $P_{\Delta_N}$ replaced by the softmax. A caveat is that it requires to handle the boundary of $\Delta_N$ as a $\mathcal{Z}$-set, in the sense of \cite[Section 5]{NiceBook1}, as in Theorem~\cite[Theorem 37 and Example 13]{kratsios2021_GDL}. }
$$
\operatorname{Softmax}_N:x\mapsto \Biggl( \frac{e^{x_n}}{\sum_{i=1}^N\,e^{x_i}}\Biggr)_{n=1}^N.
$$ 
If we replace $P_{\Delta_N}$ with the $\operatorname{Softmax}_N$ then $\hat{t}$ becomes
\begin{equation}
\label{eq:transformerapprox}
    \hat{t}\approx 
            \sum_{n=1}^N\,
                [\operatorname{Softmax}_{N}(\hat{f}(\cdot))]_n
            \,
        V_n
    .
\end{equation}
The map $u\mapsto \sum_{n=1}^N\,[\operatorname{Softmax}_{N}(u)]_n\,V_n$ is simply the the attention layer of \cite{bahdanau2014neural}, which is the main novel building block of transformer networks of~\cite{vaswani2017attention}, and the matrix $V$ is the matrix of ``values''%
\footnote{Typically one considers a more complicated attention layer which computes $\sum_{n=1}^N\,[\operatorname{Softmax}_N(K^{\top}Q)]_nV_n$ from three sources: a matrix of ``keys'' $K$, a matrix of ``queries'' $Q$, and a matrix of ``values'' $V$.  Like most mathematical analyses of deep learning models, e.g.~\cite{zhou2020universality,ZHOU_ATheoryOfConvNetswDownsampling__NeuralNetworks_2020,PetersenVoigtlaender_Equivalence_ConvNetsAMS_2020},~\eqref{eq:transformerapprox} considers a mathematically tractable simplification of the transformer network where the matrix of keys $K$ is always the $1\times 1$ matrix $K=(1)$, and we identify the softmax function's source $u\in \mathbb{R}^N$ with the $1\times N$ matrix $(u_1,\dots,u_n)$; thus, for us $u=K^{\top}Q$.  A similar simplification was assumed in the probabilistic transformer networks of \cite{Kratsios2021Transformer,AcciaioKratsiosPammer2022,kratsios2021_GCDs} when approximating regular conditional distributions.} and the right-hand side of~\eqref{eq:transformerapprox} is a simple instance of a transformer network.

\subsection{Quantitative Universal Approximation}
\label{s:MainResults__ss:Deterministic}
We now turn to the cases where $\xxx$ and $\yyy$ admit additional geometric structure. 
We will first assume that $\yyy$ is a barycentric QAS space which will allow us to construct non-randomized function approximators.  

\paragraph{The Structure of Source Spaces}
\label{s:Setup__ss:Input}

To ensure that inputs in a general metric space $(\xxx,d_{\xxx})$ are compatible with the Euclidean building block, we need to relate and compress information in $\xxx$ into Euclidean data.  We therefore require some structure of $\xxx$, akin to that of a (Banach) manifold.  Namely, we require that regions in $\xxx$ can be related to Banach spaces. However, unlike topological manifolds, we neither require that every such region is homeomorphic to the model Banach space, nor that these regions fit well together.  

\begin{figure}[ht!]
	\centering
	\includegraphics[width=0.25\textwidth]{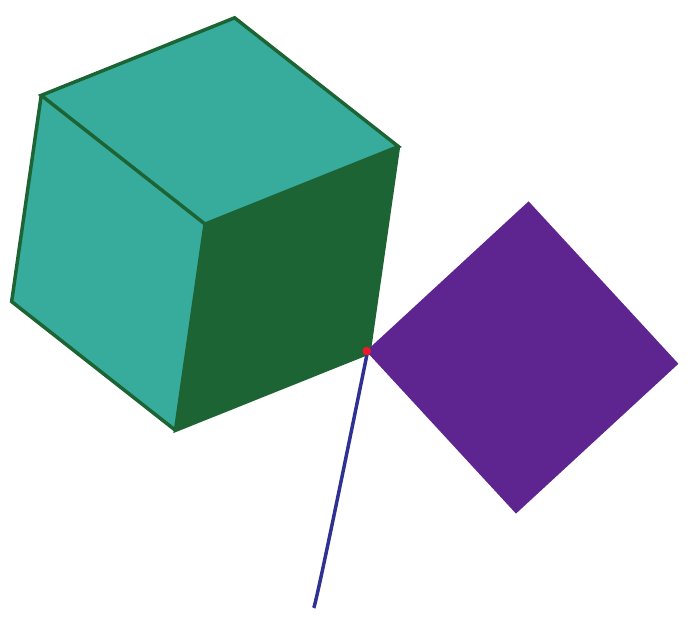}
	\caption{Feature decompositions do not need to have consistent topological dimension, nor do they need to be homeomorphic to any Banach space.}
	\label{fig_feature_manifold}
\end{figure}

Instead, as illustrated in Figure~\ref{fig_feature_manifold}, we only require that $\xxx$ can be decomposed into parts and that every such part can be embedded into a Banach space with the BAP.  The Banach spaces linearize the parts of $\xxx$ while the BAP gives us finite-dimensional approximations of points in possibly infinite-dimensional parts with arbitrarily small approximation error.  We also require that the parts be disjoint, up to a negligible subset of $\xxx$, which we quantify by equipping $\xxx$ with a measure.  

\begin{definition}[Feature Decomposition]
\label{defn:CombinatorialStructure_X}
A feature decomposition of a metric measure space $(\xxx,d_{\xxx},\mu)$, where $\mu$ is a Borel probability measure on $\xxx$, is a countable collection of pairs $\{(\xxx_n,\phi_n)\}_{n\le N}$ \footnote{$N = \infty$ is allowed.} of non-empty closed subsets $\xxx_n\subseteq \xxx$ and continuous injective maps $\phi_n:\xxx_n\rightarrow F_{n}$ into Banach ``feature spaces'' $F_n$ with the BAP satisfying
\begin{itemize}
    \item[(i)] \textbf{Almost Disjoint:} each
    $\partial \xxx_n 
    \eqdef 
        \xxx_n\bigcap \underset{{m\neq n,\,m\le N}}{\bigcup}\, \xxx_m$
    is a $\mu$-null set;
    \item[(ii)] \textbf{Covering:} there exist $C_{\mu},r>0$ 
    such that
    $
        \mu\big(
            \bigcup_{m>n} \, 
                \xxx_m
        \big)
            \le 
        C_{\mu}\, 
        n^{-r} 
    $ for all $n$;\footnote{When $N < \infty$, we only require that $\xxx = \cup_{n \le N} \xxx_n$.} 
    \item[(iii)] \textbf{Ahlfors Regularity Near Boundaries: } 
    For a subset $A \subseteq \xxx$, let $A_R$ denote its $R$-neighborhood. There is a $C>0$ such that for every $n\le N$ there is an $r_n>0$ such that
    \[
            \mu\big(
                (\partial \xxx_n)_R \cap \xxx_n
            \big)
        \le 
            C
                \mu(\xxx_n)\,
                R^{r_n}
    \]
    whenever $0\le R\le \operatorname{diam}(\xxx_n)$.
    \item[(iv)] \textbf{Non-Collapsing Diameter:} $0<\inf_{n\le N}\, \operatorname{diam}(\xxx_n)$.
\end{itemize}
Each $\phi_n$ is called a \textit{feature map} on \textit{part} $\xxx_n$.  
\end{definition}

The following example illustrates that metric spaces can have a complicated and infinite-dimensional global geometry while admitting feature decompositions into simple finite-dimensional parts.  

\begin{example}[Infinitely Many Euclidean Spaces Glued at Origin]
\label{ex_Feature_Manifold_LargebutSmall}
Consider the set $\xxx\eqdef\sqcup_{n\in \nn_+}\, \rr^{n}/\sim$ with the identification $x\sim y$ if and only if $x,y\in \{0_{\rr^n}\}_{n\in \nn_+}$. Equip $\xxx$ with the standard quotient metric
\[
    d_{\xxx}([x],[y])
        \eqdef
    \begin{cases}
        \|x-y\|_{\rr^n} & 
        : x,y\in \rr^n
    \\
        \|x\|_{\rr^n} + \|y\|_{\rr^{\tilde{n}}} & 
        : 
            x \in \rr^n,\, y \in \rr^{\tilde{n}} 
                \mbox{ and } 
                n\neq \tilde{n}.
    \end{cases}
\]
Let $\mu_n$ denote the $n$-dimensional Lebesgue measure; we make $(\xxx,d_{\xxx})$ into a metric measure space by equipping it with the measure 
\[
\mu(A)\eqdef \sum_{n=1}^{\infty} \, 
    \frac{
        1
    }{
        2^{3n/2} \pi^{n/2}
    }\int_{\rr^n} 1_{A\cap \rr^n}(x)\, 
        e^{-\|x\|^2/2}
            \mu_n(dx)
.
\]
Since $\mu(\{[0]\})=0$ we then conclude that $\{(\rr^n-\{0_{\rr^n}\},\text{Id}_{\rr^n})\}_{n\in \nn_+}$ is a feature decomposition of $(\xxx,d_{\xxx},\mu)$. 
\end{example}

Our next result applies to source spaces with  combinatorial structure in the sense of Setting~\ref{setting:theorems_Quantitative__CombinatorialX__BarycentricQAS__Y} (i), barycentric QAS target spaces, and $\omega$-H\"{o}lder-like target functions.  In this case, via an application of the barycenter map, we simply obtain universal approximators $\hat t$ from $\xxx$ to $\yyy$, as opposed to $\hat T$ in Theorem~\ref{theorem:Unstructured_Case} which maps from $\xxx$ to $\mathcal{P}_1(\yyy)$.

When the feature decomposition of $\xxx$ has more than one piece, the approximation guarantee is of a probably approximately correct (PAC)-type, meaning that it holds on a high-probability subset of $\xxx$ with probability which depends quantitatively on $\xxx$'s geometry.  

Although in this setting we proved a quantitative guarantee, here we give a qualitative statement for simplicity and because the rates coincide with our main quantitative result for general combinatorial $\yyy$ in Section~\ref{s:Main__ss:Combinatorial_XandY}.  A detailed quantitative version of the result for barycentric $\yyy$ in this section is given in Lemma~\ref{lemma:determinsitic_transferprinciple} (see also Table~\ref{tab:parameterestimates}).

\begin{setting}[Quantitative Setting: Combinatorial $\xxx$ and Barycentric QAS $\yyy$]
\label{setting:theorems_Quantitative__CombinatorialX__BarycentricQAS__Y}
Assumptions:
\begin{enumerate}[label=(\roman*)]
    \item $(\xxx,d_{\xxx},\mu)$ is a metric measure space with $\mu$ supported on a compact subset $K\subseteq \xxx$, feature decomposition $\{(\xxx_n,\phi_n)\}_{n\le N}$, and either:
    \begin{enumerate}
        \item[(a)] each $\phi_n$ is quasisymmetric and each $\operatorname{supp}(\mu)\cap \xxx_n$ is doubling; and $\phi_n\vert_{\phi_n(\xxx_n \cap \operatorname{supp}(\mu))}^{-1}$ is H\"{o}lder-like continuous for each $n\le N$,
        \item[(b)] each $\phi_n(\operatorname{supp}(\mu)\cap \xxx_n)$ is a doubling subset of $F_n$, and $\phi_n\vert_{\phi_n(\xxx_n \cap \operatorname{supp}(\mu))}^{-1}$ is H\"{o}lder-like continuous for each $n\le N$;
    \end{enumerate}
  \item for each $n\leq N$ let $\{T_k^{(n)}\}_{k=1}^{\infty}$ realize the BAP on $\phi_n(\operatorname{supp}(\mu)\cap \xxx_n)$;
  \item $(\yyy,d_{\yyy}, \hat \eta)$ is a barycentric QAS space with quantized mixing function $\hat \eta$;
  \item $\mathcal{F}_{\cdot}$ is a universal approximator.
\end{enumerate}
\end{setting}

We can now state the result for barycentric $\yyy$: any $\omega$-H\"{o}lder-like map from admissible $\xxx$ to a barycentric QAS $\yyy$ can be approximated by piecing together Euclidean universal approximators.  

\begin{theorem}[Transfer Principle: $\yyy$ is a Barycentric QAS Space]
\label{theorem:determinsitic_transferprinciple}
Assume Setting~\ref{setting:theorems_Quantitative__CombinatorialX__BarycentricQAS__Y}.  For any $\omega$-H\"{o}lder-like $f:K \rightarrow \yyy$ and any $\epsilon>0$, there exist  an $N^{\star} \le N$, positive integers 
$d_n, c_n, D_n, N_n
$ for $n \le N^{\star}$, vectors $Z_1,\dots,Z_{N^{\star}}$ with each $Z_n\in \rr^{N_n\times D_n}$, functions $\hat{f}_1,\dots,\hat{f}_{N^{\star}}$ with $\hat{f}_n \in \fff_{d_n,N_n,c_n}$,
a Borel subset $\xxx_{\epsilon}\subseteq \xxx$ with $\mu(\xxx_{\epsilon})
            \ge 
        1 - \epsilon$, and a Borel function $\hat{t}:\xxx\rightarrow \yyy$, such that 

\[
\hat{t}\vert_{\xxx_{\epsilon}}(x)
	        =
    	 \beta_{\yyy}\biggl(\sum_{n=1}^{N^{\star}}
    	   \,
    	       \psi_n(x)
    	    \,
    	    \delta_{
    	        \hat{\eta}\big(
        	            \hat{f}_n
        	        \circ 
        	            \phi^{(n)}
        	            (\cdot)
        	            ,
        	        Z_n
        	    \big)
    	    }
\biggr)
,
\] 
where \begin{equation}
\label{eq:compressedfeaturemap}
      \phi^{(n)}
\eqdef 
	\iota_{
	    T^{(n)}_{d_n}
	}^{-1}
\circ 
	T^{(n)}_{d_n}
\circ 
    \phi_n
:(\xxx_n,d_{\xxx})
    \rightarrow 
(\rr^{d_n},\|\cdot\|_{F_n:d_n})
,
\end{equation}
$\beta_{\yyy}$ denotes the barycenter map on $\yyy$, $
        \psi_n 
    \eqdef 
            \dfrac{
                d_{\xxx}(x,\xxx_n^c)
	        }{
	            \sum_{i\leq N^{\star}}\,
	                d_{\xxx}(x,\xxx_i^c)
	        }
,
$
and $\hat{t}$ satisfies
\begin{equation}
\label{eq:theorem_determinsitic_transferprinciple__PAC_Bound}
    \sup_{x\in \xxx_{\epsilon}}\,
        d_{\yyy}\big(
            \hat{t}(x)
                ,
            f(x)
        \big)
        <
            \epsilon.
\end{equation}
Moreover, if $N=1$ then $
    \sup_{x\in K}\,
        d_{\yyy}\big(
            \hat{t}(x)
                ,
            f(x)
        \big)
    <
        \epsilon$ where $K$ denotes the support of $\mu$.
\end{theorem}
\begin{proof}
The proof of a quantitative version of Theorem~\ref{theorem:determinsitic_transferprinciple}, namely Lemma~\ref{lemma:determinsitic_transferprinciple}, is given in Section~\ref{s:Proofs}.  
\end{proof}

Choosing a small value of $\varepsilon>0$ in~\eqref{eq:theorem_determinsitic_transferprinciple__PAC_Bound} simultaneously improves approximation quality and increases the size of $\xxx_\epsilon$, but requires a more complex model $\hat{f}$ as quantified in Table~\ref{tab:parameterestimates}.  

\subsection{{Quantitative Approximation: Combinatorial \texorpdfstring{$\mathcal{X}$ and $\mathcal{Y}$}{X and Y}}}
\label{s:Main__ss:Combinatorial_XandY}

\begin{figure}[H]
\centering
\includegraphics[width=0.25\textwidth]{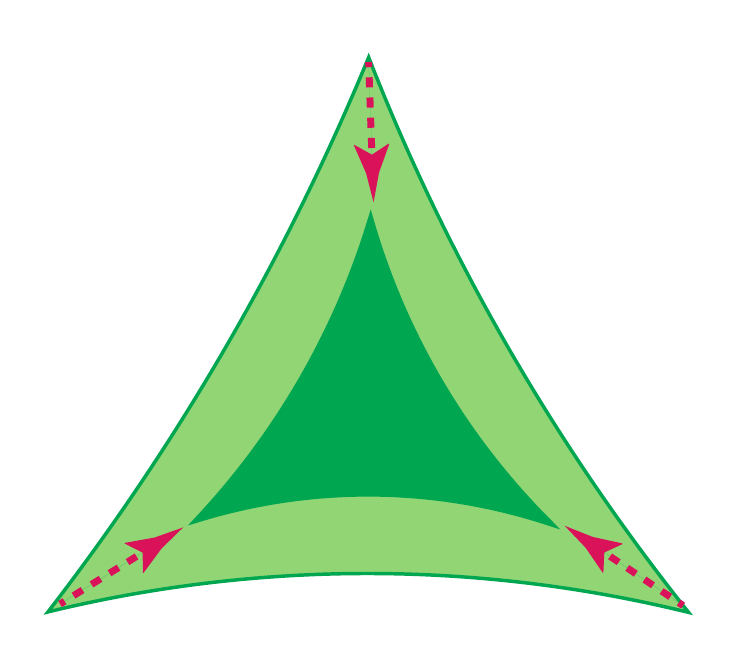}
\caption{Contracting points sets in $\yyy$ along ``$\eta$-curves''.}
\label{fig_LSet}
\end{figure}

\paragraph{The Structure of Target Spaces}
Let $(\yyy,d_{\yyy},\hat{\eta})$ be a QAS space with quantized mixing\footnote{See Section~\ref{s:Introduction__ss:QASSpaces} for details on QAS spaces and Equation~\eqref{eq:quantizedmixingmap} for definition of a quantized mixing function.} function $\hat \eta$. Fix a subset $A\subseteq \yyy$ and a reference point $\bar{y}$ in $A$, so that $(A,\bar{y})$ is a \textit{pointed subset} of $\yyy$.  As illustrated in Figure~\ref{fig_LSet}, the mixing map $\eta$ allows us to contract $A$ towards $\bar{y}$,
\[
    A^{\delta}
        \eqdef 
    \big\{
        \eta((1-\delta,\delta),(\bar{y},a)):\, a\in A
    \big\},
\]
where $0<\delta \le 1$ is a parameter quantifying how much $A$ is ``pulled towards'' $\bar{y}$.  We call a pointed subset $(A,\bar{y})$ of $\yyy$ $\eta$-\textit{geodesically convex} or simply $\eta$-\textit{convex} if for all $k \in \nn_+$, $w \in \Delta_k$ and $(y_1, \ldots,y_k) \in A^k$ one has $\eta(w, (y_1, \ldots,y_k)) \in A$. In particular, if $A$ is $\eta$-convex, then $A^{\delta}$ is contained in $A$ for every parameter $0\le \delta \le 1$.  

We are interested in retractions of sets along geodesics because they allow us to subdivide $\yyy$ into pieces which are well-behaved and can be pulled away from one another by shrinking all geodesics in $\yyy$.  
The subdivisions that we consider are such that $\yyy$ is split into nearly disjoint pieces and the boundary of each piece is a probabilistic analogue of ``topologically negligible boundaries'' studied in the geometric topology literature \cite{Anderson_1967,torunczyk1978concerning}.

\begin{definition}[Quantized Geodesic Partition]
\label{defn:CombinatorialStructure_Y}
Let $(\yyy,d_{\yyy}, \hat \eta)$ be a QAS metric space.  
A \textit{quantized geodesic partition} of $(\yyy,(Q_{q})_{q\in \nn}, \hat{\eta})$, whenever it exists, is a finite collection of closed and $\eta$-convex pointed subsets $\{(\yyy_m,\bar{y}_m)\}_{m\le M}$ satisfying
\begin{enumerate}[label=(\roman*)]
    \item for every $m,\tilde{m}\leq M$ and $0 \le \delta <1$ if $m\neq \tilde{m}$ then $\yyy_m^{\delta} \cap \yyy_{\tilde{m}} = \emptyset$;
    \item $\bigcup_{m\le M}\, \yyy_m = \yyy$;
    \item each $(\yyy_m,d_{\yyy})$ is a barycentric metric space,
\end{enumerate}
and admitting a monotonically decreasing continuous function $\boldsymbol{S}:[0,1]\rightarrow [0,\infty)$ such that $\boldsymbol{S}(1) = 0$, $\boldsymbol{S}(0)=\min_{m,\tilde{m}\le M;\, m\neq \tilde{m}}\,d_{\yyy}(\bar{y}_m,\bar{y}_{\tilde{m}})$, and
\[
        \min_{m,\tilde{m}\le M;\, m\neq \tilde{m}}\,
            d_{\mathbb{H}(\yyy)}(\yyy_m^{\delta},\yyy_{\tilde m}^{\delta}) 
    \ge 
            \boldsymbol{S}(\delta) 
,
\]
for every $\delta \in [0,1]$. The pairs $(\yyy_m,\bar{y}_m)$ are called \textit{parts} of $\yyy$'s quantized geodesic partition. 
\end{definition}

The function $\boldsymbol{S}$ in Definition~\eqref{defn:CombinatorialStructure_Y} provides a lower-bound on the rate at which the retracted parts $\{\yyy_m^{\delta}\}_{m\le M}$ are pulled apart as the parameter $\delta$ varies.  When the parameter $\delta$ approaches either 0 or 1 the bound is tight.

Using geodesic partitions, we can quantify how likely a function is to cross the intersection ``boundary-like region'' $\yyy_m \cap \yyy_{\tilde{m}}$ between any two parts $\yyy_m$ and $\yyy_{\tilde{m}}$.  The maps for which this can be quantified are called \textit{geometrically stable maps}.   

\begin{definition}[Geometrically Stable Map]
\label{defn_con_morphi}
Let $(\xxx,d_{\xxx},\mu)$ be a metric measure space, and $\{(\yyy_m,\bar{y}_m)\}_{m\leq M}$ a quantized geodesic partition of a QAS metric space $\yyy$.  A uniformly continuous $f:\xxx\rightarrow \yyy$ with H\"{o}lder-like modulus of continuity is said to be \textit{geometrically stable} if there is a constant $C_f>0$ such that for every $m\leq M$ there is a constant $q_m>0$ with
\[
	f_{\#}\mu\Big(
	    \yyy_m-\yyy_m^{\delta}
	\Big)
	    \leq
	    C_f\,
	(1-\delta)^{q_m}\,
	f_{\#}\mu(\yyy_m)
\]
for any $\delta \in [0,1)$.
\end{definition}

Our main theorem in this subsection requires the following conditions.  

\begin{setting}[Quantitative Setting: Combinatorial $\xxx$ and $\yyy$]
\label{setting:theorems_Quantitative}
Assumptions:
\begin{enumerate}[label=(\roman*)]
    \item $(\xxx,d_{\xxx},\mu)$ is a metric measure space with compactly supported $\mu$, feature decomposition $\{(\xxx_n,\phi_n)\}_{n\le N}$, and either:
    \begin{enumerate}
        \item[(a)] each $\phi_n$ is quasisymmetric and each $\operatorname{supp}(\mu)\cap \xxx_n$ is doubling; and $\phi_n\vert_{\phi_n(\xxx_n \cap \operatorname{supp}(\mu))}^{-1}$ is H\"{o}lder-like continuous for each $n\le N$,
        \item[(b)] each $\phi_n(\operatorname{supp}(\mu)\cap \xxx_n)$ is a doubling subset of the feature space\footnote{See Definition~\ref{defn:CombinatorialStructure_X}.} $F_{n}$ and $\phi_n\vert_{\phi_n(\xxx_n \cap \operatorname{supp}(\mu))}^{-1}$ is H\"{o}lder-like continuous for each $n\le N$;
    \end{enumerate}
  \item for each $n\leq N$, $\{T_k^{(n)}\}_{k=1}^{\infty}$ realizes the BAP on $\phi_n(\operatorname{supp}(\mu)\cap \xxx_n)$; 
  \item the QAS space $(\yyy,d_{\yyy},\hat{\eta})$ admits a quantized geodesic partition $\{(\yyy_m,\bar{y}_m)\}_{m\le M}$;
  \item $\mathcal{F}_{\cdot}$ is a universal approximator.
\end{enumerate}
\end{setting}

We can now state our main quantitative result for $\omega$-H\"{o}lder-like functions between metric spaces satisfying the combinatorial conditions in Setting~\ref{setting:theorems_Quantitative}.  

\begin{theorem}[Transfer Principle: Structure Case]
\label{theorem:Structured}
Assume Setting~\ref{setting:theorems_Quantitative}.  Let $f:\xxx\rightarrow \yyy$ be a $\omega$--H\"older--like continuous, geometrically stable map.  For any quantization, encoding and approximation error $\epsilon_Q$, $\epsilon_E$ and $\epsilon_A>0$ and any confidence level $0<\delta\le 1$, there exist a positive number $0<{\epsilon_{\star:A}} \le \min \{\epsilon_A,\inf_{n\le N}\, \operatorname{diam}(\xxx_{n}),1/2\}$, an $N^\star\in\mathbb{N}_+$, a Borel subset $\xxx_{\epsilon_{\star : A}}$, and a map $\hat{T}:\xxx\rightarrow \mathcal{P}_1(\yyy)$ satisfying
\begin{equation}
\label{eq:theorem_determinsitic_transferprinciple__general}
        \mu\big(
           \xxx_{\epsilon_{\star:A}}
        \big)
    \ge
        1
            -
        C_0
        \,
        \sum_{m\leq M} \,
            (1-\delta_\star)^{q_m}\,
	        f_{\#}\mu(\yyy_m)
    -
        C_1\, 
        (N^{\star})^{-r}
    -
        \sum_{n \le N^{\star}}\,
            C_2\,
            \mu\big(
                \xxx_n
            \big)
            \,
            \epsilon_{\star:A}^{r_n} \ge 1- \delta
,
\end{equation}
such that the following estimate holds on $\xxx_{\epsilon_{\star:A}}$
\[
    \sup_{x \in \xxx_{\epsilon_{\star:A}}}
        W_1\big(
                \hat{T}(x)
                    ,
                \delta_{f(x)}
            \big)
        <
            \epsilon_A + \epsilon_Q + \epsilon_E
    ,
\] 
where $\delta_\star = \boldsymbol{S}^\dagger (3 \epsilon_{\star:A})$,  $\boldsymbol{S}$ is the function specified in Definition \ref{defn:CombinatorialStructure_Y},
                $r,r_1,\dots,r_{N^{\star}}>0$ are as in Definition~\ref{defn:CombinatorialStructure_X}, and $q_1,\dots,q_M>0$ are as in Definition~\ref{defn_con_morphi}. The quantities $\epsilon_{\star:A}$, $N^{\star}$, $\delta_\star$  depend only on $\epsilon_A$ and $\delta$, and $C_0,C_1,C_2>0$ are independent of $\delta,\epsilon_A,\epsilon_E$ and $\epsilon_Q$.  
Moreover, $\hat{T}$ admits the following representation on $\xxx_{\epsilon_{\star:A}}$:
\begin{equation}
\label{eq:representationTransformer}
        \hat{T}(x) 
    = 
        \sum_{n\le N^{\star}}\,
            \sum_{m\le M}\,
                [C^{\xxx}(x)]_n\,
                [C^{\yyy}(x)]_m\,
                    \delta_{f^{(n,m)}(x)}
,
\end{equation}
where the precise expressions for the maps $([C^{\xxx}]_n)_{n\le N^{\star}}$, $([C^{\yyy}]_m)_{m\le M}$, and $f^{(n,m)}$ are recorded in Table~\ref{tab:modelsummary}.  Estimates of the number of model parameters in terms of $\epsilon_A$, $\epsilon_Q$ and $\epsilon_E$ are recorded in Table~\ref{tab:parameterestimates}.
Finally, let $\beta_{\yyy_m}$ be the contracting barycenter map on $(\yyy_m, d_{\yyy})$.  For any $x\in \xxx_{\epsilon_{\star:A}}$, there exists an $m \le M$ such that $C^{\yyy}(x)_m=1$,   $\beta_{\yyy_m}\circ \hat{T}(x)$ is well-defined and it satisfies 
\[
        d_{\yyy}\big(
                \beta_{\yyy_m}\circ \hat{T}(x)
                    ,
                f(x)
            \big)
            <
                \epsilon_A + \epsilon_Q + \epsilon_E.
\]
\end{theorem}

	\begin{table}[ht]
    \centering
    \caption{Breakdown of $\hat{T}$'s components.All constants and their quantitative estimates are recorded in Table~\ref{tab:parameterestimates}.  }
    \label{tab:modelsummary}
	\ra{1.3}
    \resizebox{\columnwidth}{!}{%
    		\begin{tabular}{@{}lllll@{}}
    			\cmidrule[0.3ex](){1-3}
    			\textbf{Component} & \textbf{Notation} & \textbf{Expression} 
    			\\    
    			\midrule
    			Partition of $\xxx$ & $[C^{\xxx}(\cdot)]_n$ & $\left(
        			\frac{d_{\xxx}(\cdot,\xxx_n^c)}{
        			\sum_{i\le N^{\star}} \, d_{\xxx}(\cdot,\xxx_i^c)}
        		\right)_{n\le N^{\star}}$
    		    \\
    			Approx. Partition of $\yyy$ & $[C^{\yyy}(\cdot)]_m$ & $\left(
        			\frac{
                        I(\hat{C}_m(\cdot)\le 2^{-2}\epsilon_A)
                    }{
                        \sum_{\tilde{m}\le M}\, 
                            I(\hat{C}_{\tilde{m}}(\cdot)\le 2^{-2}\epsilon_A)
                    }
                \right)_{m\le M}$
    			\\
    			Approx. of $f$ Near $\xxx_n\cap f^{-1}[\yyy_m]$ & $f^{(n,m)}(\cdot)$ & $\hat{\eta}\big(
            	            \hat{f}^{(m)}_n
            	        \circ 
            	            \phi^{(m)}_n
            	            (\cdot)
            	            ,
            	        Z^{(m)}_n
            	    \big)$
            	\\ 
                \midrule
            	Approx. Feature Map%
                    \tablefootnote{{Recall, $\iota_{T^{(n)}_{d_n^{(m)}}}$ is the linear isomorphism given just above~\eqref{eq:finitedimensionalcopyBanach}, $P_{\Delta_{\tilde{N}^{(m)}_n}}$ is the orthogonal projection on to Euclidean $\tilde{N}^{(m)}_n$-simplex.}}%
                on $\xxx_n$ & $\phi^{(m)}_n(\cdot)$ & 
            	$
                	\iota_{
                	    T^{(n)}_{d_n^{(m)}}
                	}^{-1}
                \circ 
                	T^{(n)}_{d_n^{(m)}}
                \circ 
                    \phi_n(\cdot)
            	$
            	\\
            	Approx. dist $f$ to $\yyy_m$ & $\hat{C}_m(\cdot)$ & 
            	$
            	\sum_{n\le N^\star}\,
                        \frac{d_{\xxx}(\cdot,\xxx_i^c)}{\sum_{j\le N^\star}\, d_{\xxx}(\cdot,\xxx_j^c) }\,
                        \sum_{i=1}^{\tilde{N}^{(m)}_n}[P_{\Delta_{\tilde{N}^{(m)}_n}}({g}^{(m)}_n\circ \tilde{\phi}^{(m)}_n(\cdot))]_i\, \vert z^{(m,n)}_i \vert
                $ 
    			\\
                Approx. Feature Maps Used to define each $\hat{C}_m$ & 
                $\tilde{\phi}^{(m)}_n(\cdot)$ & 
                $\iota_{
                	    T^{(n)}_{\tilde{d}_n^{(m)}}
                	}^{-1}
                \circ 
                	T^{(n)}_{\tilde{d}_n^{(m)}}
                \circ 
                    \phi_n(\cdot)$
                \\
    			\bottomrule
    		\end{tabular}
    }
    \caption*{%
    Here all the Euclidean universal approximators $\hat{f}^{(m)}_n$ and $g^{(m)}_n$  belong to $\mathcal{F}_{\cdot}$ with respective parameters $(d_n^{(m)},N^{(m)}_n,c^{(m)}_n)$ and $(\tilde{d}^{(m)}_n,\tilde{N}^{(m)}_n,\tilde{c}^{(m)}_n)$, 
    $T^{(n)}_{\cdot}$ realizes the BAP on $\phi(\xxx_n\cap K)$, and the parameters $Z^{(m)}_n$ and $z^{(m,n)}$ respectively belong to the Euclidean spaces of dimension $N^{(m)}_n\times D^{(m)}_n$ and $1$; for each $n=1,\dots, N^{\star}$ and $m=1,\dots,M$.    Table~\ref{tab:parameterestimates} below records all quantitative parameter estimates.  We use the notation $C_{a_1, \ldots, a_k}$ or $C_{(a_1, \ldots, a_k)}$ to highlight the dependence of the constant on $(a_1, \ldots, a_k)$.
    }
    \end{table}
	
	\begin{remark}[Different QAS Structures on Each $\yyy_m$]
	\label{remark:differentQAS_Structures}
	   Theorem~\ref{theorem:Structured} also holds when the quantized mixing function $\hat{\eta}$ is not defined globally on all of $(\yyy,d_{\yyy})$, if we instead give a family of mixing functions $\{\eta^m\}_{m=1}^M$ and quantizations $\{\mathcal{Q}_{\cdot}^m\}_{m=1}^M$, with $\mathcal{Q}_{\cdot}^m$ defined on each pointed subset $\{\yyy_m,\bar{y}_{m}\}_{m=1}^M$ satisfying Definition~\ref{defn:CombinatorialStructure_Y} (i)-(iii) (after slight modifications), such that, each $(\yyy_m,d_{\yyy}, \hat{\eta}^m)$ is a QAS space,  where the quantized mixing function $\hat{\eta}^{m}$ on $\yyy_m$ is defined via the mixing function $\eta^m$ and the quantization $\mathcal{Q}_{\cdot}^m$, as in~\eqref{eq:quantizedmixing_Wasserstein},  for $m=1,\dots,M$.  
    The only modifications one would make is to define instead each $f^{(n,m)}\eqdef \hat{\eta}^m(\hat{f}_n^{(m)}\circ \phi_n^{(m)},Z_n^{(m)})$ and $Z_n^{(m)}$ depending on $\mathcal{Q}_{\cdot}^m$.  In fact this is the case for any closed smooth submanifold in Euclidean space (which admits a triangulation), see Section~\ref{s:Manifold_Targets}.
	\end{remark}

    Next, we record quantitative estimates of all parameters used to define $\hat{T}$ in our main quantitative results.
            \begin{table}[ht!]
            \raggedleft
            \caption{Quantitative bounds on the parameters defining $\hat{T}$.  }
            \label{tab:parameterestimates}
            \centering
            \ra{1.3}
            \resizebox{\columnwidth}{!}{%
            		\begin{tabular}{@{}lllll@{}}
            			\cmidrule[0.3ex](){1-2}
            			\textbf{Parameter} & \textbf{Expression} \\
            			\midrule
            			$c_n^{(m)}$ & $c^{(m)}_n
        	\le
            	\left\lceil
                        r^{\dagger}(\omega_n,K\cap \xxx_n,d^{(m)}_n,N^{(m)}_n,s^{(m)}_n)
                    \right\rceil
        	$
        	\\
                    $\tilde{c}^{(m)}_n$ & $\tilde{c}^{(m)}_n
        	\le
            	\left\lceil
                        r^{\dagger}(\omega_n,K\cap \xxx_n,
                        \tilde{d}^{(m)}_n,
                        \tilde{N}^{(m)}_n,\tilde{s}^{(m)}_n)
                    \right\rceil
        	$
        	\\
            			$N_n^{(m)}$ & $\ln(N^{(m)}_n)
        	\le 
            C_{\omega_n}
        	       \ln\big(
                	    C_{(K\cap \xxx_n,\|\cdot\|_{F_n:d^{(m)}_n})}
            	    \big)
        	\Big\lceil
        	            - 
        	        \log_2\big(
        	           C_{\omega_n, K \cap \xxx_n, F_n, d^{(m)}_n} \omega_n^{\dagger}
                        	\big(
                        	C^\prime_{\eta,\omega_n, K\cap \xxx_n,F_n, d^{(m)}_n}\,
                            \epsilon_Q
                            \big)
        	        \big)
        	    \Big\rceil$ \\
             $\tilde{N}^{(m)}_n$ & $\ln(\tilde{N}^{(m)}_n)
        	\le 
            C_{\omega_n}
        	       \ln\big(
                	    C_{(K\cap \xxx_n,\|\cdot\|_{F_n:\tilde{d}^{(m)}_n})}
            	    \big)
        	\Big\lceil
        	            - 
        	        \log_2\big(
        	           C_{\omega_n, K \cap \xxx_n, F_n, \tilde{d}^{(m)}_n} \omega_n^{\dagger}
                        	\big(
                        	C^\prime_{\eta,\omega_n, K\cap \xxx_n,F_n, \tilde{d}^{(m)}_n}\,
                            \epsilon_A
                            \big)
        	        \big)
        	    \Big\rceil$ \\
            			$d_n^{(m)}$ & $ d^{(m)}_n\le
        	    R^{T_{\cdot}^{(n)}:\phi_n(K\cap \xxx_n)}\Big(
        	        \omega^{\dagger}_{
        	            \phi_n\vert _{\phi_n(K\cap \xxx_n)}^{-1}
        	        }
        	            \circ
        	        \omega^{\dagger}
        	        \Big(
            	        \frac{
            	            \epsilon_E
            	        }{
            	            C_{(
        	                   K\cap \xxx_n,\omega,\phi_n^{-1},F_{n},T_{\cdot}
            	            )}
            	        }
        	        \Big)
        	    \Big)$ \\
             $\tilde{d}^{(m)}_n$ & $\tilde{d}^{(m)}_n\le
        	    R^{T_{\cdot}^{(n)}:\phi_n(K\cap \xxx_n)}\Big(
        	        \omega^{\dagger}_{
        	            \phi_n\vert _{\phi_n(K\cap \xxx_n)}^{-1}
        	        }
        	            \circ
        	        \omega^{\dagger}
        	        \Big(
            	        \frac{
            	            \epsilon_A
            	        }{
            	            C_{(
        	                   K\cap \xxx_n,\omega,\phi_n^{-1},F_{n},T_{\cdot}
            	            )}
            	        }
        	        \Big)
        	    \Big)$ \\
               $D_n^{(m)}$ & $\mathscr{Q}_{f(K \cap \xxx_n)}(\epsilon_Q)$ \\
               $Z_n^{(m)}$ & $Z^{(m)}_n \in \rr^{N^{(m)}_n \times D^{(m)}_n}$ \\
            			$z_i^{(m,n)}$ & $z^{(m,n)}_i \in \rr$, $i \le \tilde{N}^{(m)}_n$ \\
            			\midrule
            			${\epsilon_{\star:A}} \text{ and 
         }N^\star$ & $\epsilon_{\star:A} \le \min \{\epsilon_A,\inf_{n\le N}\, \operatorname{diam}(\xxx_{n}),1/2\},
                C_0
                \,
                \sum_{m\leq M} \,
                    (1-\delta_\star)^{q_m}\,
        	        f_{\#}\mu(\yyy_m)
            +
                C_1\, 
                (N^\star)^{-r}
            +
                \sum_{i\le N^\star}\,
                    C_2\,
                    \mu\big(
                        \xxx_i
                    \big)
                    \,
                    \epsilon_{\star:A}^{r_i} < \delta$ \\
            			$ \delta_{\star}$ & $
            			                S^{\dagger}(3\epsilon_{A:\star})$\\
                    $\omega_n$ & $\omega_n \eqdef C_{(\epsilon_A, \omega, K\cap \xxx_n, \phi_n^{-1}, F_n, T^{(n)}_\cdot)}\omega \circ \omega_{\phi_n\vert_{\phi_n(K \cap \xxx_n)}^{-1}}$ 
                    \\
                    \midrule
                    $s^{(m)}_n$ & $C_{\eta,\omega_n,K \cap \xxx_n, F_n, d^{(m)}_n}\,
                    	   \frac{
                    	        \epsilon_A
                    	   }{
                    	        (N^{(m)}_n)^{1/2}
                    	   }$ \\
                    $\tilde{s}^{(m)}_n$ & $C_{\eta,\omega_n,K \cap \xxx_n, F_n, \tilde{d}^{(m)}_n}\,
                    	   \frac{
                    	        \epsilon_A
                    	   }{
                    	        (\tilde{N}^{(m)}_n)^{1/2}
                    	   }$
                    \\
            			\bottomrule
            		\end{tabular}
        }
        \end{table}

	\section{Applications}
	\label{s:Applications}
    
    We consider four different classes of geometries that fit our theoretical framework:
    \begin{enumerate}
        \item finite geometries, such as weighted graphs arising in computational geometry and theoretical computer science;
        \item non-smooth geometries arising in rough differential equations;
        \item infinite-dimensional linear geometries arising in inverse problems and partial differential equations;
        \item compact, smooth manifold geometries.
    \end{enumerate}
        
    \subsection{Finite Geometries}
    \label{s:Applications_ss:GraphTheory}
    
\begin{figure}[H]
    \centering
    \includegraphics[width=0.5\textwidth]{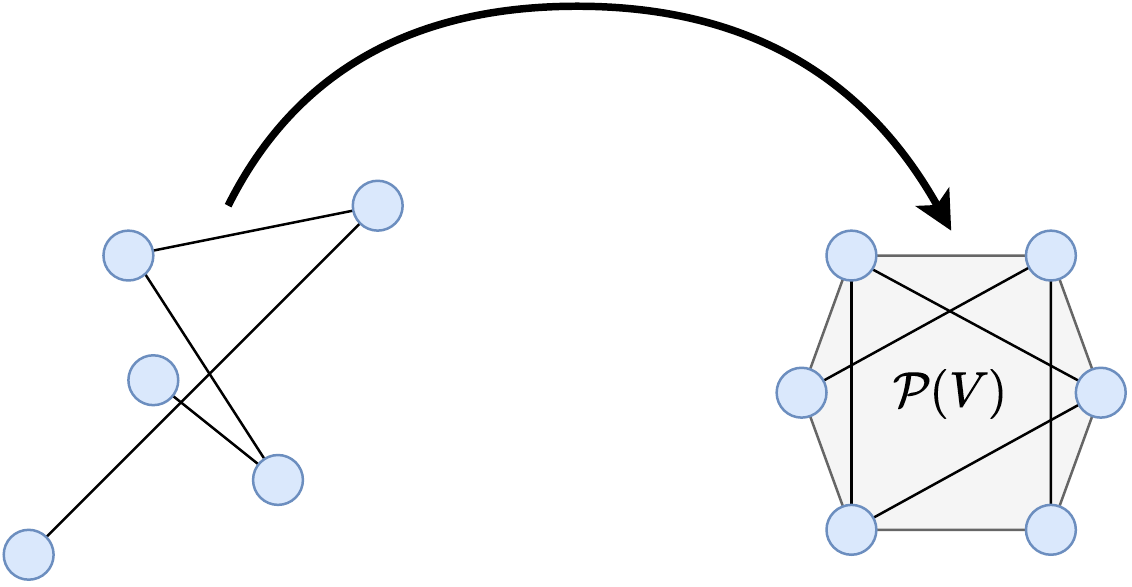}
    \caption{Illustration of a randomized function between finite graphs.}
    \label{fig:Randomized_Graph_to_Graph}
    \end{figure}

    In this section, we apply our results to approximate functions between finite metric spaces induced by weighted graph structures, illustrated in Figure~\ref{fig:Randomized_Graph_to_Graph}.  To this end, we briefly review some  terminology.
    
    Consider two weighted graphs $(V,E,W)$, where $V$ are \textit{vertices}, $E\subseteq V\times V$ are \textit{edges}, and $W:E\rightarrow (0,\infty)$ are \textit{edge weights}.  We assume that both graphs are \textit{connected}, meaning that for every pair of distinct vertices $v,u\in V$ there is a sequence of edges $\big\{(v_t,v_{t+1})\big\}_{t=1}^T\in E$ with $v=v_1$ and $v_T=u$; such sequences are called \textit{paths} from $u$ to $v$.  In this case, both graphs can meaningfully be metrized with their shortest path metric, defined for any two nodes $u,v$ in $V$ by
    \[
            d_E(u,v)
        \eqdef 
            \min\,
                \sum_{t=1}^T\,
                    W\big(\{v_t,v_{t+1})\big)
        ,
    \]
    where the minimum is computed over all paths from $u$ to $v$.  
    
    \subsubsection{Discretization of Riemannian Manifolds}
    \label{s:Applications_ss:GraphTheory___sss:Discreteizations_Riemannian}
\begin{figure}[H]
    \centering
    \includegraphics[width=0.5\textwidth]{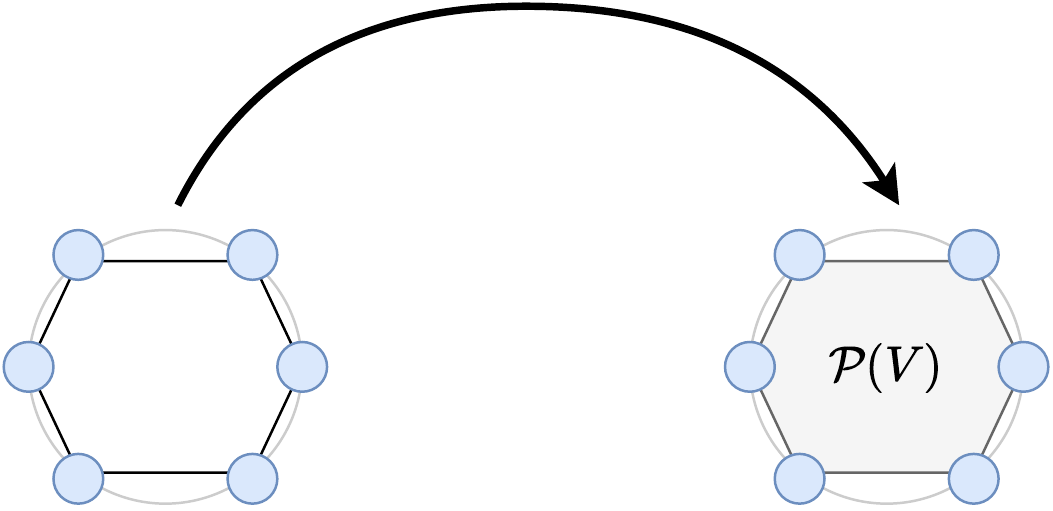}
    \caption{A function between discretizations of Riemannian manifolds, expressed as weighted graphs.}
    \label{fig:Randomized_Riemmanian_discretization}
    \end{figure}

    Illustrated in Figure~\ref{fig:Randomized_Riemmanian_discretization}, finite weighted graph approximations to compact and connected Riemannian manifolds are a classical tool in computational geometry when approximating the metric geometry of such manifolds.  Currently, approximation rates are known \cite{SingerGraphLapalcianConvergence_2006}, and there are several streamlined algorithmic implementations \cite{balasubramanian2002isomap} of graph approximation to manifolds equipped with a Riemannian distance function.  
    
    We take this as our starting point and consider two finite, connected weighted graphs $(\tilde{V},\tilde{E},\tilde{W})$ and $(V,E,W)$ together with a function $f:{V}\rightarrow \tilde{V}$ between their vertices.  If both weighted graphs can be discretizations of compact and connected Riemannian manifolds, where the vertices are points in these manifolds, the edges connect nearby points in the manifold, the weight given to each edge is the Riemannian distance between those pairs of points, and $f$ can be taken to be a restriction of a smooth function between these spaces to the vertex sets.  Since every function between finite metric spaces is Lipschitz, we deduce that $f$ is Lipschitz.  
    
    Fix a positive integer $k$ and let $\#\tilde{V} = k$.
    The $1$-Wasserstein space $\mathcal{P}_1\big(\{1,\dots,k\}\big)$ is bi-Lipschitz equivalent to the $\#\tilde{V}$-simplex with the $\ell_1$ metric%
    \footnote{This follows from the total-variation control of the $1$-Wasserstein (see \citep[Theorem 6.15]{VillaniOptTrans}) and the isometry between the total variation distance and the $\|\cdot\|_1$ distance under the set map~\eqref{eq:isometry_simplex}.}
    \begin{equation}
    \label{eq:isometry_simplex}
    \big(\mathcal{P}\big(\tilde{V},d_{\tilde{E}}),W_1\big)
    \ni
        \sum_{u\in \tilde{V}}\,w_u\,\delta_{u}
            \mapsto 
        (w_u)_{u \in \tilde{V}}
    \in (\Delta_{\#\tilde{V}},\|\cdot\|_1)
    .
    \end{equation}
    Therefore, up to identification with the map~\eqref{eq:isometry_simplex}, $f$'s ``lift'' $F\eqdef \delta_{f(\cdot)}:(\tilde{V},d_{\tilde{E}})\rightarrow (\Delta_{\#\tilde{V}},\|\cdot\|_1)$ must also be Lipschitz with the same Lipschitz constant as $f$. There are various possible choices of injective feature maps, all of which will be bi-Lipschitz.  However, a straightforward isometric feature map with finite-dimensional co-domain is given by the Fr\'{e}chet--Kuratowski type embedding
    \begin{equation}
    \label{eq:Finite_FrechetKuratowskiEmbedding}
    \begin{aligned}
    \varphi:(V,d_E) & \rightarrow (\mathbb{R}^{\#V},\|\cdot\|_{\infty})\\
    x & \mapsto \big(d_E(x,v)\big)_{v \in V}
    \end{aligned}
    .
    \end{equation}
    In this case, Theorem~\ref{theorem:Structured} implies the following approximation result.  
    \begin{corollary}[Universal Approximation of Maps Between Finite Graphs]
    \label{cor:Manifold_Discretization}
    \hfill\\
    Let $(V,E,W)$ and $(\tilde{V},\tilde{E},\tilde{W})$ be finite, connected weighted graphs, and consider $f:V\rightarrow \tilde{V}$.  
    Fix a universal approximator $\mathcal{F}_{\cdot}$.  
    For every $\epsilon>0$ there is a $\hat{T}:(V,d_E)\rightarrow \big(\mathcal{P}\big(\tilde{V},d_{\tilde{E}}),W_1\big)$ with representation 
    \[
            \hat{T}(x)
        \eqdef 
            \sum_{u\in \tilde{V}}\,
                [P_{\Delta_{\#\tilde{V}}}\circ \hat{f}\circ \big(d_E(x,v)\big)_{v \in V}]_u\,
                \delta_{u}
    ,
    \]
    satisfying the uniform estimate
    \[
        \max_{x\in V}\, 
            W_1\big(
                \hat{T}(x)
            ,
                \delta_{f(x)}
            \big)
        <
        \epsilon
    ,
    \]
    where $\hat{f}\in \bigcup_{c\in \mathbb{N}_+}\, \mathcal{F}_{\#V,\#\tilde{V},c}$.
    \end{corollary}
    In Corollary~\ref{cor:Manifold_Discretization} there is not an obvious partition of $(V,E,W)$. The next example shows how such partitions arise for planar graphs.  The approximation-theoretic advantage of partitioning is that small graphs embed into smaller Euclidean feature spaces with low distortion, leading to better approximation rates.
    
    \subsubsection{Approximating Colourings of Planar Graphs}
    \label{s:Applications_ss:GraphTheory___sss:PlanarGraphcoloring}
    
\begin{figure}[H]
    \centering
    \includegraphics[width=0.5\textwidth]{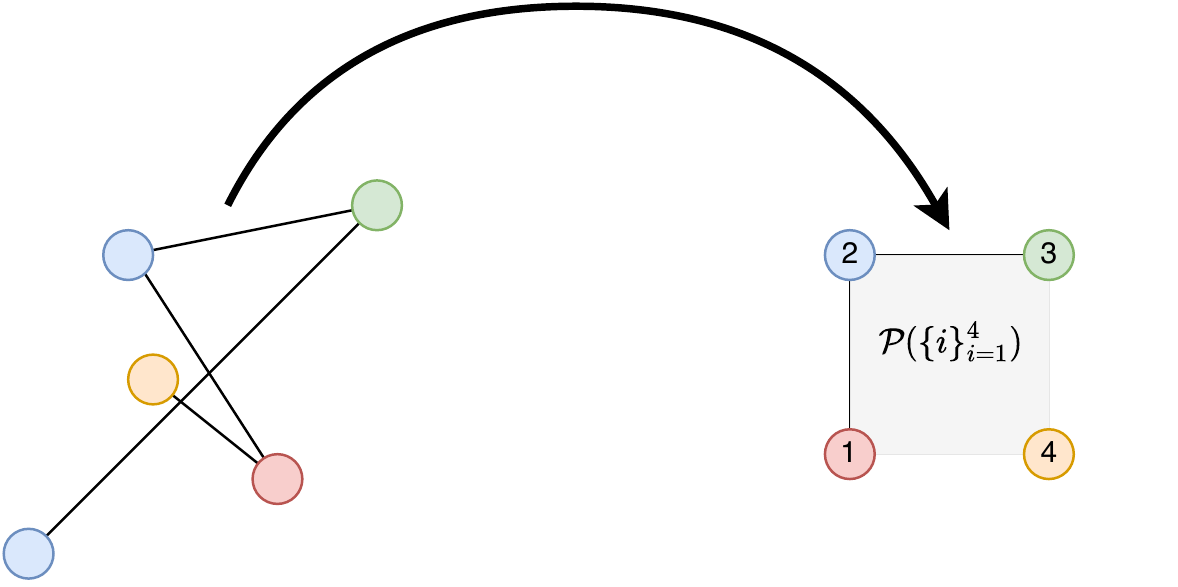}
    \caption{Randomized approximations of graph colourings.}
    \label{fig:Randomized_Graph_coloring}
    \end{figure}

    A classical problem in graph theory, illustrated in Figure~\ref{fig:Randomized_Graph_coloring}, is that of $k$-colouring a finite planar graph $G=(V,E,W)$ where $W(\{u,v\})=1$ for every $\{u,v\}\in E$.  We seek a function $f:V\rightarrow \{1,\dots,k\}$ whose outputs we interpret as colours, such that no two adjacent vertices have the same colour. 
    If a colouring exists,\footnote{This is known for three and four colours \cite{Robin_4colourings_2014,3colourings_2016}.} then our framework provides a means to approximate it.  
    
    Note that $\{1,\dots,k\}$ with the metric $\vert\cdot\vert$ can be encoded as a weighted graph $(\tilde{V},\tilde{E},\tilde{W})$ with vertices $\tilde{V}=\{i\}_{i=1}^k$, edges $\{(i,i+1)\}_{i=1}^{k-1}$, and constant weights $W(\{i,i+1\})=1$ for every $i=1,\dots,k-1$.  Therefore, the previous section's considerations apply and, up to identification with the map~\eqref{eq:isometry_simplex}, if a colouring $f:V\rightarrow \{1,\dots,K\}$ exists, we may identify its ``lift'' with the map $F\eqdef \delta_{f(\cdot)}:V\rightarrow \Delta_k$, to bi-Lipschitz equivalence of the target via~\eqref{eq:isometry_simplex}.  
    We henceforth assume that a colouring $f$ exists.  
    
    Feature decompositions of the source metric space organically arise in this problem, and their advantage is both approximation-theoretic and computational.  To see the later advantage, observe that in practice, most graphs $G$ contain a large number of vertices and edges connecting those vertices; thus, it can be computationally challenging even to compute $d_G$.  However, the under within structure on $G$ implies (\citep[Theorem 4]{PilipczukSiebertz_2021}) that there exist more than disjoint two subsets of vertices $V_1,\dots,V_I$ such that $\bigcup_{i=1}^I\, V_i = V$ and with the striking property if we define sub-graphs $G_i\eqdef (V_i,E_i)$ where $E_i\eqdef \{\{v,u\}\in E:\, u,v\in V_i\}$ then 
    \begin{equation}
    \label{eq:graph_partitioning}
        d_G(v,u) = d_{G_i}(v,u)
    ,
    \end{equation}
    for every $u,v\in V_i$ for each $i=1,\dots,I$.  Moreover, such partitions can be computed in quadratic time.  Now since each $G_i$ can contain far fewer points then the original graph $G$ and since the inclusion $(V_i,d_{G_i})$ into $(V,d_G)$ is isometric, there is no change to the original problem from a metric theoretic perspective.  
    
    When we build our feature maps, the approximation-theoretic advantage emerges from simple dimensional considerations.  For every $i\le I$, we consider the feature spaces $F_{\phi_i}\eqdef \big(\mathbb{R}^{\# V_i},\|\cdot\|_{\infty}\big)$ where the feature maps $\phi_i:V_i\rightarrow F_{\phi_i}$ are given by the analogous embeddings to~\eqref{eq:Finite_FrechetKuratowskiEmbedding} but only performed locally on each sub-graph $G_1,\dots,G_I$; that is,
    \begin{equation}
    \label{eq:frechetembedding}
            \phi_i(u)
            \eqdef 
        \big(
            d_{G_i}(u,v)
        \big)_{v\in V_i}.
    \end{equation}
    We note that $\phi_i(u) = \big(d_{G}(u,v)\big)_{v\in V_i}$.  
    
        The approximation-theoretic advantage of partitioning can naturally be explained in the context of deep learning with inputs on planar graphs.  
        Suppose that $\mathcal{F}_{\cdot}$ is the set of deep feedforward neural networks with ReLU activation function like in Example~\ref{ex:deepfeedfowardnetworks}.  Sufficiently wide neural networks in $\mathcal{F}_{\cdot}$ of depth $\mathcal{O}(c)$ approximate arbitrary Lipschitz functions from $\mathbb{R}^d$ to $\mathbb{R}$ at a rate of $\mathcal{O}(c^{2/d})$, uniformly on compact sets \cite{shen2022optimal}.  Since each sub-graph $G_i$ of $G$ contains strictly fewer vertices, approximating $f|_{G_i}\circ \varphi_i^{-1}$ with $I$ sufficiently wide neural networks of depth $\mathcal{O}(c)$ achieves an approximation error of $\mathcal{O}(c^{2/{\#V_i}})$, using a total of $\mathcal{O}(I c)$ parameters across all the $I$ neural networks.  In contrast, without partitioning,  approximating $f\circ \varphi$ by a sufficiently wide neural network of depth $\mathcal{O}(c)$ only achieves an error of $\mathcal{O}(c^{2/d})$.  Thus, by partitioning, a \textit{linear} increase in the number of parameters used to approximate $f$ on each of the sub-graphs $G_i$ results in an exponential reduction in uniform approximation error.
    
    \begin{corollary}[Universal Approximation of Stochastically Continuous $k$-Colouring's]
    \hfill\\
    Let $G=(V,E)$ be a finite planar graph, let $(G_i)_{i=1}^I$,
     $G_i=(V_i,E_i)$ be sub-graphs of $G$ satisfying~\eqref{eq:graph_partitioning}, $k$ be a positive integer, and suppose that there exists a $k$-colouring  $f:V\rightarrow \{1,\dots,k\}$ of $G$.  For every $\epsilon>0$ there is a $\hat{T}:V\rightarrow \mathcal{P}_1(\{1,\dots,k\})$ with representation
    \[
            \hat{T}(x)
        =
            \sum_{i=1}^I\,
                \psi_i(x)
            \,
            \Big(
                \sum_{v\in V_i}\,
                    [P_{\Delta_{\#V_i}}\,f_i\circ d_{G_i}(x,v)]_v\,\delta_v
            \Big)
        \qquad
        \mbox{ and }
        \qquad
            \psi_i(x)
        =
            \frac{
                    d_G(x,V_i^c)
                }{
                    \sum_{j=1}^I\,
                        d_G(x,V_j^c)
                }
    \]
    satisfying the uniform estimate
    \[
        \max_{x\in V}\, 
        W_1\big(
            \hat{T}(x)
                ,
            \delta_{f(x)}
        \big)
        <
        \epsilon
    .
    \]
    \end{corollary}
    
    \subsubsection{Classification}
    \label{s:Applicatoins__ss:Finite___sss:ClassificationRevisited}
    
\begin{figure}[ht!]
    \centering
    \includegraphics[width=0.5\textwidth]{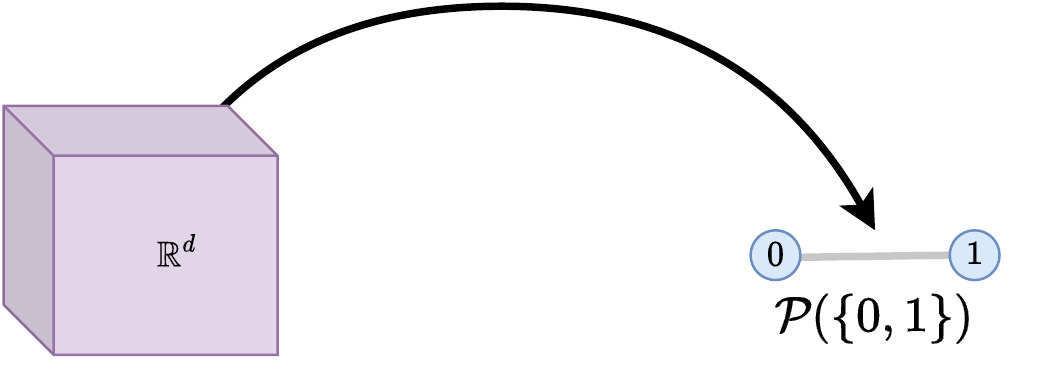}
    \caption{Classification is typically performed by lifting the target space $\{0,1\}$ to probability measures on $\{0,1\}$ and then approximating the lifted map.}
    \label{fig:Classification_Illustration}
    \end{figure}

    We now consider the binary classification problem in classical (Euclidean) machine learning, illustrated in Figure~\ref{fig:Classification_Illustration}. Suppose we are given a pair of random variables $X$ and $Y$, with $Y$ taking values in the discrete metric space $\{0,1\}$, with distance $d_{\yyy}(0,1)=1$, and $X$ taking values in the $d$-dimensional Euclidean space $\mathbb{R}^d$, for some positive integer $d$.
    By the disintegration theorem (see \citep[Theorem 6.3]{Kallenberg_FMP_Book}), we know that there exists a regular conditional distribution function (that is, a Markov kernel) $\mathbb{P}(Y\vert X=\cdot)$ describing the conditional law of $Y$ given $X$, that is a measurable function from $\mathbb{R}^d$ into the space of probability measures on $\yyy$.  
    The Bayes classifier $f:\mathbb{R}^d\rightarrow \{0,1\}$ is defined as a measurable selection\footnote{E.g.\ Suppose that $d=1$, $(W_t)_{t\ge 0}$ is a Brownian motion, $X=W_1$, and that $Y = I_{[0,\infty)}(W_2-X)$.  I.e.\ $Y$ classifies that the Brownian motion goes up or down at the next increment.  Since $(W_t)_{t\ge 0}$ is a martingale, then both states $0$ or $1$ can happen with equal probabilities; thus, there is not a unique Bayes classifier for this problem.}
    \begin{equation}
    \label{eq:ClassificationProblem}
        \mathbb{P}(Y|X=x)(\{f(x)\}) = \max_{i=0,1}\,\mathbb{P}(Y\vert X=x)(\{i\})
    .
    \end{equation}
    Since all continuous functions from $\mathbb{R}^d$ to $\{0,1\}$ are constants, no Bayes classifier can be uniformly approximated on \textit{all} compact sets by any set of continuous functions. However, the Markov kernel $\mathbb{P}(Y\vert X=\cdot)$ is always a measurable function from $\mathbb{R}^d$ to the space of probability measures $\mathcal{P}(\{0,1\})$ and it is often continuous when $\mathcal{P}(\{0,1\})$ is metrized by the Wasserstein metric\footnote{In information theory, one often encounters the total variation. We note that they are bi-Lipschitz equivalent in this case.}.   
    Thus most formulations of the classification problem instead consider the relaxed problem, illustrated by Figure~\ref{fig:Classification_Illustration}, of approximating the regular conditional distribution/Markov kernel $\mathbb{P}(Y\vert X=\cdot)$; under the implicit assumption that $x\mapsto \mathbb{P}(Y\vert X=x)$ is continuous or even Lipschitz\footnote{E.g.\ Suppose that $d=1$, $(W_t)_{t\ge 0}$ is a Brownian motion, $X=W_1$, and that $Y = I_{[0,\infty)}(W_2)$.  Then, $\mathbb{P}(Y\vert X=\cdot)$ is Lipschitz.  A similar argument holds for most strong solutions to stochastic differential equations with uniform Lipschitz dynamics, and this follows from classical stability estimates (see \cite[Propositions 8.15 and 8.16]{DaPratoMalliavin_2014}).}.  
    
    We note that the source space, $\xxx=(\mathbb{R}^d,\|\cdot\|_{\ell^2_d})$, admits the trivial feature decomposition $\{(\mathbb{R}^d,1_{\mathbb{R}^d})\}$; wherein $\mathbb{R}^d$ is its own feature space. The target space is $(\mathcal{P}_1(\{0,1\},\mathcal{W}_1)$ on $\{0,1\}$, whose elements are in correspondence with the Euclidean $2$-simplex via the bi-Lipschitz\footnote{This follows from the fact that $\mathcal{P}_1(\{0,1\}$ metrized with the total variation metric is isometric to the Euclidean $2$-simplex and from standard estimates between the total variation Wasserstein distances in such contexts (see \citep[]{VillaniOptTrans}).} identification $w_1\delta_0 + w_2\delta_1 \leftrightarrow (w_1,w_2)$.  This is a QAS space with mixing function $\eta(w,\mathbb{P}_1,\dots,\mathbb{P}_N)\eqdef \sum_{n=1}^N\, w_n\mathbb{P}_n$ and quantized by
    \[
        \mathcal{Q}_1(z)\eqdef \mathcal{Q}_1(z) \eqdef  \sigma(z)\,\delta_0 + (1-\sigma(z)) \,\delta_1 
        ,
    \]
    where $\sigma$ is a re-scaled ``hard sigmoid'' function of \cite{courbariaux2015binaryconnect} given by $\sigma(u)\eqdef \min\{\max\{u,1\},0\}$.  
    Moreover, $\yyy$ admits the  ``trivial'' quantized geodesic partition $\{([0,1],0)\}$.  
    Thus, if $\mathcal{F}_{\cdot}$ is the universal approximator of Example~\ref{ex:deepfeedfowardnetworks}, then for Lipschitz Markov kernels, Theorem~\ref{theorem:Structured} implies the following ``universal classification theorem''.
    \begin{corollary}[Universal Classification]
    \label{cor:UClassification}
    In the notation of this subsection, assume that the Markov kernel $\mathbb{P}(Y\vert X=\cdot)$ is Lipschitz.  For every $\epsilon,r>0$ there exist a map $\hat{f}$ (a ReLU network), as in Example~\ref{ex:deepfeedfowardnetworks}, satisfying
    \[
        \max_{\|x\|\le r}\,
            W_1\Big(
                \mathbb{P}(Y\vert X=x)
            ,
                \sigma(\hat{f}(x))\,\delta_{0}
                    +
                (1-\sigma(\hat{f}(x))\,\delta_{1}
            \Big)
            <
            \epsilon
    .
    \]
    \end{corollary}
    \begin{remark}
    In the binary classification literature, one typically only considers the weight of $\delta_1$ produced by the typical ``deep classifier'' model in Corollary~\eqref{cor:UClassification}, since the weight assigned to $\delta_1$ can be automatically inferred.  
    \end{remark}
    \begin{remark}[Lower Regularity Kernels]
    The Markov kernel being approximated in Corollary~\ref{cor:UClassification} is always measurable.  Therefore, if one cannot assume that it is Lipschitz, then for any ``prior probability measure'' $\mu$ on $\mathbb{R}^d$, for example the standard Gaussian probability measure, one can infer from Lusin's theorem that for $\epsilon\in (0,1]$ there is a compact subset $K$ of $\mathbb{R}^d$ on which the Markov kernel is continuous and for which $\mu(K)>1-\epsilon$.  Therefore, Theorem~\ref{theorem:Unstructured_Case} applies, from which we deduce\footnote{This type of guarantee is called a PAC (probably approximately correct) approximability guarantee in machine learning.} that
    \[
        \mu\big(
            W_1\Big(
                \mathbb{P}(Y\vert X=x)
            ,
                \sigma(\hat{f}(x))\,\delta_{0}
                    +
                (1-\sigma(\hat{f}(x))\,\delta_{1}
            \Big)
            >\epsilon
        \big)
        \ge 1-\epsilon
        .
    \]
    \end{remark}
    
    Next, we consider the implications of our theory for non-finite metric spaces when at least one of the involved geometries is a finite-dimensional (in the sense of Assouad) non-smooth metric space.  
    	
	\subsection{Non-Smooth Geometries}
    \label{s:Applications__ss:NonSmooth}
    
    We consider two classes of examples. First, we consider a broad class of finite-dimensional non-smooth metric geometries on the target space, which can arise in the context of $\operatorname{CAT}(0)$ spaces, injective metric spaces, and ultralimits thereof.  
    Second, we consider a non-smooth metric geometry on the source space, which arises from rough path theory. In this case, we approximate the solution operator to rough differential equations. 
    
    \subsubsection{Quantizable Metric Spaces with Conical Geodesic Bicombings are QAS Spaces}
    \label{s:Applications__ss:NonSmooth__sss:ConicalGeodesicBicombings}
    
    In a (separable and infinite-dimensional) Hilbert space or on a Cartan--Hadamard manifold, any two points can be joined by a unique distance-minimizing geodesic. This is not the case in general geodesic metric spaces, for which, at best, one must choose which of the multiple distance-minimizing geodesics to use when connecting a pair of points.  
    \begin{figure}[ht!]
    \centering
    \includegraphics[width=0.5\textwidth]{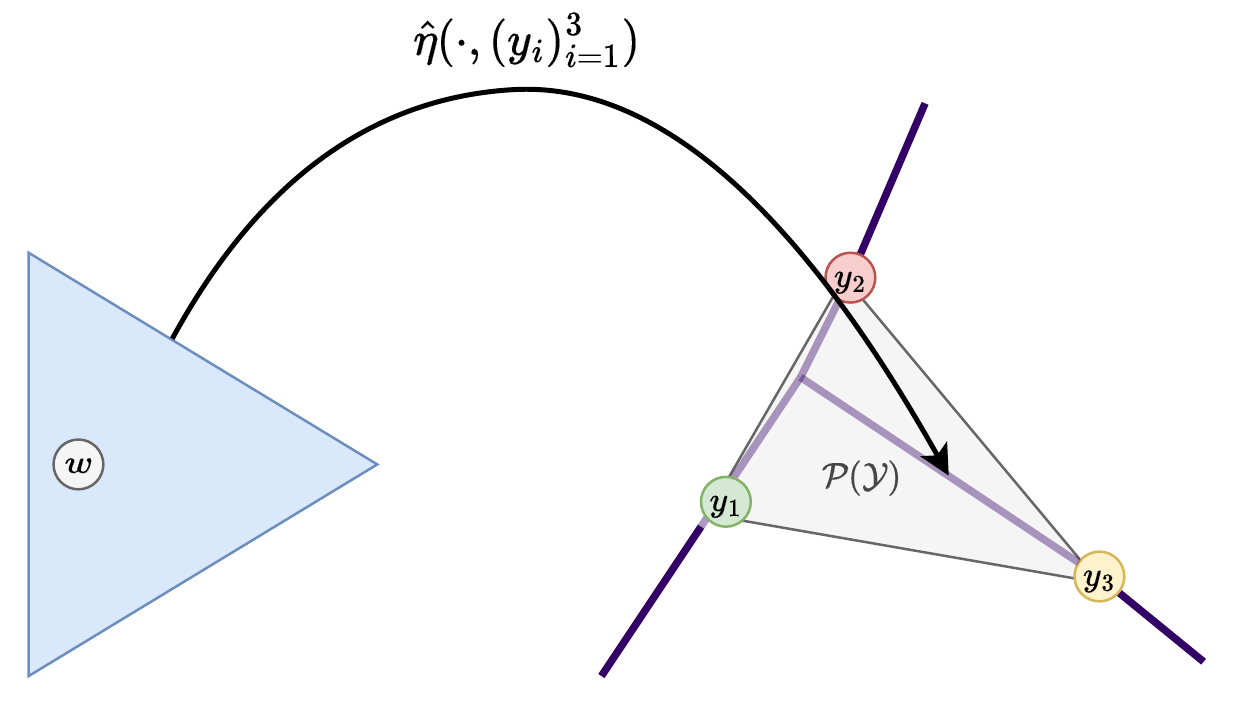}
    \caption{Visualization of the mixing function $\hat{\eta}$ in~\eqref{eq:ConicalGeodesicMixingFunction} when $(\yyy,d_{\yyy})$ is an $\mathbb{R}$-trees.  The map $\eta$ considers the unique point on $\yyy$ which is closest to the given points $y_1,y_2,y_3$ in $\yyy$, where the relative importance of each point is quantified by the ``weight'' $w\in\Delta_3$.  This is done by first going up to the $1$-Wasserstein space over $\yyy$ and mixing the lifted points $\delta_{y_1},\delta_{y_2},\delta_{y_3}$ according to weight $w$ via the mixing function~\eqref{eq:WassersteinMixingFunctionPrototype}; resulting in the measure $w_1\delta_{y_1}+w_2\delta_{y_2}+w_3\delta_{y_3}$.  Then, we go back down to $\yyy$ by identifying that measure's barycenter.}
    \label{fig:conicalgeodesicbicombings}
    \end{figure}
    Naturally, this leads to working with a distinguished selection of geodesics of a metric space $(\yyy,d_{\yyy})$, called a \textit{geodesic bicombing} \cite{BusemanPhadke_1987GeodesicBicombing}. A geodesic bicombing is a map $\sigma:\yyy\times \yyy \times [0,1]\rightarrow \yyy$ with the property that for every $x,\tilde{x}\in \yyy$ we have $\sigma(x,\tilde{x},0)=x$, $\sigma(x,\tilde{x},1)=\tilde{x}$ and $d_{\yyy}\big(\sigma(x,\tilde{x},t) , \sigma(x,\tilde{x},\tilde{t}) \big) = \vert t-\tilde{t}\vert  d_{\yyy}(x,\tilde{x})$.  
    The barycentricity condition imposes the main geometric constraint on the space $(\yyy,d_{\yyy})$.  One is typically interested in complete metric spaces admitting a \textit{conical} geodesic bicombing by which we mean a geodesic bicombing satisfying the convexity-like property%
    \footnote{The convexity of the map $t\mapsto d_{\yyy}(\eta(x,\tilde{x},t),\eta(x',\tilde{x}',t))$, for all $x,\tilde{x},x',\tilde{x}'\in \yyy$ implies but is not equivalent to $\eta$ being conical (see \cite[Proposition 3.8]{Lang_2013_InjectiveHullsMetricSpacesDiscrete}).  }
    \begin{equation}
    \label{eq_bicombing}
    	d_{\yyy}(\sigma(x,\tilde{x},t),\sigma(x',\tilde{x}',t)) \le (1-t) d_{\yyy}(x,x') + t d_{\yyy}(\tilde{x},\tilde{x}')
    	.
    \end{equation}
    This is because \cite[Theorem 2.6]{basso2020extending}  characterizes metric spaces admitting a $1$-Lipschitz barycenter map $\beta:\mathcal{P}_1(\yyy)\rightarrow \yyy$ as precisely being those which admit a conical geodesic bicombing. Examples of metric spaces admitting conical geodesic bicombings, are injective metric spaces, real trees, Cartan-Hadamard manifolds, Banach spaces, $\operatorname{CAT}(0)$ spaces, and several other spaces (see \cite{LangDescombes_2015_GeomDedic,Miesch_CartanHadamard_2017_EMS,BassoFixedPoint_2018}). 
    
    Access to such a barycenter map allows us to define a mixing function $\eta:\hat{\yyy}\rightarrow \yyy$ in two phases.  For any given $(w,(y_i)_{i=1}^N)\in \hat{\yyy}$, we first lift the mixing problem to the $1$-Wasserstein space $\mathcal{P}_1(\yyy)$ where it can be solved by assembling the finitely supported measure $\sum_{i=1}^N\,w_i\,\delta_{y_i}$.  Then, the barycenter map $\beta$ uniquely identifies a point on $\yyy$ which is closest to each $y_1,\dots,y_N$, where the relative importance of each point is quantified by the measure $\sum_{i=1}^N\, w_i\delta_{y_i}$.  
    As illustrated by Figure~\ref{fig:conicalgeodesicbicombings}, for each $(w,(y_i)_{i=1}^N)\in \hat{\yyy}$ we define
    \begin{equation}
    \label{eq:ConicalGeodesicMixingFunction}
            \eta\big(w,(y_i)_{i=1}^N\big)
        \eqdef 
            \beta\left(
                \sum_{i=1}^N\,
                    w_i\,\delta_{y_i}
            \right)
    .
    \end{equation}
    The $1$-Lipschitzness of the barycenter map $\beta$ implies that \eqref{eq:ass_approximately_simplicial} holds with $C_{\eta}=1$ and $p=1$.
    
    If, moreover, $(\yyy,d_{\yyy})$ is quantizable via some quantization $\mathcal{Q}_{\cdot}\eqdef\{\mathcal{Q}_q:\mathbb{R}^{D_q}\rightarrow \yyy\}$, then~\eqref{eq:ConicalGeodesicMixingFunction} can be used to construct a quantized mixing function $\hat{\eta} : \bigcup_{N,q\in \mathbb{N}_+} \rightarrow \yyy$, defined by,
    \begin{equation}
    \label{eq:ConicalGeodesicMixingFunction_quantized}
            \hat{\eta}(w,(z_i)_{i=1}^N)
        \eqdef 
            \beta\left(
                \sum_{i=1}^N\,
                    w_i\,\delta_{Q_q(z_i)}
            \right)
    .
    \end{equation}
    From these considerations, we deduce that any complete quantizable metric space admitting a conical geodesic bicombing is a viable target space in the context of Theorem~\ref{theorem:determinsitic_transferprinciple}. 
    \begin{proposition}[Quantizable Metric Spaces with Conical Geodesic Bicombings are QAS Spaces]
    \label{prop:QAS_Concial}
        Let $(\yyy,d_{\yyy})$ be a complete metric space with a conical geodesic bicombing $\eta$ and admitting a quantization $\mathcal{Q}$.  Then the triple $(\yyy,d_{\yyy})$ is a QAS space, with quantized mixing function $\hat{\eta}$ defined in~\eqref{eq:ConicalGeodesicMixingFunction_quantized}.  
    \end{proposition}
    A simple non-trivial example arises from the finite-dimensional space $(\mathbb{R}^d,\|\cdot\|_1)$ used in sparse learning.  
    \begin{example}[$\mathbb{R}^d$ with $\ell^1$ norm]
    \label{ex:Ell1}
    Let $d$ be a positive integer.  The finite-dimensional Banach space $(\mathbb{R}^d,\|\cdot\|_1)$ does not admit unique geodesics between any two distinct points, but $\sigma(w,y_1,y_2)\eqdef w_1y_1+w_2y_2$ is a conical geodesic bicombing, where $w\in [0,1]$ and $y_1,y_2$ are vectors in $\mathbb{R}^d$.  If we consider the ``trivial'' quantization $\mathcal{Q}_{\cdot}$ where for each $q$ we set $D_q\eqdef d$ and $\mathcal{Q}_q(z)=z$, then the quantized mixing function in~\eqref{eq:ConicalGeodesicMixingFunction_quantized} is nothing more than the weighted average of any finite set of vectors $z_1,\dots,z_N$ in $\mathbb{R}^d$ with weight $w \in \Delta_N$
    \[
            \hat{\eta}(w,(z_i)_{i=1}^N)
        =
            \sum_{i=1}^N\,
                w_i\, z_i
    .
    \]
    This is because the $1$-Lipschitz barycenter map $\beta$ on a Banach space is the Bochner integral (see \cite{BruHeinicheLootgieter1993}).  
    \end{example}
    We now consider a non-linear example relevant in stochastic analysis \cite{ContFournier_2013_FunctionalItoCalculus} and computer vision \cite{pennec2006riemannian}.  
    \begin{example}[Symmetric Positive Definite Matrices]
    \label{ex:PSD}
    Let $d$ be a positive integer.  Let $\operatorname{SPD}_d$ be the smooth manifold of $d\times d$ symmetric positive definite matrices equipped with Riemannian metric $g(X,Y)_{\Sigma}\eqdef \operatorname{tr}\big(\Sigma^{-1}X\Sigma^{-1}Y\big)$ where $\Sigma \in \operatorname{SPD}_d$ and where $X,Y$ belong to $\Sigma$'s tangent space which is identified with the set of $d\times d$ symmetric matrices $\operatorname{Sym}_d$.  This is a Cartan--Hadamard manifold, with a unique geodesic bicombing given by the weighted geometric mean $\sigma(w,\Sigma_1,\Sigma_2)\eqdef \Sigma_1^{1/2}(\Sigma_1^{-1/2}\Sigma_2\Sigma_1^{-1/2})^{w_1}\Sigma_1^{1/2}$, where $\Sigma_1,\Sigma_2\in \operatorname{SPD}_d$ and $w\in \Delta_2$ (see \cite{HiaiLim_2017_SPD_GMFlowCartanBarycenter}).  By \citep[Propositions 3.1 and 6.1]{Sturm_2003_ProbabilityNPCSpaces}, $(\operatorname{SPD}_d,d_g)$ is barycentric for its geodesic distance
    \[
    d_g(\Sigma_1,\Sigma_2) = \big\|\log(\Sigma_1^{-1/2}\Sigma_2\Sigma_1^{-1/2})\big\|_F,
    \]
    where $\Sigma_1,\Sigma_2\in \operatorname{SPD}_d$, $\log$ is the matrix logarithm, and $\|\cdot\|_F$ is the Fr\"{o}benius norm.  Exploiting the fixed-point characterization of its barycenter map in \cite[Theorem 3.1]{HiaiLim_2017_FixedPointBarycenerSPD}%
    \footnote{The barycenters in $(\operatorname{SPD}_d,g)$ are solutions to the Karcher equation $\int_{\Sigma\in \operatorname{SPD}_d}\, \log(\Sigma^{1/2} \tilde{\Sigma} \Sigma^{1/2})\,\mathbb{P}(d\tilde{\Sigma})=0$ where $\mathbb{P}$ is the finitely supported probability measure $\mathbb{P}\eqdef \sum_{n=1}^N\, w_i\, \Sigma_i$ (see \cite{LawsonLim_2014_KMeansPSDO_TAMSB,KimLeeLim_2016_OrderInequalityCharacterizationBarycenterSymCones_2016_JMAA,HiaiLim_2017_FixedPointBarycenerSPD}). }%
     implies that a mixing function $\eta$ on $\operatorname{SPD}_d$ is
    \[
            \eta\big(w,(\Sigma_i)_{i=1}^N\big)
        \eqdef 
            \min_{\Sigma \in \operatorname{SPD}_d}\,
            \biggl\|
                \sum_{i=1}^{N}
                    w_i\,
                    \log\big(
                        \Sigma^{-1/2}
                            \Sigma_i
                        \Sigma^{-1/2}
                    \big)
            \biggr\|_F
    ,
    \]
    where $\Sigma_1\dots,\Sigma_N\in \operatorname{SPD}_d$ and $w\in \Delta_N$.  We obtain a quantized mixing function by the Cartan-Hadamard Theorem since the Riemannian exponential map at the identity matrix is a diffeomorphism from $\operatorname{Sym}_d$ to $\operatorname{SPD}_D$ which is equal to the \textit{matrix exponential} $\exp$.  Upon identifying%
    \footnote{
    The identification sends any vector $z\in \operatorname{R}^{d(d+1)/2}$ to the symmetric matrix whose upper-triangular part is populated by $z$'s entries.
    }%
    $\operatorname{Sym}_d$ with $\mathbb{R}^{d(d+1)/2}$, we deduce that
    \[
            \hat{\eta}(w,(z_i)_{i=1}^N) 
        \eqdef 
            \min_{\Sigma \in \operatorname{SPD}_d}\,
            \Big\|
                \sum_{i=1}^{N}
                    w_i\,
                    \log\big(
                        \Sigma^{-1/2}
                            \exp(z_i)
                        \Sigma^{-1/2}
                    \big)
            \Big\|_F
    \]
    is a quantized mixing function on $(\operatorname{SPD}_d,d_g)$; where $z_1,\dots,z_N\in \operatorname{R}^{d(d+1)/2}$ and $w\in \Delta_N$.  
    \end{example}
 
    We now apply this to construct universal approximators of solutions to rough differential equations.  
    \subsubsection{Carnot Groups and Rough Differential Equations}
	\label{s:Applications__ss:NonSmooth__sss:RoughPath}
    Before showing how our results can be used to approximate solution operators to rough differential equations, we begin by summarizing the driving concepts behind the theory of rough paths.  
\paragraph{Rough Path Theory} The core idea of rough path theory \cite{Lyons1998} is to make precise the meaning of the (rough) differential equation
\begin{equation}\label{eq: CDE}
    dy_t = V(y_t)dx_t, \quad y(0) = y_0 \in  \mathbb{R}^e,
\end{equation}
when the driving signal $x$ is not smooth, for example, when $x \in C^\alpha([0,T], \mathbb{R}^d)$ for $0<\alpha<1/2$. Formally, the Euler scheme shows that given any sufficiently smooth vector field $V = (V_i)_{1 \le i \le d} \in \text{Lip}^{\gamma}(\mathbb{R}^e, L(\mathbb{R}^d, \mathbb{R}^e))$\footnote{$V \in \text{Lip}^{\gamma}$ means that $V$ is $[\gamma]$--times continuously differentiable and its $[\gamma]^{th}$ derivative is $(\gamma - [\gamma])$--H\"older continuous.} with $\gamma > 1/\alpha$, an integer \footnote{Here $\lfloor a \rfloor$ stands for the integer part of a real $a$.}  $N \eqdef \big\lfloor \frac{1}{\alpha} \big\rfloor\ge 2$, and any time $s <t \in [0,T]$, it holds that
\begin{equation}\label{eq: remainder estimate in Euler scheme}
   \Bigg|y(t) - y(s) - \sum_{k=1}^N\sum_{i_1, \ldots,i_k=1,\ldots, d} V_{i_1}\ldots V_{i_k}I(y(s))\int_{s<r_1< \ldots<r_k <t}dx^{i_1}_{r_1} \ldots dx^{i_k}_{r_k}\Bigg| \lesssim (t - s)^{\frac{\gamma}{\alpha}};
\end{equation}
here, the iterated integrals $\int_{s<r_1< \ldots<r_k <t}dx^{i_1}_{r_1} \ldots dx^{i_k}_{r_k}$ are well--defined for the rough signal, $x$. Since $\gamma > 1/\alpha$, the remainder term of the left--hand side of \eqref{eq: remainder estimate in Euler scheme} is of order $o(t-s)$ and, consequently, one can approximate the solution $y(t)$ to \eqref{eq: CDE} by a Riemannian sum, $y(t) = y(0) + \lim_{|\mathcal{P}|\to 0}\sum_{s<t \in \mathcal{P}}\mathcal{E}(y(s),\mathbf{X}_{s,t})$, where $|\mathcal{P}|$ is the mesh size of a partition $\mathcal{P}$ of the time-window $[0,T]$, where
\begin{equation}\label{eq:iterated_integrals_defined_rough_paths}
    \mathbf{X}_{s,t} = 1 + \sum_{k=1}^N \sum_{i_1, \ldots,i_k=1,\ldots, d}\int_{s<r_1< \ldots<r_k <t}dx^{i_1}_{r_1} \ldots dx^{i_k}_{r_k} e_{i_1}\otimes \ldots \otimes e_{i_k} 
\end{equation}
belongs to the truncated free (tensor) algebra $\in T^{N}(\mathbb{R}^d) = \bigoplus_{k=0}^N (\mathbb{R}^d)^{\otimes k}$ and where
\[
\mathcal{E}(y(s),\mathbf{X}_{s,t}) = \sum_{k=1}^N\sum_{i_1, \ldots,i_k=1,\ldots, d} V_{i_1}\ldots V_{i_k}I(y(s))\int_{s<r_1< \ldots<r_k <t}dx^{i_1}_{r_1} \ldots dx^{i_k}_{r_k}
.
\]
From this observation,~\cite{Lyons1998} introduced the notion of a rough differential equation (RDE).  Briefly, the author asserts that solving an ODE \eqref{eq: CDE} driven by rough signal $x\in C^\alpha([0,T], \mathbb{R}^d)$\footnote{For ``roughness'' we usually mean $\alpha \in (0,1/2)$ so that the Young's integral cannot be defined.} on Euclidean space $\mathbb{R}^e$ is equivalent to solving a so-called (full) RDE
\begin{equation}\label{eq:Full_RDE}
    d\mathbf{Y}_t = V(y_t)\, d \mathbf{X}_t 
\end{equation}
driven by a (geometric) rough path $\mathbf{X} \in C^\alpha([0,T], G^{N}(\mathbb{R}^d))$ on a certain manifold $G^{N}(\mathbb{R}^e) \subset T^{N}(\mathbb{R}^e)$.  The algebraic properties which the iterated integral in~\eqref{eq:iterated_integrals_defined_rough_paths} must satisfy imply that $G^{N}(\mathbb{R}^d)$ is a certain Lie group, namely, it must be the free nilpotent Carnot group over $\mathbb{R}^d$ of step $N$.  Thus, the ``rough signal'' $\mathbf{X}$ in~\eqref{eq:Full_RDE}, which mimics the iterated integrals \eqref{eq:iterated_integrals_defined_rough_paths}, takes values in this specific manifold. 
We point the reader to \cite[Chapter 7]{Friz-Victoir2010} for a precise description of these Carnot groups. Here we mainly need the fact that $G^N(\mathbb{R}^d)$ possesses a natural sub--Riemannian structure which induces the Carnot--Carath\'eodory metric $d_{CC}$. 

Consider a ``low-regularity'' $\alpha$--H\"older continuous path in $\mathbb{R}^d$, with $0<\alpha<1/2$.  Then, the iterated integrals $\int_{s<r_1< \ldots<r_k <t}dx^{i_1}_{r_1} \ldots dx^{i_k}_{r_k}$ cannot exist in any classical sense. One of the main contributions of the rough path theory is to provide a method for lifting every such rough curve $x \in C^\alpha([0,T],\mathbb{R}^d)$ to a so--called (geometric) rough path $\mathbf{X}$.  In \cite{Lyons-Victoir2007}, it is shown that these geometric rough paths take values in $G^N(\mathbb{R}^d)$, just like the classical iterated integrals do for smooth curves while also preserving the regularity of $x$. 

\begin{theorem}[{\cite[Theorem 14]{Lyons-Victoir2007}}]\label{thm:FritzVictoir2007}
For every regularity $0<\alpha<1/2$ and each path $x \in C^\alpha([0,T],\mathbb{R}^d)$, there is an $\mathbf{X} \in C^\alpha([0,T],(G^N(\mathbb{R}^d),d_{CC}))$ such that $\pi_1(\mathbf{X}) = x$, where $\pi_1: T^N(\mathbb{R}^d) = \oplus_{k=0}^N(\mathbb{R}^d)^{\otimes k} \to \mathbb{R}^d$ is the canonical projection onto the first component of any tensor therein. 
\end{theorem}

Any $\mathbf{X}$ guaranteed by  Theorem~\ref{thm:FritzVictoir2007} is referred to as a (geometric) rough path over $x$. The solution to RDE \eqref{eq: CDE} and \eqref{eq:Full_RDE} can be meaningfully constructed under the following assumptions.  
\begin{definition}\label{defn: solution to RDE}
Assume that the following hold
 \begin{enumerate}
     \item[(i)] $x \in C^\alpha([0,T],\mathbb{R}^d)$ with $\alpha \in (0,\frac{1}{2})$, $\frac{1}{\alpha} \notin \mathbb{N}$, be the driven signal in \eqref{eq: CDE};
     \item[(ii)] $V = (V_i)_{1\le i \le d}$ be a vector field such that $V_i \in \text{Lip}^\gamma(\mathbb{R}^e, \mathbb{R}^e)$ for $i=1,\ldots, d$ for some $\gamma >0$; 
     \item[(iii)] $y_0 \in \mathbb{R}^e$, $\mathbf{y}_0 \in G^N(\mathbb{R}^e)$ with $\pi_1(\mathbf{y}_0) = y_0$;
     \item[(iv)] $\mathbf{X} \in C^\alpha([0,T],(G^N(\mathbb{R}^d),d_{CC}))$ is a (geometric) rough path above $x$. 
 \end{enumerate}
Then 
 \begin{enumerate}
     \item[(a)] A (geometric) rough path $\mathbf{Y} \in C^\alpha([0,T],(G^N(\mathbb{R}^e),d_{CC}))$ is called a full RDE solution to \eqref{eq:Full_RDE} with initial value $\mathbf{y}_0$, if there exists a sequence $(x^n) \subset \text{Lip}^1([0,T],\mathbb{R}^d)$ with $\sup_n \|S_N(x^n)\|_{\alpha} < \infty$\footnote{Here $\| \cdot \|_\alpha$ denotes the $\alpha$--H\"older norm of $G^N(\mathbb{R}^d)$--valued curve with respect to $d_{CC}$.}, $\lim_{n \to \infty} \sup_{s<t \in [0,T]}d_{CC}(S_N(x^n)_{s,t}, \mathbf{X}_{s,t}) = 0 $, so that the sequence of solutions $y^n$ to the ODEs $dy^n_t = V(y^n_t) d x^n_t$, $y^n_0 = y_0$ satisfy that $\lim_{n \to \infty}\sup_{n}d_{CC}(\mathbf{y}_0 \otimes S_N(y^n)_t, \mathbf{Y}_t) = 0$. 
     \item[(b)] Let $\mathbf{Y}$ be a full RDE solution with initial value $\mathbf{y}_0$ in the above sense. Then $y = \pi_1(\mathbf{Y})$ is called a RDE solution to \eqref{eq: CDE} with initial value $y_0$.
 \end{enumerate}
\end{definition}

The following theorem regarding the regularity of the flow map associated with full RDEs, also called the \textit{It\^o--Lyons map} in stochastic analysis,  is well known in the rough path theory, see, e.g.\cite[Chapter 10]{Friz-Victoir2010}.

\begin{theorem}[{\cite[Theorem 10.41]{Friz-Victoir2010}}]\label{thm: Ito-Lyons map is Lipschitz}
 With the notations and assumptions from Definition \ref{defn: solution to RDE}  and assuming that $V \in \text{Lip}^\gamma$ with $\gamma > \frac{1}{\alpha}$, there exists a unique full RDE solution to \eqref{eq:Full_RDE} (and therefore a unique RDE solution to \eqref{eq: CDE}). Moreover, let $|\cdot|_{T^N(\mathbb{R}^e)}$ denote the normal Euclidean norm on $T^N(\mathbb{R}^e)$, the flow $\mathcal{I}^{V,\mathbf{X}}_{0 \to T}: (G^N(\mathbb{R}^e), d_{CC}) \to (G^N(\mathbb{R}^e), |\cdot|_{T^N(\mathbb{R}^e)})$, $\mathbf{y} \mapsto \mathbf{Y}^{\mathbf{y}}_T$ for $\mathbf{Y}^{\mathbf{y}}_T$ being the full RDE solution to \eqref{eq:Full_RDE} with initial value $\mathbf{y}$, is Lipschitz continuous: there exists a constant $C = C(N, d, \|\mathbf{X}\|_{\alpha}, |V|_{\text{Lip}^\gamma})$ such that for all $\mathbf{y},\mathbf{z} \in G^N(\mathbb{R}^e)$ it holds that
 $$
 \Big|\mathcal{I}^{V,\mathbf{X}}_{0 \to T}(\mathbf{y}) - \mathcal{I}^{V,\mathbf{X}}_{0 \to T}(\mathbf{z})\Big|_{T^N(\mathbb{R}^e)} \le C d_{CC}(\mathbf{y}, \mathbf{z}).
 $$
\end{theorem}
The map $\mathcal{I}^{V,\mathbf{X}}_{0 \to T}$ is referred to as the \textit{flow of the rough differential equation}~\eqref{eq:Full_RDE}. 

\begin{remark}\label{remark: importance of rough path theory}
The solution to a RDE or a full RDE depends crucially on the choice of the rough path lift $\mathbf{X}$ of the underlying $\rr^d$--valued path $x$; in general, this choice is not unique. However, if $x$ is a realization of a stochastic process $X$ with sufficiently regular trajectories (e.g.\ if $x$ is a continuous semimartingale) then, as a classical choice, $\mathbf{X}$ can simply be taken to be the iterated Stratonovich integrals.  In such cases, RDE's solution coincides with the classical strong solution to stochastic differential equation (SDE) a.s.. The advantage of the rough path theoretic approach is that it provides a pathwise stability estimate on an SDE's solution which are unavailable by classical probabilistic tools. For a comprehensive treatment of rough path theory, we refer readers to \cite{Lyons2007}, \cite{Friz-Victoir2010}, or \cite{Friz-Hairer2020} and for applications of rough path theory in stochastic analysis, time series, mathematical finance, and machine learning we refer to the 2014 ICM expository monograph \cite{Lyons2014}.
\end{remark}

Theorem~\ref{theorem:Unstructured_Case} is the first result showing that the flow of an RDE, $\mathcal{I}^{V,\mathbf{X}}_{0 \to T}$.$\mathcal{I}^{V,\mathbf{X}}_{0 \to T}$, can be uniformly approximated on compact sets, quantitatively. 

\paragraph{Source space} The set $\mathcal{X} = G^N(\mathbb{R}^e)$, $d_{CC}$ is the Carnot--Carath\'{e}odory metric, and $\mu$ is a Borel probability measure on $\mathcal{X}$ with compact support. Since the Carnot group $(G^N(\mathbb{R}^e), d_{CC})$  is a doubling metric space \cite{Hajlasz_DoublingCarnot}, if $K$ denotes the doubling constant, Naor and Neimann's quantitative formulation of Assouad's embedding theorem (\citep[Theorem 1.2]{naor2012assouad}) implies that, for every $\varepsilon \in (0,\frac{1}{2})$ there exist an $n = n(K)$, a $D = D(K,\varepsilon)$ and an embedding $\varphi_{n, \varepsilon}: G^N(\mathbb{R}^e) \to R^n $ such that for all $\mathbf{y}, \mathbf{z} \in G^N(\mathbb{R}^e)$,
    $$
   d_{CC}^{1 - \varepsilon}(\mathbf{y},\mathbf{z}) \le |\varphi_{n, \varepsilon}(\mathbf{y}) - \varphi_{n, \varepsilon}(\mathbf{z})|_{\mathbb{R}^n} \le D d_{CC}^{1 - \varepsilon}(\mathbf{y},\mathbf{z}).
    $$ 
    Fix some $\varepsilon \in (0, \frac{1}{2})$ and set $\beta := 1 - \varepsilon$,  then the above shows that $G^N(\mathbb{R}^e)$ with the snowflake metric $d_\mathcal{X} = d_{CC}^{\beta}$ admits a bi-Lipschitz feature map, namely $\varphi_{n, \varepsilon}$, with feature space $\mathbb{R}^{n}$ and the usual Euclidean norm $|\cdot|_{\mathbb{R}^n}$.  Note that $\mathbb{R}^{n}$ clearly has the BAP and, for any compact subset thereof, we can take the finite-rank operators $(T_k)_{k\in \mathbb{N}}$ to be the identity maps thereon.  
Lastly, fix any non-empty compact subset $\mathcal{K}$ of $G^N(\mathbb{R}^e)$ and let $\mu$ be any Borel probability measure thereon. 

\paragraph{Target space} The set $\mathcal{Y} = G^N(\mathbb{R}^e)$. Then we know that $\log : G^N(\mathbb{R}^e) \to \mathfrak{g}^N(\mathbb{R}^e)$ is a diffeomorphism onto a finite-dimensional Euclidean space (see \cite[Theorem 7.30]{Friz-Victoir2010}), therefore is locally bi--Lipschitz when both $G^N(\mathbb{R}^e) \subset T^N(\mathbb{R}^e)$ and $\mathfrak{g}^N(\mathbb{R}^e) \subset T^N(\mathbb{R}^e)$ are equipped with the Euclidean norm $|\cdot|_{T^N(\mathbb{R}^e)}$. Now we define $d_{\mathcal{Y}}(\mathbf{y}, \mathbf{z}) \eqdef |\log(\mathbf{y}) - \log(\mathbf{z})|_{T^N(\mathbb{R}^e)}$. Clearly, $(\mathcal{Y}, d_{\mathcal{Y}})$ is a QAS space with one single quantization map $Q = \exp: \mathbb{R}^q \cong \mathfrak{g}^N(\mathbb{R}^e)$ for $q = \text{dim}(\mathfrak{g}^N(\mathbb{R}^e))$ and the conical geodesic bicombing computed its ``Euclidean tangent space'' as
	        \[
	            \eta(w,\mathbf{y}, \mathbf{z})
	            \eqdef \exp(w_1 \log(\mathbf{y}) + w_2 \log(\mathbf{z}))
	               .\]
The mixing function $\eta$ and its quantized version can be chosen as in~\eqref{eq:ConicalGeodesicMixingFunction}. 

\paragraph{Regularity of the target function} The target function to be approximated is the flow $\mathcal{I}^{V,\mathbf{X}}_{0 \to T}: (G^N(\mathbb{R}^e), d_{CC}) \to (G^N(\mathbb{R}^e), |\cdot|_{T^N(\mathbb{R}^d})$ of the RDE~\eqref{eq:Full_RDE}.  This map fulfills the regularity requirements of the quantitative approximation result in Theorem~\ref{theorem:determinsitic_transferprinciple}: since 
$\mathbf{y} \mapsto \mathbf{Y}^{\mathbf{y}}_T$ is Lipschitz continuous by  Theorem \ref{thm: Ito-Lyons map is Lipschitz}, $d_{\mathcal{X}} = d_{CC}^\beta$, and $\text{Id}: (G^N(\mathbb{R}^e), |\cdot|_{T^N(\mathbb{R}^e)}) \to (\mathcal{Y}, d_{\mathcal{Y}})$ is locally bi--Lipschitz and therefore Lipschitz on $\mathcal{I}^{V,\mathbf{X}}_{0 \to T}(\mathcal{K})$, we see that $\mathcal{I}^{V,\mathbf{X}}_{0 \to T}: (\mathcal{K}, d_\mathcal{X}) \to (\mathcal{Y},d_\mathcal{Y})$ is actually $\beta$--H\"older continuous. Thus, we may apply Theorem \ref{theorem:Unstructured_Case}, from which we deduce the following
    \begin{corollary}[Universal Approximation of the Flow of an RDE] 
    \label{corollary: universal approx of RDE solutions}
    Let $\phi : (G^N(\mathbb{R}^e), d_{CC}^\beta) \to \mathbb{R}^n$ be an Assouad embedding defined as before and $q = \text{dim}(\mathfrak{g}^N(\mathbb{R}^e))$. For every approximation error $\varepsilon>0$ and every non-empty compact subset $\mathcal{K}\subseteq G^N(\mathbb{R}^e)$
    , there is some $m \in \mathbb{N}$ large enough such that there are $Z \in \mathbb{R}^{m \times q}$, $\hat f \in \mathcal{F}_\cdot$, 
    $\hat t(\mathbf{y})  = \hat \eta(\hat f \circ \phi(\mathbf{y}), Z)$ for which
      \[
      \sup_{\mathbf{y} \in \mathcal{K}
      } |\hat t(\mathbf{y}) - \mathcal{I}^{V,\mathbf{X}}_{0 \to T}(\mathbf{y})|_{T^N(\mathbb{R}^e)} < \varepsilon
        .
      \]
    \end{corollary}
    Although we have applied our result to approximate the flow of rough differential equations, the above universal approximation result actually holds true for any H\"older--like function $f: (G^N(\mathbb{R}^e), d_{CC}) \to (G^N(\mathbb{R}^k), |\cdot|_{T^N(\mathbb{R}^k)})$ for any $e,k \in \mathbb{N}$.
    \begin{remark}[Why not use the Carnot-Carath\'{e}odory Metric on $\yyy$?]
    Due to the Ball--Box estimate (see e.g. \cite[Proposition 7.49]{Friz-Victoir2010}), if we equip the target space $\yyy = G^N(\mathbb{R}^e)$ also with the Carnot-Carath\'{e}odory $d_{CC}$, then the flow $\mathcal{I}^{V,\mathbf{X}}_{0 \to T}: (G^N(\mathbb{R}^e), d_{CC}) \to (G^N(\mathbb{R}^e), d_{CC})$ becomes $1/N$--H\"older continuous on compact $\mathcal{K} \subset G^N(\mathbb{R}^e)$, which still fits our framework as our main theorems in Section~\ref{s:Main} hold for any H\"older--like function. Therefore one may wonder whether it is possible to use the Carnot-Carath\'{e}odory metric $d_{CC}$ on $\yyy$ to apply our main results.

    To see why it is more convenient to use the ``Euclidean metric on tangent space'' $d_{\yyy}$ note that even the simplest non-commutative Carnot group, namely the Heisenberg group,  $G^2(\mathbb{R}^2)$ equipped with the Carnot-Carath\'{e}odory metric $d_{CC}$, does not meet the criteria to apply Proposition~\ref{prop:QAS_Concial} since it does not admit a conical geodesic bicombing. A metric space admits a conical geodesic bicombing only if it has trivial Lipschitz homotopy groups; this, however, is not the case for the Heisenberg group \cite{WengerYoung_2014_GAFA,Hajlasz_2018_AMS} with its Carnot metric. 
    \end{remark}
    
    A standard \textit{infinite-dimensional} (in the sense of Assouad) example of a quantizable metric space admitting a conical geodesic bicombing are Banach spaces.  In the next subsection, we show how regular maps between infinite-dimensional linear geometries can also be approximated within our framework.  
	\subsection{Infinite-Dimensional Linear Geometries}
	\label{s:Applications__ss:InverseProblemsPDE}
	
	Theorem~\ref{theorem:determinsitic_transferprinciple} shows us how to generically construct universal approximators between infinite-dimensional Banach spaces $\xxx$ and $\yyy$ with  Schauder bases $\{x_n\}_{n=1}^{\infty}$ and $\{y_n\}_{n=1}^{\infty}$.  
	\begin{figure}[ht!]
	\centering
	\includegraphics[width=1\textwidth]{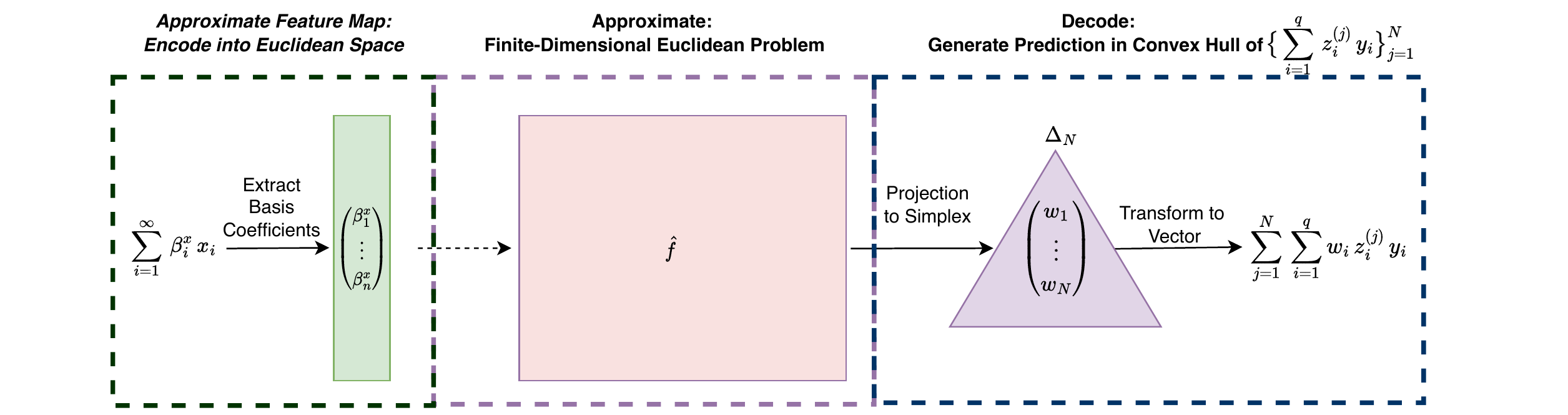}
	\caption{Summary of the simplest model constructible from approximating non-linear operators between infinite-dimensional Banach spaces $\xxx$ and $\yyy$ with Schauder bases $\{x_n\}_{n=1}^{\infty}$ and $\{y_m\}_{m=1}^{\infty}$.  First one approximates any input in $\xxx$ by its first few basis coefficients, these are transformed by a function in the universal class $\mathcal{F}$, the function's outputs are projected onto a simplex and then assembled into a vector in $\yyy$ lying in the convex hull of $\{\sum_{j=1}^q\, z_q^{(i)}\,y_q\}_{i=1}^N$ for given parameters $z^{(1)},\dots,z^{(N)}\in \mathbb{R}^q$.}
	\label{fig:OperatorApproximation}
    \end{figure}
	We will first describe a basic application of that result. We take $\varphi$ to be the identity map on $\xxx$ and let $\{T^{(n)}\}_{n=1}^{\infty}$ be finite-rank operators taken to be the truncation maps sending any $x\in \xxx$ with unique basis representation $x=\sum_{i=1}^{\infty}\, \beta_i^x\,x_i$ to its truncated Schauder basis expansion
	\begin{equation}
	\label{eq:linearmodel_featuremap__Naive}
    	T^{(n)}\Big(
    	\sum_{i=1}^{\infty}\beta_i^x \, x_i \Big)
	\eqdef 
        \sum_{i=1}^n\,\beta_i^x\,x_i
	.
	\end{equation}
	Thus, $\{\xxx,\text{Id}_{\xxx}\}$ trivially satisfies Assumptions~\ref{setting:theorems_Quantitative__CombinatorialX__BarycentricQAS__Y} (i) and (ii).  
	
	In \cite{BruHeinicheLootgieter1993}, it is shown that any Banach space $\yyy$ is barycentric with a unique barycenter map given by Bochner integration $\mathbb{P}\mapsto \int_{y\in \yyy}\, y\,\mathbb{P}(dy)$. As shown in \citep[Example 5.1]{AcciaioKratsiosPammer2022}, any such $\yyy$ is a QAS space with the following structure. For  $q\in \mathbb{N}$, the $q^{\text{th}}$ quantization map $\mathcal{Q}_q:\mathbb{R}^q\to \yyy$ assembles a parameter $z\in \mathbb{R}^q$ into a linear combination of the first $q$ Schauder basis elements in $\yyy$ by 
	$
	    \mathcal{Q}_q(z)
	\eqdef 
	    \sum_{i=1}^q\,z_i\, y_i
	.
	$  
	A mixing function $\eta$ on $\yyy$ can be defined, similarly to Example~\ref{ex:Ell1}, by assembling any finite set of points $\tilde{y}_1,\dots,\tilde{y}_N$ in $\yyy$ and any weight $w$ in the $N$-simplex $\Delta_N$ via the convex combination
	$
	\eta(w,(\tilde{y}_j)_{j=1}^N)\mapsto \sum_{j=1}^N\, w_j\, \tilde{y}_j
	.
	$  Combining the quantization $\mathcal{Q}_{\cdot}\eqdef (\mathcal{Q}_q)_{q=1}^{\infty}$ with the mixing function $\eta$ produces a QAS space structure on $\yyy$, with quantized mixing function, $\hat{\eta}$, defined for  positive integers $N,q$, weights $w$ in the $N$-simplex $\Delta_N$, and parameters $z^{(1)},\dots,z^{(N)}$ in $\mathbb{R}^q$, according to
	\begin{equation}
	\label{eq:linearmodel_quantizedmixingfunction}
	        \hat{\eta}(w,(z_j)_{j=1}^N)
	    \eqdef
    	    \sum_{j=1}^N\,
    	    \sum_{i=1}^q
    	        w_j
    	        \,
        	        z^{(j)}_{i}
        	        \,
        	            y_i
	    .
	\end{equation}
    Therefore, Assumption~\ref{setting:theorems_Quantitative__CombinatorialX__BarycentricQAS__Y} (iii) is satisfied.  The resulting class of function approximators, illustrated in Figure~\ref{fig:OperatorApproximation}, consists of all non-linear operators $\hat{t}:\xxx\rightarrow \yyy$ with representation
    \begin{equation}
    \label{eq:linearmodel__operatornetwork}
        \hat{t}(x)
    \mapsto 
        \sum_{j=1}^N\,
            \sum_{i=1}^q\,
                \Big[
                    P_{\Delta_N}
                        \circ
                            \hat{f}
                            \Big(
                                (\beta_k^x)_{k=1}^n
                            \Big)
                \Big]_j
                \,
                z^{(j)}_{i}
                    y_i
        ,
    \end{equation}
    with $n$ and $\beta^x_k$ as in~\eqref{eq:linearmodel_featuremap__Naive}, $N$ and $z^{(j)}_i$, $j=1,\ldots,N$, $i=1,\ldots, q$ as in~\eqref{eq:linearmodel_quantizedmixingfunction}, and $\hat{f} \in \cup_{c}\,\mathcal{F}_{n,N,c}$ for some universal approximator $\mathcal{F}$.
    
    Theorem~\ref{theorem:determinsitic_transferprinciple} guarantees that the class of maps with representation~\eqref{eq:linearmodel__operatornetwork} can approximate any non-linear H\"{o}lder operator from $\xxx$ to $\yyy$.
	\begin{corollary}[Quantitative Approximation of H\"{o}lder Operators]
	\label{cor:Operator_Quantitative}
	Let $f:\xxx\rightarrow \yyy$ be a non-linear H\"{o}lder continuous operator between infinite-dimensional Banach spaces $\xxx$ and $\yyy$ with respective Schauder bases $\{x_n\}_{n=1}^{\infty}$ and $\{y_n\}_{n=1}^{\infty}$.  Let $\mathcal{F}$ be a universal approximator.  
	For any error $\varepsilon>0$ and any compact subset $K\subseteq \xxx$ there is a $\hat{t}$ with representation~\eqref{eq:linearmodel__operatornetwork} satisfying
	\[
	   \sup_{x\in K}\,
	   \big\|\hat{t}(x)-f(x)\big\|_{\yyy}
	   <
	   \epsilon
	,
	\]
	quantitatively, where $\yyy$ is normed by $\|\cdot\|_{\yyy}$.  
	\end{corollary}

    \begin{remark}[A ``Discretization-Invariant'' Neural Operator Paradigm]
	\label{rem:NeuralOperators}
	   The results in this subsection results are an instance of deep learning models known as neural operators \cite{li2020fourier,kovachki2021neural,kovachki2021universal,ALG_UniversalApproximationOperators_2022}. These models  approximate maps between function spaces, typically Hilbert spaces. An important requirement is that they be ``discretization invariant'' in the sense that the model parameters do not depend on the evaluation of the input function at any point. Our framework allows us to construct such ``discretization invariant'' universal approximators between suitable Banach spaces; an example is illustrated in Figure~\ref{fig:OperatorApproximation} with a Schauder basis in $C([0,1])$ given by wavelets.  In this case, the basis ``truncation levels'' and $\hat{f}$'s weights do not explicitly depend on any input of the function being evaluated.  We further note that we recover a variant of the Fourier neural operator (FNO) when $\xxx=\yyy=L^2(\mathbb{S}^1)$ and the Schauder basis is taken to be the usual orthonormal basis of $L^2(\mathbb{S}^1)$, where $\mathbb{S}^1$ denotes the circle with its usual Riemannian metric.
	\end{remark}
 
Non-linear operators from $\xxx$ to $\yyy$ of sub-H\"{o}lder regularity can be approximated by applying Theorem~\ref{theorem:Unstructured_Case}. The resulting model will be of a form similar  to~\eqref{eq:linearmodel__operatornetwork}. To show this, we note that since $\{y_i\}_{i=1}^{\infty}$ is a Schauder basis for the second-countable space $\yyy$, there is a countably-infinite dense subset of $\yyy$ of the form $\{\sum_{i=1}^{N_k}\,z^{(k)}_i\,y_i\}_{k=1}^{\infty}$. 
	Theorem~\ref{theorem:Unstructured_Case}, concerns the set of probability measure-valued functions of the form
    \begin{equation}
    \label{eq:linearmodel__operatornetwork__setup}
        \hat{T}(x)
    \mapsto 
            \sum_{j=1}^N\,
                \Big[P_{\Delta_N}\circ \hat{f}
                    \big(
                        (\beta_i^x)_{i=1}^n
                    \big)
                \Big]_j
    \,
            \biggl(
                \sum_{q=1}^Q\,
                    u^{(j)}_q
                    \delta_{
                        \sum_{s=1}^{N_{j}}\,
                            z^{(j)}_s\,y_s
                    }
            \biggr)
        ,
    \end{equation}
    where $u^{(1)},\dots,u^{(N)}$ belong to the $Q$-simplex $\Delta_Q$ and each $\sum_{s=1}^{N_{j}}\,z^{(j)}_s\,y_s$ belongs to the countably infinite dense subset $\{\sum_{i=1}^{N_k}\,z^{(k)}_i\,y_i\}_{k=1}^{\infty}$.  
    A direct application of the result implies that for any $\varepsilon>0$, any continuous non-linear operator $f:\xxx\rightarrow \yyy$, and any compact subset $K\subseteq \xxx$ there is some $\hat{T}$ with representation~\eqref{eq:linearmodel__operatornetwork__setup} satisfying the probabilistic approximation guarantee
 \begin{equation}
 \label{eq:linearmodel__qualitativeweakform}
    \sup_{x\in K}\, W_1(\delta_{f(x)},\hat{T}(x))<\varepsilon
 .
 \end{equation}
    Leveraging the unique barycenter map $\mathcal{P}_1(\yyy)\to \yyy$ given by Bochner integration, we may ``collapse'' the measure-valued map $\hat{T}$ to a $\yyy$-valued map by post-composition with $\beta$ as in~\eqref{eq:ConicalGeodesicMixingFunction_quantized}.  The Bochner integral's linearity allows us to simplify\footnote{In more detail: 
    $
            \hat{t}(x)
    \eqdef 
        \int_{y\in \yyy}\,y\,\hat{T}(x)(dy)
    =
        \sum_{j=1}^N\,
            \Big[P_{\Delta_N}\circ \hat{f}
                \big(
                    (\beta_i^x)_{i=1}^n
                \big)
            \Big]_j
        \,
        \biggl(
            \sum_{q=1}^Q\,
                u^{(j)}_q
                    \sum_{s=1}^{N_{j}}\,
                        z^{(j)}_s\,y_s
        \biggr)
    = 
        \sum_{j=1}^N\,
        \sum_{q=1}^Q\,
            \sum_{s=1}^{N_{j}}\,
                u^{(j)}_q
                z^{(j)}_s
            \Big[P_{\Delta_N}\circ \hat{f}
                \big(
                    (\beta_i^x)_{i=1}^n
                \big)
            \Big]_j
        \,
            y_s
    \eqdef  
        \sum_{j=1}^{N}\,
        \sum_{i=1}^{\tilde{Q}}\,
                \tilde{z}^{(i)}_j
            \Big[P_{\Delta_N}\circ \hat{f}
                \big(
                    (\beta_i^x)_{i=1}^n
                \big)
            \Big]_j
        \,
            y_i
        ,
    $
    } the expression of $\hat{t}(x)
    \eqdef 
        \int_{y\in \yyy}\,y\,\hat{T}(x)(dy)$ to
    \begin{equation}
    \label{eq:linearmodel__operatornetwork__general}
    \begin{aligned}
        \hat{t}(x)
    =
        \sum_{j=1}^{N}\,
        \sum_{i=1}^{\tilde{Q}}\,
            \Big[P_{\Delta_N}\circ \hat{f}
                \big(
                    (\beta_i^x)_{i=1}^n
                \big)
            \Big]_j
        \tilde{z}^{(i)}_j
        \,
            y_i
        ,
    \end{aligned}
    \end{equation}
    where $\tilde{Q},N$ are positive integers and $\tilde{z}^{(1)},\dots,\tilde{z}^{(N)}$ belong to $\mathbb{R}^{\tilde{Q}}$.  
    Since barycenter map $\beta$ is $1$-Lipschitz, \cite{BruHeinicheLootgieter1993}\eqref{eq:linearmodel__qualitativeweakform} implies the following qualitative guarantee for maps of the form~\eqref{eq:linearmodel__operatornetwork__general}.  
    \begin{corollary}[Qualitative Approximation of Continuous Operators]
    \label{cor:Operator_Qualitative}
    Let $\xxx$, $\yyy$, and $\fff$ be as in Corollary~\ref{cor:Operator_Quantitative}.  For any continuous non-linear operator $f:\xxx\rightarrow \yyy$, any error $\varepsilon>0$, any compact subset $K$ of $\xxx$ there is some $\hat{t}:\xxx\rightarrow \yyy$ with representation~\eqref{eq:linearmodel__operatornetwork__general} satisfying
    \[
        \sup_{x\in K}\,
            \|
                f(x)
                    -
                \hat{t}(x)
            \|_{\yyy}
        <
            \varepsilon
    .
    \]
    \end{corollary}
    
	The architecture used in Corollaries~\ref{cor:Operator_Quantitative} and~\ref{cor:Operator_Qualitative} was generic for any $\xxx$ with the BAP.   However, improved approximation rates and specialized constructions for continuous non-linear operator approximation can be given if more structure is available.  
	We now describe how one can modify the first part of the model in Figure~\ref{fig:OperatorApproximation} by modifying the feature map.  The first example is motivated by the analysis of inverse problems when measurements lie in an immersed (finite-dimensional) sub-manifold of the infinite-dimensional Banach space $\xxx$.  

    \begin{figure}[H]
    \centering
     \begin{subfigure}[b]{0.45\textwidth}
     \centering
        \includegraphics[width=0.95\textwidth]{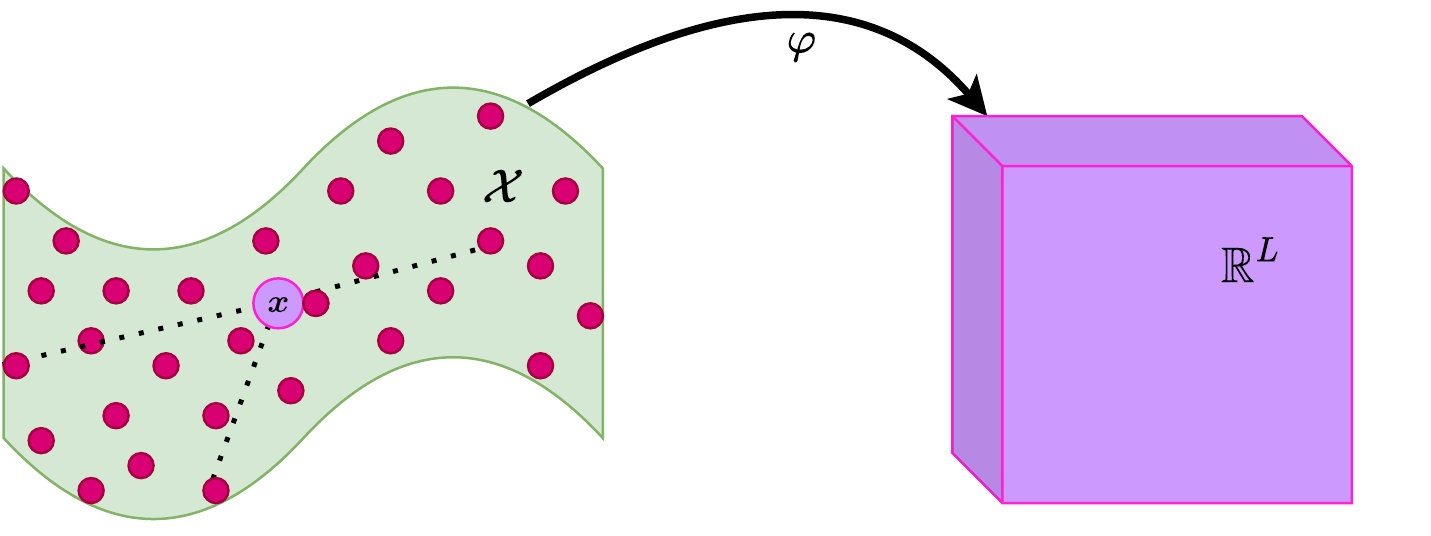}
        \caption{\textbf{Data on A Finite-Dimensional Manifold:} If data lies on a finite-dimensional closed and connected Riemannian submanifold $\xxx$ of the source Banach space then, a bi-Lipschitz feature map can be constructed by considering the distance of any input to a finite number of ``reference points'' in $\xxx$.}
        \label{fig:GromovMap}
     \end{subfigure}
     \hfill
     \begin{subfigure}[b]{0.45\textwidth}
         \includegraphics[width=0.95\textwidth]{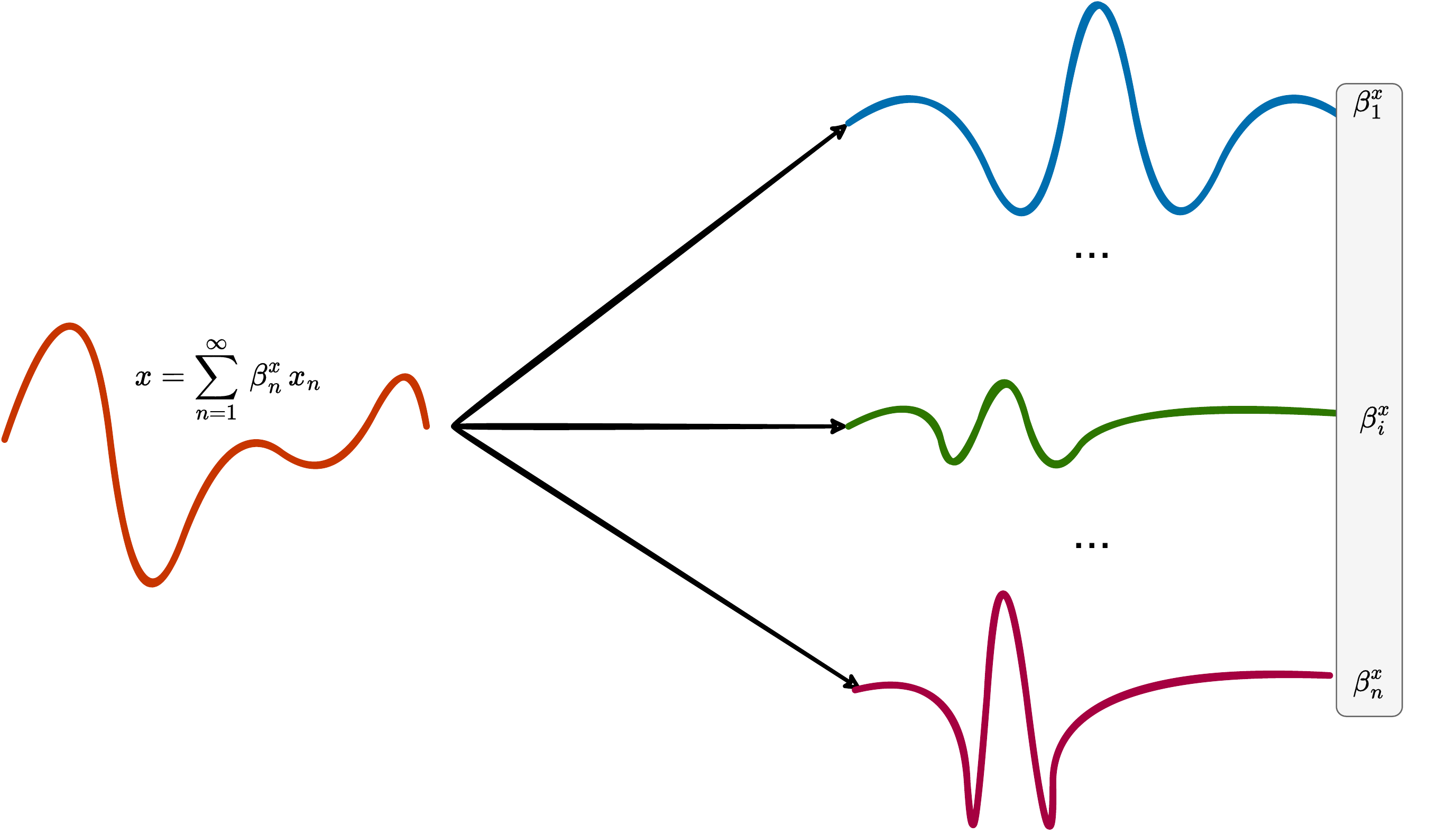}
        \caption{\textbf{No A-Priori Knowledge of Problem:} If we do not know any structure in the approximation problem, then one can always default to the feature map described in~\eqref{eq:linearmodel_featuremap__Naive}.}
        \label{fig:Frequencies_FNO}
     \end{subfigure}
    \caption{Our framework can leverage additional structure about a learning problem by customizing the feature map in the first segment in Figure~\ref{fig:OperatorApproximation}.  }
    \end{figure}

	\subsubsection{Feature Maps When Data Lies on a Smooth Compact Manifold}
\label{s:SmoothManifolds__ss:Triangulatable}
Consider two Banach spaces $B$ and $E$ with norms $\|\cdot\|_B$ and $\|\cdot\|_E$ and a continuous linear map $F:B\rightarrow E$. A common assumption in computational practice is that data in $B$ is  contained in a finite-dimensional topological submanifold $\yyy$ of $B$ (see e.g.\ \cite{VincentPascaletalManifoldDenoising2008}). In the context of inverse problems, it was recently shown that this ``manifold hypothesis'' guarantees H\"older stability of the inverse restricted to $\yyy$ \cite{alberti2020inverse}.  


We now show how this fits with our results. Suppose that $\yyy$ is a topological submanifold of $B$ and that it is equipped with a $C^{\infty}$-smooth structure $\{(\phi_{\gamma},U_{\gamma})\}_{\gamma\in \Gamma}$.  We do not require that $\yyy$ is a smooth submanifold of $B$.  We say that $\yyy$ is \textit{$\beta$-H\"{o}lder in $B$} if there is a constant $0<c\le 1$ such that
\[
        \frac{
            \|\phi_{\gamma}(x)-\phi_{\gamma}(\tilde{x})\|_2^{\beta}
        }{
            \|x-\tilde{x}\|_B
        }
    \ge
        c
,
\]
holds for every $\gamma \in \Gamma$ and every $x,\tilde{x}\in U_{\gamma}$.  The forward operator $F:B\rightarrow E$ is said to be \textit{differentiable} on $\yyy$ if $F\circ \phi_{\gamma}^{-1}:\phi_{\gamma}(U_{\gamma})\rightarrow E$ is Fr\'{e}chet differentiable and its Fr\'{e}chet derivative $\nabla(F\circ \phi_{\gamma}^{-1})$ is continuous for every $\gamma \in \Gamma$.  A fortiori, we call $F$ an \textit{immersion} if it is differentiable on $\yyy$, injective when restricted to $\yyy$, and if the differentials $dF_x: T_x(\yyy) \rightarrow E$ are injective linear maps for every $x\in \yyy$.  

If $F$ is an immersion on a smooth manifold $\yyy$ which is $\beta$-H\"{o}lder in $B$ then~\citep[Theorem 2.2]{alberti2020inverse} implies that $F\vert_{F(\yyy)}^{-1}:F(\yyy)\rightarrow B$ exists on $F(\yyy)$ and that the inverse-operator is of $\beta$-H\"{o}lder regularity.  Moreover, since $F$ is an immersion, then $\xxx\eqdef F(\yyy)$ is compact and by \citep[Theorem 3.2]{HirschDifferentialTopology1994} it is a smooth manifold.  

Let us suppose that we know $\xxx$, but we do not assume that we know $F^{-1}$.  Assuming that $\yyy$ is connected, we have that $\xxx$ is a closed and connected smooth manifold.  We may therefore endow it with some complete Riemannian metric $g$ so that $(\xxx,g)$ is a Riemannian manifold.  If we can identify a bi-Lipschitz feature map from $\xxx$ to a suitable feature space, then we may apply Theorem~\ref{theorem:Structured} to approximate the inverse-operator $F\vert_{\xxx}^{-1}:\xxx\rightarrow B$ so long as $B$ also admits a Schauder basis.  
Let us construct such a feature map using a modification of the idea in~\eqref{eq:Finite_FrechetKuratowskiEmbedding} originally due to Gromov \cite{Gromov_FillingRiemannianManifolds_1983__JDiffGeo} and illustrated in Figure~\ref{fig:GromovMap}.

\paragraph{Featurizing the Source Smooth Manifold}
Let $(\xxx,g)$ be a closed, connected Riemannian manifold.  As usual, we view $(\xxx,d_g)$ as the metric space metrized by the geodesic distance $d_g$, defined for any $x,\tilde{x}\in \xxx$ as the infimal length of any piecewise smooth curve $\gamma:[0,1]\rightarrow \xxx$ joining those two points (that is, $\gamma(0)=x$ and $\gamma(1)=\tilde{x}$)
\[
    d_g(x,\tilde{x}) 
        \eqdef 
    \inf_{\gamma}\,\int_0^1
        \sqrt{  g_{\gamma(t)}(\dot\gamma(t),\dot\gamma(t)) }
    dt
.
\]
To describe the (global) feature map we recall the notion of a \textit{systole} of $\xxx$, denoted by $\operatorname{Sys}(\xxx)$. The systole is defined as the length of the shortest non-contractible loop in $\xxx$.  That is,
\[
    \operatorname{Sys}(\xxx)
        \eqdef
    \inf \{ \operatorname{Length}(\gamma([0,1])) : \gamma:[0,1]\rightarrow \xxx,\, \gamma(0)=\gamma(1)\mbox{ and } \gamma \mbox{ non-contractible} \}
.
\]
We set $\delta \eqdef \frac{\operatorname{Sys}(\xxx)}{11}$ and let $\{x_l\}_{l=1}^L$ be any maximal $\delta$-separated subset of $\xxx$. For every $i,j=1,\dots,L$ it holds that $d_g(x_i,x_j)\ge \delta$ whenever $i\neq j$.  
As shown in the proof of \cite[Theorem 1.1]{katz2011bi}, the map $\phi:\xxx\rightarrow \rr^L$ defined by
\begin{equation}
\label{ex:FeatureMap_Gromov}
    \phi(x)
        \eqdef 
    \big(
        d_g(x,x_i)
    \big)_{i=1}^L
\end{equation}
is a bi-Lipschitz embedding of $\xxx$ into $\rr^L$. It  follows that $\phi$ is quasi-symmetric (see \cite[page 78]{heinonen2001lectures}). Hence, $\{(\xxx,\phi)\}$ is a feature decomposition of $(\xxx,d_g)$ satisfying conditions (i) and (ii) in Settings \ref{setting:theorems_Quantitative__CombinatorialX__BarycentricQAS__Y} and \ref{setting:theorems_Quantitative}. 

We often have more detailed information about the source and target metric spaces on which the non-linear operator $f:\xxx\rightarrow \yyy$ is defined.  We now provide an inverse-problem-theoretic example when this additional structure can be leveraged to construct a feature map.

\subsubsection{Feature Maps From Inverse Problems}
\label{s:Applications__ss:InverseProblemsPDE___sss:DirichletToNeumanMap}
	
	\def \beq {\begin {eqnarray}}
	\def \eeq {\end {eqnarray}}
	\def \ba {\begin {eqnarray*}}
	\def \ea {\end  {eqnarray*}}
	\def \R {{\mathbb R}}
	\def \p {{\partial}}
	\def \e {{\epsilon}}
	\def\bra{\langle}
\def\cet{\rangle}
\def\ket{\rangle}
    In a typical inverse problem, a parameter fuction varying inside a manifold with a boundary, often a spatially varying coefficient function in a partial differential equation, must be recovered from boundary  data. The modulus of continuity of the inverse map may be significantly worse than Lipschitz or H\"older \cite{Alessandrini_logStabilityConductivityBoundaryMeasurments_1988,Mandache_2001,MattisBook_InverseBoundarySpectralProblems__2001,BellassouedYamamoto_2006,RulandSalo__LogStability_NonlinearAnal__2020}. H\"older stability can sometimes be obtained by imposing additional assumptions~\cite{StefanovUhlmann_1998,StefanovUhlmann_StabilityXray_2004,covi2022global}. Theorem~\ref{lemma:Unstructured_Case} applies both to the weakly stable and to H\"older stable inverses, but Theorems~\ref{theorem:determinsitic_transferprinciple} and~\ref{theorem:Structured} give sharper, quantitative guarantees when the inverse is H\"older.

  We consider an example for a wave equation where the direct map is Lipschitz-stable and determining an unknown coefficient function is H\"older-stable. We will construct a feature map from the Dirichlet-to-Neumann map which corresponds to the measurements on the 
  boundary of an unknown body.


    Let us now formulate rigorously the inverse problem for a wave equation.
	Let $M\subset \R^n$, $n\ge 2$ be a simply connected bounded open set with $C^\infty$-smooth boundary, $H^{k,p}(M)$ be the Sobolev space of functions having $k$ weak derivatives in $L^p(M)$, see \cite{Lions} and \cite[Sec. 4.2]{Taylor1}. We denote $H^{k}(M)=H^{k,2}(M)$, $H^{k,p}_0(M)=\hbox{cl}_{H^{k,p}(M)}(C^\infty_0(M))$, and $H^{k}_0(M)=\hbox{cl}_{H^{k}(M)}(C^\infty_0(M))$. 
	In particular,  $H^{1}_0(M)=\{u\in H^1(M):\ u\vert _{\p M}=0\}$ and 
 $\|u\|_{H^{r}_0(M)}=\|u\|_{H^{r}(M)}$ for $u\in H^{r}_0(M)$. 
	
	Let $g=(g_{jk}(x))_{j,k=1}^n$ be a $C^\infty$-smooth Riemannian metric on $M \subset \mathbb R^n$. We assume that $(M,g)$ is a simple Riemannian manifold,
	that is, $M$ is simply connected, its boundary is strictly convex, and the geodesics of $(M,g)$  have no conjugate points.
	Moreover, assume that 
 \begin{equation}\label{eps 0 condition}
     |g_{jk}(x)-\delta_{jk}|\le \varepsilon_0,\quad \hbox{for $x\in M$, $j,k=1,2,\dots,n$}
 \end{equation}
 and
 $T> \hbox{diam}(M,g)$.
We	let $1<p<\infty$, $k\in  \mathbb{N}_+$ be large enough, $\epsilon_0>0$ be small enough, $R_0>0$, and 
	\beq\label{def xxx}
	\xxx=\bigg\{q:M\to \mathbb R\ \,\bigg|\ q \in H^{k+1,p}_0(M),\  \|q\|_{H^{k+1,p}(M)}\le R_0
	\bigg\}.
	\eeq
	We define in $\xxx$ the distances by
	$
	d_{\xxx}(q_1,q_2)=\|q_1-q_2\|_{H^{k,p}(M)}.
	$

	\begin{lemma}
	\label{lem:CompletenessofX}
	The metric space
	$(\xxx,d_{\xxx})$ is complete.
	\end{lemma}

Next, 
		we consider a classical inverse problem for the wave equation, that is the determination of the (lower order) coefficient function $q(x)$ from the boundary observations. Later
		we will construct a feature map $\varphi:\xxx\rightarrow F_{\varphi}$ related to these boundary observations.  We begin by considering the wave equation
	\begin{equation}
	\label{eq:WaveEquation}
	\begin{aligned}
	    & &(\p_{t}^2-\Delta_g+q)u(x,t)=0\quad\hbox{on }(x,t)\in  M \times \R_+,\\
	& & u(x,t)\vert _{\p  M\times \R_+}=f(x,t),\\
	& &u(x,t)\vert _{t=0}=0,\quad \p_tu(x,t)\vert _{t=0}=0,
	\end{aligned}
	\end{equation}
	where $\Delta_g $ is the Laplace operator associated to a Riemannian metric $g$,
	\ba
	& &\Delta_g u=\sum_{j,k=1}^n \vert g(x)\vert ^{-1/2}\frac \p{\p x^j}\bigg( \vert g(x)\vert ^{1/2}g^{jk}(x)\frac \p{\p x^k} u(x)\bigg),
	\ea
	where $\vert g\vert =\det (g_{jk}),$ $(g^{jk})=(g_{jk})^{-1}$.
	Function $f \in H^r_0(\p M \times (0,T))$ can be considered as a boundary source. 
	
 The Dirichlet-to-Neumann map  corresponding to the potential $q$ is defined to be the linear operator
	\ba
	\Lambda_q f=\p_\nu u_{q,f}(x,t)\big\vert _{(x,t)\in \p  M \times (0,T)},
	\ea
    where $u_{q,f}$ denotes the solution to the wave equation~\eqref{eq:WaveEquation}, $\nu$ is the unit outward normal to $\partial M$ and $\p_\nu u=\nu\cdot \nabla u$ is the normal derivative of function $u$.  The Dirichlet-to-Neumann map models all possible measurements that one can do on the boundary of $M$.
For $q\in \xxx$,   \cite[Sec. 2.8]{LaLiTr} or \cite{KKL}, Theorem 2.45 imply that the Dirichlet-to-Neumann map is a bounded map
	 		\beq\label{DN Map def.}
		\Lambda_{q}:H^s_0 (\p M\times (0,T))\to H^{s-1} (\p M\times (0,T)),\quad \hbox{for }1\le s\le k-\frac pn,
		\eeq
 
	We next construct the feature map and its associated feature space.	
 When $L(X,Y)$ is the  the Banach space of
	bounded linear maps $X\to Y$, we have  $\Lambda_{q}\in L(H^s_0 (\p M\times (0,T)), H^{s-1} (\p M\times (0,T)))$. Thus one could consider choosing the feature space be the Banach space of linear maps,
 $L(H^s_0 (\p M\times (0,T)), H^{s-1} (\p M\times (0,T)))$. However,
 the complicated structure of this Banach space makes it difficult to study learning theory if this space is used as the feature space. Our aim is to construct a feature map that takes values in a Hilbert space.
 To this end, we introduce some auxiliary operators. 
	Let
	\ba
	& &A: L^2(\p M\times (0,T))\to L^2(\p M\times (0,T)),\\
	& &Au=(I-\p_t^2-\Delta_{\p M})u,\quad u\in \mathcal D(A)=H^{2}(\p M\times (0,T))\cap H^{1}_0(\p M\times (0,T)),
	\ea
	be the unbounded selfadjoint operator with the domain $\mathcal D(A)$.
	Here, $\Delta_{\p M}$ is the Laplace-Beltrami
	operator of ${\p M}$ and $\p_t^2+\Delta_{\p M}:H^{2}(\p M\times (0,T))\cap H^{1}_0(\p M\times (0,T))\to L^2(\p M\times (0,T))$ is the Laplace operator on $\p M\times (0,T)$ defined with the initial and final conditions, $f|_{t=0}=0$ and  $f|_{t=T}=0$. 

For  an even integer $m\in\mathbb{N}_+$,  $m>\frac {n+1}2$,
let us consider the quadratic form 
\beq
Q(u,v)=\bra u,v\ket_{H^{m}_0(\p M\times (0,T))},\quad u,v\in \mathcal D(Q)=H^{m}_0(\p M\times (0,T)),
\eeq
where $\bra u,v\ket_{H^{m}_0(\p M\times (0,T))}$ is the inner product of $u$ and $v$ in $H^{m}_0(\p M\times (0,T))$. The quadratic form $Q$ defines a positive, closed, unbounded quadratic form in $L^2(\p M\times (0,T))$.
Further, the quadratic form $Q$ defines an unbounded selfadjoint operator $N$ in  $L^2(\p M\times (0,T))$ by the Friedrichs extension theorem, see  \cite[Ch.\ VI, Theorem 2.1 and section VI.3]{Kato}. When the inner product $\bra u,v\ket_{H^{m}_0(\p M\times (0,T))}$
is chosen in the suitable way (from the equivalent inner products in $H^{m}_0(\p M\times (0,T))$)
We see that the operator $N$ 
is the elliptic partial differential operator 
	\ba
	& &N:L^2(\p M\times (0,T))\to L^2(\p M\times (0,T)),\\
	& &Nu=(I-\p_t^2-\Delta_{\p M})^m u,\quad u\in \mathcal D(N),
    \ea
	that is an unbounded selfadjoint operator with the domain 
	\ba
\mathcal D(N)&=&H^{2m}(\p M\times (0,T))\cap H^{m}_0(\p M\times (0,T))\\
&=&
	\{u\in H^{2m}(\p M\times (0,T)):\ u|_{\p M\times \{0\}}=0,\ \p_t u|_{\p M\times \{0\}}=0,\dots ,
	\p_t^m u|_{\p M\times \{0\}}=0,\\ & &\hspace{35mm}  u|_{\p M\times \{T\}}=0,\ \p_t u|_{\p M\times \{T\}}=0,\dots ,
	\p_t^m u|_{\p M\times \{T\}}=0\}.
	\ea
By the Friedrichs second represenation theorem, see \cite[Ch.\ VI, Theorem 2.23]{Kato}), the square root $N^{1/2}$ of the selfadjoint operator $N$ is defined in the domain $\mathcal D(N^{1/2})=D(Q)=H^{m}_0(\p M\times (0,T))$.
Note that the inverse operator $N^{-1}:L^2(\p M\times (0,T))\to H^{2m}(\p M\times (0,T)) \cap H^{m}_0(\p M\times (0,T))$
and its square root $N^{-1/2}:L^2(\p M\times (0,T))\to  H^{m}_0(\p M\times (0,T))$ are bounded linear operators.

	
	By Weyl's asymptotics of eigenvalues of the Laplace operator $A$ (see \cite{Ivriuv__WeylAsymptotics_SmoothManifoldwithBoundary__1980}), 
	the eigenvalues of the unbounded self-adjoint operator $A$ in $L^2(\p M\times (0,T))$ have the asymptotics 
	$$
	\lambda_j = c_{T,M} j^{2/(n+1)}(1+o(1)).
	$$
%
%
%
%
	
		As $m>\frac {n+1}2$, the eigenvalues of $A^{-m/2}$ have the asymptotics 
	$$
	\lambda_j^{-m/2} = c_{T,M} j^{\,- \frac m2\cdot \frac 2{n+1}}(1+o(1)),
	$$
	and the sequence $(\lambda_j^{-m/2})_{j=1}^\infty$ is in $\ell^2$. Thus, when  $m>\frac {n+1}2$
	we have that   $A^{-m/2}$ is a Hilbert-Schmidt
	operator  in $L^2(\p M\times (0,T))$.

As $	A^{m/2}N^{-1/2}:L^2(\p M\times (0,T))\to L^2(\p M\times (0,T))$ is bounded and
the  Hilbert-Schmidt operators are an operator ideal,   the operator
	$N^{-1/2}=A^{-m/2}A^{m/2}N^{-1/2}:L^2(\p M\times (0,T))\to L^2(\p M\times (0,T))$  a Hilbert-Schmidt operator.
	
	Let now $2\le r\le m$ be an integer. 
	Below we consider bounded operators $B\in L(H^r_0(\p M\times (0,T)),L^2(\p M\times (0,T)))$. Then,
	 $BN^{-1/2}\in L(L^2(\p M\times (0,T)),L^2(\p M\times (0,T)))$ is bounded.Thus, $BN^{-1}=BN^{-1/2}N^{-1/2}$
	 is a Hilbert-Schmidt operator in $L^2(\p M\times (0,T))$. 
	Thus $BN^{-1}$ is an integral operator with the integral (Schwartz) kernel $k_{BN^{-1}}\in L^2((\p M\times (0,T))\times (\p M\times (0,T)))$.
	
	The feature space in this example is
	\begin{equation}
	\label{eq: feature-M}
	F_{\varphi}=HS(L^2(\p M\times (0,T)), L^2(\p M\times (0,T))),
	\end{equation}
	that is, the space of Hilbert-Schmidt operators $L^2(\p M\times (0,T))\to L^2(\p M\times (0,T)))$.
	We note that $HS(L^2(\p M\times (0,T)), L^2(\p M\times (0,T)))$ can be identified with the Hilbert space $L^2((\p M\times (0,T))\times (\p M\times (0,T)))$
	by identifying operators and their Schwartz kernels.

	To define the feature map for the inverse problem for the wave equation we consider 
	the composed map of the Dirichlet-to-Neumann operator $\Lambda_q$ and the smoothing operator $N^{-1}$, $\Lambda_qN^{-1}=\Lambda_q\circ N^{-1}:L^2 (\p M\times (0,T))\to L^2(\p M\times (0,T))$. We define the feature map to be the function 
 \beq
 \varphi:\xxx\to F_{\varphi},\quad 
 \varphi(q) = \Lambda_qN^{-1}.
	\eeq	
	\begin{proposition}[{A Lipschitz Dirichlet-to-Neumann Feature Map with H\"{o}lder Inverse}]\label{prop:DNmapisLipschitz} Let 
	$m>\frac {n+1}2$ be an even integer, $2\le r\le m$ be an integer, and 
	     {$k\ge r+\frac p n$}.
	The feature space $F_{\varphi}$, defined in \eqref{eq: feature-M}, has the BAP and the feature map $\varphi:\xxx\rightarrow F_{\varphi}$, $\varphi(q) = \Lambda_qN^{-1}$ satisfies
	\begin{enumerate}
	    \item[(i)] 
	    The map $
	{\varphi}:\xxx\to F_{\varphi}
	$ is Lipschitz;
 \item[(ii)] Assume that  the parameters defining the space $\xxx$ are such that $k \in \mathbb{N}_+$ is large enough, $p=2$, $\varepsilon_0 > 0$ is small enough and  $R_0 > 0$. 
  Then,
 the feature map $\varphi: \xxx\to F_{\varphi}$ is injective and it has a left inverse function which extends 
 to a  H\"{o}lder continuous map $\Psi:F_{\varphi}\to \xxx$.  
	\end{enumerate}
	\end{proposition}
	
We conclude our examples with finite-dimensional spaces having complicated topological structures.

\subsection{Target Spaces with Combinatorial Geometries: Closed Smooth Manifolds}
\label{s:Manifold_Targets}

For completeness, we show how our theory can be used to approximate continuous functions taking values in $\yyy$ which is a closed smooth submanifold of $\mathbb{R}^m$\footnote{More broadly, these derivations apply to any topological manifold admitting a triangulation.}. We endow $\yyy$ with a metric $d_{\yyy}$ compatible with its topology. In particular, we construct a $d_{\yyy}$ such that $(\yyy,d_{\yyy})$ satisfies Setting~\ref{setting:theorems_Quantitative__CombinatorialX__BarycentricQAS__Y}. This metric will allow us to nearly partition $\yyy$ in barycentric QAS spaces $\{\yyy_m\}_{m=1}^M$ whose number of parts $M$ on $\yyy$ is equal to the number of simplices associated with a certain simplicial complex of $\yyy$.


One example is the geometric realization of a manifold's triangulation.  For instance, if $\yyy$ is not only $C^2$ but smooth, classical results in geometric and algebraic topology \cite{SmoothmanifoldareTriangulatable1940Whitehead} confirm that $\yyy$ can be triangulated. In either case, we only require that the target topological manifold $\yyy$ be compact and that it admit a triangulation $K(\yyy)$, which we may assume is embedded as a subset of the Euclidean space $\mathbb{R}^{2 d+1}$ (\citep[Chapter 3, Theorem 9]{SpanierTop1995}), where we set $d\eqdef \dim(\yyy)$ and $k\eqdef d+1$.  By passing to a barycentric subdivision $bK(\yyy)$ of the triangulation $K(\yyy)$, we may ensure that each pair of distinct simplices $\triangle_1$ and $\triangle_2$ in the simplicial complex $bK(\yyy)$ are disjoint up to their boundaries.  Consider the geometric realization $\vert bK(\yyy) \vert$ of the simplicial complex $bK(\yyy)$, which we recall is given by the set of finitely supported probability measures $\sum_{i=1}^k\, w_i\delta_{y_i}$ on $\yyy$ such that  $y_1,\dots,y_k$ are distinct vertices of $bK(\yyy)$'s underlying set of vertices.  

We define a metric on $\yyy$ using the geometric realization $\vert bK(\yyy) \vert$ of the simplicial complex $bK(\yyy)$ as follows.  By identifying each subset $\Delta_{[y_1,\dots,y_k]}\eqdef \{\sum_{i=1}^k\, w_i\delta_{y_i}:\, w\in \Delta_k\}$ of $\vert bK(\yyy) \vert$ with the corresponding standard simplex $\Delta_k\eqdef \{w \in [0,1]^k:\, \sum_{i=1}^k\, w_i=1\}$ via the map $\sum_{i=1}^k\, w_i\delta_{y_i}\mapsto w$ we can safely pullback the Euclidean metric on $\Delta_k$ to $\Delta_{[y_1,\dots,y_k]}$; that is,
\[
d_{[y_1,\dots,y_k]}\biggl(\sum_{i=1}^k\, w_i\delta_{y_i}
,
\sum_{i=1}^k\, u_i\delta_{y_i}\biggr) \eqdef \|
w-u
\|_2.
\]
Since each $\Delta_{[y_1,\dots,y_k]}\subseteq \vert bK(\yyy) \vert$ is a geodesic space, then we may metrize $\vert bK(\yyy) \vert$ via the standard quotient metric obtained by gluing every pair of convex subsets $\Delta_{[y_1,\dots,y_k]}$ and $\Delta_{[y_1',\dots,y_k']}$ of $\vert bK(\yyy) \vert$ by identifying
\[
\sum_{i=1}^k\, w_i\delta_{y_i} \sim \sum_{i=1}^k\, u_i\delta_{y_i'}
\]
in the respective subsets $\{\sum_{i=1}^k\, w_i\delta_{y_i}:\, w\in \Delta_k; w_{j_1}=\dots=w_{j_l}=0\}$ and $\{\sum_{i=1}^k\, u_i\delta_{y_i'}:\, u \in \Delta_k; u_{j_1}=\dots=u_{j_l}=0\}$ whenever the set $\{y_1,\dots,y_k\}\setminus \{y_{j_1},\dots,y_{j_l}\}$ and $\{y_1',\dots,y_k'\}\setminus \{y_{j_1}',\dots,y_{j_l}'\}$ coincide and $w=u$, possibly up to a permutation of indices; see \citep[Theorem 3.1.27]{Burgao2IVanov_CourseMetricGeometry_2001} for a definition of this type of gluing construction.  We denote this metric on $\vert bK(\yyy)\vert $ by $d_{\vert bK(\yyy)\vert }$ and by \citep[Corollary 3.1.24]{Burgao2IVanov_CourseMetricGeometry_2001} we note that it makes $\big(\vert bK(\yyy)\vert ,d_{\vert bK(\yyy)\vert }\big)$ into a geodesic space.  
This metrizes the topology on the geometric realization $\vert bK(\yyy) \vert$ (see \citep[Chapter 3, Theorem 9]{SpanierTop1995}).  Since $K(\yyy)$ triangulates $\yyy$ then, by definition, there is a homeomorphism $h$ from $\yyy$ to $\vert bK(\yyy) \vert$ (see \citep[page 121]{SpanierTop1995}).  Pulling back $d_{\vert bK(\yyy)\vert }$ along the homeomorphism $h$ via
\[
        d_{\yyy}(y,\tilde{y})
    \eqdef 
        d_{\vert bK(\yyy)\vert }\big(h(y),h(\tilde{y})\big)
    ,
\]
properly metrizes $\yyy$'s topology.  By construction, $(\yyy,d_{\yyy})$ is isometric to $\big(\vert bK(\yyy)\vert,d_{\vert bK(\yyy)\vert}\big)$, wherefrom we conclude that $(\yyy,d_{\yyy})$ is also a geodesic space.  
Denote the barycenter $\frac1{k}\,\sum_{i=1}^k\, \delta_{y_k} \in \mathcal{P}_1(\mathbb{R}^{2d+1})$ of each simplex $\triangle_{[y_1,\dots,y_k]}$ in $bK(\yyy)$.  For any $0\le t <1$, define the associated ``retracted'' set 
\[
        \triangle_{[y_1,\dots,y_k]}^{t}
    \eqdef 
        \biggl\{\sum_{i=1}^k\, 
        \Big(
            t (w_i-\frac1{k}) + \frac1{k}
        \Big)\,\delta_{y_i}
        :\, w \in \Delta_k\biggr\}
. 
\]
Because we have passed to a barycentric subdivision of the triangulation $K(\yyy)$ of $\yyy$, any two sets $\triangle_{[y_1,\dots,y_k]}^{t}$ and $\triangle_{[y_1',\dots,y_k']}^{t}$, $0 \le t <1$, are disjoint whenever the sets $\{y_i\}_{i=1}^k$ and $\{y_i'\}_{i=1}^k$ do not coincide.  Setting $\{\yyy_m\}_{m\le M}$ to be the pre-images $h^{-1}(\triangle_{[y_1,\dots,y_k]})$ in $\yyy$ of the geodesic triangles in $\vert bK(\yyy) \vert$ with vertices belonging to the vertex set of $bK(\yyy)$, the above implies that the collection $\{\yyy_m\}_{m\le M}$ satisfies Definition~\ref{defn:CombinatorialStructure_Y} (i) and (ii).   

As in Remark~\ref{remark:differentQAS_Structures}, we do not need to define $\eta$ globally, but we can instead only define it on each $\yyy_m\eqdef h^{-1}(\Delta_{[y_1,\dots,y_k]})$, where $M$ is the number of such geodesic triangles (where geodesics are given under the glued metric).  Here, each $\eta^m$ is given by
\[
        \eta^m(w,(u_n)_{n=1}^N)
    \eqdef 
        h^{-1}\Big(
            \sum_{n=1}^N\,
                w_n h(u_n)
        \Big)
    ,
\]
where $N\in \mathbb{N}_+$, $w\in \Delta_N$, and $u_1,\dots,u_N\in \yyy_m$; for $m=1,\dots,M$. 
It remains to quantize $(\yyy,d_{\yyy})$ in order to conclude that $\big(\{\yyy_m\}_{m\le M},(Q_q)_{q\in \mathbb{N}_+}, \eta^m\big)$ is a geodesic partition of $\yyy$.  For every $q\in \mathbb{N}_+$ and set $D_q\eqdef d+1$, we define the quantization maps $Q_q\eqdef Q_1$ by
\[
    Q_1(z_1,\dots,z_{d+1}) \eqdef 
    h^{-1}\big(
        \sum_{i=1}^{d+1}\,
            [P_{\Delta_{d+1}}(z_1,\dots,z_{d+1})]_i\, y_i^{(m)}
    \big),
\]
where $\{y_i^{(m)}\}_{i=1}^{d+1}$ is the set of vertices defining the $m^{th}$ simplex in $\vert bK(\yyy)\vert$.

\section{Proof of Main Results}
\label{s:Proofs}
This section is organized as follows.  First, we present several technical lemmata in Section~\ref{s:Proofs__ss:technical_lemmata}, on which the other results are built.  
Section~\ref{s:Proofs__ss:qualitative} contains the proof of our main qualitative approximation results.  
Section~\ref{s:Proofs__ss:quantitative} is devoted to the proof of our quantitative approximation results.  
\subsection{Technical Lemmata}
\label{s:Proofs__ss:technical_lemmata}
	\begin{lemma}[Embedding of the $1$-Wasserstein Space into the Lipschitz-Free Space]
	\label{lemma_closedconvex_embedding_Wasserstein}
	Fix an arbitrary base-point $y_0$ in a Polish metric space $\yyy$.  
	Let $\left({\text{\AE}}(\yyy,d_{\yyy},y_0),\|\cdot\|_{{\text{\AE}}}\right)$ denote the Lipschitz-free Banach space over the pointed metric space $(\yyy,d_{\yyy},y_0)$; see \cite{GodefroyandKalton2003} or \cite{WeaverLipschitzAlgebras_2018} for a an overview of the topic.  Consider the isometric embedding
	\[
	\begin{aligned}
	\Phi: 
	(\mathcal{P}_1(\yyy),W_1) 
	& \rightarrow
	\left({\text{\AE}}(\yyy,d_{\yyy},y_0),\|\cdot\|_{{\text{\AE}}}\right)
	\\
	\pp & \mapsto 
	\left[
	g \mapsto \int_{y \in \yyy} \, g(y)\, \pp(dy) - g(y_0)
	\right]
	.
	\end{aligned}
	\]
	$\Phi\left(\mathcal{P}_1(\yyy)\right)$ is a separable, closed and convex subset of $\left({\text{\AE}}(\yyy,d_{\yyy},y_0),\|\cdot\|_{{\text{\AE}}}\right)$.  
	\end{lemma}
	\begin{proof}[Proof of Lemma~\ref{lemma_closedconvex_embedding_Wasserstein}]
	By linearity of integration, the image of $(\mathcal{P}_1(\yyy),W_1)$ under $\Phi$ is convex.  Since $(\mathcal{Y},d_{\yyy})$ is complete and separable, then so is $(\mathcal{P}_1(\yyy),W_1)$; hence, $\Phi\left(\mathcal{P}_1(\yyy)\right)$ is a separable, closed and convex subset of $\left({\text{\AE}}(\yyy,d_{\yyy},y_0),\|\cdot\|_{{\text{\AE}}}\right)$.  
	\end{proof}

    The following Lemma is a quantitative version of Lemma~\ref{lem_snowflake}.  It is also a variant of \citep[Theorem 10.18]{heinonen2001lectures} with explicit constants and additionally ensures the generalized snowflake isn't only a quasisymmetric map but that $d^{\omega}$ also defines a genuine metric.  
    \begin{lemma}[Generalized Snowflakes are Quasisymmetric to Their Original Space]
    \label{lem_snowflake__quantitative}
    Let $\omega$ be a H\"{o}lder-like modulus and $(\xxx,d_{\xxx})$ be a metric space.  Then, $d_{\xxx}^{\omega}\eqdef \omega \circ d_{\xxx}$ defines a metric and the map $(\xxx,d_{\xxx})\ni x \mapsto x\in (\xxx,d_{\xxx}^{\omega})$ is quasisymmetric.  Furthermore, if $K\Subset \xxx$ has doubling constant $C_{(K,d_{\xxx})}$ for the metric $d_{\xxx}$, then $K$ has doubling constant
    $
    C_{(K,d_{\xxx})}^{
    \lceil
    -\log_2(h_{\omega}^{\dagger}(\frac{1}{4}))/4
    \rceil
    }
    $
    under the snowflaked metric $d_{\xxx}^{\omega}$.  
    \end{lemma}
    \begin{proof}[Proof of Lemma~\ref{lem_snowflake__quantitative}]
    Because $\omega$ was assumed to be sub-additive,$d_{\xxx}^{\omega}$ satisfies the triangle inequality, and since $\omega(0)=0$ and $\omega$ is injective (since it is strictly increasing), $d_{\xxx}^{\omega}(x_1,x_2)=0=d_{\xxx}(x_1,x_2)$ if and only if $x_1=x_2$. Hence, $d_{\xxx}^{\omega}$ is a metric.
    Now, let $x_1,x_2,x_3\in K$ and $t>0$ be such that $d_{\xxx}(x_1,x_2) \le t d_{\xxx}(x_1,x_3)$.  Since $\omega$ is strictly increasing,
    \[
    d_{\xxx}^{\omega}(x_1,x_2) 
        \eqdef 
    \omega\Big(
    d_{\xxx}(x_1,x_2)
    \Big)
    \le 
    \omega\Big(
    t\,d_{\xxx}(x_1,x_3)
    \Big)
    \le 
    h_{\omega}(t)
    \omega(d_{\xxx}(x_1,x_3))
    \eqdef 
    \eta(t) d_{\xxx}^{\omega}(x_1,x_3)
    ,
    \]
    where $\eta\eqdef h_{\omega}$. But then the map $(\xxx,d_{\xxx})\ni x \mapsto x\in (\xxx,d_{\xxx}^{\omega})$ is quasisymmetric with $\eta=h_{\omega}$.  
    \hfill\\
    The last claim follows from Lemma~\ref{lemma_quantiative_doublinginvariance_of_quasisymmetricmaps}.  
    \end{proof}
    
	\begin{lemma}[Extension]
	\label{lemma_extension_Lemma}
    	Let $K$ be a closed doubling subset of a metric space $(\xxx,d_{\xxx})$, with doubling constant $C_K>0$, and let $(\yyy,d_{\yyy})$ be a barycentric Polish metric space. Let $C^{\omega-\text{H\"ol}}((K,d_{\xxx}), (\yyy,d_{\yyy}))$ denote the space of all $\omega$--H\"older--like functions from $(K,d_{\xxx})$ to $(\yyy,d_{\yyy})$. There exists an extension operator
    	\[
    	\mathcal{E}:\, 
    	C^{\omega-\text{H\"ol}}((K,d_{\xxx}), (\yyy,d_{\yyy}))
    	\rightarrow
    	C\big(
    	(\xxx,d_{\xxx}),(\yyy,d_{\yyy})
    	\big)
    	\]
    	such that, for each $f\in C^{\omega-\text{H\"ol}}((K,d_{\xxx}), (\yyy,d_{\yyy}))$ the map $\mathcal{E}(f) \in C\big((\xxx,d_{\xxx}),(\yyy,d_{\yyy})\big)$ has modulus of continuity (with respect to $d_{\xxx}$)
    	\[
        	    \omega_{\mathcal{E}(f)} 
        	=
                \operatorname{Lip}(f, d_{\xxx}^\omega)
                c \log(C_{K,d_{\xxx}^{\omega}})\omega (\cdot)
    	,
    	\]
    	where $C_{K,d_{\xxx}^{\omega}}= C_{(K,d_{\xxx})}^{
                \lceil
                -\log_2(h_{\omega}^{\dagger}(\frac{1}{4}))/4
                \rceil
                } >0
        $ is the doubling constant of $K$ under $d_{\xxx}^{\omega}$ and $c>0$ is an absolute constant.  
	\end{lemma}
	
	\begin{proof}[Proof of Lemma~\ref{lemma_extension_Lemma}]
	Since $\yyy$ is barycentric then, there exists a $1$-Lipschitz barycenter map $\beta_{\yyy}:\mathcal{P}_1(\yyy)\rightarrow \yyy$ satisfying
	\[
    	\beta_{\yyy}(\delta_y) = y
    	,
	\]  
	for all $y\in \yyy$.  
	By Lemma~\ref{lem_snowflake}, the map $(\xxx,d_{\xxx})\ni x \mapsto x \in (\xxx,d_{\xxx}^{\omega})$ is quasisymmetric.  Since $K$ is doubling then by \cite[Theorem 10.18]{heinonen2001lectures} $K$ is a doubling subset of $(\xxx,d_{\xxx}^{\omega})$.  
   Let $f:(K,d_{\xxx}) \rightarrow (\yyy,d_{\yyy})$ be a (generalized) $\omega$--H\"older function so that it is a Lipschitz function on $(K,d_{\xxx}^\omega)$.
   
Since $K$ is closed, \cite[Theorem 3.2]{brue2021linear} implies that there exists a random 
    projection $\Pi \in \operatorname{Lip}((\xxx,d_{\xxx}^\omega),\mathcal{P}_1(K))$
    (that is, $\Pi_x=\delta_x$ whenever $x\in K$) with Lipschitz constant $\operatorname{Lip}(\Pi) \le c\log(C_{K,d_{\xxx}^{\omega}})$, where $c>0$ is an absolute constant and $C_{K,d_{\xxx}^{\omega}}$ is the doubling constant of $K$ with respect to the (generalized) snowflaked metric $d_{\xxx}^{\omega}$. (This constant is explicitly computed in Lemma~\ref{lem_snowflake__quantitative}).  Fix an arbitrary $y_0\in \yyy$ and let $\Phi$ be the isometric embedding of Lemma~\ref{lemma_closedconvex_embedding_Wasserstein}.  
    In a similar spirit to \cite[Theorem 2.4]{ambrosio2019linear} and \cite[Remark 3.3]{brue2021linear}, we define the extension map
    \[
    \begin{aligned}
    \mathcal{E}:\operatorname{Lip}((K,d_{\xxx}^\omega),\yyy) & \rightarrow \operatorname{Lip}((\xxx,d_{\xxx}^\omega),\yyy)\\
    f & \mapsto 
        \left[
            x
        \mapsto
            \beta_{\yyy}\Biggl( 
                    \Phi\vert_{\Phi\big(\mathcal{P}_1(\yyy)\big)}^{-1}
                    \biggl(
                        \int_{u \in K}\,
                            \Phi\circ \delta_{f(u)}
                        \Pi_x(du)
                \biggr)
            \Biggr)
        \right]
        ,
    \end{aligned}
    \]
    where $\int \cdot \Pi_x$ denotes the Bochner integral on the Lipschitz-free space $(\mbox{{\text{\AE}}}(\yyy,d_{\yyy},y_0),\|\cdot\|_{\mbox{{\text{\AE}}}})$ with respect to the measure $\Pi_x$.  
    First, we note that $\mathcal{E}$ is well-defined as $\Phi(\mathcal{P}_1(\yyy))$ is (non-empty) closed and convex by Lemma~\ref{lemma_closedconvex_embedding_Wasserstein} (noting that $\yyy$ is Polish); hence, for every $x \in \xxx$, the quantity $\int_{u \in K}\,\Phi\circ \delta_{f(u)}\Pi_x(du)$ belongs to the image of $\Phi(\mathcal{P}_1(\yyy))$ in $({\text{\AE}}(\yyy,d_{\yyy},y_0),\|\cdot\|_{{\text{\AE}}})$ for some $y_0 = f(u_0), u_0 \in K$. Therefore, as $\Phi$ is injective, $\Phi$ is invertible on its image and $\beta_{\yyy}$ is defined on $\mathcal{P}_1(\yyy)$.  
    
    Next, since $\Pi$ is a random projection then, for every $x\in K$, we have $\Pi_x=\delta_x$. Thus,
    \[
    \begin{aligned}
    \mathcal{E}(f)(x) 
        \eqdef & 
        \beta_{\yyy}\Biggl( 
                \Phi\vert_{\Phi\big(\mathcal{P}_1(\yyy)\big)}^{-1}
                \biggl(
                    \int_{u \in K}\,
                        \Phi\circ \delta_{f(u)}
                    \Pi_x(du)
            \biggr)
        \Biggr)
    \\
    = &
    \beta_{\yyy}\Biggl( 
                \Phi\vert_{\Phi\big(\mathcal{P}_1(\yyy)\big)}^{-1}
                \biggl(
                    \int_{u \in K}\,
                        \Phi\circ \delta_{f(u)}
                    \delta_x(du)
            \biggr)
        \Biggr)
    \\
    = & 
        \beta_{\yyy}\biggl( 
                \Phi\vert_{\Phi\big(\mathcal{P}_1(\yyy)\big)}^{-1}
                (\Phi\circ \delta_{f(x)})
        \biggr)
    \\[0.25cm]
    = & f(x)
    .
    \end{aligned}
    \]
    Then, for every $x\in K$ we have that $\mathcal{E}(f)(x)=f(x)$.  
    
    It remains to examine $\mathcal{E}(f)$'s regularity.  For any $x,\tilde{x}\in \xxx$, we have
    \begin{align}
    \nonumber
    d_{\yyy}(\mathcal{E}(f)(x),\mathcal{E}(f)(\tilde{x}))
        \le & 
        \omega_{\beta}\circ\omega_{\Phi\vert _{\Phi\big(\mathcal{P}_1(\yyy)\big)}^{-1}}
        \left(\sup_{g \in \operatorname{Lip}(\yyy): \|g\|_{\operatorname{Lip}} = 1}
        \int_{u \in K}\,
        g(f(u))
        \, (\Pi_x-\Pi_{\tilde{x}})(du)
        \right)
    \\
    \label{PROOF_lemma_extension_Lemma__KRduality}
    \le &
    \operatorname{Lip}(f,d_{\xxx}^\omega)W_1(\Pi_x,\Pi_{\tilde{x}})
    \\
    \label{PROOF_lemma_extension_Lemma__Rand_Proj}
    \le &
    \operatorname{Lip}(f,d_{\xxx}^\omega)
    c \log(C_{K,d_{\xxx}^{\omega}})
    \big(
        d_{\xxx}^{\omega}(x,\tilde{x})
    \big),
    \end{align}
   where we have used the fact that ${\text{\AE}}(\yyy,d_{\yyy},y_0)$ is the predual of $\operatorname{Lip}_{y_0}(\yyy)$\footnote{$\operatorname{Lip}_{y_0}(\yyy)$ is the space of Lipschitz functions on $\yyy$ which vanish at $y_0$.} in the first inequality and the Kantorovich--Rubinstein duality for the Wasserstein metric in the second inequality.
    Applying the definition of the $\omega$-snowflake, we find that 
    $\mathcal{E}(f)$ has modulus of continuity 
    $
    \operatorname{Lip}(f,d_\xxx^\omega)
    c \log(C_{K,d_{\xxx}^{\omega}})\omega (\cdot).$  
    Appealing to Lemma~\ref{lem_snowflake__quantitative}, we obtain $$
        C_{K,d_{\xxx}^{\omega}}
    =
        C_{(K,d_{\xxx})}^{\lceil
        -\log_2(h_{\omega}^{\dagger}(\frac{1}{4}))/4
        \rceil
        },
    $$ which yields the result.  
	\end{proof}
	
    Using the definition of a feature map, we use Lemma~\ref{lemma_extension_Lemma} to factor any function defined on a compact doubling subset $K$ of $(\xxx,d_{\xxx})$ through a feature space $F_{\phi}$.  The commutative diagram in Figure~\ref{fig:ExtensionFactorization_Lemma} summarizes the next lemma.  
    
    \begin{figure}[H]
        \centering
        \begin{tikzcd}
            (K,d_{\xxx}) \arrow[d,"\phi"] \arrow[r, "f"] & (\yyy,d_{\yyy}) \\
            (F_{\phi},\|\cdot\|_{F_{\phi}}) \arrow[ur, "f_{\phi}"] &  
        \end{tikzcd}
    \caption{Extension-Factorization Lemma}
    \label{fig:ExtensionFactorization_Lemma}
    \end{figure}

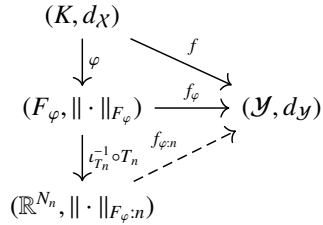
\begin{figure}[H]
            \centering
            \begin{tikzcd}
                (K,d_{\xxx}) \arrow[d,"\phi"] \arrow[rd, "f"] & \\
                (F_{\phi},\|\cdot\|_{F_{\phi}}) \arrow[r, "f_{\phi}"] \arrow[d, "\iota_{T_n}^{-1}\circ T_n"] & (\yyy,d_{\yyy})   \\
                (\rr^{N_n},\|\cdot\|_{F_{\phi}:n}) \arrow[ur, dashed, "f_{\phi:n}"] &  \\
            \end{tikzcd}
            \caption{Approximate Extension-Factorization Lemma}
            \label{fig:ApproximateExtensionFactorization_Lemma}
        \end{figure}

	\begin{lemma}[Extension-Factorization]
	\label{lemma_ExtensionFactorization}
	In the setting of Lemma~\ref{lemma_extension_Lemma}, let $\omega$ be a H\"{o}lder-like modulus, and either suppose that:
	\begin{enumerate}
	    \item[(i)] $\phi:(\xxx,d_{\xxx})\hookrightarrow (F_{\phi},\|\cdot\|_{F_{\phi}})$ be a quasisymmetric map into a separable Banach space $F_{\phi}$ and for every compact subset $K\subseteq \xxx$ the restriction of $\phi^{-1}$ to $\phi(K)$ has a H\"{o}lder-like modulus of continuity $\omega_{\phi\vert _{\phi(K)}^{-1}}$,
	    \item[(ii)] $\phi(K)$ is a doubling subset of $F_{\phi}$ (esp. if $K$ is finite) and the restriction of $\phi^{-1}$ to $\phi(K)$ has a H\"{o}lder-like modulus of continuity $\omega_{\phi\vert _{\phi(K)}^{-1}}$.
	\end{enumerate}
	Then, for every compact doubling set $K$ and every $f:K\rightarrow \yyy$ with H\"{o}lder-like modulus of continuity $\omega$, there exists a constant $C_{(K,\omega,\phi^{-1},F_{\phi})}>0$ (depending only on $K$, $\omega$, $\phi^{-1}$, and on $F_{\phi}$) and a map $f_{\phi}:F_{\phi}\rightarrow \yyy$ satisfying
	\begin{enumerate}
	\item[(1)] \textbf{Factorization:} $
	    f_{\phi}\circ \phi(x) = f(x)
	$ for all $x\in K$,
	\item[(2)] \textbf{Stability:} $
	    f_{\phi}
	$ has the following H\"{o}lder-like modulus of continuity:
	\[
	    C_{(K,\omega,\phi^{-1},F_{\phi})}
	    \omega
	        \circ 
	    \omega_{\phi\vert _{\phi(K)}^{-1}}
	.
	\]
	\end{enumerate}
	\end{lemma}
	\begin{proof}[{Proof of Lemma~\ref{lemma_ExtensionFactorization}}]
	Since $\phi$ is injective,  $\phi^{-1}$ exists on $\phi(\xxx)$. Because $K$ is compact and $\phi$ is continuous, $\phi(K)$ is itself compact; as a consequence, $\omega_{\phi\vert _{\phi(K)}^{-1}}$ is uniformly continuous on $K$. We denote its modulus of continuity thereon by $\omega_{\phi\vert _{\phi(K)}^{-1}}$. Then the map $f\circ \phi^{-1}:\phi(K)\rightarrow \yyy$ is well-defined and uniformly continuous, with modulus of continuity $
	\omega \circ \omega_{\phi\vert _{\phi(K)}^{-1}}
	.
	$  
	
	\hfill\\
	\textbf{If Condition (i) holds}: Since $\phi$ is a quasisymmetric map and since $K$ is doubling then, \cite[Theorem 10.18]{heinonen2001lectures} implies that $\phi(K)$ is doubling.  Thus the conditions of Lemma~\ref{lemma_extension_Lemma} are met and there is a uniformly continuous extension, $\mathcal{E}(f\circ \phi\vert _{\phi(K)}^{-1}):F_{\phi}\rightarrow \yyy$, of $f$ from $\phi(K)$ to all of $F_{\phi}$ with modulus of continuity
	\begin{equation}
	\label{PROOF_eq__lemma_ExtensionFactorization___modulus_of_continuity}
        	\omega_{
        	   \mathcal{E}(f\circ \phi\vert _{\phi(K)}^{-1})
        	}
	    =
            c \log(
    	        C_{(K,\omega\circ \omega_{\phi\vert _{\phi(K)}^{-1}} \circ \|\cdot\|_{F_{\phi}})}
    	    )
    	    \omega\circ \omega_{\phi\vert_{\phi(K)}^{-1}}(t)
	 .
	\end{equation}
	We set $f_{\phi}\eqdef \mathcal{E}(f\circ \phi\vert _{\phi(K)}^{-1})$ and set $
	C_{(K,\omega,\phi^{-1},F_{\phi})}
	    \eqdef 
	c \log\big(
	    C_{
	        (K,\omega\circ \omega_{\phi\vert _{\phi(K)}^{-1}} \circ \|\cdot\|_{F_{\phi}})
	   }
    	    \big)
	$.
	\hfill\\
	\textbf{If Condition (ii) holds}: Then $\phi(K)$ is doubling. We argue precisely as in case (i).  
    In either case, by construction. we have that for every $x\in K$ since $\phi(x)\in \phi(K)$ then
    \[
    f_{\phi}
    \circ \phi(x) = f\circ \phi\vert_{\phi(K)}^{-1}\circ \phi(x) = f(x).
    \]
	\end{proof}
	The next lemma is illustrated in Figure~\ref{fig:ApproximateExtensionFactorization_Lemma} (where the dashed line represents $\epsilon$-approximate commutation). The result is an approximate version of Lemma~\ref{lemma_ExtensionFactorization} where the maps $\phi$ and $f_{\phi}$ are approximated by passing through the finite-dimensional subspace of $F_{\phi}$ defined in~\eqref{eq:finitedimensionalcopyBanach}.

	\begin{lemma}[Approximate Extension-Factorization]
	\label{lemma_Approximate_ExtensionFactorization}
	Let $K\subseteq \xxx$ be non-empty and compact, $\phi:\xxx\rightarrow F_{\phi}$ be a feature map, $(T_n)_{n=1}^{\infty}$ be a sequence of bounded linear operators on $F_{\phi}$ implementing the $C_{BAP:T_\cdot}$-BAP on $\phi(K)$, and
	assume that any of conditions (i)-(ii) in Lemma~\ref{lemma_ExtensionFactorization} holds.  
	For every $\omega$-H\"{o}lder-like map $f:\xxx\rightarrow \yyy$ and each ``encoding error'' $\epsilon_E>0$, there is an $n_{\epsilon}\in \mathbb{N}$, a map $\phi_{\epsilon_E}:(\xxx,d_{\xxx})
	    \rightarrow 
	(
	    \rr^{N_{n_{\epsilon_E}}}
	  ,
	    \|\cdot\|_{F_{\phi}:n_{\epsilon_E}}
	)$, and a uniformly continuous map
	$
	f_{\phi:n_{\epsilon_E}}: 
	(\rr^{N_{n_{\epsilon_E}}},\|\cdot\|_{F_{\phi}:n_{\epsilon_E}})
	\rightarrow 
	(\yyy,d_{\yyy})
	$
    :
	\begin{enumerate}
	\item[(i)] \textbf{Approximation:} $
	\sup_{x\in K}\,
	    d_{\yyy}\left(
	        f(x)
	    ,
	        f_{\phi:n_{\epsilon_E}}\circ \phi_{\epsilon_E}(x)
	    \right)
	        \le 
	    \epsilon_E 
	,
	$
	\item[(ii)] \textbf{Regularity:} The map $
	\omega_{f_{\phi:n_{\epsilon_E}}}(t)
	    \eqdef 
    C_{(\omega,K,\phi^{-1},F_{\phi})}\,
    \omega
        \circ
    \omega_{\phi\vert _{\phi(K)}^{-1}}\big(
        C_{\operatorname{BAP}:T_{\cdot}}
        t
    \big)
	$ is a modulus of continuity for $f_{\phi:\epsilon_E}$,
    \item[(iii)] \textbf{Approximate Feature-Map Representation:} $\phi_{\epsilon_E}
	\eqdef 
    	\iota_{
    	    T_{
    	        n_{\epsilon_E}
    	    }
    	}^{-1}
	\circ 
    	T_{
    	    {n_{\epsilon_E}}
    	}
	\circ 
	    \phi
	.$
	\end{enumerate}
	Moreover, the integer $n_{\epsilon_E}$ is given by 
	\[
	    n_{\epsilon_E} 
	        \eqdef 
	    \min\Big\{
        n\in \nn_+:\, 
            \max_{u\in \phi(K)}\,
            \big\|
                u
                    -
                T_{n}(u)
            \big\|_{F_\phi}
                \le 
            \Big(
	        \omega^{\dagger}_{
	            \phi\vert _{\phi(K)}^{-1}
	        }
	            \circ
	        \omega^{\dagger}
	        \Big(
	        \frac{
	            \epsilon_E
	        }{
	            C_{(K,\omega,\phi^{-1},F_{\phi})}
	        }
	        \Big)
	    \Big)
        \Big\}
	\]
	where $
        C_{
            (K,f,\phi^{-1},F_{\phi},T_{\cdot})
        }
            \eqdef
    	C_{
    	    (K,f,\phi^{-1},F_{\phi})
    	   }
>0
	$ and $C_{(K,f,\phi^{-1},F_{\phi})}>0$ is defined in Lemma~\ref{lemma_ExtensionFactorization}, it depends only on $K,f,\phi^{-1},F_{\phi}$, and on $(T_n)_{n=1}^{\infty}$ and on which of the conditions Lemma~\ref{lemma_ExtensionFactorization} (i) or (ii) hold.  
    Furthermore, $N_{n_{\epsilon_E}}\in \nn_+$ denotes the rank of $T_{n_{\epsilon_E}}$.
	\end{lemma}
	\begin{proof}[Proof of Lemma~\ref{lemma_Approximate_ExtensionFactorization}]
	By Lemma~\ref{lemma_ExtensionFactorization}, there is a map $f_{\phi}:(F_{\phi},\|\cdot\|_{F_{\phi}})\rightarrow (\yyy,d_{\yyy})$ satisfying $f_{\phi}\circ \phi(x)=f(x)$ for every $x\in K$.  For every $n\in \nn_+$ define the map $f_{\phi:n}\eqdef f_{\phi}\circ \iota_{T_n}:(\rr^{N_n},\|\cdot\|_{F_{\phi}:n})\rightarrow (\yyy,d_{\yyy})$ and define the map $\phi^{(n)}:\iota_{T_n}^{-1}\circ T_n\circ \phi:(K,d_{\xxx})\rightarrow (\rr^{N_n},\|\cdot\|_{F_{\phi}:n})$.  

	Since the operators $(T_n)_{n=1}^{\infty}$ realize the $C_{BAP:T_\cdot}$-BAP on $\phi(K)$ and since $\phi$ is injective, we have that for every $x \in K$,
	\[
	\begin{aligned}
    	d_{\yyy}\Big(
    	    f(x)
    	,
    	    f_{\phi:n}\circ \phi^{(n)}(x)
    	\Big)
    &=
    	d_{\yyy}\Big(
    	    f_{\phi}\circ \phi(x)
    	,
    	    f_{\phi:n}\circ \phi^{(n)}(x)
    	\Big)
    \\
    &=
    	d_{\yyy}\Big(
    	    f_{\phi}\circ \phi(x)
    	,
    	    \big(
    	        f_{\phi}\circ \iota_{T_n}
    	    \big)
    	        \circ 
    	    \big(
    	        \iota_{T_n}^{-1}\circ T_n \circ \phi(x)
    	    \big)
    	\Big)    
    \\
    &= 
    	d_{\yyy}\Big(
    	   f_{\phi}\circ \phi(x)
    	,
    	   f_{\phi}\circ T_n \circ \phi(x)
    	\Big)
    \\
    &\le
    \omega_{f_{\phi}}\left(
        \big\|
            \phi(x)
                -
            T_n\circ \phi(x)
        \big\|_{F_{\phi}}
    \right)
    \\
    &=
        C_{(K,\omega,\phi^{-1},F_{\phi})}
        \omega
            \circ 
        \omega_{\phi\vert_{\phi(K)}^{-1}}\left(
            \big\|
                \phi(x)
                    -
                T_n\circ \phi(x)
            \big\|_{F_{\phi}}
        \right)
    .
	\end{aligned}
	\]
	By the monotonicity of the modulus of continuity $\omega_{f_{\phi}}$, we deduce that
	\[
	\begin{aligned}
        	d_{\yyy}\Big(
        	    f(x)
        	,
        	    f_{\phi:n}\circ \phi^{(n)}(x)
        	\Big)
        \leq &
        	\sup_{u\in \phi(K)}\,
                C_{(K,\omega,\phi^{-1},F_{\phi})}
                \omega
                    \circ 
                \omega_{\phi\vert _{\phi(K)}^{-1}}\big(
                    \|
                        u
                            -
                        T_n(u)
                    \big\|_{F_{\phi}}
                \big)
	.
	\end{aligned}
	\]
	Setting $
	    n_{\epsilon_E}
	\eqdef 
	   	R^{T_{\cdot}:\phi(K)}\Big(\omega_{f_{\phi}}^{-1}\big(
	   	        \epsilon_E
	   	\big)\Big)
	$, $\phi_{\epsilon_E}\eqdef \phi^{(n_{\epsilon_E})}$, where $\omega_{f_\phi} =  C_{(K,\omega,\phi^{-1},F_{\phi})}
                \omega
                    \circ 
                \omega_{\phi\vert _{\phi(K)}^{-1}}$, and noting that $\operatorname{Lip}(T_{n_{\epsilon}})=\|T_{n_{\epsilon}}\|_{op}\le C_{BAP:F_{\phi}}$, yields the conclusion. 
	\end{proof}
    
	To obtain the ``local version'' of our main result, it only remains to approximate the maps $f_{\phi:\epsilon}$ constructed in Lemma~\ref{lemma_Approximate_ExtensionFactorization}.  However, since all norms on $\rr^n$ are equivalent, under the assumption that $(\yyy,d_{\yyy})$ is a QAS space, we only need a mild variant of the main ``static result'' of \cite[Theorem 3.6]{AcciaioKratsiosPammer2022}. We now present this variant, adapted to our context.  We note that if $A$ is a subset of a Euclidean space, then $A$ is doubling; see \cite[Section 10.13]{heinonen2001lectures} or \cite[Theorem 12.1]{heinonen2001lectures} for details).
	
	\begin{lemma}[{Approximation of Functions on Finite-Dimensional Banach Space if $\yyy$ is a Barycentric QAS Space}]
	\label{lemma_metric_transformers_generalization}
	Let $K$ be a non-empty compact subset of the finite-dimensional Banach space $(\rr^n,\|\cdot\|_{F_{\phi}:n})$ and let $C_{(K,\|\cdot\|_{F_{\phi}:n})} \ge 0$ be $K$'s doubling constant, $(\yyy,d_{\yyy})$ admit a quantized mixing function $\hat{\eta}$, $\mathcal{F}_{\cdot}$ be a universal approximator, and $f:(K, \|\cdot\|_{F_\phi:n}) \rightarrow (\yyy, d_\yyy)$ be an $\omega$-H\"{o}lder-like function.  For each ``quantization error'' $\epsilon_Q>0$  and each ``approximation error'' $\epsilon_A>0$, there exists: $c,N,q\in \nn_+$, an $\hat{f}_{n,N} \in \mathcal{F}_{n,N,c}$, and $
	    Z = (Z_1, \ldots, Z_N)\in \mathbb{R}^{
	        N\times D_q
	    }
	$ such that
	\[
    	\sup_{x\in K}
    	\,
    	d_{\yyy}\Big(
        	    f(x)
        	,
            	\hat{\eta}\big(
                	        \hat{f}_{n,N}(x)
                	,
                	    \big(
                    	    Z_n
                    	 \big)_{n=1}^N
                	\big)
    	\Big)
	\le 
	    \epsilon_A 
	        + 
	    \epsilon_Q
	.
	\]
	Furthermore, $c,N,D\in \nn_+$ are bounded above by
	\begin{itemize}
	\item[(i)] $
	 c
	\le
    	\left\lceil
                r^{\dagger}(C_{\omega,F_\phi,n}\omega,K,
                n,
                N,\cdot)\left(
                    C_{\eta,\omega,K, F_\phi, n}\,
            	   \frac{
            	        \epsilon_A
            	   }{
            	        N^{1/2}
            	   }
                \right)
            \right\rceil
	$,
	\item[(ii)] 
	
	$
      \ln(N)
	\le C_{\omega}
	       \ln\big(
        	    C_{(K,\|\cdot\|_{F_\phi:n})}
    	    \big)
	    \Big\lceil
	            - 
	        \log_2\big(
	           C_{\omega,K, F_\phi,n} \omega^{\dagger}
                	\big(
                	C^\prime_{\eta,\omega,K,F_\phi,n}
                        \,
                    \epsilon_A
                    \big)
	        \big)
	    \Big\rceil
$
	
	\item[(iii)] $
	D \le \mathscr{Q}_{f(\xxx)}(\epsilon_Q)
	$
	.
	\end{itemize}
	The constants $C_{\eta,\omega,K, F_\phi, n},\,C_\omega,\,C_{\omega,F_\phi,n}, \,C_{\omega,K,F_\phi,n}, \,C^\prime_{\eta,\omega,K, F_\phi, n} >0$ each depends only on the quantities appearing in their respective indices.  
	\end{lemma}

	\begin{proof}[{Comment on the proof of Lemma~\ref{lemma_metric_transformers_generalization}}]
	The steps in proving this lemma are essentially identical to those of \cite[Theorem 3.6]{AcciaioKratsiosPammer2022}'s proof up to two minor modifications, which we now discuss. 
	
	\hfill\\
	\textit{First Modification of \cite[Theorem 3.6]{AcciaioKratsiosPammer2022}'s proof - Steps $1$ and $2$}
	\hfill\\
	Since we do not require that $f$ is $\alpha$--H\"{o}lder continuous \textit{(which is the case in \cite[Theorem 3.6]{AcciaioKratsiosPammer2022})} but rather that it is uniformly continuous with a H\"{o}lder-like modulus of continuity $\omega$, Lemma~\ref{lem_snowflake__quantitative} implies that the upper-bound in \cite[Equation (6.21)]{AcciaioKratsiosPammer2022} (on the Lipschitz constant of the random projection of $(K,\|\cdot\|_{F_\phi:n}^{\omega})$ to any closed subset in $K$; in the sense of \cite{ohta2009extending}) is
	\[
	2 c \log\big(
	            C_{(K,\|\cdot\|_{F_\phi:n}^{\omega})} 
	        \big)
	    \le 
	2 c 
	\lceil
	    -\log_2(h_{\omega}^{\dagger}(1/4))/4
	\rceil
	\log(C_{(K,\|\cdot\|_{F_\phi:n})})
	\eqdef C_{\Pi}
	,
	\]
	for some absolute constant $c>0$.  Accordingly, we modify the constant $\delta>0$ defined in \cite[Equation (6.23)]{AcciaioKratsiosPammer2022} to be 
	$
	\delta 
	    \eqdef 
	\omega^{\dagger}
	\Big(
    	\frac{
    	    \epsilon_A
    	}{ 3 C_{\eta} C_{\Pi}}
    \Big).
	$ 
	Together, Lemmata~\ref{lem_snowflake__quantitative} and \cite[Lemma 6.1]{AcciaioKratsiosPammer2022} therefore 
	imply that the estimate on the external covering number of $f(K)\subset (\yyy,d_\yyy)$ related to $\delta$ (which we denote by $N$) in \cite[Equation (6.21)]{AcciaioKratsiosPammer2022} is at most
	\[
	N 
	    \le 
	\left(
	    C_{(K,\|\cdot\|_{F_\phi:n})}^{
        \lceil
        -\log_2(h_{\omega}^{\dagger}(\frac{1}{4}))/4
        \rceil
    	}
	\right)^{
	    \Big\lceil
	        \log_2(\operatorname{diam}(K, \|\cdot\|_{F_\phi:n})) 
	            - 
	        \log_2\Big(
	            \omega^{\dagger}
                	\big(
                    	\frac{
                    	    \epsilon_Q
                    	}{ 3 C_{\eta} C_{\Pi}}
                    \big)
	        \Big)
	    \Big\rceil
	}
	.
	\]

	\hfill\\
	\textit{Second Modification of \cite[Theorem 3.6]{AcciaioKratsiosPammer2022}'s proof - Step $5$}
	\hfill\\
Following Steps $3$ and $4$ of the proof of \cite[Theorem 3.6]{AcciaioKratsiosPammer2022} verbatim, we can construct an $\omega$-H\"{o}lder-like function%
	\footnote{Remark that $f^{\star}$ is denoted by $f_N$ in \cite{AcciaioKratsiosPammer2022}.}%
	$f^{\star}:(\mathbb{R}^{n}, \|\cdot\|_{F_\phi:n}) \rightarrow (\Delta_{N}, \|\cdot\|_2)$ (defined straightaway after \cite[Equation (6.31)]{AcciaioKratsiosPammer2022} and $N$ is as in (ii) of the statement of Lemma~\ref{lemma_metric_transformers_generalization}) such that its modulus of conitnuity $\omega_{f^\star}$ satisfies
 \[
	\omega_{f^\star}(t)
	    \eqdef 
	12 c^2 \omega(t) \lceil -\log_2(h_{\omega}^{\dagger}(1/4))/4\rceil^2 \log_2(C_{(K,\|\cdot\|_{F_\phi:n})})^2 C_{\eta}/\epsilon_A,
	\]
 and
	\begin{equation}
	\label{PROOF_eq___lemma_metric_transformers_generalization___fqApprox_of_f}
	\sup_{x\in \xxx}\,
	d_{\yyy}\left(
    	    f(x)
    	,
    	\hat{\eta}\left(
        	f^{\star}(x)
        	    ,
        	Z
    	\right)
	\right)
	\le \epsilon_A + \epsilon_Q
	,
	\end{equation}
	where $Z$ is some aptly-chosen matrix in $\mathbb{R}^{N\times D}$ (with $N$ and $D$ as in the statement of Lemma~\ref{lemma_metric_transformers_generalization}).  The remaining step in their proof, namely Step $5$, reduces to the approximation of $f^{\star}$ using a neural network.  It is this step which we now modify (noting that their final Step $6$ remains unaltered in our context); starting from their \cite[Equation (6.37)]{AcciaioKratsiosPammer2022}.  
	Let $
	    C_1
	\eqdef 
	    \max_{u \in \rr^n:\,\|u\|_{F_{\phi}:n}=1}
	                \frac{
	                    \|u\|_2
	                }{
	                    \|u\|_{F_{\phi}:n}
	                }
	$ and $
	C_2 \eqdef 
	    \max_{ u \in \rr^n:\,\|u\|_2=1 }
	                \frac{
	                    \|u\|_{F_{\phi}:n}
	                }{
	                    \|u\|_2
	                }
	$.  
 Then the $\omega$--H\"older continuous function $f^\star: (\rr^n, \|\cdot\|_{F_\phi:n}) \to (\Delta_N, \|\cdot\|_2)$ is an $\hat \omega$--H\"older continuous function with respect to $\|\cdot\|_2$ on $\rr^n$ with $\hat \omega(t) \eqdef h_\omega(C_2)\omega(t)$. 
Now, because $\mathcal{F}_{\cdot}$ is a universal Euclidean approximator in the sense of Definition \ref{definition:UniversalApproximators}, by setting 
	$%
    	c^{\epsilon_A}
	\eqdef
        \left\lceil
                r^{\dagger}(\hat \omega,K,
                n,
                N,\cdot)\left(
            	\frac{\epsilon_A}{
            	    C_{\eta}
            	        \omega\big(
            	            \operatorname{diam}(K,\|\cdot\|_{F_\phi:n})
            	        \big)
            	    \sqrt{N}}
                \right)
            \right\rceil
	$
	as in \cite[(6.37)]{AcciaioKratsiosPammer2022}, we deduce the following estimate
	\begin{equation}
	\label{PROOF_eq___lemma_metric_transformers_generalization___Approximation_of_star}
	\begin{aligned}
	\inf_{ 
	    \hat f_{n,N} \in \mathcal{F}_{n,N,c^{\epsilon_A}}
	}
	\sup_{x\in K}\,
	\big\|
    	f^{\star}(x)
            -
    	\hat{f}_{n,N}(x)
	\big\|_{2}
	\le &
    r\left(
    \hat \omega,K,
        n,
        N,
        \left\lceil
            r^{\dagger}(\hat \omega,K,
            n,
            N,\cdot)\left(
        	\frac{\epsilon_A}{
        	    C_{\eta}
        	        \omega\big(
        	            \operatorname{diam}(K,\|\cdot\|_{F_\phi:n})
        	        \big)
        	    \sqrt{N}}
            \right)
        \right\rceil
    \right)
	\\
	\le &
	\frac{
	  \epsilon_A
	}{
	    C_{\eta}\sqrt{N}
	    \omega\big(C_1 \operatorname{diam}(K,\|\cdot\|_2)\big)
	}
	.
	\end{aligned}
	\end{equation}
	\textit{The remainder of the proof is identical (mutatis mutandis) to the proof of \cite[Theorem 3.6]{AcciaioKratsiosPammer2022};} lastly, we relabel $\epsilon_A$ as $\epsilon_A/2$ (because we replace $\epsilon_{\mathcal{N}\mathcal{N}}$ appearing in the proof of \cite[Theorem 3.6]{AcciaioKratsiosPammer2022} by $\epsilon_A/2$).  
	\hfill\\
	Lastly, we can define $
	C_{\omega,F_\phi,n} 
	    \eqdef h_\omega(C_2)
	$ (recall $C_2 \eqdef 
	    \max_{ u \in \rr^n:\,\|u\|_2=1 }
	                \frac{
	                    \|u\|_{F_{\phi}:n}
	                }{
	                    \|u\|_2
	                }$), $
	C_{\eta,\omega,K, F_\phi, n} \eqdef \Big(
        C_{\eta}
        \omega\big(C_1 
           \operatorname{diam}(K,\|\cdot\|_2)
       \big)
    \Big)^{-1}$ for $C_1 \eqdef 
	    \max_{u \in \rr^n:\,\|u\|_{F_{\phi}:N}=1}
	                \frac{
	                    \|u\|_2
	                }{
	                    \|u\|_{F_{\phi}:n}
	                }$, 
    $C_\omega \eqdef \lceil
           -\log_2(h_{\omega}^{\dagger}(\frac{1}{4}))/4
          \rceil$, $C_{\omega, K, F_\phi,n} \eqdef \omega(\operatorname{diam}(K, \|\cdot\|_{F_\phi:n}))^{-1}$ and $C^\prime_{\eta,\omega,K, F_\phi, n} \eqdef \Big(
        	6c
        	C_{\eta} 
            	\lceil
            	    -\log_2(h_{\omega}^{\dagger}(1/4))/4
            	\rceil
            	\ln(C_{(K,\|\cdot\|_{F_\phi:n})})
        \Big)^{-1}$. 
    The conclusion then follows.  
	\end{proof}

	We present the proof of the special case of our main result when both $\xxx$ and $\yyy$ admit a a simple global structure.  By this, we mean that the singleton $\{(\xxx,\phi)\}$ is a feature decomposition of $(\xxx,d_{\xxx},\mu)$ and $(\yyy,\eta)$.  We obtain this result by applying Lemma~\ref{lemma_metric_transformers_generalization} to approximate the map $f_{\phi:\epsilon_E}:(\mathbb{R}^{n_{\epsilon_E}},\|\cdot\|_{F_{\phi}:n_{\epsilon_E}}) \rightarrow (\yyy,d_{\yyy})$ constructed in Lemma~\ref{lemma_Approximate_ExtensionFactorization}. Henceforth, for notational simplicity, given a BAP family $\{T_n\}_{n=1}^\infty$, we will assume that $\operatorname{rank}(T_n) = n$ for all $n$.
	    
    \begin{lemma}[$\mathcal{F}_{\cdot}$-Approximate Extension-Factorization]
	\label{lemma_SPECIALCASE__theorem_transferprinciple}
	Let $K$ be a non-empty compact doubling subset of a metric space $(\xxx,d_{\xxx})$ with doubling constant $C_{(K,d_{\xxx})} \ge 0$, $(\yyy,d_{\yyy})$ be a (barycentric) QAS space, $\mathcal{F}_\cdot$ be a universal approximator, $\phi:\xxx\rightarrow F_{\phi}$ be a quasisymmetric feature map such that $\phi \vert_{\phi(K)}^{-1}$ admits a (generalized) H\"older modulus of continuity $\omega_{\phi\vert_{\phi(K)}^{-1}}$, and $(T_n)_{n=1}^{\infty}$ realize the $C_{BAP:\phi(K)}$-BAP on $\phi(K)$. For every ``encoding error'', $\epsilon_E>0$, there is an ``encoding dimension'' $,
	    n = n_{\epsilon_E}
	\in \nn_+
	$,
	and an ``approximate feature map'', 
	\begin{equation}
	\label{eq:compressedfeaturemap_E}
	       \phi_{\epsilon_E}
	\eqdef 
    	\iota_{
    	    T_{
    	        n
    	    }
    	}^{-1}
	\circ 
    	T_{
    	    n
    	}
	\circ 
	    \phi
	:(\xxx,d_{\xxx})
	    \rightarrow 
	(\rr^n,\|\cdot\|_{F_{\phi}:n})
	,
	\end{equation}
    with the property that for every $\omega$-H\"{o}lder-like map $f:K \rightarrow \yyy$, every ``quantization error'' $\epsilon_Q>0$ and every ``approximation error'' $\epsilon_A>0$ there are $c,N,D\in \nn_+$, an $\hat{f}_{n,N}\in \mathcal{F}_{n,N,c}$, and $
	    Z\in \mathbb{R}^{
	        N\times D
	    }
	$ such that the following uniform estimate holds
    \[
        \sup_{x\in K}\,
        d_{\yyy}\Big(
            f(x)
        ,
            \hat{\eta}(\hat{f}_{n,N}\circ \phi_{\epsilon_E}(x),Z)
        \Big)
            \le 
        \epsilon_E + \epsilon_Q + \epsilon_A
        .
    \]
	Furthermore, $n,c,N,D\in \nn_+$ are bounded above by
	\begin{enumerate}
 \item[(i)] $ n \le 
	    R^{T_{\cdot}:\phi(K)}
	    \Big(
	        \omega^{\dagger}_{\phi^{-1}\vert_{\phi(K)}}
	            \circ
	        \omega^{\dagger}
	        \Big(
    	        \frac{
    	            \epsilon_E
    	        }{
    	            C_{(K,\omega,\phi^{-1},F_{\phi},T_{\cdot})}
    	        }
	        \Big)
	    \Big)
    $,
	\item[(ii)] $c
	\le
    	\left\lceil
                r^{\dagger}(C_{\tilde \omega,F_\phi,n}\tilde\omega,K,
                n,
                N,\cdot)\left(
                    C_{\eta,\tilde \omega,K, F_\phi, n}\,
            	   \frac{
            	        \epsilon_A
            	   }{
            	        N^{1/2}
            	   }
                \right)
            \right\rceil
	$,
	
	\item[(iii)] $
	D
	\le 
	\mathscr{Q}_{f(\xxx)}(\epsilon_Q),
	$
	\item[(iv)] $\ln(N)
	\le C_{\tilde \omega}
	       \ln\big(
        	    C_{(K,\|\cdot\|_{F_\phi:n})}
    	    \big)
	    \Big\lceil
	            - 
	        \log_2\big(
	           C_{\tilde \omega,K, F_\phi,n} \tilde{\omega}^{\dagger}
                	\big(
                	C^\prime_{\eta,\tilde \omega,K,F_\phi,n}
                        \,
                    \epsilon_A
                    \big)
	        \big)
	    \Big\rceil$,
	\end{enumerate}
    where $\tilde \omega(t) = C_{(K,\omega,\phi^{-1},F_{\phi},T_{\cdot})}\omega \circ \omega_{\phi\vert_{\phi(K)}^{-1}}(t)$.
    Moreover, the above constants $
    C_{
            (K,\omega,\phi^{-1},F_{\phi},T_{\cdot})
        }
    ,\, C_{\tilde \omega,F_\phi, n}, \ldots,
    $ depend only on the quantities appearing in their respective subscripts.
	\end{lemma}
	
	\begin{proof}[{Proof of Lemma~\ref{lemma_SPECIALCASE__theorem_transferprinciple}}]
	For a fixed $\epsilon_E>0$, let $\phi_{\epsilon_E}$ and $n = n_{\epsilon_E}$ be as in Lemma~\ref{lemma_Approximate_ExtensionFactorization}.  Then, by Lemma~\ref{lemma_Approximate_ExtensionFactorization} we have that for every $c,N,D\in \nn_+$, each $\hat{f}_{n,N}\in \mathcal{F}_{n,N,c}$, and every $Z\in \rr^{N \times D}$ it holds that
    \begin{equation}\label{PROOF__lemma_SPECIALCASE__theorem_transferprinciple___ApproximateExtensionFactorization_Estimate}    
    \begin{aligned}
        d_{\yyy}\big(
            f(x)
                ,
            \hat{\eta}(\hat{f}_{n,N}\circ \phi_{\epsilon_E}(x),Z)
        \big)   
    \le & 
            d_{\yyy}\big(
                f(x)
                    ,
                f_{\phi:\epsilon_E}\circ \phi_{\epsilon_E}(x)
            \big)   
        \\
        &+
            d_{\yyy}\big(
                f_{\phi:\epsilon_E}\circ \phi_{\epsilon_E}(x)
                    ,
                \hat{\eta}(\hat{f}_{n,N}\circ \phi_{\epsilon_E}(x),Z)
            \big)   
        \\
    \le &
        \epsilon_E
        +
            d_{\yyy}\big(
                f_{\phi:\epsilon_E}\circ \phi_{\epsilon_E}(x)
                    ,
                \hat{\eta}(\hat{f}_{n,N}\circ \phi_{\epsilon_E}(x),Z)
            \big)   
    .
      \end{aligned}
    \end{equation}
    
    Next, we claim that $(K,\|\cdot\|_{F_{\phi}:n})$ is doubling.  
    Since all finite-dimensional Banach spaces are bi-Lipschitz equivalent and bi-Lipschitz maps are quasisymmetric, $\phi_{\epsilon_E}(K)$ is a doubling subset of $(\rr^{n},\|\cdot\|_{F_\phi:n})$ if and only if it is a doubling subset of $(\rr^{n},\|\cdot\|_2)$.  However, all subsets of a Euclidean space are doubling (see, for example, \cite[Theorem 12.1]{heinonen2001lectures}).  Furthermore, as $\phi_{\epsilon_E}$ is continuous and since $K$ is compact, we conclude that $\phi_{\epsilon_E}(K)$ is a compact doubling subset of $(\rr^n,\|\cdot\|_{F_\phi:n})$; hence, from Lemma~\ref{lemma_metric_transformers_generalization} we deduce that there are $c,N,D\in \nn_+$, an $\hat{f}_{n,N} \in \mathcal{F}_{n,N,c}$, and a $Z\in \rr^{N \times D}$ (as specified in Lemma~\ref{lemma_metric_transformers_generalization}) such that the following estimate holds
    \begin{equation}
    \label{PROOF__lemma_SPECIALCASE__theorem_transferprinciple___Generalized_APK_Estimate}    
	\sup_{u\in \phi_{\epsilon_E}(K)}
	\,
        d_{\yyy}\big(
                f_{\phi:\epsilon_E}(u)
                    ,
                \hat{\eta}(\hat{f}_{n,N}\circ \phi_{\epsilon_E}(x),Z)
            \big) 
    \le 
        \epsilon_A + \epsilon_Q
    .
    \end{equation}
    Combining~\eqref{PROOF__lemma_SPECIALCASE__theorem_transferprinciple___ApproximateExtensionFactorization_Estimate} and~\eqref{PROOF__lemma_SPECIALCASE__theorem_transferprinciple___Generalized_APK_Estimate} yields the desired inequality,
    \[
    \begin{aligned}
            \sup_{x\in K}\,
            d_{\yyy}\big(
                f(x)
                    ,
                \hat{\eta}(\hat{f}_{n,N}\circ \phi_{\epsilon_E}(x),Z)
            \big)   
        \le &
        \epsilon_E
            +
            \sup_{u\in \phi(K)}\,
                d_{\yyy}\big(
                    f_{\phi:\epsilon_E}(u)
                        ,
                    \hat{\eta}(\hat{f}_{N,n}\circ \phi_{\epsilon_E}(x),Z)
                \big)   
        \\
        \le &
        \epsilon_E + \epsilon_A +\epsilon_Q.
     \end{aligned}
    \]
Lastly, we note that the upper bound for $n = n_{\epsilon_E}$ can be found in the statement of Lemma \ref{lemma_Approximate_ExtensionFactorization}, and the upper bounds for $c$, $N$ and $D$ follow directly from Lemma \ref{lemma_metric_transformers_generalization} applied to the function $f_{\phi:\epsilon_E}$ whose modulus of continuity is $\tilde \omega(t) = C_{(K,\omega,\phi^{-1},F_{\phi},T_{\cdot})}\omega \circ \omega_{\phi\vert_{\phi(K)}^{-1}}(t) $ by Lemma \ref{lemma_Approximate_ExtensionFactorization}.
	\end{proof}

The next lemma shows that the partition of unity which we use in the definition of our approximating class is with high probability well-defined, quantitatively.  

\begin{lemma}[Probabilistic Partition of Unity on a Featurization]
\label{lemma:hp_partition_of_unity}
Let $\{(\xxx_n,\phi_n)\}_{n\le N}$ be a feature decomposition of $(\xxx,d_{\xxx},\mu)$ (cf. Definition \ref{defn:CombinatorialStructure_X}).  For every $n\in \nn_+$ with\footnote{Note $n$ \textit{must be finite}.} $n\le N$ and every $0<R\le \inf_{n\le N}\, \operatorname{diam}(\xxx_n) $ define the Borel subset $\xxx^{(n,R)}$ of $\xxx$ by
\begin{equation}
\label{lemma:hp_partition_of_unity__GoodSet}
\xxx^{(n,R)}
        \eqdef 
    \bigcup_{i=1}^n
    \,
    \Big(
        \xxx_i
    \setminus
                [
                    (\partial \xxx_i)_R \cap \xxx_i
                ]
    \Big)
.
\end{equation}
For every $\tilde{n}\le n$ the map
\[
    \Pi^{(\tilde{n})}(x)
        \eqdef 
    \frac{
        d_{\xxx}(x,\xxx_{\tilde{n}}^c)
    }{
        \sum_{i\le n} \,
            d_{\xxx}(x,\xxx_{i}^c)
    }
\]
is well-defined.  
Moreover, for every $x\in \xxx^{(n,R)}$ it holds that $\sum_{\tilde{n}\leq n} \Pi^{(\tilde{n})}(x)=1$.  
\end{lemma}
\begin{proof}[Proof of Lemma~\ref{lemma:hp_partition_of_unity}]
Fix $1\le n \le N$, $0<R\le \inf_{n\le N}\operatorname{diam}(\xxx_n) $, the set $\xxx^{(n,R)}$ defined by \eqref{lemma:hp_partition_of_unity__GoodSet} is a Borel set.
By construction of $\xxx^{(n,R)}$, for every $x\in \xxx^{(n,R)}$, by definition there is some $i_x\le n$ such that $x\in \xxx_{i_x} - [(\partial \xxx_{i_x})_R \cap \xxx_{i_x}]$ and therefore $d_{\xxx}(x,\xxx_{i_x}^c)\ge R$; whence,
\[
    \sum_{i\le n}\, d_{\xxx}(x,\xxx_i^c)
        \ge 
    d_{\xxx}(x,\xxx_{i_x}^c)
        \ge 
    R.
\]
Therefore, for every $1\le \tilde{n}\le n$, the map $\Pi^{(\tilde{n})}:\xxx^{(n,R)}\rightarrow [0,1]$ is well-defined.  
By construction, for every $x\in \xxx^{(n,R)}$ it holds that $\sum_{\tilde{n}\leq n} \Pi^{(\tilde{n})}(x)=1$.  
\end{proof}
    
	Next, we derive a ``deterministic'' version of our main result, namely Theorem \ref{theorem:Structured}, in the case where $\yyy$ admits a quantized geodesic partition in the sense of Definition \ref{defn:CombinatorialStructure_Y}. In the lemma below, we will consider a special case of a slightly more general version of the assumptions in our main result, which corresponds to $M=1$, that is, the quantized geodesic partition of $\yyy$ consists of a singleton $\{\yyy\}$. We also allow $\yyy$ to have a ``topologically negligible boundary'' as in \cite{Anderson_1967,torunczyk1978concerning} where the boundary can be ``deleted'' quantitatively.  This can be visualized analogously to the illustration in Figure~\ref{fig_LSet}.  

	\begin{definition}[$\boldsymbol{L}$-Boundaries]
	\label{defn_L_set}
	Let $\boldsymbol{L}:(0,1]\rightarrow [0,\infty)$ be a decreasing continuous map with $\boldsymbol{L}(1)=0$.  
	A (possibly empty) subset $\zzz\subset \yyy$ is called an $\boldsymbol{L}$-boundary if there exists a homotopy $H\in C(\yyy\times [0,1],\yyy)$ satisfying:
	\begin{enumerate}
	\item[(i)] $H\left(\yyy,(0,1]\right)\subseteq \yyy \setminus \zzz$,
	\item[(ii)] $
	    \sup_{y \in \yyy}\, 
	    d_{\yyy}(H(y,t),y)
	        \le 
	   \boldsymbol{L}(1-t)$,
	\item[(iii)] For each $t \in [0,1)$, the map $y\mapsto H_t(y) \eqdef H(y,t)$ is Lipschitz.  
	\end{enumerate}
	\end{definition}

    The next result is a quantitative version of Theorem~\ref{theorem:determinsitic_transferprinciple} where the target space is a barycentric QAS space after removing an $\boldsymbol{L}$-boundary.  

	\begin{lemma}[Transfer Principle: Case $M=1$ up to an $\boldsymbol{L}$-Boundary]
	\label{lemma:determinsitic_transferprinciple}
	Let $K$ be a non-empty compact subset of a metric measure space $(\xxx,d_{\xxx},\mu)$ with feature decomposition $\{(\xxx_n,\phi_n)\}_{n\leq N}$ and suppose that $\mu$ is supported on $K$, 
	$(\yyy,d_{\yyy},\hat{\eta})$ be a metric space with $\boldsymbol{L}$-boundary $\zzz$ such that $(\yyy \setminus \zzz,d_{\yyy},\hat{\eta})$ is barycentric QAS space with a contracting barycentric map $\beta_\yyy$, 
	let $\mathcal{F}_{\cdot}$ be a universal approximator, 
    and for each $n\leq N$ let $\{T_k^{(n)}\}_{k=1}^{\infty}$ realize the BAP on $\phi_n(K\cap \xxx_n)$.  
    Suppose that either (or both) of the following hold:
    \begin{enumerate}
        \item For every $n\le N$, $\phi_n(K\cap \xxx_n)$ be doubling, and $\phi_n\vert_{\phi_n(K \cap \xxx_n)}^{-1}$ admits a H\"{o}lder-like modulus of continuity $\omega_{\phi_n\vert_{\phi_n(K \cap \xxx_n)}^{-1}}$;
        \item $K$ is doubling and for each $n\leq N$, the feature map $\phi_n$ is quasisymmetric, and $\phi_n\vert_{\phi_n(K \cap \xxx_n)}^{-1}$ admits a H\"{o}lder-like modulus of continuity $\omega_{\phi_n\vert_{\phi_n(K \cap \xxx_n)}^{-1}}$.
        \end{enumerate}
    For every ``encoding error'' $\epsilon_E>0$ and every $n\le N$, there is an ``encoding dimension'' $d_n \in \nn_+
	$
	and an ``approximate feature map'' 
	\begin{equation}
	\label{eq:compressedfeaturemap_1}
	       \phi^{(n)}
	\eqdef 
    	\iota_{
    	    T^{(n)}_{d_n}
    	}^{-1}
	\circ 
    	T^{(n)}_{d_n}
	\circ 
	    \phi_n
	:(\xxx_n,d_{\xxx})
	    \rightarrow 
	(\rr^{d_n},\|\cdot\|_{F_n:d_n})
	,
	\end{equation}
    with the property that: 
    for every $\omega$-H\"{o}lder-like continuous function $f:K \rightarrow \yyy$, every ``quantization error'' $\epsilon_Q>0$, every ``approximation error'' $\epsilon_A>0$ and every ``confidence level'' $\delta \in (0,1]$ 
	there exist an integer $N^\star \in \nn_+$, a finite family $\{(c_n,D_n,N_n)\}_{n\leq N^{\star}}$ of triples of positive integers, $Z_1,\dots,Z_{N^{\star}}$ with each $Z_n\in \rr^{N_n\times D_n}$, a family of approximators $\hat{f}_1,\dots,\hat{f}_{N^{\star}}$ with each $\hat{f}_n \in \mathcal{F}_{d_n,N_n,c_n}$, 
	a Borel-function $\hat{T}:\xxx\rightarrow \mathcal{P}_1(\yyy)$ and a Borel 
	subset $\xxx_{\delta}\subseteq \xxx$ such that:
	\begin{enumerate}
	    \item \textbf{Deterministic Uniform Approximation:} $\sup_{x\in K \cap \xxx_{\delta}}\,
	        d_{\yyy}\big(
	            \beta_{\yyy}(\hat{T}(x))
	                ,
	            f(x)
	        \big)
	    <
	       \epsilon_E + \epsilon_Q + \epsilon_A
	    $,
	    \item \textbf{Randomized Uniform Approximation:} $
        \sup_{x\in K \cap \xxx_{\delta}}\,
        	        W_1\big(
        	                \hat{T}(x)     
        	            ,
        	                \delta_{f(x)}
        	        \big)
              <
                    \epsilon_E + \epsilon_Q + \epsilon_A
            ,
    	$
	    \item \textbf{Representation:} 
    	$\hat{T}\vert_{\xxx_{\delta}}
    	        \eqdef 
        	 \sum_{n\leq N^{\star}}\,
        	       \frac{
    	                d_{\xxx}(x,\xxx_n^c)
        	        }{
        	            \sum_{i\leq N^{\star}}\,
        	                d_{\xxx}(x,\xxx_i^c)
        	        }
        	    \,
        	    \delta_{
        	        \hat{\eta}\big(
            	            \hat{f}_n
            	        \circ 
            	            \phi^{(n)}
            	            (x)
            	            ,
            	        Z_n
            	    \big)
        	    }
    	,
    	$
    	\item \textbf{{Probability of Sampling Inputs From $\xxx_{\delta}$}:} $\mu(\xxx_{\delta})\ge 1- \delta$.
	\end{enumerate}
	Furthermore, for every $n\leq N^{\star}$, the following holds:
	\begin{enumerate}
    \item[(i)] $
	    c_n
	\le
    	\left\lceil
                r^{\dagger}(\omega_n,K\cap \xxx_n,
                d_n,
                N_n,s_n)
            \right\rceil
	$ where $s_n=C_{\eta,\omega_n,K \cap \xxx_n, F_n, d_n}\,
            	   \frac{
            	        \epsilon_A
            	   }{
            	        N_n^{1/2}
            	   }$,
	\item[(ii)] $ d_n
	        \le
	    R^{T_{\cdot}^{(n)}:\phi_n(K\cap \xxx_n)}\Big(
	        \omega^{\dagger}_{
	            \phi_n\vert _{\phi_n(K\cap \xxx_n)}^{-1}
	        }
	            \circ
	        \omega^{\dagger}
	        \Big(
    	        \frac{
    	            \epsilon_E
    	        }{
    	            C_{(
	                   K\cap \xxx_n,\omega,\phi_n^{-1},F_{n},T_{\cdot}
    	            )}
    	        }
	        \Big)
	    \Big)
    ,
	$
	\item[(iii)] $
	D_n
	\le 
	\mathscr{Q}_{f(K \cap \xxx_n)}(\epsilon_Q)
	$
	,
	\item[(iv)] $
        \ln(N_n)
	\le 
    C_{\omega_n}
	       \ln\big(
        	    C_{(K\cap \xxx_n,\|\cdot\|_{F_n:d_n})}
    	    \big)
	\Big\lceil
	            - 
	        \log_2\big(
	           C_{\omega_n, K \cap \xxx_n, F_n, d_n} \omega_n^{\dagger}
                	\big(
                	C^\prime_{\eta,\omega_n, K\cap \xxx_n,F_n, d_n}\,
                    \epsilon_Q
                    \big)
	        \big)
	    \Big\rceil
	$,
	\item[(v)] $N^{\star}
    	        \eqdef 
    	    \inf\big\{
        N\in \mathbb{N}_+:\,
        \mu\big(
          \xxx^{(N,1/N)}
        \big) \, \ge \, 1-\delta
        \big\}$,
    \item[(vi)] $\xxx_{\delta}
    	    \eqdef \xxx^{(N^\star,1/N^\star)}
    ,
    $
	\end{enumerate}
 where $\omega_n \eqdef C_{(\epsilon_A, \omega, K\cap \xxx_n, \phi_n^{-1}, F_n, T^{(n)}_\cdot)}\omega \circ \omega_{\phi_n\vert_{\phi_n(K \cap \xxx_n)}^{-1}}$ and $\xxx^{(N,1/N)}$ is defined as in \eqref{lemma:hp_partition_of_unity__GoodSet}. 
	In particular, when $N=1$, then $\xxx_{\delta}\eqdef \xxx$, independently of the choice of $0<\delta\le 1$; also, $\mu(\xxx_{\delta})=1$.  
	\end{lemma}
	\begin{proof}[{Proof of Lemma~\ref{lemma:determinsitic_transferprinciple}}]
    \hfill\\
	\textbf{Outline:} \textit{We prove the result in three steps.  First, we restrict each part $\xxx_n$ of $\xxx$, and we perturb $f$'s output to lie in $\yyy$ minus the $\zzz$-boundary.  Next, we locally approximate each of these perturbations on each piece $\xxx_n$.  Finally, we use a Lipschitz-partition of unity of $\xxx_{\delta}$ to glue each approximation back together, forming our model.}
	\paragraph{Step 1 - Perturbing $f$ on Pieces of $\xxx$ and $\yyy$'s:}
	By our hypotheses, there exists an $\boldsymbol{L}$-boundary $\zzz$ of $\yyy$ such that $\yyy \setminus \zzz$ is barycentric and admits a quantizable mixing function $\hat{\eta}$.  Therefore, there exists a homotopy $H:[0,1]\times \yyy \rightarrow \yyy$ satisfying: 
	for every $t\in (0,1]$
	\begin{equation}
	\label{PROOF_eq__theorem_transferprinciple___ZboundaryRemoval}
	\sup_{y \in \yyy}\,
	d_{\yyy}\left(
	H_{t}(y)
	,
	y
	\right)
	    \le 
	\boldsymbol{L}(1-t)
	\quad 
	    \mbox{ and } \quad
	H_{t}(\yyy)\subseteq \yyy \setminus \zzz
	.
	\end{equation}
	For every $\boldsymbol{L}(0) > \epsilon_C'>0$ set $t^{\star}\eqdef 1 - \boldsymbol{L}^{\dagger}(\epsilon_C')$ and note that $t^{\star}$ belongs to $(0,1)$.  By Definition~\ref{defn_L_set} (iii), the map $[y\mapsto H_{t^{\star}}(y)]\in \operatorname{Lip}((\yyy,d_{\yyy}),(\yyy \setminus \zzz,d_{\yyy}))$ is Lipschitz.  
	For each $n\le N$ define the map
	\[
	    \bar{f}^{(n)}
	\eqdef
    	H_{t^{\star}}\circ f\vert _{\xxx_n}: \xxx_n
    	    \rightarrow
    	\yyy \setminus \zzz
	.
	\]
	
	We note that since $\xxx_n$ is closed and since $K$ is compact and doubling then for every $n\le N$ the subset $
	K_{n}
	    \eqdef 
	K\cap \xxx_n$ of $K$ is both compact and doubling\footnote{This follows from \cite[Lemma 9.3]{Robinson_2009} together with the observation that the inclusion of $K\cap \xxx_n$ into $K$ is bi-Lipschitz).}. Because $K_n\subseteq \xxx_n$, $\bar{f}^{(n)}(K_n)$ is a well-defined subset of $\yyy \setminus \zzz$.  Furthermore, by the definition of $t^{\star}$ and the maps $\{\hat{f}^{(n)}\}_{n\leq N}$, and~\eqref{PROOF_eq__theorem_transferprinciple___ZboundaryRemoval} we have that
	\begin{equation}
	    \label{PROOF_eq__theorem_transferprinciple___ZboundaryRemoval____functional_version}
	    \max_{n\leq N}\,
	    \max_{x\in K_n}\,
    	    d_{\yyy}\big(
    	            f
    	            (x)
    	        ,
    	            \bar{f}^{(n)}
    	            (x)
    	    \big)
    	\le 
    	    \sup_{x\in f(X)}\, d_{\yyy}\big (y,H_{t^{\star}}(y)\big)
        \le 
    	\sup_{y\in \yyy}\, 
    	    d_{\yyy}\big(
    	            y
    	                ,
    	           H_{t^{\star}}
    	           (y)
    	    \big)
    	 \le \boldsymbol{L}(1 - t^\star) \le
    	    \epsilon_C'.
	\end{equation}
	\paragraph{Step 2 - Approximating $f$ on Each Part of $\xxx$:}
	For every $n\le N$, suppose that we are given any family of bounded linear operators $\big\{T_k^{(n)}\big\}_{k=1}^{\infty}$ on $F_{\phi_n}$ approximating the identity on $\phi_n(K_n)$, that each $\phi_n(K_n)$ is a doubling subset of $F_n$ respectively, and suppose that we are given errors $\epsilon_C'',\epsilon_E,\epsilon_Q>0$.  
	Then, we may apply Lemma~\ref{lemma_SPECIALCASE__theorem_transferprinciple} to the H\"{o}lder-like continuous function $\bar{f}^{(n)}$ whose modulus of continuity $\bar{\omega}_n$ only depends on $\epsilon^\prime_C$ and $\omega$, to conclude that there exist a family $\{(c_n,d_n,D_n,N_n)\}_{n\leq N}$ of $4$-tuples of positive integers (each as in Lemma~\ref{lemma_SPECIALCASE__theorem_transferprinciple} (i)-(iv)) 
	such that for the family of maps $\{\phi^{(n)} :\xxx_n \rightarrow (\rr^{d_{n}},\|\cdot\|_{F_{n}:d_n})\}_{n\le N}$, defined by
    $\phi^{(n)}
	\eqdef 
    	\iota_{
    	    T_{d_n}^{(n)}
    	}^{-1}
	\circ 
    	T_{d_n}^{(n)}
	\circ 
	    \phi_n
	:(\xxx_n,d_{\xxx})
	    \rightarrow 
	(\rr^{d_n},\|\cdot\|_{F_n:d_n})
    $
    ,
    there exist $Z_1,\dots,Z_N\in \rr^{N_n\times D_n}$ and a family of approximators $\hat{f}_1,\dots,\hat{f}_{N}\in \mathcal{F}_{\cdot}$, with each $\hat{f}_n\in \mathcal{F}_{d_n,N_n,c_n}$, such that the following holds
	\begin{equation}
	\label{PROOF_theorem_transferprinciple__eq_application_localversion___homotopedboundaryversion}
	    \sup_{x\in K_{n}}\,
    	    d_{\yyy}\Big(
    	        \bar{f}^{(n)}(x)
    	    ,
    	       \hat{\eta}\big(
        	            \hat{f}_n
        	        \circ 
        	            \phi^{(n)}(x)
        	            ,
        	        Z_n
        	    \big)
    	    \Big)
    	        <
    	   \epsilon_C'' + \epsilon_E + \epsilon_Q
    ,
	\end{equation}
    for each $n=1,\dots,N$.  
	Together, equation~\eqref{PROOF_eq__theorem_transferprinciple___ZboundaryRemoval}, the definition of $t^{\star}$, and the uniform estimate in~\eqref{PROOF_theorem_transferprinciple__eq_application_localversion___homotopedboundaryversion}, we find that for all $n \le N$
	\allowdisplaybreaks
	\begin{equation}
	\label{PROOF_theorem_transferprinciple__eq_application_localversion___completedversion}
    \begin{aligned}
        \sup_{x\in K_{n}}\,
    	    d_{\yyy}\Big(
    	        f(x)
    	    ,
    	       \hat{\eta}\big(
        	            \hat{f}_n
        	        \circ 
        	            \phi^{(n)}(x)
        	            ,
        	        Z_n
        	    \big)
    	    \Big)
    \le &
            \sup_{x\in K_{n}}\,
                d_{\yyy}\Big(
        	        f
        	        (x)
        	    ,
        	        \bar{f}^{(n)}
        	        (x)
        	    \Big)
    \\	 
    	 & +
    	    \sup_{x\in K_{n}}\,
        	    d_{\yyy}\Big(
        	        \bar{f}^{(n)}(x)
    	    ,
    	       \hat{\eta}\big(
        	            \hat{f}_n
        	        \circ 
        	            \phi^{(n)}(x)
        	            ,
        	        Z_n
        	    \big)
        	    \Big)
    \\  
    \le & 
            \epsilon_C'
        +
    	    \epsilon_C'' + \epsilon_E + \epsilon_Q
    .
    \end{aligned}
	\end{equation}
	Setting $\epsilon_C''\eqdef \frac{\epsilon_A}{2}$ and $\epsilon_C'\eqdef \epsilon_C''$ we find that~\eqref{PROOF_theorem_transferprinciple__eq_application_localversion___completedversion} yields
	\begin{equation}
	\label{PROOF_theorem_transferprinciple__eq_application_localversion___CleanEstimate}
    \begin{aligned}
        \sup_{n\leq N}\,
        \sup_{x\in K_{n}}\,
    	    d_{\yyy}\Big(
    	        f(x)
    	   ,
    	       \hat{\eta}\big(
        	            \hat{f}_n
        	        \circ 
        	            \phi^{(n)}(x)
        	            ,
        	        Z_n
        	    \big)
        	    \Big)
    \le 
            \epsilon_A + \epsilon_E + \epsilon_Q
    .
    \end{aligned}
	\end{equation}
	For notational simplicity, we henceforth abbreviate
	$
	\hat{f}^{(n)}_{\star}
	    \eqdef
	\hat{\eta}\big(
        	            \hat{f}_n
        	        \circ 
        	            \phi^{(n)}(x)
        	            ,
        	        Z_n
        	    \big)
    .
	$
    
	\paragraph{Step 3 - Gluing the Approximations Together via the Partition of Unity:}
	We fix $\delta \in (0,1]$.  Using the notation of Lemma~\ref{lemma:hp_partition_of_unity}, we define the positive integer $N^{\star}$ by
	\[
	        N^{\star} 
	    \eqdef
	        \inf\big\{
            1\le \tilde{N} \le N:\,
            \mu\big(
              \xxx^{(\tilde{N},1/{\tilde{N}})}
            \big) \, \ge \, 1-\delta
            \big\}
    .
	\]
 Thanks to the properties of Feature Decomposition (cf. Definition \ref{defn:CombinatorialStructure_X}) we can easily see that $N^\star$ is always finite for any $\delta >0$. 
	We set $X_{\delta}\eqdef X^{(N^{\star},1/N^{\star})}$.
	By Lemma~\ref{lemma:hp_partition_of_unity}, the family of maps $\{\psi_n:\xxx_{\delta}\rightarrow [0,1]\}_{n\le N^{\star}}$ given by
	\[
	        \psi_n(x) 
	  \eqdef 
	        \frac{
	            d_\xxx(x,\xxx_n^c)
	         }{ 
	            \sum_{i=1}^{N^{\star}}\, d_\xxx(x,X_i^c) 
	        }
	\]
	forms a well-defined locally Lipschitz partition of unity of the metric subspace $(\xxx_{\delta},d_{\xxx})$.  Moreover, for every $n\leq N^{\star}$, $\inf_{x\in \xxx_n\cap \xxx_{\delta},\, \tilde x \in \xxx_n^c\cap \xxx_{\delta}}\,d(x,\tilde{x})\ge 1/N^{\star}>0$; whence, the $\xxx_n\cap \xxx_{\delta}$ is both closed and open.  Therefore, for every $n\le N^{\star}$ and any $x\in \xxx_{\delta}$, $\psi_n(x)>0$ if and only if $x \in \xxx_n \cap \xxx_{\delta}$. As a consequence, the map $\tilde{T}:\xxx_{\delta}\rightarrow \mathcal{P}_1(\yyy)$ defined by 
    \begin{equation}
    \label{eq:lemma_transferprinciple_Mis1__TtildeExplicitForm}
    	    \tilde{T}(x)
	        \eqdef 
	    \sum_{n=1}^{N^{\star}}\,
	       \psi_n(x)\,
	   \delta_{
	    \hat{f}^{(n)}_{\star}(x)
	   },
    \end{equation}
	is well-defined on $\xxx_{\delta}$.  
	Fixing any $y_0\in \yyy$, we extend $\tilde{T}$ to a Borel function $\hat{T}$ defined on all of $\xxx$ by
	\[
    	\hat{T}(x) 
    	    \eqdef
    	I_{\xxx_{\delta}}(x)\, 
    	    \tilde{T}(x) 
    	    +
        I_{\xxx_{\delta}^c}(x)\,
            \delta_{y_0},
	\]
	where $I_{\xxx_{\delta}}$ is the measure-theoretic\footnote{not in the sense of convex analysis} indicator function.  Therefore, the following estimate holds
	\begin{align}
	\label{eq:lemma_transferprinciple_Mis1__CompletedEstimate___BEGIN}
            \max_{x\in K\cap \xxx_{\delta}}\,
                        W_1\big(
                            \hat{T}(x)
                        ,
                            \delta_{f(x)}
                        \big)
        \eqdef &
            \max_{x\in K\cap \xxx_{\delta}}\,
                W_1\Big(
                    \sum_{n=1}^{N^{\star}}\,
            	   \psi_n(x)
            	   \,
            	   \delta_{
                      \hat{f}^{(n)}_{\star}(x)
            	   }
                ,
                    \delta_{f(x)}
                \Big)
        \\ 
        \nonumber
        = &
            \max_{x\in K\cap \xxx_{\delta}}\,
            \Big\|
                    \Big(
                            \sum_{n=1}^{N^{\star}}\,
                    	       \psi_n(x)
                    	       \,
                    	   \delta_{
                    	   \hat{f}^{(n)}_{\star}(x)
                    	   }
                    	-
                    	    \delta_{y_0}
                	\Big)
                -
                    \big(
                            \delta_{f(x)}
                        -
                            \delta_{y_0}
                    \big)
            \Big\|_{{\text{\AE}}(\yyy,y_0)}
    \\
    \nonumber
        = &
            \max_{x\in K\cap \xxx_{\delta}}\,
            \Big\|
                    \sum_{n=1}^{N^{\star}}\,
                    	   \psi_n(x)    
                    	   \,
                    	   \delta_{
                    	   \hat{f}^{(n)}_{\star}(x)
                    	   }
                -
                            \delta_{f(x)}
            \Big\|_{{\text{\AE}}(\yyy,y_0)}
    \\
    \nonumber
        = &
            \max_{x\in K\cap \xxx_{\delta}}\,
            \Big\|
                    \sum_{n=1}^{N^{\star}}\,
                    	   \psi_n(x)
                    	       \,
                    	   \delta_{
                    	  \hat{f}^{(n)}_{\star}(x)
                    	   }
                -
                        \sum_{n=1}^{N^{\star}}\,
                    	   \psi_n(x)
                    	   \,
                    	   \delta_{f(x)}
            \Big\|_{{\text{\AE}}(\yyy,y_0)}
    \\
    \nonumber
        \le &
            \max_{x\in K\cap \xxx_{\delta}}\,
            \sum_{n=1}^{N^{\star}}\,
    	   \psi_n(x)    
    	   \,
            \Big\|
        	   \delta_{
        	       \hat{f}^{(n)}_{\star}(x)
        	   }
                    -
                \delta_{f(x)}
            \Big\|_{{\text{\AE}}(\yyy,y_0)}
  \\
  \nonumber
        = &
            \max_{x\in K\cap \xxx_{\delta}}\,
            \sum_{n=1}^{N^{\star}}\,
    	    \psi_n(x)
    	    \,
            \Big\|
        	   \Big(
            	    \delta_{
            	        \hat{f}^{(n)}_{\star}(x)
            	    }
            	        -
            	   \delta_{y_0}
        	   \Big)
                    -
                \Big(
                    \delta_{f(x)}
                        -
                    \delta_{y_0}
                \Big)
            \Big\|_{{\text{\AE}}(\yyy,y_0)}
    \\
    \nonumber
        = &
            \max_{x\in K\cap \xxx_{\delta}}\,
            \sum_{n=1}^{N^{\star}}\,
    	    \psi_n(x)
    	    \,
            W_1\Big(
            	    \delta_{
            	   \hat{f}^{(n)}_{\star}(x)
            	    }
                        ,
                    \delta_{f(x)}
                \Big)
    \\
    =& 
    \sup_{x \in K\cap \xxx_{\delta}} \,
    \sum_{n=1}^{N^{\star}}\, 
      \psi_n(x)
      \,
      d_{\yyy}\big(\hat{f}_{\star}^{(n)}(x),f(x)\big)
    \\
    \nonumber
        < &
            \max_{x\in K\cap \xxx_{\delta}}\,
            \sum_{n=1}^{N^{\star}}\,
    	   \psi_n(x)
    	   \,
          ( \epsilon_E + \epsilon_Q + \epsilon_A)
    \\
	\label{eq:lemma_transferprinciple_Mis1__CompletedEstimate___END}
        = &
        \epsilon_E + \epsilon_Q + \epsilon_A
    .
	\end{align}
Since $\yyy\setminus \zzz$ is barycentric, we let  $\beta_{\yyy}: \mathcal{P}_1(\yyy\setminus \zzz) \to \yyy\setminus \zzz$ be the contracting barycentric map with Lipschitz constant $1$ such that $\beta_\yyy(\delta_y) = y$.  Therefore, estimate in~\eqref{eq:lemma_transferprinciple_Mis1__CompletedEstimate___BEGIN}-\eqref{eq:lemma_transferprinciple_Mis1__CompletedEstimate___END} implies that
\allowdisplaybreaks
\begin{align}
\nonumber
    \max_{x\in K\cap \xxx_{\delta}}\,
            d_{\yyy}\big(
                \beta_{\yyy}\circ \hat{T}(x)
            ,
                f(x)
            \big)
    = &
    \max_{x\in K\cap \xxx_{\delta}}\,
            d_{\yyy}\big(
                \beta_{\yyy}\circ \hat{T}(x)
            ,
                \beta_{\yyy}\circ \delta_{f(x)}
            \big)
\\
\nonumber
    \le &
    \max_{x\in K\cap \xxx_{\delta}}\,
        W_1\big(
            \hat{T}(x)
        ,
            \delta_{f(x)}
        \big)
\\[0.25cm]
\nonumber
    < &
        \epsilon_E + \epsilon_Q + \epsilon_A
        .
\end{align}
We note that if $N=1$ then we may set $\xxx_{\delta}\eqdef \xxx$, independently of the choice of $\delta$, and $\psi_1$ is identically equal to $1$.  Finally, the statements on the bounds for $c_n$, $d_n$, $D_n$ and $N_n$ can be obtained by using the corresponding estimates in Lemma \ref{lemma_SPECIALCASE__theorem_transferprinciple}.
\end{proof}

\subsection{{Proof of Theorem~\ref{theorem:Unstructured_Case}}}
\label{s:Proofs__ss:qualitative}
We now derive our main \textit{qualitative} universal approximation theorem, which is valid for general metric target spaces and general topological source spaces.  
The following is a mild generalization of Theorem~\ref{theorem:Unstructured_Case}.  
Since the result is qualitative, we opt for a streamlined formulation, in which we aggregate the ``quantization error'' $\varepsilon_D$, ``approximation error'', and ``encoding error'' $\varepsilon_E$ into a single error $\varepsilon>0$.
\begin{lemma}[Transfer Principle: Polish $\xxx$ and $\yyy$]
\label{lemma:Unstructured_Case}
Assume Setting~\ref{setting:theorems_Quantitative} (i),(ii), and (iv) and suppose that 
$(\yyy,d_{\yyy})$ is a complete separable metric space, and let $K\eqdef \operatorname{supp}(\mu)$.  
Then, for every 
continuous $f:(K, d_\xxx) \rightarrow \yyy$ 
there is a metric $d_K$ on $K$ generating the subspace topology such that for any $\epsilon > 0$ there exists a Lipschitz function $\tilde{f}: (K,d_K) \to \yyy$ with $ \max_{x\in K}\,
        d_{\yyy}\big(
            f(x)
        ,
            \tilde{f}(x)
        \big)
    <
        \epsilon$. Moreover,
for every ``error'' $\epsilon>0$ and every $n\le N$, there exist an $N^\star\in \nn_+
$, a family of integers $\{(c_n, d_n,N_n,D_n)\}_{n=1}^{N^\star}$, a family of approximators
$\hat{f}_1,\dots,\hat{f}_{N^{\star}}$ with $\hat f_n \in \fff_{d_n,N_n,c_n}$, a Borel set $\xxx_{\epsilon} \subset \xxx$ and a Borel map $\hat{T}:\xxx\rightarrow \mathcal{P}_1(\yyy)$ satisfying
\begin{equation}
\label{eq:theorem_determinsitic_transferprinciple__PAC_Bound___RandomizedVersion}
    \mu\big(
            \xxx_\epsilon
        \big)
    \ge 
        1 - \max\{\epsilon,1\},
\end{equation}
$ \sup_{x \in \xxx_\epsilon} d_{\yyy}(
                \hat{T}(x)
                    ,
                \delta_{f(x)}
            )
        < \epsilon$
and when restricted to $\xxx_{\epsilon}$, the map $\hat{T}$ is of the form
\[
        \hat{T}(x)
    =
        \sum_{n\le N^\star}\, 
            \psi_n(x)
            \,
            \cdot
            \,
                \sum_{i=1}^{N_n}\,
                [P_{\Delta_{N_n}}(
                    \hat{f}_n\circ \phi^{(n)}(x)
                )]_i
                \,
                    \biggl(
                        \sum_{j=1}^{Q_n}\,
                            [P_{\Delta_Q}(u^n_{i})]_j\,
                            \delta_{\tilde{y}_{\lceil 
                                            z_{i,j}^n 
                                      \rceil}}
                    \biggr),
\]
where
$(u_{i}^n,z_{i}^n)_{i=1}^{N_n}$ are tensors each in $\mathbb{R}^{2 Q_n}$ for $2Q_n = D_n$, $\{\tilde y_i\}_{i \in \nn_+}$ is a fixed dense subset in $(\mathcal{P}_1(\yyy), W_1)$, $\phi^{(n)} = \iota_{T^{(n)}_{d_n}}^{-1} \circ T^{(n)}_{d_n} \circ \phi_n$ and $
        \psi_n(x)
    \eqdef 
        \frac{d_{K}(x,\xxx_n^c)}{\sum_{j=1}^{N^{\star}}\,d_{K}(x,\xxx_j^c)}
$.  
\hfill\\
Moreover, if $f$ is $\omega$-H\"{o}lder-like then we may take $d_K=d_{\xxx}$ and if $N=1$ then it holds that
\[
    \sup_{x\in K}\, W_1(\hat{T}(x),\delta_{f(x)})<\epsilon
.
\]
\end{lemma}
\begin{proof}[{Proof of Lemma~\ref{lemma:Unstructured_Case}}]
Fix $\epsilon>0$.
We first introduce some notation.  
Let $K\eqdef \operatorname{supp}(\mu)$, $(\tilde{\yyy},d_{\tilde{\yyy}})\eqdef (\mathcal{P}_1(\yyy),W_1)$, and denote the $1$-Wasserstein metric on $
        \mathcal{P}_1(\tilde{\yyy},d_{\tilde{\yyy}})
    =
        \mathcal{P}_1\big(\mathcal{P}_1(\yyy,d_{\yyy}),W_1\big)
$ by $\mathcal{W}_1$. 
\paragraph{Step 1 - Metrizing $\xxx$ so that $f$ Can be Approximated by Lipschitz Functions on $K$:}
Since $f:\xxx\rightarrow \yyy$ is continuous when restricted to $\mu$'s support, it belongs to the topological space $f\vert_{K}\in C(K,\mathcal{Y})$ with the uniform convergence on compact sets topology.  Since $\mu$ is a compactly-supported measure on $\xxx$ and $\xxx$ is a metrizable Polish space, then $f\vert_{K}$ is a continuous function defined on a compact metrizable Polish space with outputs in a separable metric space $(\yyy,d_{\yyy})$. Hence, Lemma~\ref{lemma:change_of_metric} applies.  Thus, there is a metric $d_{K}$ on $K$ generating the subspace topology on $K$ such that there is a Lipschitz map $\tilde{f}:(K,d_K) \rightarrow \yyy$ satisfying the uniform estimate,
\begin{equation}
\label{eq:PROOF__theorem:Unstructured_Case__LipschitzRedux}
    \max_{x\in K}\,
        d_{\yyy}\big(
            f(x)
        ,
            \tilde{f}(x)
        \big)
    <
        \frac{\epsilon}{2}
    .
\end{equation}
\paragraph{Step 2 - $(\mathcal{P}_1(\yyy),W_1)$ is a Barycentric QAS Space:}
First we observe that the map $\yyy\rightarrow \tilde{\yyy}$, given by $y\mapsto \delta_{y}$, is an isometric embedding, whence the map $F:(K,d_K)\rightarrow \tilde{\yyy}$, given by $x\mapsto \delta_{\tilde{f}(x)}$, is a Lipschitz map.  
We consider the geodesic bicombing $\gamma:\tilde{\yyy}\times \tilde{\yyy}\times [0,1]\rightarrow \tilde{\yyy}$ sending any $\mathbb{P},\mathbb{Q}\in \tilde{Y}$ and each $t\in [0,1]$ to
\[
        \gamma(\mathbb{P},\mathbb{Q},t) 
    \eqdef 
        (1-t)\mathbb{P} + t\mathbb{Q}  
\]
and fix some $y_0 \in \yyy$.  By the isometric embedding $\Phi$ in Lemma~\ref{lemma_closedconvex_embedding_Wasserstein}, for every $t \in [0,1]$ and each $\mathbb{P},\mathbb{Q},\tilde{\mathbb{P}},\tilde{\mathbb{Q}}\in  \tilde{\yyy}$, we have
\allowdisplaybreaks
\begin{align}
\nonumber
    W_1\big(
        \gamma(\mathbb{P},\mathbb{Q},t)
            ,
        \gamma(\tilde{\mathbb{P}},\tilde{\mathbb{Q}},t)
    \big)
        = &
    \big\|
       \Phi (\gamma(\mathbb{P},\mathbb{Q},t)-\delta_{y_0})
            -
        \Phi(\gamma(\tilde{\mathbb{P}},\tilde{\mathbb{Q}},t)-\delta_{y_0})
    \big\|_{{\text{\AE}}(\yyy,y_0)}
    \\
\nonumber
        = &
    \big\|
       \Phi \circ \gamma(\mathbb{P},\mathbb{Q},t)
            -
        \Phi \circ \gamma(\tilde{\mathbb{P}},\tilde{\mathbb{Q}},t)
    \big\|_{{\text{\AE}}(\yyy,y_0)}
    \\
\nonumber
        = &
    \big\|
        \big(
            (1-t)\Phi(\mathbb{P}) + t\Phi(\mathbb{Q})
        \big)
            -
        \big(
            (1-t)\Phi(\tilde{\mathbb{P}}) + t\Phi(\tilde{\mathbb{Q}})
        \big)
    \big\|_{{\text{\AE}}(\yyy,y_0)}
    \\
\nonumber
        = &
    \big\|
        (1-t)
            (\Phi(\mathbb{P})-\Phi(\tilde{\mathbb{P}}))
            -
        t
        (\Phi(\mathbb{Q})-\Phi(\tilde{\mathbb{Q}}))
    \big\|_{{\text{\AE}}(\yyy,y_0)}
    \\
\nonumber
        \le &
    (1-t)\big\| \Phi(\mathbb{P})-\Phi(\tilde{\mathbb{P}})\big\|_{{\text{\AE}}(\yyy,y_0)}
        +
        t
        \big\| \Phi(\mathbb{Q})-\Phi(\tilde{\mathbb{Q}})\big\|_{{\text{\AE}}(\yyy,y_0)}
    \\
\nonumber
        = &
        (1-t)
        W_1(\mathbb{P},\tilde{\mathbb{P}})
            +
        t
        W_1(\mathbb{Q},\tilde{\mathbb{Q}})
\end{align}
Hence, $\gamma$ is a conical geodesic bicombing on $(\tilde{\yyy},d_{\tilde{\yyy}})$.  Since $\yyy$ is complete and separable then \citep[Theorem 6.18]{VillaniOptTrans} implies that $(\tilde{\yyy},d_{\tilde{\yyy}})$ is also complete.  By \cite[Theorem 2.6 and Definition 2.4]{basso2020extending}, we deduce that $(\tilde{\yyy},d_{\tilde{\yyy}})$ admits a $1$-Lipschitz barycenter map $\beta_{\tilde{\yyy}}:\mathcal{P}_1(\tilde{\yyy})\rightarrow \tilde{\yyy}$.  Also, by the example in Section \ref{s:Introduction__ss:QASSpaces} we know that the Wasserstein space $(\tilde{\yyy}, d_{\tilde{\yyy}})$ is quantizable. Hence, by Proposition \ref{prop:QAS_Concial} we conclude that $(\tilde{\yyy}, d_{\tilde{\yyy}})$ is a barycentric QAS space. From now on, we fix a dense subset $\{\tilde{y}_i\}_{i \in \nn_+}$ of $\tilde{\yyy}$ and define the mixing function $\hat{\eta}$ of the same form as in \eqref{eq:quantizedmixing_Wasserstein}.
\paragraph{Step 3 - Approximating $F$ on Each Part $\xxx_n$:}
For each $n\le N$, we consider the restricted map $F\vert_{\xxx_n \cap K}:\xxx_n\cap K \rightarrow \tilde{\yyy}$ and note that $\{(\xxx_n,\phi)\}$ is a feature decomposition of $(\xxx_n\cap K,d_{K})$. Because $(\tilde{\yyy},d_{\tilde{\yyy}})$ is a barycentric QAS space, all conditions of Lemma~\ref{lemma:determinsitic_transferprinciple} are satisfied. Thus, for any $\epsilon>0$ there is a map $\hat{T}_n:\xxx_n\cap K \rightarrow \tilde{\yyy}$ satisfying
\begin{equation}
\label{eq:corollary_Unstructured_Case__partwise_estimate}
    \sup_{x\in \xxx_n\cap K}\, 
        W_1\big(
            \beta_{\tilde{\yyy}}\circ \hat{T}_n(x)
        ,
            \beta_{\tilde{\yyy}}\circ \delta_{F(x)}
        \big)
        \le 
    \sup_{x\in \xxx_n\cap K}\, 
        \mathcal{W}_1\big(
            \hat{T}_n(x)
        ,
            \delta_{F(x)}
        \big)
            <
        \epsilon/2
    .
\end{equation}
In view of Step 2 in the proof of Lemma~\ref{lemma:determinsitic_transferprinciple}, each $\hat{T}_n\vert_{\xxx_n\cap K}$ is of the form
\begin{equation}
\label{eq:corollary_Unstructured_Case__explicitform}
    \hat{T}_n\vert_{\xxx_n\cap K} = \delta_{\hat{\eta}\big(
                \hat{f}_n\circ \phi^{(n)}(\cdot), Z_n
            \big)},
\end{equation}
where each $Z_n\in \rr^{N_n\times D_n}$, while $\hat{f}_1,\dots,\hat{f}_{N^{\star}}$ with each $\hat{f}_n \in \fff_{d_n,N_n,c_n}$, $\{(c_n,d_n, D_n,N_n)\}_{n\leq N^{\star}}$ is a family of positive integers, and noting that $\phi^{(n)}$ is defined in~\eqref{eq:compressedfeaturemap_1}.  Therefore, by the definition of the $1$-Lipschitz barycenter map $\beta_{\tilde{\yyy}}: \big(\mathcal{P}_1(\tilde{\yyy},d_{\tilde{\yyy}}),\mathcal{W}_1\big) \rightarrow (\tilde{\yyy},d_{\tilde{\yyy}})$ and by the definition of the ``lifted map'' $F$, we have that~\eqref{eq:corollary_Unstructured_Case__partwise_estimate} implies that
\begin{equation}
\label{eq:corollary_Unstructured_Case__partwise_estimate___simplified}
    \sup_{x\in \xxx_n\cap K}\, 
        W_1\big(
            \hat{\eta}\big(
                \hat{f}_n\circ \phi^{(n)}(x), Z_n
            \big)
        ,
            F(x)
        \big)
    \le 
    \sup_{x\in \xxx_n\cap K}\, 
        W_1\big(
            \beta_{\tilde{\yyy}}\circ \hat{T}_n(x)
        ,
            \beta_{\tilde{\yyy}}\circ \delta_{F(x)}
        \big)
    <
        \frac{\epsilon}{2}
\end{equation}
holds for every $n\le N$.  
\paragraph{Step 4 - Gluing Together the Local Approximators from Step 3:}
Step 4 is nearly identical, mutatis mundais, to Step 3 in the proof of Lemma~\ref{lemma:determinsitic_transferprinciple}.  That is, the integer $N^\star$ and $\xxx_{\epsilon}$ are defined in the same way (with $\delta$ replaced by $\epsilon$) and $\tilde{T}:\xxx_{\delta}\cap K\rightarrow \tilde{\yyy}$ is defined similarly to~\eqref{eq:lemma_transferprinciple_Mis1__TtildeExplicitForm} by
\begin{equation}
\label{eq:theorem:Unstructured_Case__tildeTdefined}
        \tilde{T}(x) 
    \eqdef 
        \sum_{n\le N}\, 
            \psi_n(x)
            \,
            \hat{\eta}\big(
                \hat{f}_n\circ \phi^{(n)}(x), Z_n
            \big)
\end{equation}
for a partition of unity $\{\psi_n\}_{n=1}^{N^\star}$.
As in the previous Lemma's proof, we take $\hat{T}$ to be any Borel extension of $\tilde{T}$ from $\xxx_{\epsilon}\cap K$ to all of $\xxx$.  
\paragraph{Step 5 - Explicit Form of $\hat{T}$:}
The explicit form of $\hat{T}$ follows from the quantized mixing function $\hat{\eta}$ defined on $\tilde{\yyy}$, which was given in~\eqref{eq:quantizedmixing_Wasserstein} as
\[
    \hat{\eta}\big(w,Z_n\big)
    \eqdef 
        \hat{\eta}\big(w,(u^n_{i,j},z^n_{i,j})_{i}^{N_n,Q_n}\big)
    \eqdef
            \sum_{i=1}^{N_n}\,
                [P_{\Delta_{I_n}}(w)]_i
    \,
            \biggl(
                \sum_{j=1}^{Q_n}\,
                    [P_{\Delta_{Q_n}}(u^n_{i,j})]_j\,
                    \delta_{\tilde{y}_{\lceil 
                                    z_{i,j}^n 
                              \rceil}}
            \biggr)
,
\]
where $Z_n\eqdef (u_{i,j}^n,z_{i,j}^n)_{i,j=1}^{N_n,Q_n} \in \mathbb{R}^{2 N_n Q_n}$, $D_n = 2 Q_n$, and $\{\tilde{y}_i\}_{i\in \nn_+}$ is a specified dense subset in $(\tilde{\yyy},d_{\tilde{\yyy}})$.  
Together~\eqref{eq:theorem:Unstructured_Case__tildeTdefined} and~\eqref{eq:quantizedmixing_Wasserstein} imply that for any $x\in \xxx_{\delta}$, the Borel function $\hat{T}$ is of the form
\allowdisplaybreaks
\begin{align}
\nonumber
        \hat{T}(x) 
    \eqdef &
        \sum_{n\le N^\star}\,
            \psi_n(x)
            \,
            \hat{\eta}\big(
                \hat{f}_n\circ \phi^{(n)}(x)
                ,
                Z_n
            \big)
    \\
    \nonumber
    =& 
        \sum_{n\le N}\, 
            \psi_n(x)
            \,
                \sum_{i=1}^{N_n}\,
                [P_{\Delta_{N_n}}(
                    \hat{f}_n\circ \phi^{(n)}(x)
                )]_i
                \,
                    \biggl(
                        \sum_{j=1}^{Q_n}\,
                            [P_{\Delta_{Q_n}}(u^n_{i})]_j\,
                            \delta_{\tilde{y}_{\lceil 
                                            z_{i,j}^n 
                                      \rceil}}
                    \biggr)
\end{align}
Combining the estimates in~\eqref{eq:PROOF__theorem:Unstructured_Case__LipschitzRedux} and~\eqref{eq:corollary_Unstructured_Case__partwise_estimate___simplified} implies that
\allowdisplaybreaks
\begin{align}
\nonumber
    \sup_{x\in \xxx_n\cap K}\, 
        W_1\big(
            \hat{\eta}\big(
                \hat{f}_n\circ \phi^{(n)}(x), Z_n
            \big)
        ,
            \delta_{f(x)}
        \big)
    \le &
        \sup_{x\in \xxx_n\cap K}\, 
        W_1\big(
            \hat{\eta}\big(
                \hat{f}_n\circ \phi^{(n)}(x), Z_n
            \big)
        ,
            F(x)
        \big)
    \\
    \nonumber
    &+
        \sup_{x\in \xxx_n\cap K}\, 
        W_1\big(
            F(x)
        ,
            \delta_{f(x)}
        \big)
    \\
    = &
        \sup_{x\in \xxx_n\cap K}\, 
        W_1\big(
            \hat{\eta}\big(
                \hat{f}_n\circ \phi^{(n)}(x), Z_n
            \big)
        ,
            F(x)
        \big)
    \\
    \nonumber
    & +
        \max_{x\in K}\,
        d_{\yyy}\big(
            f(x)
        ,
            \tilde{f}(x)
        \big)
    \\
    \nonumber
    < & \epsilon.
\end{align}   
Finally, if $f$ is a (generalized) $\omega$--H\"older function, we can indeed just take $d_K = d_{\xxx}^\omega$ in this case so that $f: (\xxx, d_{\xxx}^\omega) \to \yyy$ is a Lipschitz function. If  $N=1$, when there is a feature decomposition of $\xxx$ with only one chart, then, as in the proof of Lemma \ref{lemma:determinsitic_transferprinciple}, we have $N^\star = 1$ and $\xxx_\epsilon = \xxx$ for any $\epsilon > 0$. This concludes the proof.  
\end{proof}

\begin{proof}[{Proof of Theorem~\ref{theorem:Unstructured_Case}}]
The setting of Theorem \ref{theorem:Unstructured_Case} corresponds to the setup in Lemma \ref{lemma:Unstructured_Case} above with $N=1$ and without the assumptions on the doubling space property of $(K, d_K)$. However, noting that the doubling space assumption is only needed for the quantitative estimates of the upper bounds of parameters $\{c_n,d_n, N_n, D_n\}_{n \le N}$, the ``only qualitative'' Theorem \ref{theorem:Unstructured_Case} follows immediately from Lemma \ref{lemma:Unstructured_Case} by taking $N=1$ and any probability measure $\mu$ compactly supported by the compact set $K$.

On the other hand, 
we note that the set $\mathcal{K}$ of all non-empty finite subsets $\xxx$ is dense in the space of compact subsets of $(\xxx,d_{\xxx})$ equipped with the Hausdorff metric; that is, $\mathcal{K}$ is typical.  Since every finite metric space has the doubling property, for every $K\in \mathcal{K}$, $K$ is a doubling subset for the metric $d_K$ and $\phi(K)$ is a doubling subset of $F$.  
Then for any compact subset $\tilde{K} \subset \xxx$, any continuous function $f: (\tilde{K}, d_{\xxx}) \to (\yyy,d_{\yyy})$  and any $\epsilon > 0$, using the denseness of $\mathcal{K}$ in the space of compact subsets of $(\xxx,d_{\xxx})$ with respect to the Hausdorff metric, one can find a $K_\epsilon \in \mathcal{K}$ with $K_\epsilon \subset \tilde{K}$ such that $\sup_{x \in \tilde{K}} \inf_{x^\prime \in K_\epsilon} d_{\yyy
}(f(x), f(x^\prime)) \le \epsilon$. Now, applying the quantitative part of Lemma \ref{lemma:Unstructured_Case} to $f\vert_{K_\epsilon}$, we can find a Borel mapping $\hat{T}_\epsilon : K_\epsilon \to \mathcal{P}_1(\yyy)$ with $\sup_{x \in K_\epsilon}W_1(\hat{T}_\epsilon (x), \delta_{f(x)}) \le \epsilon$ and obtain quantitative estimates of the parameters $(c,d,N,D)$ used in the construction of $\hat{T}_\epsilon$ (as now $K_\epsilon$ is a doubling set).  Thus we also obtain a ``quantitative version'' of Theorem \ref{theorem:Unstructured_Case}.
\end{proof}

\subsection{{Proof of Theorem~\ref{theorem:Structured}}}
\label{s:Proofs__ss:quantitative}

\begin{proof}[{Proof of Theorem~\ref{theorem:Structured}}]
Denote the support of $\mu$ by $K$. We fix an ``approximation error'' $\epsilon_A>0$, an ``encoding error'' $\epsilon_E>0$ and a quantization error $\epsilon_Q>0$; we let a confidence level $\delta > 0$ be given.
We define the ``modified approximation error'',
\[
        {\epsilon_{\star:A}} 
    \le 
        \min \{\epsilon_A,\inf_{n\le N}\, \operatorname{diam}(\xxx_{n}),1/2\},
\]
which will be specified later. We note that $\epsilon_{\star:A}>0$ can be satisfied because of Definition~\ref{defn:CombinatorialStructure_X} (iv). We fix a positive integer $n\le N$, which we will choose later. 
\paragraph{Step 1 - Removing a Pathological Low-Probability Region in $K$:}
Definition~\ref{defn:CombinatorialStructure_Y} (i) implies that $\{(\yyy_m^{\delta},\bar{y}_m)\}_{m\le M}$ is a pairwise disjoint family of subsets of $\yyy$ whenever $\delta \in [0,1)$. The $\eta$-convexity of $\yyy_m$ implies that for every $\delta \in [0,1]$, $\yyy_m^{\delta}\subseteq \yyy_m$.  Consequentially, $K\cap f^{-1}[\yyy_m^{\delta}]\subseteq K\cap f^{-1}[\yyy_m]$ for each $m\le M$ and every $\delta \in [0,1]$. As ${\epsilon_{\star:A}} \in (0,1/2]$, we may define
\[
    \delta_{\star}
        \eqdef 
    \max_{m\le M}\,
    \sup\Big\{
        \delta \in [0,1
            -
            {\epsilon_{\star:A}}
        ]
        :\,
            \inf_{y\in (\yyy_m^{\delta})_{\epsilon_{\star:A}}}\,
            d_{\yyy}(y,\yyy_m^c)
            \ge 
            \epsilon_{\star:A}
        \mbox{ and }
            (\yyy_m^{\delta})_{\epsilon_{A:\star}}
                \subseteq 
            \yyy_m
    \Big\}.
\]
We note that the continuity of map $\boldsymbol{S}$ in Definition~\ref{defn:CombinatorialStructure_Y} implies that $\delta_{\star}$ exists. In fact, we can choose $\delta_\star = \boldsymbol{S}^\dagger(3 \epsilon_{\star:A})$ close to $1$ for the map $\boldsymbol{S}$.  

Upon applying Lemma~\ref{lemma:hp_partition_of_unity}, we have that for every $\tilde{n}\le n$, the partition of unity map
\[
    \Pi^{(\tilde{n})}(x)
        \eqdef 
    \frac{
        d_{\xxx}(x,\xxx_{\tilde{n}}^c)
    }{
        \sum_{i\le n} \,
            d_{\xxx}(x,\xxx_{i}^c)
    }
\]
is well-defined on $x\in \xxx^{(n,{\epsilon_{\star:A}})}$. We will work on the compact set $K_{\epsilon_A}$ defined by
\begin{equation}
\label{eq__theorem_transferprinciple___definition_of_good_set__Kepsilon}
    K_{\epsilon_{\star:A}}
        \eqdef 
    \xxx^{(n,{\epsilon_{\star:A}})} 
        \bigcap 
    \biggl(
        \bigcup_{m\leq M} f^{-1}\big[\yyy_m^{{\delta_{\star}}}\big]
        \bigcap K
    \biggr)
,
\end{equation}
where the set $\xxx^{(n,\epsilon_{\star:A})}$ is defined in~\eqref{lemma:hp_partition_of_unity__GoodSet}.  
The probability of $K_{\epsilon_{\star:A}}$ can be approximated from below as follows.  Since $\mu$ is a probability measure with $K = \operatorname{supp}(\mu)$, it holds that 
\begin{align}
\nonumber
        \mu\big(
            K_{\epsilon_{\star:A}}
        \big)
    = &
        \mu\Big(
            \xxx^{(n,{\epsilon_{\star:A}})} 
                \bigcap 
            \biggl(
                \bigcup_{m\leq M} f^{-1}\big[\yyy_m^{{\delta_{\star}}}\big]
            \biggr)
    \Big)
    \\
    = &
        \mu\biggl(
                \xxx \cap 
                \bigcup_{m\leq M} f^{-1}\big[\yyy_m^{{\delta_{\star}}}\big]
        \biggr)
    -
        \mu\biggl(
            \big[
                \xxx \setminus \xxx^{(n,{\epsilon_{\star:A}})} 
            \big]
                \bigcap 
            \Big[
                \bigcup_{m\leq M} \,
                f^{-1}\big[\yyy_m^{{\delta_{\star}}}\big]
            \Big]
        \biggr)
    \\
    \label{eq:estimating_probability_badset__excision}
    = &
        \mu\biggl(
                \bigcup_{m\leq M} f^{-1}\big[\yyy_m^{{\delta_{\star}}}\big]
        \biggr)
    -
        \mu\biggl(
            \bigcup_{m\leq M} \,
            \Big[
                f^{-1}\big[\yyy_m^{{\delta_{\star}}}\big]
            \cap
                \big[
                    \xxx \setminus \xxx^{(n,{\epsilon_{\star:A}})} 
                \big]
            \Big]
        \biggr).
\end{align}
Because $\{f^{-1}[\yyy_m^{{\delta_{\star}}}]\}_{m\le M}$ is a disjoint family of subsets of $\xxx$, we can rewrite the right-hand side of~\eqref{eq:estimating_probability_badset__excision} in the form
\allowdisplaybreaks
\begin{align}
    \nonumber
        \mu\big(
            K_{\epsilon_{\star:A}}
        \big)
    = &
        \sum_{m\leq M} 
        \,
        f_{\#}\mu\bigl(
            \yyy_m^{{\delta_{\star}}}
        \bigr)
    -
        \mu\biggl(
            \bigcup_{m\le M}
            \,
            \big[
                    f^{-1}\big[\yyy_m^{{\delta_{\star}}}\big]
                \cap
                    \big[
                        \xxx \setminus \xxx^{(n,{\epsilon_{\star:A}})} 
                    \big]
            \big]
        \biggr)
    \\
    \nonumber
    \ge &
        \sum_{m\leq M} 
        \,
        f_{\#}\mu\bigl(
            \yyy_m^{{\delta_{\star}}}
        \bigr)
    -
        \mu\bigl(
            \xxx \setminus \xxx^{(n,{\epsilon_{\star:A}})} 
        \bigr)
    \\
    \nonumber
    = &
        \sum_{m\leq M} 
        \,
            f_{\#}\mu\bigl(
                \yyy_m
            \bigr)
            -
        \sum_{m\leq M} \,
            f_{\#}\mu\bigl(
                \yyy_m 
                    \setminus 
                \yyy_m^{{\delta_{\star}}}
            \bigr)
    -
        \mu\bigl(
            \xxx \setminus \xxx^{(n,{\epsilon_{\star:A}})} 
        \bigr)
    \\
    \label{eq:estimating_probability_badset__partitioning}
    = &
        1
            -
        \sum_{m\leq M} \,
            f_{\#}\mu\bigl(
                \yyy_m 
                    \setminus 
                \yyy_m^{{\delta_{\star}}}
            \bigr)
    -
        \mu\bigl(
            \xxx \setminus \xxx^{(n,{\epsilon_{\star:A}})} 
        \bigr)
.
\end{align}
Since $g$ is geometrically stable (Definition~\ref{defn_con_morphi}) then there are constants $C_0,q_1,\dots,q_M>0$ (depending only on $\mu$, $f$, and on $\{\yyy_m,\bar{y}_m\}_{m\le M}$) such that the right-hand side of \eqref{eq:estimating_probability_badset__partitioning} can be re-expressed as
\allowdisplaybreaks
\begin{align}
        \mu\big(
            K_{\epsilon_{\star:A}}
        \big)
    \ge &
        1
            -
        C_0
        \,
        \sum_{m\leq M} \,
            (1-\delta_{\star})^{q_m}\,
	        f_{\#}\mu(\yyy_m)
    -
        \mu\bigl(
            \xxx \setminus \xxx^{(n,{\epsilon_{\star:A}})} 
        \bigr)
.
\end{align}
It remains to control the term $\mu\big(\xxx \setminus \xxx^{(n,{\epsilon_{\star:A}})}\big)$. Definition~\eqref{defn:CombinatorialStructure_X} (i)-(iii) imply the following union-bound (that is, subadditivity of measure)
\allowdisplaybreaks
\begin{align}
\label{eq:estimating_probability_badset__featuredecompositiondisjoitness___BEGIN}
        \mu\big(
            \xxx \setminus \xxx^{(n,{\epsilon_{\star:A}})}
        \big)
   \le &\,
        \mu\biggl(
            \bigcup_{i>n}\,
               \xxx_i
        \biggr)
    +
       \sum_{i\le n}\,
           \mu\big(
               (\partial \, \xxx_i)_{\epsilon_{\star:A}} \cap \xxx_i
            \big)
   \\
    \le &
        C_1\, 
        n^{-r}
    +
        \sum_{i\le n}\,
            C_2\,
            \mu\big(
                \xxx_i
            \big)
            \,
            \epsilon_{\star:A}^{r_i}
\label{eq:estimating_probability_badset__featuredecompositiondisjoitness___END}
,
\end{align}
where $C_1,C_2>0$ are constants depending only on $\mu$ and on the feature decomposition $\{(\xxx_n,\phi_n)\}_{n \le N}$ of the metric measure space $(\xxx,d_{\xxx},\mu)$.  
Incorporating~\eqref{eq:estimating_probability_badset__featuredecompositiondisjoitness___BEGIN}-~\eqref{eq:estimating_probability_badset__featuredecompositiondisjoitness___END} into~\eqref{eq:estimating_probability_badset__geometricstability} implies that
\begin{align}
\label{eq:estimating_probability_badset__geometricstability}
        \mu\big(
            K_{\epsilon_{\star:A}}
        \big)
    \ge &
        1
            -
        C_0
        \,
        \sum_{m\leq M} \,
            (1-\delta_\star)^{q_m}\,
	        f_{\#}\mu(\yyy_m)
    -
        C_1\, 
        n^{-r}
    -
        \sum_{i\le n}\,
            C_2\,
            \mu\big(
                \xxx_i
            \big)
            \,
            \epsilon_{\star:A}^{r_i}
.
\end{align}
Clearly, as $\epsilon_{\star:A} \to 0$ and $n \to N$ (noting that if $N < \infty$, then for $n = N$ the term $C_1n^{-r}$ will vanish), we have $\delta_\star = \boldsymbol{S}^\dagger(3\epsilon_{\star:A}) \to 1$ and $n^{-r} \to 0$, which ensures that $$1
            -
        C_0
        \,
        \sum_{m\leq M} \,
            (1-\delta_\star)^{q_m}\,
	        f_{\#}\mu(\yyy_m)
    -
        C_1\, 
        n^{-r}
    -
        \sum_{i\le n}\,
            C_2\,
            \mu\big(
                \xxx_i
            \big)
            \,
            \epsilon_{\star:A}^{r_i} \to 1.$$
Hence, for any given confidence level $\delta \in (0,1)$, we can find a $\epsilon_{\star:A} \le 
        \min \{\epsilon_A,\inf_{n\le N}\, \operatorname{diam}(\xxx_{n}),1/2\} $ small enough and an $N^\star \le N$ large enough such that 
$\mu\big(
            K_{\epsilon_{\star:A}}
        \big) \ge 1 - \delta$ for $K_{\epsilon_{\star:A}}  \eqdef 
    \xxx^{(N^\star,{\epsilon_{\star:A}})} 
        \bigcap 
    \bigl(
        \bigcup_{m\leq M} f^{-1}\big[\yyy_m^{{\delta_{\star}}}\big]
        \bigcap K
    \bigr) $. From now on let us fix such $\epsilon_{\star:A}$, $N^\star$ and $K_{\epsilon_{\star:A}}$. 
\paragraph{Step 2 - Building the Approximate Partition of Unity, $\hat{C}$, on $f(K_{\epsilon_{\star:A}})$:}
We first observe that $\big([0,\infty),\vert\cdot\vert\big)$ is quantized by the family of maps $\{\mathcal{Q}_q\}_{q\in \mathbb{N}_+}$, where $\mathcal{Q}_q:\mathbb{R}^{D_q}\eqdef \mathbb{R}\ni z \mapsto \vert z \vert\in [0,\infty)$ and 
$\boldsymbol{\eta}(w,(z_k)_{k=1}^N) \eqdef \sum_{k=1}^N\, w_k\,z_k$ defines a mixing function thereon. Thus, $([0,\infty),\vert \cdot \vert)$ can be endowed with the structure of a QAS space with quantized mixing function $\hat{\boldsymbol{\eta}}$ given for any positive integer $k$ and any $w\in \mathbb{R}^k\times [0,\infty)^k$ by
\[
    \hat{\boldsymbol{\eta}}(w,z_1,\dots,z_k)\eqdef \sum_{i=1}^k\, w_i\,\vert z_i \vert
.
\]
We recall the notation $\xxx_{\epsilon}$ in Lemma~\ref{lemma:determinsitic_transferprinciple} for $0<\epsilon < \inf_{n \le N} \operatorname{diam}(\xxx_n)$.  Next, we observe that for each $m\le M$, the map $C_m:\xxx_{\epsilon}\rightarrow \big([0,\infty),\vert\cdot\vert\big)$ defined by
\[
    C_m(x)\eqdef d_{\yyy}\big(f(x),\yyy_m\big),
\]
is the composition of a $1$-Lipschitz map (namely $y\mapsto d_{\yyy}\big(y,\yyy_m\big)$) with the $\omega$-H\"{o}lder-like map, $f$. Hence, each $C_m$ is itself an $\omega$-H\"{o}lder map.  Therefore, we may independently apply Lemma~\ref{lemma:determinsitic_transferprinciple} to functions $C_1, \ldots, C_M$ once for each $m=1,\dots,M$, to conclude that there are maps $\hat{C}_1$, $\dots,$ $\hat{C}_M:\xxx_{\epsilon}\rightarrow \big([0,\infty),\vert\cdot\vert\big)$ each with representations,
\begin{equation}
\label{eq:representation_hatC}
    \hat{C}_m(x)
        =\sum_{n\le N^{\star}}\,
        \sum_{i\le \tilde{N}^{(m)}_n}\,
        \frac{d_{\xxx}(x,\xxx_n^c)}{\sum_{j\le N^\star}\, d_{\xxx}(x,\xxx_j^c) }\,
        [P_{\Delta_{\tilde{N}^{(m)}_n}}(g_n^{(m)}\circ \tilde{\phi}^{(m)}_n(x))]_i\, \vert z^{(m,n)}_i \vert
,
\end{equation}
where each $g_n^{(m)} \in \fff_{\tilde{d}_n^{(m)}, \tilde{N}_n^{(m)}, \tilde{c}_n^{(m)}}$, $z_i^{(m,n)} \in \mathbb{R}$, $N^{\star}$ is defined as in Step 1, and for each $m\le M$ the ``approximate feature maps'' $\tilde{\phi}^{(m)}_n:\xxx_n\rightarrow (\mathbb{R}^{\tilde{d}_n^{(m)}},\|\cdot\|_{F_n:\tilde{d}_n^{(m)}})$ are defined by
\begin{equation}
\label{eq:compressedfeaturemap_2}
          \tilde{\phi}^{(m)}_n
    \eqdef 
    	\iota_{
    	    T^{(n)}_{\tilde{d}_n^{(m)}}
    	}^{-1}
    \circ 
    	T^{(n)}_{\tilde{d}_n^{(m)}}
    \circ 
        \phi_n
,
\end{equation}
such that, the following uniform estimate holds
\begin{equation}
\label{eq:uniformeClassificationEstimate}
    \max_{m\le M}\,
        \sup_{x\in \xxx_{\epsilon}}\,
        \big\vert
            C_m(x)
                -
            \hat{C}_m(x)
        \big\vert
    <
        2^{-3}\epsilon_A
.
\end{equation}
Because all of mappings $C_m(\cdot)$, $m \le M$ are $\omega$--H\"older continuous, the estimates for the family of parameters $(\tilde{c}^{(m)}_n, \tilde{d}^{(m)}_n, \tilde{N}^{(m)}_n)$ satisfy the same upper bounds specified in (i), (ii) and (iv) in Lemma~\ref{lemma:determinsitic_transferprinciple} for all $m\le M$, where we may set the encoding error $\tilde{\epsilon}_E$, the approximation error $\tilde{\epsilon}_A$ and quantization error $\tilde{\epsilon}_Q$ for $\hat{C}_m$ by $\tilde{\epsilon}_E = \tilde{\epsilon}_A = \tilde{\epsilon}_Q = \frac{1}{3}2^{-3}\epsilon_A$.
Since $\{\yyy_m^{\delta_{\star}}\}_{m\le M}$ forms a disjoint closed cover of $f(\xxx_{\epsilon}\cap \bigcup_{m\le M}\,f^{-1}[\yyy_m^{\delta_{\star}}])$ then for every $x\in \xxx_{\epsilon}\cap \bigcup_{m\le M}\,f^{-1}[\yyy_m^{\delta_{\star}}]$ there is exactly one integer $1\le m\le M$ for which $C_m(x)$ vanishes.  Moreover, the estimate in~\eqref{eq:uniformeClassificationEstimate} implies that: if $\hat{C}_m(x)\le 2^{-2}\epsilon_A$ then $C_m(x)\le 2^{-1}\epsilon_A$.  
However, by the fact that no point in any distinct $\yyy_{\tilde{m}}^{\delta_{\star}}$ and $\yyy_m^{\delta_{\star}}$ at $d_{\yyy}$-distance of less than $\epsilon_A/2$ \textit{(part of the definition of $\delta_{\star}$)}, and note that $\epsilon_{\star:A} \le \epsilon_A$, we deduce that for every $x\in \xxx_{\epsilon}\cap \cup_{m\le M}\,f^{-1}[\yyy_m^{\delta_{\star}}]$
\begin{equation}
\label{eq:ClassifyingCm_inclusionbound}
  \hat{C}_m(x) \le 2^{-2} \epsilon_A \Rightarrow  C_m(x)\le 2^{-1}\epsilon_A \iff f(x) \in (\yyy_m^{\delta_{\star}})_{2^{-1}\epsilon_A} 
        \subseteq 
            \yyy_m^{\delta_{\star}}
\end{equation}
holds.  
Since the estimate~\eqref{eq:uniformeClassificationEstimate} implies that for every $\tilde{m}\le M$ distinct from $m$ we have that $C_{\tilde{m}}(x)>2^{-1}\epsilon_A$, wherefrom we conclude that
\begin{equation}
\label{eq:ClassifyingCm_exclusionbound}
   \hat{C}_m(x) \le 2^{-2} \epsilon_A \Rightarrow  C_m(x)\le 2^{-1}\epsilon_A \iff f(x) \not\in 
    \bigcup_{
        \underset{\tilde{m}\neq m}{\tilde{m}\le M}
    }\, 
    (\yyy_{\tilde{m}}^{\delta_{\star}})_{2^{-1}\epsilon_A} 
.
\end{equation}
Together,~\eqref{eq:ClassifyingCm_inclusionbound} and~\eqref{eq:ClassifyingCm_exclusionbound} imply that the map $\hat{C}:\xxx_{\epsilon}\rightarrow \big(\Delta_M,\|\cdot\|_1\big)$ defined by
\[
\hat{C}(x) \eqdef \biggl(
    \frac{
        I\big(\hat{C}_m(x)\le 2^{-2}\epsilon_A\big)
    }{
        \sum_{\tilde{m}\le M}\, 
            I\big(\hat{C}_{\tilde{m}}(x)\le 2^{-2}\epsilon_A\big)
    }
\biggr)_{m\le M}
\]
is such that $\hat{C}(x)_m=1$ if and only if $f(x)\in \yyy_m^{\delta_{\star}}$, whenever $x\in \xxx_{\epsilon}\cap \cup_{m\le M}\,f^{-1}[\yyy_m^{\delta_{\star}}]$.  
We also note that exactly one component of $\hat{C}(x)$ is non-zero, for any $x\in K_{\epsilon_{\star:A}} \subset \xxx_{\epsilon}$ for some suitable $\epsilon$.  

\paragraph{Step 3 - Approximating $f$ on Pieces of $\xxx$ and $\yyy$ and Assembling to Approximator via $\hat{C}$:}
Since each $\yyy_m$ was $\eta$-convex, then the restriction of the mixing function $\eta$ to each $\bigcup_{k \in \mathbb{N}_+}\, (\yyy_m^k \times \Delta_k)$ is a mixing function on $\yyy_m$.  Therefore, for every positive integer $m\le M$, $(\yyy_m,d_{\yyy}\vert_{\yyy_m\times \yyy_m},\hat{\eta}\vert_{\bigcup_{k \in \mathbb{N}_+}\, (\yyy_m^k \times \Delta_k)})$ is a barycentric QAS space, which by a mild abuse of notation we denote by $(\yyy_m,d_{\yyy},\hat{\eta})$. As a consequence, Lemma~\ref{lemma:determinsitic_transferprinciple} applied to each $\omega$--H\"older continuous function $f\vert_{K\cap f^{-1}[\yyy_m]}$ implies that there are maps $\hat{T}^{(m)}:\xxx\rightarrow \mathcal{P}_1(\yyy_m)\subseteq \mathcal{P}_1(\yyy)$, $m \le M$, satisfying
\begin{equation}
\label{EQ_PROOF__theorem_transferprinciple__GeneralCase___individualestimates}
        \sup_{x\in K \cap f^{-1}[\yyy_m]\cap \xxx^{(N^\star,\epsilon_{\star:A})}}\, 
        W_1\big(
            \hat{T}^{(m)}(x)
                ,
            \delta_{f(x)}
        \big)
    <
        \epsilon_A + \epsilon_Q + \epsilon_E
        ,
\end{equation}
where we have used the fact that each $K\cap f^{-1}[\yyy_m]$ is compact (because $f$ is continuous, $K$ is compact, and $\yyy_m$ is closed).  
Again in view of Lemma \ref{lemma:determinsitic_transferprinciple}, we know that on $\xxx^{(N^\star,\epsilon_{\star:A})}$ the mapping $\hat{T}^{(m)}$ admits the following representation,
$$
\hat{T}^{(m)}(x) = \sum_{n\le N^{\star}}\,
                [C^{\xxx}(x)]_n\,
                    \delta_{f^{(n,m)}(x)},
$$
where $[C^{\xxx}(x)]_n = 
        			\frac{d_{\xxx}(\cdot,\xxx_n^c)}{
        			\sum_{i\le N^{\star}} \, d_{\xxx}(\cdot,\xxx_i^c)}$, 
$f^{(n,m)}(x) = \hat{\eta}(
            	            \hat{f}^{(m)}_n
            	        \circ 
            	            \phi^{(m)}_n
            	            (x)
            	            ,
            	        Z^{(m)}_n)$ for
$\phi^{(m)}_n = \iota_{T^{(n)}_{d_n^{(m)}}
                	}^{-1}
                \circ 
                	T^{(n)}_{d_n^{(m)}}
                \circ
                    \phi_n$,
$\hat{f}^{(m)}_n\in \mathcal{F}_{d_n^{(m)},N^{(m)}_n,c^{(m)}_n}$, 
    $Z^{(m)}_n \in \mathbb{R}^{N^{(m)}_n\times D^{(m)}_n}$, 
and the family of parameters $(c^{(m)}_n, d^{m}_n, N^{(m)}_n, D^{(m)}_n)$ can be estimated by the upper bounds obtained in Lemma \ref{lemma:determinsitic_transferprinciple} (i) -- (iv) for all $m \le M$.

We define the map $\tilde{T}:K_{\epsilon_{\star:A}}\rightarrow \mathcal{P}_1(\yyy)$ (noting that $K_{\epsilon_{\star:A}} \subset K \cap \xxx^{(N^\star,\epsilon_{\star:A})} \cap \cup_{m\le M}f^{-1}[\yyy_m]$) by
\begin{equation}
\label{EQ_PROOF__theorem_transferprinciple__Definition_Map}
    \tilde{T}(x)
        \eqdef 
    \sum_{m\le M}
	        \,
        [\hat{C}(x)]_m
            \,
    \hat{T}^{(m)}(x)
.
\end{equation}
As in the proof of the previous result, namely Lemma~\ref{lemma:Unstructured_Case}, we fix any $y_0\in \yyy$ and extend $\hat{T}$ of~\eqref{EQ_PROOF__theorem_transferprinciple__Definition_Map} to a Borel map $\hat{T}:\xxx\rightarrow \mathcal{P}_1(\yyy)$ by defining
\[
        \hat{T}(x) 
    \eqdef 
        \begin{cases}
            \tilde{T}(x) 
                &: 
                \mbox{if } x \in K_{\epsilon_{\star:A}}\\
            \delta_{y_0} 
                &: 
                \mbox{if } x \not\in K_{\epsilon_{\star:A}}
        .
        \end{cases}
\]
\paragraph{Step 4 - Verifying the PAC Approximation of $f$ by $\hat{T}$:}
	For every $x\in K_{\epsilon_{\star:A}}$, the following estimate holds (recalling that $\Phi$ denotes the isometric mapping from Lemma \ref{lemma_closedconvex_embedding_Wasserstein})
\allowdisplaybreaks
\begin{align}
    \label{eq_last_bound__BEGIN}
    	W_1\big(
        	    \hat{T}(x)
        	,
        	    \delta_{f(x)}
    	\big)
    = &
    W_1\Biggl(
    	\sum_{m\le M}
            \,
            [\hat{C}(x)]_m
            \,
            \hat{T}^{(m)}(x)
      ,
            \delta_{
                f(x)
            }
    	 \Biggr)
    \\
    \nonumber
    = &
    \Biggl\|
    	\sum_{m\le M}
            [\hat{C}(x)]_m
            \,
           \Phi \circ  \hat{T}^{(m)}(x)
    	-
          \Phi(  \delta_{
                f(x) )
            }
    \Biggr\|_{{\text{\AE}}(\yyy,y_0)}
    \\
    \nonumber
     = &
    \Biggl\|
            \,
          \sum_{m\le M}  [\hat{C}(x)]_m
            \,
        \Phi(    \hat{T}^{(m)}(x) )
    	-
            \sum_{m\le M}
            \,
                [\hat{C}(x)]_m
                \,
    	     \Phi(   \delta_{f(x)} )
    \Biggr\|_{{\text{\AE}}(\yyy,y_0)}
    \\
    \nonumber
    = &
    \Biggl\|
    	\sum_{m\le M}
    	\,
        [\hat{C}(x)]_m
            \,
    	\Phi\Biggl(
            \hat{T}^{(m)}(x)
    	-
            \delta_{
                f(x)
            }
      \Biggr)
    \Biggr\|_{{\text{\AE}}(\yyy,y_0)}
    \\
    \label{proof_Main:POU_AFTER}
    \le &
        \sum_{m\le M}\,
          [\hat{C}(x)]_m\,
        W_1\big(
          \hat{T}^{(m)}(x)
          ,
          \delta_f(x)
        \big)
    \\
    \label{finalbound_A}
    \le &
        \sum_{m\le M}\,
          I\big(f(x)\in \yyy_m^{\delta_{\star}}\big)
        \,
        W_1\big(
          \hat{T}^{(m)}(x)
          ,
          \delta_f(x)
        \big)
    \\
    \label{finalbound_B}
    = &
    \epsilon_A + \epsilon_Q + \epsilon_E.
\end{align}
Here, we passed from \eqref{proof_Main:POU_AFTER} to \eqref{finalbound_A} by using the equivalence relation $\hat{C}(x)_m = 1$ iff $f(x) \in \yyy^{\delta_\star}_m$, and passed from~\eqref{finalbound_A} to~\eqref{finalbound_B} by using the fact that $\{\yyy_m^{\delta_{\star}}\}_{m\le M}$ is a disjoint cover of $\bigcup_{m\le M}\, \yyy_m^{\delta_{\star}}$ together with each of the estimates in~\eqref{EQ_PROOF__theorem_transferprinciple__GeneralCase___individualestimates}.  

Finally, we set $\xxx_{\epsilon_{\star}:A}\eqdef K_{\epsilon_{\star}:A}$.  By Step 2, for any $x\in \xxx_{\epsilon_{\star}:A}$ and each $m=1,\dots,M$, we have that $\hat{C}_m(x)=1$ if and only of $f(x)\in \yyy_m^{\delta_{\star}}$.  By Step 1, $\yyy_m^{\delta_{\star}}\subseteq \yyy_m$ and the sets $\{\yyy_m^{\delta_{\star}}\}_{m=1}^M$ are disjoint.  Therefore, for any $x\in \xxx_{\epsilon_{\star}:A}$ we have that if $\hat{C}_m(x)=1$, then the measure $\hat{T}(x)$ is supported in $\yyy_m$.  Therefore $\beta_{\yyy_m}(\hat{T}(x))$ is well-defined and since $\beta_{\yyy_m}$ is a contracting barycenter map, the estimate in~\eqref{finalbound_A}-~\eqref{finalbound_B} implies that
\[
\begin{aligned}
        d_{\yyy}\big(
                \beta_{\yyy_m}\circ \hat{T}(x),
            ,
                \delta_{f(x)}
        \big)
    = &
        d_{\yyy}\big(
                \beta_{\yyy_m}\circ \hat{T}(x),
            ,
                \beta_{\yyy_m}(\delta_{f(x)})
        \big)
    \\
    \le &
        W_1\big(
        	    \hat{T}(x)
        	,
        	    \delta_{f(x)}
    	\big)
    \\
    \le &
        \epsilon_A + \epsilon_Q + \epsilon_E
.
\end{aligned}
\]
\end{proof}

\subsection{Proofs for Section~\ref{s:Applications__ss:InverseProblemsPDE___sss:DirichletToNeumanMap}}
\label{s:Proofs__ss:InverseProblems}
	
\begin{proof}[Proof of Lemma~\ref{lem:CompletenessofX}]
If $q_j\in \xxx$, $j=1,2,\dots$ is a Cauchy sequence in $\xxx$ then it converges to a limit $q\in H^{k,p}_0(M)$
as $H^{k,p}_0(M)$ a Banach space. However, as the $q_j$'s are uniformly bounded in $H^{k+1,p}_0(M)$,
by the Banach-Alaoglu theorem and the fact that $H^{k+1,p}_0(M)$ is a reflexive space, we have that
$q_j$ has a subsequence that converges weakly in $H^{k+1,p}_0(M)$ to a function $q'$.
This implies that $q'=q$ and, hence, $q\in H^{k+1,p}_0(M)$. We conclude that $q\in \xxx$ and thus $\xxx$ is a complete metric space.
\end{proof}
	
\begin{proof}[{Proof of Proposition~\ref{prop:DNmapisLipschitz}}]
Let $2\le r\le k-p/n$  be an integer and $f\in H^{r}_0(\p M\times (0,T))$.
	Let $u_{q,f}(x,t)$ be the solution of the wave equation
	\beq\label{bvp1}
	& &(\p_{t}^2-\Delta_g+q)u_{q,f}=0\quad\hbox{on }(x,t)\in  M \times \R_+,\\ 
 \label{bvp2}
	& &u_{q,f}\vert _{\p  M\times \R_+}=f(x,t),\\
 \label{bvp3}
	& &u_{q,f}\vert _{t=0}=0,\quad \p_tu_{q,f}\vert _{t=0}=0.
	\eeq
	 By Sobolev's embedding theorem, the identical embedding, $I : H^{k,p} (M)\to C^{r}(\overline M)$
	is bounded and hence $q\in C^{r}(\overline M)$.
	By \cite[Sec. 2.8]{LaLiTr}, see also \cite{KKL}, Theorem 2.45 and its proof,
	$\mathcal U_q:f\to u_{q,f}$ is a bounded linear map
	$$
	\mathcal U_q:H^{r}_0(\p M\times (0,T))\to 
	Z^r(M\times [0,T])\eqdef\bigcap_{l=0}^{r} C^l([0,T];H^{r-l}(M))\subset H^r(M\times (0,T)).
	$$
	Moreover, there is a constant $C_0$ such that for all integers $s$, $1\le s\le r$, 
		\beq\label{first a priori estimate}
	\|u_{q,f}\|_{Z^{s}(M\times [0,T])}\le C_0\|f\|_{H^{s}_0(\p M\times (0,T))}, \quad \hbox{for all }q\in \xxx,
	\eeq
	that is, the norm of the map $ \mathcal U_q$ is bounded by $C_0$
	for all $q\in \xxx$.
We emphasize that as $r\ge 2$ and $f\in H^{r}_0(\p M\times (0,T))$,
the solutions of the initial boundary value problem \eqref{bvp1}-\eqref{bvp3} satisfy
the compatibility conditions
\beq
\p_t^{2j+i}f(x,t)\bigg|_{t=0}=(\Delta_g-q)^j\p_t^iu_{q,f}(x,t)\bigg|_{t=0}=0,\quad \hbox{for }x\in \p M
\eeq
for $j\in \mathbb N$, $i\in\{0,1\}$ such that $2j+i\leq r-1$.
 
 Also,  by \cite[Sec. 2.8]{LaLiTr} or \cite{KKL}, Theorem 2.45,
	there is a $c_1>0$ such that 
		\beq\label{DN Map a priori estimate}
		\|\Lambda_{q}\|_{H^r_0 (\p M\times (0,T))\to H^{r-1} (\p M\times (0,T))}\le c_1\eeq
		 for  all $q\in \xxx$.
	
	We next consider $f\in H^{r}_0(\p M\times (0,T))$.
	For the functions
	 $q_1$ and $q_2$ the difference $w_{q_1,q_2,f}=u_{q_1,f}-u_{q_2,f}$ of solutions satisfies
	\ba
	& &(\p_{t}^2-\Delta_{g}+q_1)w_{q_1,q_2,f}=(q_2-q_1)u_{q_2,f}
	\quad\hbox{on }(x,t)\in  M \times \R_+,\\
	& &w_{q_1,q_2,f}\vert _{\p  M\times \R_+}=0,\\
	& &w_{q_1,q_2,f}\vert _{t=0}=0,\quad \p_tw_{q_1,q_2,f}\vert _{t=0}=0.
	\ea
	As $q_j\in \xxx\subset W^{k+1,p}_0(M)$, we observe that 
 \beq
 F_{q_1,q_2,f}=(q_2-q_1)u_{q_2,f}\in 
 Z^r_0(M\times [0,T])\eqdef\bigcap_{l=0}^{r} C^l([0,T];H^{r-l}_0(M)).
 \eeq
 By applying again  \cite[Sec. 2,8]{LaLiTr} or \cite{KKL}, Theorem 2.45,
	we observe that there are constants $C_1, C_2 > 0$ such that for all integers $1\le s\le r$
	\beq\nonumber
	\|u_{q_1,f}-u_{q_2,f}\|_{{Z^{s+1}(M\times [0,T])}}&\le &C_1\|F_{q_1,q_2,f}\|_{Z^{s}(M\times [0,T])}
	\\ \label{u difference}
	&\le& C_2\|q_2-q_1\|_{C^{r}(M)}\|f\|_{H^{s}_0(\p M\times (0,T))}
	\eeq
	for all $q_1,q_2\in \xxx$. By  comparing \eqref{first a priori estimate} and \eqref{u difference}, we observe that the map from the boundary value to the difference of the solutions, $f\to u_{q_1,f}-u_{q_2,f}$,
	is one degree smoother than map from the boundary value to a solution, $f\to u_{q_j,f}$. 
	
	By equation \eqref{u difference} and the trace theorem in Sobolev spaces \cite[Sec. 4, Prop.\ 4.5]{Taylor1} imply that the normal derivative $\p_\nu(u_{q_1,f}-u_{q_2,f})\vert _{\p M\times (0,T)}=(\Lambda_{q_2}-\Lambda_{q_1})f$
	satisfies  for all integers $1\le s\le r$ and $f\in H^s_0(\p M\times (0,T))$,
	\beq\label{eq: Lambda difference}
	\|(\Lambda_{q_2}-\Lambda_{q_1})f\|_{H^{(s+1)-3/2}(\p M\times (0,T))}\le  C_2\|q_2-q_1\|_{C^{r}(M)}\|f\|_{H^{s}_0(\p M\times (0,T))}.
	\eeq 
	Thus, the above shows that when $r\ge 2$,  $k\ge r+ p/n$, and $1\le s\le r$,
	we have that
	\beq\label{Lambda lip}
	\|\Lambda_{q_2}-\Lambda_{q_1}\|_{H^{s}_0(\p M\times (0,T))\to H^{s-1/2}(\p M\times (0,T))}\le  C_2\|q_2-q_1\|_{\xxx}.
	\eeq
	Combining \eqref{DN Map a priori estimate} and \eqref{Lambda lip} we see that  the map $H$ 
	\beq\label{eq: previously claim (ii)}
	H:\xxx\to L(H^{r}_0(\p M\times (0,T)),H^{r-1}(\p M\times (0,T))), \quad  q\mapsto \Lambda_{q}
	\eeq
 is Lipschitz
	when
	$r\ge 2$ and
	$k\ge r+p/n$. 
	 
	As $ \Lambda_{q}\in L(H^{r}_0(\p M\times (0,T)),H^{r-1}(\p M\times (0,T)))\subset  L(H^{r}_0(\p M\times (0,T)),L^2(\p M\times (0,T)))$,
	 the operators 
  $N^{-1/2}:L^2(\p M\times (0,T))\to H^{r}_0(\p M\times (0,T))$ and 
  $\Lambda_{q}N^{-1/2}:L^2(\p M\times (0,T))\to L^2(\p M\times (0,T))$  are bounded for  $q\in \xxx$, and $N^{-1/2}$ belongs to $HS(L^2(\p M\times (0,T)), L^2(\p M\times (0,T)))$,  the
	formula
	\eqref{eq: previously claim (ii)} implies that the map  
	\beq
	\varphi:\xxx\to F_{\varphi} = HS(L^2(\p M\times (0,T)), L^2(\p M\times (0,T))), \quad q \mapsto (f  \mapsto \Lambda_{q}N^{-1}(f)),
	\eeq
	 is  Lipschitz. This proves claim  (i).
	
	It remains to show that $\varphi$ is injective and prove H\"older-continuity of its left-sided inverse.  
	As we saw in \eqref{eq: Lambda difference},
	for all  $s\in \mathbb Z_+$, $1\le s\le r$
	\beq\label{interpolation estimate1}
	S=S_{q_1,q_2}\eqdef\Lambda_{q_1}-\Lambda_{q_2}:H^{s}_0(\p M\times (0,T))\to H^{s-1/2}(\p M\times (0,T))
	\eeq
	is bounded and that its norm is bounded by  $C_2\|q_2-q_1\|_{\xxx}$. 
	To avoid below the difficulties related to interpolation with smoothness indexes $s-\frac 12$ that are in the set $\mathbb Z+\frac 12$, we make the trivial observation that by \eqref{interpolation estimate1}, for all $0<\epsilon<\frac 12$ and $s\in \mathbb Z_+$, $1\le s\le r$, 
	\beq\label{interpolation estimate1B}
	S=S_{q_1,q_2}\eqdef\Lambda_{q_1}-\Lambda_{q_2}:H^{s}_0(\p M\times (0,T))\to H^{s-1/2-\epsilon}(\p M\times (0,T))
	\eeq
	is a bounded operator which norm is bounded by  $C_2\|q_2-q_1\|_{\xxx}$. 

	Let $X'$ denote the dual space of a Hilbert space $X$.
	When we use the extension of the standard pairing of distributions and test functions to define dual spaces of Sobolev spaces, we have that  the dual space of $H_0^{s}(\p M\times (0,T))$, $s\ge 0$, is  $H^{-s}(\p M\times (0,T))$, see \cite[Sec.\ 4, Prop.\ 5.1]{Taylor1}. 
 Below, we use the shorthand notations $L^2(\p M\times (0,T))=L^2$, $H^s(\p M\times (0,T))=H^s$, and $H^s_0(\p M\times (0,T))=H^s_0.$ By the above, $(H^{-s})'=H^s_0$ for $s\ge 0$. To simplify
 the notations used in the interpolation theory below, we use also the notation
 \beq\label{duality}
H^s_*=\begin{cases} H^s_0, & s\ge 0,\\
(H^{-s})', & s< 0,\end{cases}\quad \hbox{for }s-\frac 12\not \in \mathbb Z.
 \eeq
	By \cite [p. 72]{Bingham}, the adjoint operator of the map $S:H^{1}_0(\p M\times (0,T))\to 
	H^{1-1/2-\epsilon}(\p M\times (0,T))$, denoted $S^*: H^{-1+1/2+\epsilon}_*(\p M\times (0,T))\to H^{-1}(\p M\times (0,T))$, satisfies
	\beq\label{Sstar}
	S^*f=RSRf,\quad \hbox{for } f\in  H^{1}_0(\p M\times (0,T)),
	\eeq
	where $Rf(x,t)=f(x,T-t)$ is the time reversal operator. 
	
	Since 
 the time reversal operator defines isometric maps $R:H_*^{s}(\p M\times (0,T))\to H_*^{s}(\p M\times (0,T))$ and
 $R:H^{s}(\p M\times (0,T))\to H^{s}(\p M\times (0,T))$ for $s\in \mathbb R$,
 we see using \eqref{interpolation estimate1B} with $s=1$ that the operator 
 \beq\label{interpolation estimate2C} 
 RS^*R:H^{-1+1/2+\epsilon}_*(\p M\times (0,T))\to H^{-1}(\p M\times (0,T))
 \eeq is bounded for $0<\epsilon<\frac 12$. 
 
 As $R^2=I$, formula \eqref{Sstar} implies
 $Sf=RS^*Rf$ for $f\in  H_0^{1}(\p M\times (0,T))$. 
 Since $H_0^{1}(\p M\times (0,T))$ is a dense subset of $H_*^{-1/2+\epsilon}(\p M\times (0,T))$, we see using  \eqref{interpolation estimate2C}  that $S=S_{q_1,q_2}$ can be extended in a unique way to a bounded operator 
	\beq\label{interpolation estimate2}
	S^e_{q_1,q_2}=RS^*R: H^{-1/2+\epsilon}_*(\p M\times (0,T))\to H^{-1}(\p M\times (0,T))
	\eeq
	which norm is bounded by  $C_2\|q_2-q_1\|_{\xxx}$, that is, $S^e_{q_1,q_2}|_{H_0^{r}(\p M\times (0,T))}=S$. 
	Observe that as the extension is unique, we have by  \eqref{DN Map def.} that $S^e_{q_1,q_2}f=(\Lambda_{q_2}-\Lambda_{q_1})f$ for $f\in {H^1_0 (\p M\times (0,T))}$.

 Next, our aim is to show that the feature map $\varphi$ is Lipschitz. To this end, we will combined the mapping properties of the differences of Dirichlet-to-Neumann maps, $\Lambda_{q_2}-\Lambda_{q_1}$, which are  determined by the formula
 \eqref{DN Map def.} in the space $H^r_0$ (where one looses one degree of smoothness) and by the formula
 \eqref{interpolation estimate2} in the space $H^{-1/2+\epsilon}_*$ (where one looses $\frac 12+\e$ degrees of smoothness). Below, these two inequalities are combined using interpolation of function spaces.

 Let $[X,Y]_\theta$ denote the interpolation space of Banach spaces $X$ and $Y$ obtained
	by using the complex interpolation methods with parameter $0\le \theta\le 1$, see \cite{BerghLofstrom_1976_InterpolationSpaces_Introduction} for the detailed definition and properties of the interpolation spaces.
	We recall that if a linear operator $B$  defines a bounded operator $B:X_0\to X_1$  and  $B:Y_0\to Y_1$ and
	for the Banach spaces $X_0,X_1,Y_0$ and $Y_1$ the interpolation spaces $[X_0,X_1]_\theta$
	and $[Y_0,Y_1]_\theta$ are defined for $0\le \theta\le 1$, then by  \cite[Sec.\ 3, Prop.\ 2.3]{Lions}, see also \cite{BerghLofstrom_1976_InterpolationSpaces_Introduction},
	see Sect.\ 2, formula (6), the linear operator $B:[X_0,X_1]_\theta\to [Y_0,Y_1]_\theta$ is bounded and
	\beq\label{basic interpolation}
	\|B\|_{[X_0,X_1]_\theta\to [Y_0,Y_1]_\theta}\le C'\|B\|_{X_0\to Y_0}^{1-\theta}\|B\|_{X_1\to Y_1}^{\theta}
	\eeq
 with an interpolation constant $C'>0$ that depends only on spaces $X_j$ and $Y_j$, $j=1,2$. 
	Next, we observe that the set $\p M\times (0,T)$ is diffeomorphic to an $h$-neighborhood
	of the set $\p M$ in $\R^n$ when $h>0$ is sufficiently small, and thus we can apply interpolation
	results concerning bounded subsets of $\R^n$ to $\p M\times (0,T)$. Next we recall some of these 
	results.
	By \cite[Thm.\ 11.6]{Lions}, for $s_1,s_2\ge 0$ and $0\le \theta\le 1$,
		\beq\label{inte 1}
	[ H^{s_1}_0,H^{s_2}_0]_{\theta}
	= H^{(1-\theta)s_1+\theta s_2}_0,\quad \hbox{when }(1-\theta)s_1+\theta s_2\not \in \mathbb Z+\frac 12,
	\eeq
 and 
 by \cite[Thm.\ 12.6]{Lions}, for $s_1,s_2\ge 0$, $s_1\not \in \mathbb Z+\frac 12$, 
 and $0\le \theta\le 1$,
		\beq\label{inte 1A}
	& &[ H^{s_1}_0,(H^{s_2})']_{\theta}
	= H^{\mu}_0,\quad \hbox{when }\mu=(1-\theta)s_1-\theta s_2\ge 0,\ \mu\not \in  \mathbb Z+\frac 12,
	\\  \label{inte 1B}
& &[ H^{s_1}_0,(H^{s_2})']_{\theta}
	= (H^{-\mu})',\quad \hbox{when }\mu=(1-\theta)s_1-\theta s_2\le 0.
 \eeq
 In other words, \eqref{inte 1}-\eqref{inte 1B} mean that
 \beq\label{inte 1D}
 [ H^{s_1}_*,H^{s_2}_*]_{\theta}=H^{\mu}_*\quad\hbox{for $s_1\in \R_+,$ $s_2\in \R$ and
 $\mu=(1-\theta)s_1+\theta s_2$, when $s_1,s_2,\mu\not \in \Z+\frac 12$.}
 \eeq
        By \cite[Theorems 9.6 and 12.4]{Lions} for $s_1>0$, $-\infty<s_2<s_1$, $s_2 \not \in \mathbb Z+\frac 12$ and $0\le \theta\le 1$,
	\beq\label{inte 2}
	[H^{s_1},H^{s_2}]_{\theta}=H^{(1-\theta)s_1+\theta s_2},\quad \hbox{when }(1-\theta)s_1+\theta \cdot s_2\not \in \mathbb Z+\frac 12.
		\eeq
		Moreover, by \cite[Def.\ 2.3]{Lions} or \cite[Sec. 4, Prop.\ 2.2]{Taylor1}, for $\theta_1=\frac 12$ 
	\beq\label{inte 3}
	[H^{2m}\cap H^{m}_0,L^2]_{\theta_1}=[\mathcal D(N),L^2]_{\theta_1}=\mathcal D(N^{1/2})
	= H^{m}_0.
	\eeq 
	Then, using
	by the  reiteration theorem for the complex interpolation method,
	see \cite{Calderon-interpolation}, 
	and \eqref{inte 1} and \eqref{inte 3}, we obtain for  $0\le s<m$ and $0<\theta<1$,  such that $(1-\theta)\cdot 2m+\theta \cdot s\le m$ and $(1-\theta)\cdot 2m+\theta \cdot s\not \in \mathbb Z+\frac 12$,
		\beq\label{inte reiterated}
	[H^{2m}\cap H^{m}_0,H^s_0]_{\theta}
	= H^{(1-\theta)\cdot 2m+\theta \cdot s}_0.
	\eeq
	Below, we use $C'$ to denote an interpolation constant which is the maximum
 of the interpolation constants for all the above used Sobolev spaces.

	Using interpolation of function spaces, see \eqref{basic interpolation},
 \eqref{inte 1D}, and \eqref{inte 2},
	as well as formulas \eqref{interpolation estimate1B} and \eqref{interpolation estimate2} , we see that the operator, $S^e_{q_1,q_2}$ is a bounded operator
	\beq\label{interpolation estimate}
	S^e_{q_1,q_2}: H^{s+\e}_0(\p M\times (0,T))\to H^{s-1/2}(\p M\times (0,T)),& & \hbox{for  $0<\e<\frac 12,$ $ s\ge 0,$ $s+\e\le r,$} \\
 \nonumber &&\hbox{such that $s-\frac 12 ,s+\epsilon\not \in \mathbb Z+\frac 12,$}
	\eeq
	  and its norm is bounded by  $C'C_2\|q_2-q_1\|_{\xxx}$.
    %
	%
	%
	%
	Next, we consider the stability of the solution of the inverse problem.
	Assuming that  $k>1$ is large enough and the metric $g$ satisfies \eqref{eps 0 condition} with  $\epsilon_0>0$ small enough,
 \cite{StefanovUhlmann_2005_Stabledetermination_metricfrom_DtNmap} yields that
 	the Dirichlet-to-Neumann map, $\Lambda_q$, determines $q$ in a H\"older stable way under a priori bounds on $q$ that we have assumed above,
	that is,  
	\beq\label{stability ip}
	\| q_1-q_2\|_{C(M)}\le C_0\|\Lambda_{q_2}-\Lambda_{q_1}\|^{\,\beta_1}_{H^1_0 (\p M\times (0,T))\to L^2 (\p M\times (0,T))},\quad \hbox{for }
	q_1,q_2\in \xxx
	\eeq
 with some $0<\beta_1<1$ and $C_0>0$.
	This implies that the map $\varphi:\xxx\to F_\varphi$ is one-to-one.  
	
	Next, we consider the map, ${\Psi}:{\varphi}(\xxx)\to \xxx$, that solves the inverse problem,
	\ba
	{\Psi}:\Lambda_qN^{-1}\to q.
	\ea
	As
	\ba
	F_{\varphi}=HS(L^2(\p M\times (0,T)), L^2(\p M\times (0,T)))\subset L(L^2(\p M\times (0,T)), L^2(\p M\times (0,T))),
	\ea
	we then have that
	\beq\label{Lambda and varphi estimate}
	\|\Lambda_{q_2}-\Lambda_{q_1}\|_{H^{2m}\cap H^{m}_0\to L^2} &\le &
	\|(\Lambda_{q_2}-\Lambda_{q_1})N^{-1}\|_{L^2\to L^2}\\ \nonumber
	&\le &  \|\varphi({q_2})-\varphi({q_1})\|_{L^2\to L^2}\\ \nonumber
	&\le &  \|\varphi({q_2})-\varphi({q_1})\|_{HS(L^2, L^2)}\\ \nonumber
	&\le &  \|\varphi({q_2})-\varphi({q_1})\|_{F_{\varphi}}.\nonumber
	\eeq
	As the identical embedding $H^s (\p M\times (0,T))\to  L^2(\p M\times (0,T))$, $s>0$ is continuous, the interpolation results for function spaces, see  \cite{BerghLofstrom_1976_InterpolationSpaces_Introduction},
	and \eqref{interpolation estimate} with $s_\e=\frac 12+\e$, where  $0<\e<\frac 12$,
 and \eqref{inte reiterated}
	imply
		\beq\label{eq: interpolation of S}
	\|S^e_{q_1,q_2}\|_{H^1_0 (\p M\times (0,T))\to L^2 (\p M\times (0,T))}	&\le & C'
	\|S^e_{q_1,q_2}\|_{H^{s_\e}_0\to L^2}^{1-\theta_\e}
		\|S^e_{q_1,q_2}\|_{H^{2m}\cap H^{m}_0\to L^2}^{\theta_\e}\\
		\nonumber
		&\le & C'
	\|S^e_{q_1,q_2}\|_{H^{s_\e}_0(\p M\times (0,T))\to H^{s_\e-1/2}(\p M\times (0,T)))}^{1-\theta_\e}
		\|S^e_{q_1,q_2}\|_{H^{2m}\cap H^{m}_0\to L^2}^{\theta_\e },
	\eeq
	where $C'>0$ is the interpolation constant and  $\theta_\e=(1-s_\e)/(2m- s_\e)$, so that $(1-\theta_\e)s_\e+\theta_\e \cdot 2m=1$.
		By \eqref{DN Map def.}, $S^e_{q_1,q_2}f=(\Lambda_{q_2}-\Lambda_{q_1})f $ for $f\in H_0^{1}(\p M\times (0,T))$, see \eqref{DN Map def.}, and thus the above
		formulas \eqref{Lambda and varphi estimate} and
		\eqref{eq: interpolation of S} imply
				\ba
	\|\Lambda_{q_2}-\Lambda_{q_1}\|_{H^1_0 (\p M\times (0,T))\to L^2 (\p M\times (0,T))}	\le 
	 C_3 
\|\Lambda_{q_2}-\Lambda_{q_1}\|_{H^{2m}\cap H^{m}_0\to L^2}^{\theta_\e},	 
		\ea
		where $$
	C_3=C'\cdot \sup_{q_1,q_2\in \xxx} \|S^e_{q_1,q_2}\|_{H^{s_\e}_0(\p M\times (0,T))\to H^{s_\e-1/2}(\p M\times (0,T)))}^{1-\theta_\e} \le
	C'\cdot (C'C_2\|q_2-q_1\|_{\xxx})^{1-\theta_\e}.
	$$
	Thus by \eqref{Lambda and varphi estimate},
				\ba
	\|\Lambda_{q_2}-\Lambda_{q_1}\|_{H^1_0 (\p M\times (0,T))\to L^2 (\p M\times (0,T))}	\le 
	 C_3  \|\varphi({q_2})-\varphi({q_1})\|_{F_{\varphi}}^{\theta_\e}.
	\ea

    Then \eqref{stability ip} implies that
	\ba
	\| {\Psi}( \varphi(x_1))-{\Psi}( \varphi(x_2))\|_{L^2(M)}\le C_4\|\varphi(x_1)-\varphi(x_2)\|^{\beta_2}
	_{F_{\varphi}},\quad
	x_1,x_2\in \xxx,
	\ea
	with some $C_4>0$ and  $\beta_2\in (0,1)$.  %
	By the definition of the space $\xxx$, $\| {\Psi}( \varphi(x_1))-{\Psi}( \varphi(x_2))\|_{H^{k+1,p}(M)}\le 2R_0$
	for all $x_1,x_2\in \xxx$. 
Thus by using interpolation of Sobolev spaces $W^{k,p}$, see
\cite[Sec.\ 4.3.1, Thm.\ 2]{Triebel}, we conclude that
	\ba
	\| {\Psi}( \varphi(x_1))-{\Psi}( \varphi(x_2))\|_{H^{k,p}(M)}\le C_5\|\varphi(x_1)-\varphi(x_2)\|_{F_{\varphi}}^{\beta_3},\quad
	x_1,x_2\in \xxx,
	\ea
	with some $C_5>0$ and $\beta_3\in (0,1)$.
	
	As $F_{\varphi}$ is a Hilbert space and $p=2$, 
	the Benyamini-Lindenstrauss theorem, \cite[Theorem 1.12]{BenyaminiLindenstrauss_2000_NonlinearFunctionalAnalysis}, see also \cite{LancienRandrianatoanian_2005_Extension_Holdermapsin_CX} and \cite{Naor_2001_Phasetransition_isometricisomorphic_extension_Holderfunctionsbetween_Lpspaces}, implies that
	 the map ${\Psi}: \varphi(\xxx)\to H^{k,p}(M)$, 
 defined  a priori in the subset $\varphi(\xxx)\subset F_{\varphi}$, has 
 a H\"older-smooth extension to the vector space $F_{\varphi}$.
Moreover, by  \cite[Theorem 1.12]{BenyaminiLindenstrauss_2000_NonlinearFunctionalAnalysis}, there is a $C_6>0$ such that the extension ${\Psi}:F_{\varphi}\to H^{k,p}(M)$ satisfies 
	\ba
	\| {\Psi}(y_1)-{\Psi}( y_2)\|_{\xxx}\le C_6C_5\|y_1-y_2\|^{\beta_3}_{F_{\varphi}},\quad
	y_1,y_2\in F_{\varphi}.
	\ea	
 Finally, we remark that as $F_{\varphi}$ is a separable Hilbert space, it has a $\lambda$-BAP.
	\end{proof}

\subsection{Supporting Results and Proofs}
\label{s:SupportingResults_Proofs}
The following lemma shows that the generalized snowflake transform of a metric space is again a metric space and it shows that the generalized snowflake transform of a doubling metric space is again doubling, quantitatively.  The second result is demonstrated by showing that the generalized snowflake transform is a quasi-symmetric map\footnote{Thus the second part of our argument is an explicit version of \cite[Theorem 10.18]{heinonen2001lectures} which shows the quasi-symmetric image of a doubling metric space is doubling.}.  In particular, this lemma generalizes \cite[Lemma 6.1]{AcciaioKratsiosPammer2022} which shows provides exact estimates on the doubling number of the snowflaked doubling metric spaces. 
Following \cite{PaulMarius_usefulnote_generalizedinverse_2013}, we define the generalized inverse of an increasing function $\eta:\rr\rightarrow \rr$ by $\eta^{\dagger}(t)\eqdef \inf\{s\in \rr:\, \eta(s)\ge t\}$, where $t \in \rr$.  
\begin{lemma}
\label{lemma_quantiative_doublinginvariance_of_quasisymmetricmaps}
Let $\phi:\xxx\rightarrow \yyy$ be a quasisymmetric map and let $K\Subset \xxx$ be doubling with doubling constant $C_{(K,d_{\xxx})}>0$. Then $\phi(K)$ is doubling with doubling constant at most $C_{(K,d_{\xxx})}^{
\lceil
-\log_2(\eta^{\dagger}(\frac{1}{4}))/4
\rceil
}$. 
\end{lemma}
\begin{proof}
Since $\phi$ is a quasisymmetric map, it is a homeomorphism onto its image; thus $\phi\vert _{\phi(K)}^{-1}:\phi(K)\rightarrow K$ is well-defined and uniformly continuous. We denote its modulus of continuity by $\omega_{\phi\vert _{\phi(K)}^{-1}}$.  We fix a $y\in \phi(K)$ and $0<r\le \operatorname{diam}(\phi(K))$. For every $u \in \operatorname{Ball}_{\yyy}(y,r)\cap \phi(K)$ we then have that 
$
d_{\xxx}(\phi^{-1}(y),\phi^{-1}(u)) \le \omega_{\phi\vert _{\phi(K)}^{-1}}(r);
$ 
therefore, $\operatorname{diam}(\phi^{-1}(B_{\yyy}(y,r)\cap K))= 2 \omega_{\phi\vert _{\phi(K)}^{-1}}(r)$. By \cite[Lemma 6.1]{AcciaioKratsiosPammer2022}, there are $x_1,\dots,x_N\in K$ with $C_{(K,d_{\xxx})}^{\lceil \log_2(-\log_2(\eta^{\dagger}(1/4)/4)\rceil}$ covering $\phi^{-1}(B_{\mathcal{Y}}(y,r)\cap K)$ with $\operatorname{diam}(\phi(B(x_i,r)))\le r \eta(4 \eta^{\dagger}(1/4)/4) \le r/4$ by \cite[Proposition 10.8]{heinonen2001lectures} and \cite[Proposition 2.3 (3)]{PaulMarius_usefulnote_generalizedinverse_2013} (using that $\eta$ is strictly increasing).
\end{proof}

The following argument is due to \textit{John Von Name}. We refer the reader to \cite{JohnVonName_UniformDensity__FullPost} for the original argument and for a lengthier discussion on the topic.  
\begin{proof}[Proof of Lemma~\ref{lemma:change_of_metric}]
Since $\xxx$ and $\yyy$ are separable, the topology of uniform convergence on compact sets is a separable topology on $C(\xxx,\yyy)$.  Therefore, we may pick a dense countable family $(f_i)_{i\in I}$ of functions in $C(\xxx,(\yyy,d_{\yyy}))$, where $I$ is a non-empty countable (possibly infinite) indexing set. We define the real positive sequence $\{k_i\}_{i\in I}$ by
\[
    k_i
        \eqdef
    \frac{1}{
        2^i\,\big(\max_{j=1,\dots,i}\max_{y,\tilde{y}\in f_i(\xxx)}\, \max\{d_{\yyy}(y,\tilde{y}),1\}\big)
    }
.
\]
If $I$ is finite then $k_i\cdot\text{diam}(f_i(\xxx))$ is bounded and if $I$ is infinite then, by construction, $\lim\limits_{i\mapsto \#I}\,k_i\cdot\text{diam}(f_i(\xxx))= 0$.  In either case, the metric
\[
        \tilde{d}_{\xxx}(x,\tilde{x})
    \eqdef 
        d(x,\tilde{x})
        +
        \sup_{i\in I}\,k_i\cdot d_{\yyy}(f_i(x),f_i(\tilde{x}))
\] 
is a well-defined metric on $\xxx$.  
Furthermore, by continuity of each $f_i$, the metric $d_{\xxx}$ generates the topology on $\xxx$; consequently,
$C((\xxx,d_{\xxx}),(\yyy,d_{\yyy}))$ $=$ $C((\xxx,\tilde{d}_{\xxx}),(\yyy,d_{\yyy}))$.  Moreover, by construction, for each $i\in I$ we have  $d_{\yyy}(f_i(x),f_i(\tilde{x}))\leq k_i^{-1}\cdot \tilde{d}_{\xxx}(x,\tilde{x})$, for every $x,\tilde{x}\in \xxx$; hence, each member of the dense countable family $(f_i)_{i\in I}$ is Lipschitz and therefore the set of Lipschitz functions from $\xxx$ to $\yyy$ is dense in $C((\xxx,d_{\xxx}),(\yyy,d_{\yyy}))$.  
\end{proof}

\begin{proof}[{Proof of Proposition~\ref{prop:phi_typical}}]
Fix an arbitrary base point $x_0\in \xxx$ and consider the isometric embedding $\varphi_1:(\xxx,d_{\xxx})\rightarrow \mbox{\AE}(\xxx,x_0)$ given by $\phi_1(x)\mapsto \delta_x-\delta_{x_0}$ with $\text{\AE}(\xxx,x_0)$ being the Lipschitz free space.  

\paragraph{Case 1 - $\xxx$ is Finite:}
Suppose, first, that $\xxx$ is finite.  Since $\text{\AE}(\xxx,x_0)$ is spanned by the finite set $\{\delta_x\}_{x\in \xxx}$, it is a finite-dimensional normed space. As all norms are equivalent on finite-dimensional Banach spaces, there is a linear bi-Lipschitz map $\phi_2:(\text{\AE}(\xxx,x_0),\|\cdot\|_{\text{\AE}(\xxx,x_0)})\rightarrow (\mathbb{R}^{\#\xxx},\|\cdot\|_2)$.  Since $\phi_3:(\mathbb{R}^{\#\xxx},\|\cdot\|_2):x\mapsto (x_1,\dots,x_{\#\xxx},0,0,0,\dots)\in (\ell^2,\|\cdot\|_2)$ is an isometric embedding, $\phi\eqdef \phi_3\circ \phi_2\circ \varphi_1$ leads to the conclusion.

\paragraph{Case 2 - $\xxx$ is Infinite}\hfill\\
Now, suppose that $\xxx$ is infinite.  
Since $\xxx$ is separable then $\text{\AE}(\xxx,x_0)$ is also separable, simply consider a countable dense subset $\{x_i\}_{i\in I}$ of $\xxx$ then observe that the span of $\{x_i\}_{i\in I}$ with rational coefficients is dense in $\text{\AE}(\xxx,x_0)$.  Since $\{\delta_{x}\}_{x\in \xxx}$ is a linearly independent set in $\text{\AE}(\xxx,x_0)$, then $\text{\AE}(\xxx,x_0)$ is infinite-dimensional since $\xxx$ is infinite.  Thus, the Anderson-Kadec Theorem (see \citep[Theorem 5.1]{BessagaPelczynski_1975_SelectTopicsInfDimTopology_Book}) applies; whence there is a homeomorphism $\phi_2:\text{\AE}(\xxx,x_0)\rightarrow (\ell^2,\|\cdot\|_2)$.  Taking $\varphi\eqdef \varphi_2\circ \varphi_1$ yields the conclusion. 
\end{proof}

    \begin{proof}[{Proof of Proposition~\ref{prop:simple_Interpretation}}]
    Fix an $x\in \xxx$.  The $1$-Wasserstein distance $\mathcal{W}_1(\hat{T}(x),\delta_{f(x)})$ is defined as in~\eqref{eq:Wassersteindistance} by minimizing over all Radon probability measures $\pi$ on $\yyy\times \yyy$ with ``marginals'', that is, push-forwards by the canonical projections of the Cartesian product $\xxx\times \xxx$ onto its two components, $\hat{T}(x)$ and $\delta_{f(x)}$, respectively. However, the only such $\pi$ is the product measure $\hat{T}(x) \otimes \delta_{f(x)}$ (see the proof of \citep[Lemma 6.4]{kratsios2022universal} for a standard argument). Therefore,
    \allowdisplaybreaks
    \begin{align}
    \label{eq:def_pi}
            \mathcal{W}_1\big(
                \hat{T}(x)
                    ,
                \delta_{f(x)}
            \big)
        = &
            \inf_{\pi} \int_{(y_1,y_2)\in \yyy\times \yyy}\,
                d_{\yyy}(y_1,y_2)
                \, 
                \pi(d(y_1,y_2))
    \\
    \notag
        = &
            \int_{(y_1,y_2)\in \yyy\times \yyy}\,
                d_{\yyy}(y_2,y_1)
                \, 
                [\hat{T}(x)\otimes \delta_{f(x)}](d(y_1,y_2))
    \\
    \label{eq:fubini}
        = &
            \int_{y_2\in \yyy}
            \int_{y_1\in \yyy}
            \,
                d_{\yyy}(y_2,y_1)
                \, 
                \delta_{f(x)}(dy_1)
                \hat{T}(x)(dy_2)
    \\
    \label{eq:def_pi__RHS}
        = &
            \int_{y\in \yyy}
            \,
                d_{\yyy}(y,f(x))
                \, 
                \hat{T}(x)(dy),
    \end{align}
    where the infimum in~\eqref{eq:def_pi} is taken over all Radon probability measures on $\yyy\times \yyy$ with respective marginals $\hat{T}(x)$ and $\delta_{f(x)}$, and \eqref{eq:fubini} follows from Fubini's Theorem (see \citep[Theorem 1.27]{Kallenberg_FMP_Book}).  Taking the supremum on both sides of~\eqref{eq:def_pi} and~\eqref{eq:def_pi__RHS} yields
    \[
            \sup_{x\in \xxx}\,
                \mathcal{W}_1\big(
                    \hat{T}(x)
                        ,
                    \delta_{f(x)}
                \big)
        =
            \sup_{x\in \xxx}\,
                \int_{y\in \yyy}
                \,
                    d_{\yyy}(y,f(x))
                    \, 
                    \hat{T}(x)(dy)
        \eqdef
            \mathbb{E}_{Y\sim \hat{T}(x)}\big[
                d_{\yyy}(Y,f(x))
            \big]
        .
    \]
    \end{proof}
    \begin{proof}[{Proof of Corollary~\ref{cor:uniformW1_small_implies_hpestimates}}]
    Let $\varepsilon>0$, $Y_1,\dots,Y_N$ be independent random variables on a probability space $(\Omega,\mathcal{F},\mathbb{P})$ with $Y_n\sim \hat{T}(x_n)$ for $n=1,\dots,N$, and suppose that 
    \[
            \max_{n=1,\dots,N}\,
                \mathcal{W}_1\big(\hat{T}(x_n),f(x_n)\big)
        <
            \varepsilon
        .
    \]
    Let $\delta>0$ be a constant, to be chosen below.  By the Chebyshev--Markov inequality \citep[Lemma 4.1]{Kallenberg_FMP_Book} and by Proposition~\ref{prop:simple_Interpretation}, respectively, the following holds, for each $n=1,\dots,N$
    \allowdisplaybreaks
    \begin{align}
    \label{eq:cor:uniformW1_small_implies_hpestimates__firstestimate___BEGIN}
            \mathbb{P}\big(
                    d_{\yyy}(Y_n,f(x_n))
                \ge 
                    \delta
            \big)
        \le & 
            \frac{
                \mathbb{E}_{\mathbb{P}}\big[
                    d_{\yyy}(Y_n,f(x_n))
                \big]
            }{\delta}
        \\
        \notag
        = & 
            \frac{
                \mathbb{E}_{Y\sim \hat{T}(x_n)}\big[
                    d_{\yyy}(Y,f(x_n))
                \big]
            }{\delta}
        \\
        \notag
        = & 
            \frac{
                \mathcal{W}_1\big(\hat{T}(x_n),\delta_{f(x_n)}\big)
            }{\delta}
        \\
        < & 
            \frac{
                \varepsilon
            }{\delta}
    \label{eq:cor:uniformW1_small_implies_hpestimates__firstestimate___END}
    .
    \end{align}
    As $Y_1,\dots,Y_N$ are independent, the estimate in~\eqref{eq:cor:uniformW1_small_implies_hpestimates__firstestimate___BEGIN}-\eqref{eq:cor:uniformW1_small_implies_hpestimates__firstestimate___END} implies that
    \allowdisplaybreaks
    \begin{align*}
            \mathbb{P}\biggl(\,
                    \max_{n=1,\dots,N}
                    \,
                    d_{\yyy}(Y_n,f(x_n))
                \le 
                    \delta
            \biggr)
        = & 
           \prod_{n=1}^N\,
            \mathbb{P}\biggl(\,
                d_{\yyy}(Y_n,f(x_n)) \le \delta
            \biggr)
        \ge 
           \biggl(1-\frac{\varepsilon}{\delta}\biggr)^N
    .
    \end{align*}
    Since $N>0$, setting $\delta \eqdef \sqrt{\varepsilon}N>0$ yields the statement.
    \end{proof}

\section{Acknowledgments}
\label{s:Acknowledgments}
\paragraph{Funding Sources} AK was funded by the NSERC Discovery grant RGPIN-2023-04482; they also gratefully acknowledge the support of McMaster University via their McMaster startup grant, CL gratefully acknowledges the startup grant from ShanghaiTech University, AK and ID were funded by the European Research Council (ERC) Startup Grant 852821-SWING.  ML was partially supported by Academy of Finland, projects 312339 and 320113 and the Finnish Centre of Excellence on Inverse Modelling and Imaging.
MVdH gratefully acknowledges support from the Department of Energy under grant DE-SC0020345, the National Science Foundation under grant DMS-2108175 and the Simons Foundation under the MATH + X program.  

\paragraph{People} AK would like to thank Giuliano Basso for the insightful references surrounding geodesic bicombings on the Heisenberg group and Andrew Colinet for their example of low-regularity H\"{o}lder-like moduli of continuity, and Joseph Van Name for showing AK Lemma~\ref{lemma:change_of_metric}.  
AK would also like to thank Behnoosh Zamanlooy for her all support over the years.

\bibliographystyle{elsarticle-num} 
\bibliography{0_References}
\end{document}